 \documentclass[final,3p,times]{elsarticle}
\usepackage{ulem}
\usepackage{cancel}
\usepackage{graphicx}
\usepackage{exscale}
\usepackage{times}
\usepackage{color}
\usepackage{amssymb}
\usepackage{setspace}
\usepackage{natbib}
\usepackage{amsmath}
\font\ppppppcarac=ptmr8y at 4pt
\font\pppppcarac=ptmr8y at 5pt
\font\ppppcarac=ptmr8y at 6pt
\font\pppcarac=ptmr8y at 7pt
\font\ppcarac=ptmr8y at 8pt
\font\pcarac=ptmr8y at 9pt

\font\carac=ptmr8y at 9pt

\font\Ppcarac=ptmr8y at 10pt
\font\bf=ptmb8y at 10pt
\font\ppbf=ptmb8y at 8pt

\newcommand{\bfC}{{\textbf{C}}}

\newcommand{\bfG}{{\textbf{G}}}
\newcommand{\bfH}{{\textbf{H}}}

\newcommand{\bfK}{{\textbf{K}}}
\newcommand{\bfL}{{\textbf{L}}}

\newcommand{\bfP}{{\textbf{P}}}
\newcommand{\bfQ}{{\textbf{Q}}}

\newcommand{\bfU}{{\textbf{U}}}

\newcommand{\bfW}{{\textbf{W}}}
\newcommand{\bfX}{{\textbf{X}}}
\newcommand{\bfY}{{\textbf{Y}}}

\newcommand{\bfzero}{{ \hbox{\bf 0} }}
\newcommand{\ppbfzero}{{\hbox{\ppbf 0}}}

\newcommand{\bfb}{{\textbf{b}}}

\newcommand{\bff}{{\textbf{f}}}
\newcommand{\bfg}{{\textbf{g}}}
\newcommand{\bfh}{{\textbf{h}}}

\newcommand{\bfp}{{\textbf{p}}}
\newcommand{\bfq}{{\textbf{q}}}
\newcommand{\bfr}{{\textbf{r}}}

\newcommand{\bfu}{{\textbf{u}}}

\newcommand{\bfw}{{\textbf{w}}}
\newcommand{\bfx}{{\textbf{x}}}
\newcommand{\bfy}{{\textbf{y}}}

\newcommand{\bfGamma}{{\boldsymbol{\Gamma}}}

\newcommand{\bfgamma}{{\boldsymbol{\gamma}}}

\newcommand{\bfeta}{{\boldsymbol{\eta}}}

\newcommand{\bflambda}{{\boldsymbol{\lambda}}}

\newcommand{\bfvarphi}{{\boldsymbol{\varphi}}}
\newcommand{\bfpsi}{{\boldsymbol{\psi}}}

\newcommand{\bfsigma}{{\boldsymbol{\sigma}}}

\newcommand{\HH}{{\mathbb{H}}}

\newcommand{\KK}{{\mathbb{K}}}

\newcommand{\MM}{{\mathbb{M}}}

\newcommand{\PP}{{\mathbb{P}}}

\newcommand{\RR}{{\mathbb{R}}}

\DeclareMathAlphabet{\mathonebb}{U}{bbold}{m}{n}
\def\11{{\ensuremath{\mathonebb{1}}}}

\def\bb{{\ensuremath{\mathonebb{b}}}}

\newcommand{\curA}{{\mathcal{A}}}
\newcommand{\curB}{{\mathcal{B}}}
\newcommand{\curC}{{\mathcal{C}}}
\newcommand{\curD}{{\mathcal{D}}}
\newcommand{\curE}{{\mathcal{E}}}

\newcommand{\curJ}{{\mathcal{J}}}

\newcommand{\curL}{{\mathcal{L}}}

\newcommand{\curP}{{\mathcal{P}}}

\newcommand{\curR}{{\mathcal{R}}}

\newcommand{\curT}{{\mathcal{T}}}
\newcommand{\curU}{{\mathcal{U}}}

\newcommand{\curW}{{\mathcal{W}}}

\newcommand{\curY}{{\mathcal{Y}}}

\newcommand{\bfcurL}{{\boldsymbol{\mathcal{L}}}}
\newcommand{\bfcurN}{{\boldsymbol{\mathcal{N}}}}

\newcommand{\bfcurR}{{\boldsymbol{\mathcal{R}}}}

\newcommand{\bfcurX}{{\boldsymbol{\mathcal{X}}}}

\newcommand{\bfcurZ}{{\boldsymbol{\mathcal{Z}}}}

\newcommand{\pMC}{{\hbox{{\pppppcarac MC}}}}
\newcommand{\ppMC}{{\hbox{{\ppppppcarac MC}}}}
\newcommand{\pCFD}{{\hbox{{\pppppcarac CFD}}}}
\newcommand{\pref}{{\hbox{{\pppcarac ref}}}}
\newcommand{\err}{\hbox{{\Ppcarac err}}}
\newcommand{\errp}{\hbox{{\ppcarac err}}}
\newcommand{\algo}{\hbox{{{\textit{algo}}}}}
\newcommand{\ar}{{\hbox{{\ppppcarac ar}}}}
\newcommand{\arp}{{\hbox{{\ppppcarac ar}}}}

\newcommand{\ad}{{{\hbox{{\pppcarac ad}}}}}
\newcommand{\bfnabla}{{{\boldsymbol{\nabla}}}}
\newcommand{\pmax}{{\hbox{{\pppcarac max}}}}
\newcommand{\pmin}{{\hbox{{\pppcarac min}}}}
\newcommand{\ptime}{{\hbox{{\ppppcarac time}}}}
\newcommand{\pptime}{{\hbox{{\pppppcarac time}}}}
\newcommand{\pNL}{{\hbox{{\pppppcarac NL}}}}
\newcommand{\pNS}{{\hbox{{\pppppcarac NS}}}}
\newcommand{\pSD}{{\hbox{{\pppppcarac SD}}}}
\newcommand{\pDIV}{{\hbox{{\pppppcarac DIV}}}}
\newcommand{\KL}{{\hbox{{\ppppcarac KL}}}}
\newcommand{\pKL}{{\hbox{{\pppppcarac KL}}}}
\newcommand{\PCA}{{{\hbox{{\pppppcarac PCA}}}}}
\newcommand{\opt}{{\hbox{{\pppcarac opt}}}}
\newcommand{\popt}{{\hbox{{\ppppcarac opt}}}}
\newcommand{\ppopt}{{\hbox{{\pppppcarac opt}}}}
\newcommand{\diff}{{\hbox{{\ppppcarac diff}}}}
\newcommand{\pdiff}{{\hbox{{\pppppcarac diff}}}}
\newcommand{\mode}{{ \hbox{\pppcarac mode} }}
\newcommand{\pmode}{{ \hbox{\ppppcarac mode} }}

\newcommand{\pparam}{{\hbox{{\ppppcarac param}}}}
\newcommand{\psp}{{\hbox{{\ppppcarac sp}}}}
\newcommand{\pfreq}{{\hbox{{\ppppcarac freq}}}}

\newcommand{\pObs}{{\hbox{{\pcarac Obs}}}}
\newcommand{\ppObs}{{\hbox{{\ppcarac Obs}}}}
\newcommand{\pdB}{{\hbox{{\pcarac dB}}}}

\newcommand{\dof}{{\hbox{{\Ppcarac dof}}}}
\newcommand{\pdof}{{\hbox{{\carac dof}}}}

\newcommand{\linear}{{\hbox{{\pppcarac lin}}}}
\newcommand{\sol}{{\hbox{{\pppcarac sol}}}}
\newcommand{\psol}{{\hbox{{\ppppcarac sol}}}}
\newcommand{\tr}{\hbox{{\Ppcarac tr}}\,}
\newcommand{\cov}{\hbox{{\Ppcarac cov}}}
\newcommand{\var}{\hbox{{\Ppcarac var}}}
\newcommand{\wien}{{\hbox{{\pppcarac wien}}}}

\newcommand{\st}{{\hbox{{\pppcarac st}}}}

\newcommand{\pnorm}{{\hbox{{\ppppcarac norm}}}}
\newcommand{\pSB}{{\hbox{{\pppppcarac SB}}}}
\newcommand{\pIC}{{\hbox{{\ppppcarac IC}}}}
\newcommand{\pHS}{{\hbox{{\ppppcarac HS}}}}

\journal{arXiv}

\begin{document}

\begin{frontmatter}

\title{Probabilistic learning on manifolds constrained by nonlinear\\
  partial differential equations for small datasets}


\author[1]{C. Soize \corref{cor1}}
\ead{christian.soize@univ-eiffel.fr}
\author[2]{R. Ghanem}
\ead{ghanem@usc.edu}

\cortext[cor1]{Corresponding author: C. Soize, christian.soize@univ-eiffel.fr}
\address[1]{Universit\'e Gustave Eiffel, MSME UMR 8208 CNRS, 5 bd Descartes, 77454 Marne-la-Vall\'ee, France}
\address[2]{University of Southern California, 210 KAP Hall, Los Angeles, CA 90089, United States}

\begin{abstract}
A novel extension of the Probabilistic Learning on Manifolds (PLoM) is
presented. It makes it possible to synthesize solutions to a wide
range of nonlinear stochastic boundary value problems described by
partial differential equations (PDEs) for which a stochastic computational
model (SCM) is available and depends on a vector-valued random control parameter.
The cost of a single numerical evaluation of this SCM is assumed to be
such that only a limited number of points can be computed for
constructing the training dataset (small data). Each point of the
training dataset is made up realizations from a vector-valued
stochastic process (the stochastic solution) and the associated random
control parameter on which it depends.
The presented  PLoM constrained by PDE allows for generating a large
number of learned realizations of the stochastic process and its
corresponding random control parameter. These learned realizations are
generated so as to minimize the vector-valued random residual of the
PDE in the mean-square sense. Appropriate novel methods are developed
to solve this challenging problem. Three applications are
presented. The first one is a simple uncertain nonlinear dynamical
system with a nonstationary stochastic excitation. The second one
concerns the 2D nonlinear unsteady Navier-Stokes equations for
incompressible flows in which the Reynolds number is the random
control parameter. The last one deals with the nonlinear dynamics of a
3D elastic structure with uncertainties. The results obtained make it
possible to validate the PLoM constrained by stochastic PDE but also
provide further validation of the PLoM without constraint.
\end{abstract}


\end{frontmatter}

\section{Introduction}
\label{Section1}
%

\noindent\textit{(i) Context}.
Physics-informed machine learning (eg. neural networks and deep learning) has seen a growing interest in recent years.
Numerous works in this area of machine learning have been carried out for Schr{\"o}dinger equation \cite{Mills2017}, for reconstructing Reynolds stress model in turbulent flow \cite{Wang2017} and for augmenting turbulence model \cite{Singh2017,Wu2018}, for predicting  large eddy simulation \cite{Yang2019},  for projection-based model reduction \cite{Swischuk2019}.
Physics-informed machine learning has also been used in the context of linear  \cite{Raissi2017}, quasilinear \cite{Sirignano2018}, and nonlinear partial differential equations \cite{Raissi2018,Raissi2019}. All these approaches are based on the use of Neural Networks, which, in general, require big data  for training.
In machine learning, probabilistic or statistical learning deals with the problem of inferring a function based on data.
In particular for supervised learning, it consists in training from a set of data consisting of input-output pairs.
The statistical and probabilistic learning approaches have been extensively developed (see for instance,
\cite{Scholkopf1997,Vapnik2000,Aggarwal2012,Dalalyan2012,Murphy2012,Balcan2013,James2013,Dong2014,Ghahramani2015,Taylor2015})
and play an increasingly important role in computational physics and
engineering science, in particular for problems using large scale
computational models. In this framework, statistical  methods for
supervised learning have been developed in the form of surrogate
models from which approximations of the expensive functions can be easily evaluated
\cite{Jones1998,Du2004,Queipo2005,Byrd2011,Eldred2011,Yao2011,HomemdeMello2014,Papadopoulos2018}.
Among the many methodologies available on this topic, we are interested
here in \textit{Probabilistic Learning on Manifolds} (PLoM), which was
initially  introduced  in \cite{Soize2016} and specially developed for
solving unsupervised and supervised problems in the case of small
data, as opposed to big data. The PLoM is not based on the use of
neural networks. The first initial work published in 2016 was
completed from a theoretical point of view in
\cite{Soize2019b,Soize2020c}, extended to the case of the polynomial
chaos representation  \cite{Soize2017a}, to the case of  the learning
in presence of physics constraints \cite{Soize2020a}, and  to the
sampling of Bayesian posteriors \cite{Soize2020b}. The PLoM has been
used with success for probabilistic nonconvex optimization under
constraints and  uncertainties \cite{Ghanem2018a} that has allowed for
analyzing very complex engineering systems such as hypersonic
combustion flows \cite{Ghanem2019,Soize2019}, optimal placement of
wells \cite{Ghanem2018b},  the design optimization under uncertainties
of mesoscale implants \cite{Soize2018a}, the quantification of model
uncertainties in nonlinear computational dynamics
\cite{Farhat2019,Soize2019a}, the fracture paths in random composites
\cite{Guilleminot2020}, the digital twin for composite material
systems \cite{Ghanem2020}, the computation of Sobol indices for
complex stochastic system \cite{Arnst2020}, and the detection of
structural changes from dynamic monitoring \cite{Soize2020d}. \\

\noindent\textit{(ii) Objective of the paper}. As already mentioned,
the PLoM is not based on the use of neural networks and was developed
for the unsupervised or supervised cases of small data. For the
supervised case, these small data  constitute the  training dataset
that is constructed as $N_d$ input-output pairs. The PLoM then allows,
using only the training dataset, and by a relying on purely
probabilistic arguments, to generate a large number $n_\pMC \gg N_d$
of additional realizations (the learned dataset) while preserving the
concentration of the probability measure on an intrinsic  manifold
delineated using diffusion maps \cite{Coifman2005}.
We are then entitled to ask the following fundamental question: do the
$n_\pMC$ additional realizations generated by the PLoM satisfy, in a
"reasonable" sense, the stochastic Partial Differential Equation (PDE)
that was used to build the training dataset. As previously explained,
the PLoM has been validated for many situations and the convergence of
the learning with respect to $N_d$ has theoretically and numerically
been studied. This paper will give further confirmation of this
point. However, for certain problems, we want to be able to control
that the  additional realizations satisfy, with a quantifiable metric,
the stochastic PDE underlying the stochastic computational model used
to generate $N_d$ points of the training dataset. This is the objective
of this paper: to develop and validate the PLoM method constrained by the
stochastic PDE that has been used to build the training dataset.\\

\noindent\textit{(iii) Novelty of the paper}. Two novel extensions of
the PLoM are presented, which allow to give a constructive answer to
our objective. The first is to extend the initial PLoM for nonlinear
stochastic dynamical systems (that is to say, physics system depending on time)
such as those of the unsteady computational fluid dynamics (CFD) or the nonlinear
solid dynamics (SD) in presence of a vector-valued random control parameter.
Indeed the PLoM was developed for random vectors and it is therefore necessary
to extend it to the case of vector-valued stochastic processes for analyzing
these types of stochastic dynamical systems.
The second is to extend the PLoM method to take into account the constraint
that the generated additional realizations verify  the stochastic PDE in a certain sense.
These two novel extensions constitute a real challenge and correspond to a difficult problem,
which will necessitate the introduction of appropriate methods.\\

\noindent\textit{(iv) Organization of the paper}.
In order to facilitate the reading of this work, we have chosen to
present an abstract formulation in the first part in order to better
introduce the concepts and methods. These developments are presented
for a generic nonlinear stochastic computational dynamical model,
which results from the spatial discretization of a nonlinear
stochastic boundary value problem.
We delay the introduction of the complexity in the
stochastic boundary value problem until the sections where the
respective applications are introduced. We can thus focus our
attention initially on setting up the PLoM with PDE constraints.
Later in the paper, we consider nonlinear PDEs with random
coefficients.  We intersperse in that presentation additional new
developments required to efficiently implement our proposed
procedure.  We specifically introduce two applications that are representative of
broad classes of problems, namely a first order (CFD case) and a second
order (SD case) problems, which involve one field (SD case) or several fields (CFD case)
whose partial time derivatives are different according to the field in
question (eg. velocity field and pressure field).

\section*{Notations}
\noindent Lower-case letters such as $q$ or $\eta$ are deterministic real variables.\\
Boldface lower-case letters such as $\bfq$ or $\bfeta$ are deterministic vectors.\\
Upper-case letters such as $X$ or $H$ are real-valued random variables.\\
Boldface upper-case letters such as $\bfX$ or  $\bfH$ are vector-valued random variables.\\
Lower-case letters between brackets such as  $[x]$ or $[\eta]$ are deterministic matrices.\\
Boldface upper-case letters between brackets such as $[\bfX]$ or $[\bfH]$ are matrix-valued random variables.\\

\noindent
$[I_{n}]$: identity matrix in $\MM_n$.\\
$\MM_{n,N}$: set of all the $(n\times N)$ real matrices.\\
$\MM_N$: set of all the square $(n\times n)$ real matrices.\\
$\MM_n^{+0}$: set of all the positive symmetric $(n\times n)$ real matrices.\\
$\MM_n^+$: set of all the positive-definite symmetric $(n\times n)$ real matrices.\\
$\RR$: set of all the real numbers.\\
$\RR^n$: Euclidean vector space on $\RR$ of dimension $n$.\\
$[x]_{kj}$: entry of matrix $[x]$.\\
$[x]^T$: transpose of matrix $[x]$.\\
$\Vert\cdot\Vert$: usual Euclidean norm on $\RR^n$.\\
$<\! \bfx\, ,\bfy \!>$: usual Euclidean inner product in $\RR^n$.\\
$\delta_{\alpha\beta}$: Kronecker's symbol.\\

\noindent In this paper, any Euclidean space $\curE$ (such as $\RR^{N}$) is equipped with its Borel field $\curB_\curE$, which means that $(\curE,\curB_\curE)$ is a measurable space on which a probability measure can be defined. The mathematical expectation is denoted by $E$.

\section{Abstract nonlinear stochastic computational model, training dataset, and time sampling}
\label{Section2}
\subsection{Nonlinear stochastic equation with its initial condition}
\label{Section2.1}
We consider the abstract nonlinear stochastic computational model (SCM) on $\RR^N$ written as
\begin{align}
\bfcurN_t\big (\bfY(t),t,\bfW\big ) = & \, \bfzero \quad\,\, a.s. \quad , \quad t\in ]t_0,T] \, ,              \label{eq2.1} \\
\bfcurL^\pIC_{t_0}\big (\bfY\big ) = & \, \bfy_0 \quad a.s. \quad , \quad t = t_0     \, ,                      \label{eq2.2}
\end{align}
in which $[t_0,T]$ is a given compact interval of $\RR$  and $\bfY = \big\{\bfY(t),t\in [t_0,T]\big\}$ is the unknown $\RR^N$-valued stochastic process.
The vector $\bfW$ is the $\RR^{n_w}$-valued random control parameter of the system, which can correspond either to a set of system parameters, or to the set of generalized coordinates of a reduced-order representations based on  the Karhunen-Lo\`eve or chaos expansions, (1) of random fields that are the coefficients of partial differential operators of the problem and/or (2) of stochastic processes modeling the time-dependent excitations of the system.
In Eq.~\eqref{eq2.1}, $\bfcurN_t$ is a given nonlinear differential operator in time $t$ of order $m_d \geq 1$ with values in $\RR^N$.
Equation~\eqref{eq2.2} is the deterministic initial condition at $t=t_0$. The operator $\bfcurL^\pIC_{t_0}\big (\bfY\big )$ is a linear differential operator with values in $\RR^{N_0}$  in which $N_0\geq N$ depends on $m_d$. The deterministic vector $\bfy_0$ is given in $\RR^{N_0}$.
For instance, $\bfcurL^\pIC_{t_0}\big (\bfY\big ) = \bfY(t_0)$ and $N_0=N$ for $m_d=1$ (first-order nonlinear differential operator $\bfcurN_t$) or
$\bfcurL^\pIC_{t_0}\big (\bfY\big ) = \big (\bfY(t_0),\dot\bfY(t_0)\big )$ with $N_0=2 N$ for $m_d=2$ (second-order nonlinear differential operator).
In addition, in order to simplify the presentation,  we will assume that the first $m_d$ derivatives of $\bfY$ with respect to $t$ at initial time $t = t_0$ take deterministic values (this simplified hypothesis can easily be removed without difficulty and does not call into question the presented developments).
These operators will be detailed for each of the applications
presented in the sequel.
\subsection{Vector-valued random control parameter}
\label{Section2.2}
The random vector $\bfW= (W_1,\ldots,$ $W_{n_w})$ is defined on a
probability space $(\Theta,\curT,\curP)$. Its prior probability
distribution $P_\bfW(d\bfw) = p_\bfW(\bfw)\, d\bfw$ is defined by a
given probability density function $\bfw\mapsto p_\bfW(\bfw)$ on
$\RR^{n_w}$ with respect to the Lebesgue measure $d\bfw$ on
$\RR^{n_w}$. Its support, $\curC_w \subset \RR^{n_w}$, is the
admissible set of $\bfW$, which will be specified for each of the applications.
\subsection{Uniqueness of the stochastic solution of the SCM}
\label{Section2.3}
Let $m_d$ and $\bfy_0$ be as defined in Section~\ref{Section2.1}.
Let $\curC_\ad^{\,m_d} =\big \{\bfy \in {\rm{C}}^{m_d}([t_0,T],\RR^N) \, ; \, \bfcurL^\pIC_{t_0}(\bfy) =  \bfy_0  \big \}$ in which ${\rm{C}}^{\, m_d}([t_0,T],\RR^N)$ is the set of all the $m_d$-times continuously differentiable functions $t\mapsto \bfy(t) = (y_1(t),\ldots , y_N(t))$ defined on $[t_0,T]$ with values in $\RR^N$. For $m_d=1$, the first time derivative is denoted by $\dot\bfy(t) = d\bfy(t)/dt$  and for $m_d=2$, the second time derivative is denoted by $\ddot\bfy(t) = d^2\bfy(t)/dt^2$, which are continuous functions.
Let $\bfw^\ell = \bfW(\theta_\ell)\in\curC_w \subset\RR^{n_w}$ be any realization of $\bfW$ for $\theta_\ell\in\Theta$. It is assumed that the problem defined by
\begin{align}
\bfcurN_t(\bfy^\ell(t),t,\bfw^\ell) = & \, \bfzero  \quad , \quad t\in ]t_0,T] \, ,                                                         \label{eq2.4} \\
\bfcurL^\pIC_{t_0}(\bfy^\ell) = & \, \bfy_0  \quad , \quad t = t_0     \, .                                                                      \label{eq2.5}
\end{align}
admits a unique solution  $\bfy^\ell = \big\{t\mapsto \bfy^\ell(t)\big\}$ from $[t_0,T]$ into $\RR^N$, which is  $m_d$-times differentiable in time, that is to say,
$(\bfy^\ell, \bfw^\ell)   \in \curC_\ad^{\, m_d}\times \curC_w$.
Consistent with this hypothesis, it is assumed that Eqs.~\eqref{eq2.1}-\eqref{eq2.2} admit a unique stochastic solution
$\bfY^\sol =\big\{\bfY^\sol(t),$ $t\in[t_0,T]\big\}$ that depends on $\bfW$ and $\bfy_0$, defined on $(\Theta,\curT,\curP)$, indexed by $[t_0,T]$, with values in $\RR^N$,
which is assumed to be a second-order stochastic process whose trajectories (sample paths)
are thus $m_d$-times differentiable in time. Consequently, for all $\theta_\ell$ in $\Theta$,
$\big (\bfY^\sol(\cdot ;\theta_\ell), \bfW(\theta_\ell)\big ) \in \curC_\ad^{\, m_d}\times \curC_w $.
It should be noted that the probability measure  of $(\bfY^\sol,\bfW)$
(defined for instance by its system of marginal (or cylindrical)
distributions) cannot be exactly constructed in the considered general
case. We will limit ourselves to building an approximation of this measure using the training dataset and the PLoM without or under constraints.
\subsection{Training dataset $\curY_{N_d}^{\, d}$}
\label{Section2.3bis}
We consider $N_d$ independent realizations $\bfw^\ell\in\curC_w$ of $\bfW$ and the corresponding $N_d$ independent realizations $\bfy^\ell\in \curC_\ad^{\, m_d}$ of $\bfY^\sol$ (see Section~\ref{Section2.3}).
The training dataset $\curY_{N_d}^{\, d}$ is then defined by
\begin{equation}
\curY_{N_d}^{\, d}  = \big\{ (\bfy^1,\bfw^1),\ldots ,(\bfy^{N_d},\bfw^{N_d}) \big\} \subset (\curC_\ad^{\, m_d}\times\curC_w)^{N_d} \, .       \label{eq2.10}
\end{equation}
Thus the construction of  the training dataset $\curY_{N_d}^{\, d}$
requires solving $N_d$ samples of the SCM
defined by Eqs.~\eqref{eq2.4}-\eqref{eq2.5}. As we have explained in Section~\ref{Section1}, it is assumed that the numerical cost for one computation is very high and consequently, only a small value of $N_d$ can be considered (small data case).
\subsection{Time sampling and time integration scheme}
\label{Section2.4}
The construction of the training dataset requires the numerical
solvution of Eqs.~\eqref{eq2.4}-\eqref{eq2.5}. Therefore, we have to introduce a time sampling and a time integration scheme.
For all $\ell \in \big\{1,\ldots , N_d \big\}$, the time sampling is performed (thus, is independent from $\ell$)  with a constant time step $\Delta t$  and $n_\ptime$ sampling times $t_1,\ldots, t_{n_\pptime}$ constituting a finite part $\curJ$ of $[t_0,T]$, such that
\begin{equation}
 t_n = t_0 + n \, \Delta t \,\, , \,\,  n= 1,\ldots n_\ptime \quad , \quad \curJ = \big\{t_1,\ldots, t_{n_\pptime}\big\}\subset [t_0,T]\, ,           \label{eq2.8}
 \end{equation}
in which $\Delta t$ and $n_\ptime$ verify $T - t_0 = n_\ptime \, \Delta t$.
For all $\ell$ in $\big\{1,\ldots,N_d\big\}$, Eqs.~\eqref{eq2.4}-\eqref{eq2.5} are solved using a time integration scheme (such as central difference scheme or Newmark scheme, etc), which must be adapted to the type of nonlinear operator $\bfcurN_t$, that is to say,  which depends on each application (see Sections~\ref{SectionDO}, \ref{SectionNS}, and \ref{SectionSD}).
We then obtain the time sampling of function $\big\{t\mapsto \bfy^\ell(t)\big\}\in \curC_\ad^{\, m_d}$,
\begin{equation}
\big\{\bfy^\ell(t), t\in\curJ\big\} = \big\{\bfy^\ell(t_1),\ldots , \bfy^\ell(t_{n_\pptime})\big\} \quad , \quad \bfy^\ell(t_n)\in\RR^N \, .                       \label{eq2.9}
 \end{equation}
Using the time integration scheme, the time derivative up to order $m_d$ of $\bfy^\ell$ can be computed at all sampling times belonging to $\curJ$.
It should be noted that for each pair $(\bfy^\ell,\bfw^\ell)$ in $\curY_{N_d}^{\, d}$, function $\bfy^\ell$ is only known at the sampling times of $\curJ$.
\subsection{Definition of the stochastic process $\bfY$ and its second-order moments}
\label{Section2.5}
Let us consider the couple $(\bfY,\bfW)$ in which $\bfY
=\big\{\bfY(t),t\in [t_0,T]\big\}$ is a mean-square continuous
second-order stochastic process, defined on $(\Theta,\curT,\curP)$,
indexed by $[t_0,T]$, with values in $\RR^N$, whose trajectories are
assumed to be $m_d$-times differentiable in time, and where the $N_d$
independent realizations of  $(\bfY,\bfW)$ are those defined in
Eq.~\eqref{eq2.10}. Therefore, $(\bfY,\bfW)$ is an approximation of
$(\bfY^\sol,\bfW)$ which depends on $N_d$.
Let $t\mapsto \underline\bfy(t) = E\big\{\bfY(t)\big\}$ be the mean function
of $\bfY$, which is a continuous function from $[t_0,T]$ into $\RR^N$ and
let $(t,t')\mapsto [C_\bfY(t,t')] = E\big \{ \big (\bfY(t) - \underline\bfy(t)\big ) \big (\bfY(t') - \underline\bfy(t')\big )^T\big \}$ be
its matrix-valued covariance function that is a continuous function from $[t_0,T]\times [t_0,T]$ into $\MM_N$.
These two functions are estimated at the sampling times as follows: for $n$ and $n'$ in $\big\{1,\ldots,n_\ptime\big\}$,
\begin{equation}
\underline\bfy(t_n)  = \frac{1}{N_d} \sum_{\ell=1}^{N_d} \bfy^\ell(t_n) \quad , \quad
             [C_\bfY(t_n,t_{n'})] =  \frac{1}{N_d -1} \sum_{\ell=1}^{N_d} \bfy^\ell_c(t_n) \bfy^\ell_c(t_{n'})^T \,                                      \label{eq2.11}
\end{equation}
in which, for $\ell\in\big\{1,\ldots,N_d\big\}$, $\bfy_c^\ell$ is the centered realization such that
\begin{equation}
 \bfy^\ell_c(t_n) =\bfy^\ell(t_n) -  \underline\bfy(t_n) \quad , \quad n\in \big\{ 1,\ldots,n_\ptime \big\} \, .                                                     \label{eq2.12}
\end{equation}
\section{Reduced-order representation of the nonstationary stochastic process $\bfY$}
\label{Section3}
The construction of the reduced-order representation of stochastic process $\bfY =\big\{\bfY(t), t\in [t_0,T]\big\}$ is based on its classical Karhunen-Lo\`eve expansion that will effectively be calculated at sampling times $t_0,t_1,\ldots,t_{n_\pptime}$.
\subsection{Eigenvalue problem for the Karhunen-Lo\`eve (KL) expansion of stochastic process $\{\bfY(t),t\in [t_0,T]\}$}
\label{Section3.1}
The KL expansion of stochastic process $\big\{ \bfY(t), t\in[t_0,T] \big\}$ is based on the use of the spectral analysis of its covariance operator. Let $\HH = L^2_{\mu_0}([t_0,T],\RR^N)$ be the Hilbert space of all the square integrable functions $t\mapsto \bfy(t)$ from $[t_0,T]$ into $\RR^N$, equipped with the inner product $<\! \bfy\, ,\bfy' \!>_\HH $ and its associated norm $\Vert\,\bfy\,\Vert_\HH = <\! \bfy\, ,\bfy \!>_\HH^{1/2}$ such that
$<\! \bfy\, ,\bfy' \!>_\HH =  \int_{t_0}^T <\! \bfy(t)\, ,\bfy'(t) \!> \mu_0(dt)$,
in which the measure $\mu_0(dt)$ on $\RR$ is defined by  $\mu_0(dt) = (T-t_0)^{-1} dt$ with $dt$ the Lebesgue measure on $\RR$. Since the covariance function $(t,t') \mapsto [C_\bfY(t,t')]$ is continuous on the compact set $[t_0,T]\times [t_0,T]$, it can be seen that
$\int_{t_0}^T \int_{t_0}^T \Vert\, [C_\bfY(t,t')] \,\Vert_F^2 $ $\mu_0(dt) \, \mu_0(dt') < + \infty$ in which $\Vert \cdot \Vert_F$\ is the Froebenius norm. Therefore, the covariance linear operator $\bfC_\bfY$ such that $(\bfC_\bfY \bfy)(t)  = \int_{t_0}^T  [C_\bfY(t,t')] \, \bfy(t')\, \mu_0(dt')$ belongs to $\curL_\pHS(\HH)$ (space of the Hilbert-Schmidt operators in $\HH$) \cite{Guelfand1967}. Consequently, the eigenvalue problem $\bfC_\bfY \widetilde\bfvarphi = \widetilde\Lambda \, \widetilde\bfvarphi$ has a countable spectrum $\widetilde\Lambda_1\geq \widetilde\Lambda_2\geq \ldots$ such that $\sum_\alpha \widetilde\Lambda_\alpha^2 < +\infty$. \\

\subsection{Reduced-order representation of discrete stochastic process $\{\bfY(t),t\in\curJ\}$}
\label{Section3.2}
Instead of discretizing the continuous eigenvalue problem $\bfC_\bfY \widetilde\bfvarphi = \widetilde\Lambda \, \widetilde\bfvarphi $, we choose to approximate it by the eigenvalue problem related to the covariance matrix of the random vector
$\bfY_\curJ = \big (\bfY(t_1),\ldots , \bfY(t_{n_\pptime})\big )$ with values in $\RR^{n_\pptime N}$, which corresponds to the time sampling of stochastic process $\big\{ \bfY(t), t\in[t_0,T] \big\}$. This choice is justified by Section~\ref{Section3.1} (the spectrum of $\bfC_\bfY$ is countable) and by the fact that for applications devoted to nonlinear dynamical system, number $n_\ptime$ of sampling times is generally very high, and consequently, the constructed approximation is excellent. Consequently, the KL expansion of stochastic process $\bfY$ is replaced by the Principal Component Analysis (PCA)  of random vector $\bfY_\curJ$.\\

\noindent\textit{(i) Eigenvalue problem for the PCA expansion}. The eigenvalue problem related to the covariance matrix of random vector $\bfY_\curJ$ is written, using Eq.~\eqref{eq2.11}, as
\begin{equation}
\frac{1}{n_\pptime} \sum_{n'=1}^{n_\pptime} [C_\bfY(t_n,t_{n'})] \, \bfvarphi^\alpha(t_{n'}) =
        \Lambda_\alpha \, \bfvarphi^\alpha(t_{n}) \quad , \quad  n\in \big\{1,\ldots, n_\ptime \big\} \, .             \label{eq3.1}
\end{equation}
The eigenvalues are ordered as $\Lambda_1 \geq \Lambda_2 \geq \ldots \geq 0$ and the family of eigenvectors
$\big\{ \bfvarphi^\alpha(t_{1}),\ldots , \bfvarphi^\alpha(t_{n_\pptime})  \big\}_\alpha$ in $\RR^{ n_\pptime N }$ satisfies the orthogonality property,
$(1/n_\pptime) \sum_{n=1}^{n_\pptime} <\! \bfvarphi^\alpha(t_{n})\, ,\bfvarphi^\beta(t_{n}) \!> = \delta_{\alpha\beta}$.\\

\noindent\textit{(ii) Reduced-order representation of discrete stochastic process} $\big\{ \bfY(t),t\in\curJ \big\}$. The  reduced representation
$\big\{\bfY^{(n_q)}(t), t$ $\in\curJ\big\}$ of $\big\{\bfY(t), t\in\curJ\big\}$ of order $n_q$ is constructed using the eigenvectors associated with the first $n_q \ll n_\pptime N$ strictly positive eigenvalues, and is written as
\begin{equation}
\bfY^{(n_q)}(t_n) = \underline\bfy(t_n)  + [V(t_n)]\, \bfQ \quad , \quad  n\in \big\{ 1,\ldots, n_\ptime \big\} \, ,       \label{eq3.2}
\end{equation}
in which $\underline\bfy(t_n) \in\RR^N$ is defined by Eq.~\eqref{eq2.11} and where $[V(t_n)]$ is the matrix defined by
\begin{equation}
 [V(t_n)] = [ \Lambda_1^{1/2}\,\bfvarphi^1(t_{n}) \ldots  \Lambda_{n_q}^{1/2}\,\bfvarphi^{n_q}(t_{n})] \in \MM_{N,n_q} \, .         \label{eq3.3}
\end{equation}
In Eq.~\eqref{eq3.2}, $\bfQ = (Q_1,\ldots , Q_{n_q})$ is a second-order centered and orthogonal random variable defined on probability space $(\Theta,\curT,\curP)$ with values in $\RR^{n_q}$, which is written as
\begin{equation}
\bfQ = [\Lambda]^{-1} \frac{1}{n_\pptime} \sum_{n=1}^{n_\pptime} [V(t_n)]^T\,\big (\bfY(t_n) - \underline\bfy(t_n)\big ) \, ,  \label{eq3.4}
\end{equation}
in which $[\Lambda]$ is the positive-definite diagonal $(n_q\times n_q)$ matrix, $[\Lambda]_{\alpha\beta} = \Lambda_\alpha\, \delta_{\alpha\beta}$. The $N_d$ independent realizations $\big\{\bfq^\ell\big\}_\ell$ in $\RR^{n_q}$  of $\bfQ$, associated with the $N_d$ points of the training dataset $\curY_{N_d}^{\, d}$ defined by Eq.~\eqref{eq2.10}, are computed by
\begin{equation}
\bfq^\ell = [\Lambda]^{-1} \frac{1}{n_\pptime} \sum_{n=1}^{n_\pptime} [V(t_n)]^T\,\big (\bfy^\ell(t_n) - \underline\bfy(t_n)\big )\, .  \label{eq3.5}
\end{equation}
The empirical estimates of the mean vector  $\underline\bfq\in\RR^{n_q}$ and the covariance matrix $[C_\bfQ] \in\MM^+_{n_q}$ of random vector $\bfQ$ are such that
$\underline\bfq = (1/N_d) \sum_{\ell=1}^{N_d}\bfq^\ell =\bfzero_{n_q}$
and $[C_\bfQ] =  \big ( 1/(N_d -1)\big ) \sum_{\ell=1}^{N_d}\bfq^\ell (\bfq^\ell )^T = [I_{n_q}]$.
Using a time integration scheme and the initial condition at $t_0$ yields the $m_d$ time derivatives of $\underline\bfy(t)$ and $[V(t)]$ at
$t\in\big\{t_0,t_1,\ldots ,t_{n_\ptime}\big\}$ and consequently, allows for calculating the $m_d$ time derivatives of $\bfY^{(n_q)}(t_n)$ for $n  \in\big\{0,1,\ldots , n_\ptime\big\}$.
For this type of calculation, an adapted methodology must be developed for each application because $N$ and $n_\ptime$ can be very large, inducing possible  difficulties due to Random Access Memory (RAM) limitations.\\

\noindent\textit{(iii) Calculation of the reduction order} $n_q$. Let $n_q\mapsto \err_\KL(n_q)$ be the error function related to the relative mean-square error
between $\big\{\bfY(t), t\in\curJ\big\}$ and $\big\{\bfY^{(n_q)}(t), t\in\curJ\big\}$, defined by
\begin{equation}
\err_\KL(n_q)  = 1 - \frac{ \sum_{\alpha=1}^{n_q} \Lambda_\alpha  }
                                       { (1/n_\ptime)\sum_{n=1}^{n_\pptime} \tr [C_\bfY(t_n,t_n)] } \, .                  \label{eq3.7}
\end{equation}
The value of $n_q\ll n_\ptime N$ is calculated such  that
$\Lambda_{n_q}  > 0$ and $\err_\KL(n_q)  \leq \varepsilon_\KL$ where
$\varepsilon_\KL$ is a given small positive real number used for error
control.\\

\noindent\textit{(iv) Algorithm for solving the eigenvalue problem}.
For $\ell\in\big\{1,\ldots , N_d \big\}$ and for $\alpha\in\big\{1,\ldots , n_q\big\}$, we introduce the vectors
$\bfy_\curJ^\ell  = \big (\bfy^\ell_c(t_1),\ldots, \bfy^\ell_c(t_{n_\pptime} ) \big )$ and
$\bfvarphi_\curJ^\alpha  = \big (\bfvarphi^\alpha(t_1),\ldots, \bfvarphi^\alpha(t_{n_\pptime} ) \big )$ that belong to $\RR^{n_\pptime N}$,
in which $\bfy^\ell_c(t_n)$ is defined by Eq.~\eqref{eq2.12} and $\bfvarphi^\alpha(t_n)$ by Eq.~\eqref{eq3.1}. To these vectors, we associate the matrices $[y_\curJ] \in \MM_{n_\pptime N,N_d}$ and $[\varphi_\curJ] \in \MM_{n_\pptime N,n_q}$ such that
$[y_\curJ] = [\bfy_\curJ^1 \ldots \bfy_\curJ^{N_d}]$ and $[\varphi_\curJ] = [\bfvarphi_\curJ^1 \ldots \bfvarphi_\curJ^{n_q}]$.
Using these notations, the eigenvalue problem defined by Eq.~\eqref{eq3.1} can be rewritten as $[C_\curJ] \, \bfvarphi_\curJ^\alpha = \Lambda_\alpha \bfvarphi_\curJ^\alpha$ in which $[C_\curJ]$ is a symmetric real $(n_\ptime N \times n_\ptime N)$ matrix whose rank is
$n_q \leq N_d$, and which is written as $[C_\curJ] = \big (1/(N_d -1)\big ) (n_\pptime^{-1/2} \, [y_\curJ])  \, (n_\pptime^{-1/2} \, [y_\curJ])^T$.
In general, for the considered nonlinear SCM, $N$ is large and the number $n_\ptime$ of sampling times is also large. Consequently, the dimension $(n_\ptime N \times n_\ptime N)$  of matrix $[C_\curJ]$ of this eigenvalue problem is a full matrix that cannot be assembled (the number of entries is, for a medium problem, of order $10^{18}$). In the framework of the present work, $N_d$ is small (small size of the training dataset, for instance, of order $100$). Therefore, $n_\ptime N \gg N_d$, which allows for solving the eigenvalue problem, without assembling $[C_\curJ]$, by using a "thin SVD"  \cite{Golub1993} applied to rectangular matrix $n_\pptime^{-1/2} \, [y_\curJ] $. Keeping only the $n_q$ largest singular values $s_\alpha=[s]_{\alpha\alpha} $ ordered as $s_1 \geq s_2 \geq \ldots \geq s_{n_q}  > 0$, one obtains the following factorization,
$n_\pptime^{-1/2} \, [y_\curJ] = [b_\curJ] \, [s ] \, [v]$ in which $[v]\in \MM_{n_q}$ verifies $[v]\,[v]^T = [I_{n_q}]$ and where $[b_\curJ]\in \MM_{n_\pptime N,n_q}$ is such that $[b_\curJ]^T\, [b_\curJ] = [I_{n_q}]$. Finally, eigenvalues $\Lambda_1, \ldots ,\Lambda_{n_q}$ and matrix $[\varphi_\curJ]\in \MM_{n_\pptime N,n_q}$ are given by $\Lambda_\alpha = s_\alpha^2 /(N_d-1)$ and $[\varphi_\curJ] = n_\pptime^{1/2} \, [b_\curJ]$.
\section{Random residual associated with the nonlinear SCM and time subsampling}
\label{Section4}
In  this section, we consider any mean-square continuous second-order stochastic process $\big\{\widehat\bfY(t),$ $t\in [t_0, T] \big\} $ with values in $\RR^N$, defined on
$(\Theta,\curT,\curP)$, whose trajectories belong to $\curC_\ad^{\,m_d}$, and which is statistically dependent on a random vector $\widehat\bfW$ defined on
$(\Theta,\curT,$ $\curP)$, with values in $\curC_w\subset\RR^{n_w}$.
It is assumed that $n_r$ independent realizations $(\widehat\bfy^\ell,\widehat\bfw^\ell)$ of  $(\widehat\bfY,\widehat\bfW)$ are known, that is to say, for all $\ell\in \big\{1,\ldots , n_r\big\}$ and $\theta_\ell\in\Theta$,
\begin{equation}
(\widehat\bfy^\ell, \widehat\bfw^\ell)\in \curC_\ad^{\, m_d}\times \curC_w
\,\, , \,\, \widehat\bfy^\ell = \widehat\bfY(\cdot;\theta_\ell)
\,\, , \,\, \widehat\bfw^\ell = \widehat\bfW(\theta_\ell) \, .                                                     \label{eq4.1}
\end{equation}
As previously, $\big\{\widehat\bfY(t), t\in \curJ \big\} $ and $\big\{\widehat\bfy^\ell(t), t\in \curJ \big\} $ will denote the time sampling of $\widehat\bfY$ and $\widehat\bfy^\ell$.
In this section, we will formulate the residual calculation for the couple $(\widehat\bfY,\widehat\bfW)$. We will use it for computing the random residual
when $(\widehat\bfY,\widehat\bfW)$ will represent either $(\bfY,\bfW)$ with $n_r=N_d$, $\bfW$ defined in Section~\ref{Section2.2}, and $\bfY$ defined in Section~\ref{Section2.5} (the reference) or will represent $(\bfY^{(n_q)},\bfW)$ with $n_r= n_\pMC$, $\bfY^{(n_q)}$ defined by Eq.~\eqref{eq3.2}, and  for which  the probability measure of random vector $\bfX =(\bfQ,\bfW)$ will be estimated either with the PLoM without constraint or with the PLoM under constraints.
\subsection{Random residual}
\label{Section4.1}
%
\noindent \textit{(i) Definition of the time dependent vector-valued random residual} $\bfcurR(t)$.  For all $t$ fixed in $[t_0,T]$, the time dependent $\RR^N$-valued residual $\bfcurR(t) = (\curR_1(t),\ldots ,\curR_N(t))$, associated with the couple $(\widehat\bfY ,\widehat\bfW)$, is defined by
\begin{equation}
\bfcurR(t) = \bfcurN_t\big (\widehat\bfY(t),t,\widehat\bfW\big ) \quad , \quad t\in[t_0,T] \, .                           \label{eq4.2}
\end{equation}
It is assumed that $\big\{\bfcurR(t), t\in[t_0,T] \big\}$ is a  mean-square continuous second-order $\RR^N$-valued stochastic process defined on $(\Theta,\curT,\curP)$, whose trajectories are almost surely continuous.
 For $\ell\in\big\{1,\ldots , n_r\big\}$, and using Eqs.~\eqref{eq4.1} and \eqref{eq4.2},
the realization $\bfr^\ell(t)$ of $\bfcurR(t)$ is written as
\begin{equation}
\bfr^\ell(t) = \bfcurN_t(\widehat\bfy^\ell(t),t,\widehat\bfw^\ell)  \, .                                      \label{eq4.3}
\end{equation}
The $\RR^N$-valued function $t\mapsto \bfr^\ell(t)$ is continuous on $[t_0,T]$ and is evaluated at the sampling times of $\curJ$.\\

\noindent \noindent \textit{(ii) Definition of the time dependent real-valued random residual} $\curR_\pnorm(t)$.
 For all $t$ fixed in $[t_0,T]$, the time dependent real-valued random residual $\curR_\pnorm(t)$ is the positive-valued random variable is defined by
\begin{equation}
\curR_\pnorm(t) = \frac{1}{\sqrt{N}} \Vert \,\bfcurR(t)\,\Vert \, ,                                                     \label{eq4.4}
\end{equation}
in which $\Vert \cdot \Vert$ is the usual Euclidean norm.\\

\noindent \textit{(iii) Definition of the random residual} $\widehat\rho$.
The \textit{random residual} associated with the couple $(\widehat\bfY,\widehat\bfW)$ is then defined as follows,
\begin{equation}
\widehat\rho = \Vert \,\curR_\pnorm \,\Vert_\mu
\quad , \quad \Vert \,\curR_\pnorm \, \Vert_\mu^2 = \frac{1}{T-t_0}\int_{t_0}^T \curR_\pnorm(t)^2\, d\mu(t) \, ,                 \label{eq4.5}
\end{equation}
in which $d\mu$ is a given bounded positive measure on $\RR$, which will be defined in Section~\ref{Section4.2} when introducing the time subsampling for computing the residual.
From Eqs.~\eqref{eq4.4} and \eqref{eq4.5}, it can be seen that
\begin{equation}
\widehat\rho^{\,2} = \frac{1}{N(T-t_0)}\int_{t_0}^T \Vert \, \bfcurR(t)\, \Vert^2\, d\mu(t) \, .                              \label{eq4.6}
\end{equation}
It should be noted that, if $d\mu(t)$ was chosen as $dt$ on bounded interval $[t_0,T]$, then for $\ell\in\big\{1,\ldots , n_r\big\}$, the realization $\widehat\rho^{\,\ell}$ of random variable $\widehat\rho$ could only be computed using the time sampling $\curJ$, that is to say, by the formula $(\widehat\rho^{\,\ell})^2 =  (N\, n_\pptime)^{-1} \sum_{n=1}^{n_\pptime} \Vert\,\bfr^\ell(t_n)\,\Vert^2$. \\

\noindent \textit{(iv) Definition of the $L^2$-norm of random residual} $\widehat\rho$.
Since $\bfcurR$ is mean-square continuous on compact set $[t_0,T]$, it can be deduced \cite{Doob1953,Kree1986} that $\widehat\rho$ is a real-valued (in fact a positive-valued) second-order random variable. Consequently, $\widehat\rho$ belongs to $L^2(\Theta,\RR)$ and its norm is
$\Vert\,\widehat\rho\,\Vert_{L^2} = \sqrt{E\big\{\widehat\rho^2\big\} }$.
If $(\widehat\bfY,\widehat\bfW)$ is the couple $(\bfY^\sol,\bfW)$ defined in Section~\ref{Section2.3}, then  $\Vert\,\widehat\rho\,\Vert_{L^2} = 0$.
Using the $n_r$ independent realizations $\widehat\rho^{\,\ell}$ of $\widehat\rho$, the empirical estimate of $\Vert\,\widehat\rho\,\Vert_{L^2}$ is computed as
$\big( (1/n_r) \sum_{\ell=1}^{n_r} (\widehat\rho^{\,\ell})^2  \big )^{1/2}$. \\

\noindent \textit{(v) Definition of the normalized random residual} $\rho$.
Let $\widehat\rho_\pref$ be the random residual $\widehat\rho$ calculated for a fixed value $N_{d,\pref}$ of the number $N_d$ of points in the training dataset (thus $n_r = N_{d,\pref}$ and $(\widehat\bfY,\widehat\bfW)$ is the couple $(\bfY,\bfW)$  defined in Section~\ref{Section2.5} with $N_d = N_{d,\pref}$).
Let $\underline{\widehat\rho}_\pref$ be the empirical mean value of $\widehat\rho_\pref$ estimated with the $N_{d,\pref}$ realizations of $\widehat\rho_\pref$, computed with the KL expansion $\big\{\bfY^{(n_q)}(t),t\in [t_0,T]\big\}$ of $\big\{\bfY(t),t\in [t_0,T]\big\}$, that is to say, $(\widehat\bfY,\widehat\bfW)$ is the couple $(\bfY^{(n_q)},\bfW)$. The positive-valued normalized random residual $\rho$ is then defined by
\begin{equation}
\rho = \frac{\widehat\rho}  {\underline{\widehat\rho}_\pref} \, ,                                              \label{eq4.8}
\end{equation}
in which $\widehat\rho$ is the random residual defined by Eq.~\eqref{eq4.5}.
Consequently, if $n_r = N_{d,\pref}$ and $(\widehat\bfY,\widehat\bfW)$ is the couple $(\bfY,\bfW)$  defined in Section~\ref{Section2.5} with $N_d = N_{d,\pref}$, then the empirical mean value of $\rho$ is equal to $1$ (because, in this case $\widehat\rho$ is $\widehat\rho_\pref$).
\subsection{Time subsampling for computing the realizations of the random residual}
\label{Section4.2}
The role played by measure $d\mu$ introduced in Eq.~\eqref{eq4.5} is the following. The computation of the $n_r$ realizations $\big\{\bfr^\ell(t), t\in\curJ \big\}_\ell$ of $\big\{\bfcurR(t),t\in\curJ\big\}$ using Eq.~\eqref{eq4.3} requires  $n_r \times n_\ptime$ numerical evaluations of the nonlinear operator $\bfcurN_t$ and requires a RAM of $n_r\times n_\ptime \times N$ $64$-bit words.  The values of $N$, $n_\ptime$, and $n_r$ can be large. In general the CPU time is not a limitation because a parallel computation can be done for the loop on the number $n_r$ of realizations, but difficulties can occur with possible RAM limitations.
On the other hand, such a large value of $n_\ptime$ is not necessary for constructing the constraint on $\rho$ and a reasonable approximation of the integral on $[t_0, T]$ in the Eq.~\eqref{eq4.6} can be used to compute the $n_r$ realizations $\big\{\widehat\rho^{\,\ell}\big\}_\ell$ of random residual $\widehat\rho$. For such a computation, we introduce the following  subsampling times,
\begin{equation}
\big\{\tau_{n_s}, n_s = 1,\ldots , n_\psp\big\} \subset \big\{t_n, n = 1,\ldots , n_\ptime\big\} \quad , \quad  n_\psp \ll n_\ptime\, ,   \label{eq4.8b}
\end{equation}
in which $\tau_{1},\ldots, \tau_{n_\psp}$ are defined as the subsampling times. The measure $d\mu$ is then defined by
$d\mu(t) = \big ((T-t_0)/n_\psp\big ) \, \sum_{n_s=1}^{n_\psp} \delta_0(t-\tau_{n_s})$,
in which $\delta_0$ is the Dirac measure on $\RR$ at $0$. With this approximation, there is a big gain with respect to the RAM limitations and also a gain for the CPU time. Using this measure, Eq.~\eqref{eq4.6} yields
\begin{equation}
\widehat\rho^{\,2} = \frac{1}{N n_\psp} \sum_{n_s=1}^{n_\psp} \Vert \,\bfcurR(\tau_{n_s}) \,\Vert^2    \, .                               \label{eq4.10}
\end{equation}
The $n_r$ independent realizations $\big\{\widehat\rho^{\,\ell}\big\}_\ell$ of random residual $\widehat\rho$ are such that
\begin{equation}
(\widehat\rho^{\,\ell})^2 =  \frac{1}{N n_\psp} \sum_{n_s=1}^{n_\psp} \Vert\, \bfr^\ell(\tau_{n_s})\, \Vert^2
                                                                       \quad , \quad \ell\in\big\{1,\ldots,n_r\big\}\, ,                                  \label{eq4.11}
\end{equation}
in which $\bfr^\ell(\tau_{n_s})\in\RR^N$ is computed with Eq.~\eqref{eq4.3} for $t=\tau_{n_s}$.
The time subsampling has to be defined for each application.

\section{Definition of random vectors $\bfX$ and $\bfH$}
\label{Section5}
The formulation of PLoM under constraints for stochastic processes requires the definition of the initial dataset $\curD_{N_d}^{\, d}$ of the random vector $\bfX = (\bfQ,\bfW)$ and its PCA constructed using $\curD_{N_d}^{\, d}$.\\

\noindent \textit{(i) Definition of the initial dataset} $\curD_{N_d}^{\, d}$.
For the probabilistic learning, we introduce the initial dataset $\curD_{N_d}^{\, d}$ that is associated with the training dataset
$\curY^{\, d}_{N_d} = \big\{(\bfy^1,\bfw^1),\ldots , (\bfy^{N_d},\bfw^{N_d})\big\}$ defined by Eq.~\eqref{eq2.10} and with the reduced-order representation defined in Section~\ref{Section3.2}-(ii). Each realization $\bfy^\ell$ of $\bfY$ is associated with the realization $\bfw^\ell$ of $\bfW$ (see Section~\ref{Section2.3}). The reduced-order representation $\bfY^{(n_q)}$, defined by Eq.~\eqref{eq3.2}, allows the $N_d$ independent realizations $\big\{\bfq^\ell ,\ell=1,\ldots, N_d\big\}$ of the $\RR^{n_q}$-valued random variable $\bfQ$  to be computed by using Eq.~\eqref{eq3.5}. Let $\bfX = (\bfQ,\bfW)$ be the random variable defined on $(\Theta,\curT,\curP)$ with values in $\RR^{n_x} =  \RR^{n_q}\times \RR^{n_w}$.
Each realization $\bfq^\ell$ depends on realization $\bfw^\ell$, and $\bfx^\ell = (\bfq^\ell,\bfw^\ell)$ is a realization of $\bfX$.
The initial dataset $\curD_{N_d}^{\, d}$ is made up of the $N_d$ independent realizations $\bfx^\ell$,
\begin{equation}
\curD_{N_d}^{\, d} = \big\{\bfx^\ell = (\bfq^\ell,\bfw^\ell) \in \RR^{n_x} = \RR^{n_q}\times \RR^{n_w}
                                      \,\, , \, \, \ell = 1,\ldots , N_d \big\} \, ,                                                             \label{eq5.1}
\end{equation}

%
\noindent \textit{(ii) PCA of random vector} $\bfX$.
The first step consists of performing a PCA of $\bfX$. This step is mandatory to formulate the constraints and to extend  the PLoM under  constraints. It should be noted that the reduced-order representation of  stochastic process $\bfY$ has already introduced the random vector $\bfQ$, which is centered and whose covariance matrix is the identity matrix.
However, random vector $\bfX$ is not centered and its covariance matrix is not the identity matrix.
The purpose of the PCA is to improve the statistical computation of $\bfX$ through de-correlation and normalization.
Let $\underline{\bfx}\in \RR^{n_x}$ and $[C_\bfX]\in \MM_{n_x}^{+0}$ be the empirical estimates of the mean vector and the covariance matrix of $\bfX$ such that
$\underline{\bfx}= (1/N_d) \sum_{\ell=1}^{N_d} \bfx^\ell$ and $[C_\bfX] = \big (1/(N_d-1)\big ) \sum_{\ell=1}^{N_d} (\bfx^\ell -\underline{\bfx})\, (\bfx^\ell -\underline{\bfx})^T$.
Let $\xi_1\geq \xi_2\geq \ldots \geq \xi_{n_x} \geq 0$ be the eigenvalues and let $\bfpsi^1,\ldots,\bfpsi^{n_x}$ be the orthonormal eigenvectors of the eigenvalue problem $[C_\bfX]\, \bfpsi^\alpha = \xi_\alpha \,\bfpsi^\alpha$. For $\varepsilon_\PCA > 0$ fixed, let $\nu$ be the integer such that $n_q + 1\leq \nu \leq n_x$
and $xi_\nu > 0$, verifying,
\begin{equation}
\err_\PCA(\nu) = 1- \frac{\sum_{\alpha=1}^{\nu} \xi_{\alpha}}{\tr[C_\bfX]} \leq \varepsilon_\PCA\, .                                        \label{eq5.3}
\end{equation}
The PCA of $\bfX$ yields its representation $\bfX^{(\nu)}$ that is written as
\begin{equation}
\bfX^{(\nu)} = \underline{\bfx} +[\psi]\,[\xi]^{1/2}\, \bfH \quad , \quad
 E\big\{\Vert \,\bfX-\bfX^{(\nu)}\,\Vert^2 \big\} \leq \varepsilon_\PCA \, E\big\{ \Vert\,\bfX\,\Vert^2\big\}\, ,                                            \label{eq5.4}
\end{equation}
in which $[\psi] = [\bfpsi^1 \ldots \bfpsi^\nu] \in \MM_{n_x,\nu}$ such that $[\psi]^T\, [\psi] = [I_\nu]$ and where $[\xi]$ is the
positive-definite diagonal $(\nu\times\nu)$ matrix such that $[\xi]_{\alpha\beta} = \xi_\alpha \delta_{\alpha\beta}$. The $\RR^\nu$-valued random variable $\bfH$ is obtained by projection,
\begin{equation}
\bfH = [\xi]^{-1/2}\, [\psi]^T\,(\bfX -\underline{\bfx}) \, ,                                                                          \label{eq5.5}
\end{equation}
and its $N_d$ independent realizations  are such that
\begin{equation}
\bfeta_d^\ell = [\xi]^{-1/2}\, [\psi]^T\,(\bfx^\ell -\underline{\bfx})\in\RR^\nu \quad ,\quad  \ell\in\big\{1,\ldots , N_d\big\}\, .      \label{eq5.6}
\end{equation}
Using $\big\{\bfeta_d^1, \ldots, \bfeta_d^{N_d} \big\}$, the estimates  of the mean vector $\underline{\bfeta}$ and the
covariance matrix $[C_\bfH]$ of random vector $\bfH$ are such that $\underline{\bfeta}=\bfzero_\nu$ and $[C_\bfH] =[I_\nu]$.\\

%
\noindent \textit{(iii) Probability density function of} $\bfH$ \textit{estimated with the initial dataset}.
Let $P_\bfH(d\bfeta) = p_\bfH(\bfeta)\, d\bfeta$ be the probability distribution of  the $\RR^\nu$-valued random variable $\bfH$ defined by Eq.~\eqref{eq5.5}. The pdf $\bfeta\mapsto p_\bfH(\bfeta)$ on $\RR^\nu$ with respect to $d\bfeta$ is estimated by using the Gaussian kernel-density estimation method with the $N_d$ independent realizations $\big\{\bfeta^\ell_d , \ell=1,\ldots , N_d\big\}$ defined by Eq.~\eqref{eq5.6}. Using the modification proposed in \cite{Soize2015} of the classical formulation \cite{Bowman1997},  $p_\bfH(\bfeta)$ is estimated, for all $\bfeta$ in $\RR^\nu$,  as
\begin{equation}
p_\bfH(\bfeta) \simeq c_\nu\, \zeta(\bfeta) \quad , \quad c_\nu = \frac{1}{(\sqrt{2\pi} \, \widehat s_\nu)^\nu}  \, ,                            \label{eq5.7}
\end{equation}
\begin{equation}
\zeta(\bfeta) = \frac{1}{N_d} \sum_{\ell=1}^{N_d} exp\big \{-\frac{1}{2 \widehat s_\nu^2}
\,\Vert\,\frac{\widehat s_\nu}{s_\nu}\bfeta_d^\ell -\bfeta\,\Vert^2\big \}\, ,                                                                              \label{eq5.8}
\end{equation}
in which $s_\nu = \big \{ 4 /\big (N_d(2+\nu)\big ) \big \}^{1/(\nu+4)}$ is the usual multidimensional optimal Silverman bandwidth (taking into account that the covariance matrix of  $\bfH$ is $[I_\nu]$), and where
$\widehat s_\nu = s_\nu  \big (s_\nu^2 + (N_d-1)/N_d\big )^{-1/2}$
has been introduced to ensure that
$E\big\{\bfH\big\} = (\widehat s_\nu / s_\nu )\, \underline{\bfeta} =\bfzero_\nu$
and
$E\big\{\bfH\bfH^T\big\} = \widehat s_\nu^2\, [I_\nu]  + \widehat s_\nu^2 s_\nu^{-2} (N_d-1) N_d^{-1}\, [C_\bfH] = [I_\nu]$.
The pdf of $\bfH$, defined by Eqs.~\eqref{eq5.7} and \eqref{eq5.8}, is rewritten as
\begin{equation}
p_\bfH(\bfeta) = c_\nu\, e^{-\phi(\bfeta)} \quad , \quad \phi(\bfeta) = - \log \zeta(\bfeta)\, .                                           \label{eq5.9}
\end{equation}

\section{Probabilistic learning on manifolds (PLoM) without constraint}
\label{Section6}

\noindent \textit{(i) Generation of the learned datasets} $\curD_{n_\ppMC}^{\,\ar}$ \textit{and} $\curY^{\, ar}_{n_\ppMC}$.
Using only the realizations $\big\{\bfeta_d^1,\ldots , \bfeta_d^{N_d}\big\}$ given by Eq.~\eqref{eq5.6}, the PLoM method \cite{Soize2016,Soize2020c} allows for generating
$n_\pMC \gg N_d$ additional  realizations $\big\{\bfeta_\ar^1,\ldots , \bfeta_\ar^{n_\ppMC}\big\}$ of $\bfH$. Using Eq.~\eqref{eq5.4} allows for computing $n_\pMC$ additional  realizations $\big\{\bfx_\ar^1,\ldots , \bfx_\ar^{n_\ppMC}\big\}$ of $\bfX^{(\nu)}$ such that
$\bfx_\ar^\ell = \underline{\bfx} +[\psi]\,[\xi]^{1/2}\, \bfeta_\ar^\ell$ for all $\ell$ in $\big\{1,\ldots, n_\pMC\big\}$. For random vector $\bfX^{(\nu)}$, the learned dataset is defined by $\curD_{n_\ppMC}^{\,\ar}= \big\{\bfx^\ell_\ar \in \RR^{n_x} \,\, , \, \, \ell = 1,\ldots , n_\pMC \big\}$. From the knowledge of the $n_\pMC$ realizations $\big\{ \bfx^\ell_\ar \big\}_\ell$, we can deduce the $n_\pMC$ additional realizations
$\big\{ (\bfq^\ell_\ar,\bfw^\ell_\ar) \big\}_\ell$ in $\RR^{n_q}\times \RR^{n_w}$ of random variable $\big (\bfQ^{(\nu)},\bfW^{(\nu)}\big ) = \bfX^{(\nu)}$ such that
\begin{equation}
(\bfq^\ell_\ar,\bfw^\ell_\ar) = \bfx^\ell_\ar \quad ,\quad \ell\in\big\{1,\ldots, n_\pMC\big\} \, .                               \label{eq6.3}
\end{equation}
For each additional realization $\bfq^\ell_\ar$ given by Eq.~\eqref{eq6.3}, and using Eq.~\eqref{eq3.2}, the corresponding additional realization $\bfy^\ell_\ar = \big\{\bfy^\ell_\ar(t), t\in \curJ\big\}$  of stochastic process $\big\{\bfY^{(n_q)}(t), t\in \curJ\big\}$ is computed by $\bfy^\ell_\ar(t) = \underline\bfy(t)  + [V(t)]\, \bfq^\ell_\ar$.
The learned dataset for the couple $(\bfY^{(n_q)},\bfW)$ is then defined by
$\curY^{\ar}_{n_\ppMC}  = \big\{ ( \bfy^1_\ar,\bfw^1_\ar), \ldots , (\bfy^{n_\ppMC}_\ar, \bfw^{n_\ppMC}_\ar)  \big\}$,
in which  $\bfy^\ell_\ar = \big\{\bfy^\ell_\ar(t)\, t \in \curJ\big\}$ and where $\bfw_\ar^\ell \in\curC_w\subset\RR^{n_w}$.
As explained in Section~\ref{Section2.4},  the time derivatives up to order $m_d$ of $\bfy^\ell_\ar$ are computed at all the sampling times belonging to $\curJ$ by using the time integration scheme.
It should be noted that the PLoM method introduces a $\MM_{\nu,N_d}$-valued random variable $[\bfH]$ related to random vector $\bfH$ whose probability distribution is defined by Eq.~\eqref{eq5.9}. The method allows for generating $n_\pMC$ additional realizations $\big\{ [\eta_\ar^1], \ldots , [\eta_\ar^{n_\ppMC}]\big\}$ in $\MM_{\nu,N_d}$ of random matrix $[\bfH]$. For each $\ell$, the  additional realization $\bfeta_\ar^\ell$ is obtained  by a random extraction of one column of matrix $[\eta_\ar^\ell]$, such that, for $k\in\big\{1,\ldots,\nu\big\}$ and for $\ell \in \big\{1,\ldots ,n_\pMC\big\}$, we have
$\big\{\bfeta_\ar^\ell\big\}_k = [\eta_\ar^\ell]_{k\, j_0^{\,\ell}}$ in which $j_0^{\, 1},\ldots ,j_0^{\,n_\ppMC}$ are  $n_\pMC$  independent realizations of a uniform discrete random variable $J_0$ with values in $\big\{1,\ldots , N_d\big\}$, which are constructed at the beginning of the algorithms and are never modified. \\

%
\noindent \textit{(ii) Computing the random residual of the stochastic equations for the learned realizations}.
The $n_\pMC$ realizations $\big\{\widehat\rho_\ar^{\,\ell}\big\}_\ell$ of random residual $\widehat\rho$, corresponding
to the learned dataset $\curY^{\, ar}_{n_\ppMC}$, are computed (see Eq.~\eqref{eq4.11}), for $\ell\in\big\{1,\ldots,n_\pMC\big\}$, by
$(\widehat\rho_\ar^{\,\ell})^2 =  \big (1/(N\, n_\psp) \big ) \sum_{n_s=1}^{n_\psp} \Vert\,\bfr_\ar^\ell(\tau_{n_s})\,\Vert^2$,
in which $\bfr_\ar^\ell(\tau_{n_s})\in\RR^N$ is computed (see Eq.~\eqref{eq4.3}), for $t=\tau_{n_s}$, by
$\bfr_\ar^\ell(\tau_{n_s}) = \bfcurN_t\big (\bfy_\ar^\ell(\tau_{n_s}),\tau_{n_s},\bfw_\ar^\ell\big )$.
The corresponding additional realizations $\big\{ \rho_\ar^1, \ldots
,\rho_\ar^{n_\ppMC} \big\}$ of the normalized random residual $\rho$
can then be computed using Eq.~\eqref{eq4.8} and written as,
\begin{equation}
\rho^\ell_\ar =  {\widehat\rho_\ar^{\,\ell}} \, / \,{\underline{\widehat\rho}_\pref} \quad , \quad \ell =1,\ldots , n_\pMC \, .                         \label{eq6.7bis}
\end{equation}
\section{Definition of the constraints for the PLoM under constraints}
\label{Section7}
In a first formulation of the proposed theory, only the constraint on the normalized random residual $\rho$ of the stochastic equations was introduced in the PLoM under constraints (it will be $\algo~1$ defined in Section~\ref{Section8.3}-(i)).
Some  applications  revealed a small difficulty when the constraint on $\rho$ alone was taken into account.  As the learning is done on the random vector $\bfX = (\bfQ,\bfW)$, the calculation of the probability measure of $\bfX$, which is based on the minimization of the Kullback-Leibler cross-entropy does not guarantee the preservation of the prior probability distribution  of $\bfW$. Indeed, we obtained good results, but the prior probability distribution of $\bfW$ is altered during the minimization process. Since the objective is to find the optimal probability measure of the couple $(\bfY^{(n_q)},\bfW)$, that is to say, the optimal probability measure of $\bfX$, which preserves the prior probability distribution of $\bfW$ while minimizing the $L^2$-norm of $\rho$, we have introduced constraints on $\bfW$ in order to best preserve its prior probability distribution (it will be $\algo~2$ or $\algo~3$ defined in Section~\ref{Section8.3}-(ii-iii)). In this section, we define the constraints on $\rho$ and $\bfW$.
The PLoM under constraints is  formulated using the Kullback-Leibler principle in order to satisfy the constraints on $\rho$ and $\bfW$.
However, as this principle is formulated for the random vector $\bfH $, the constraints must be expressed in terms of a mathematical expectation of a function of $\bfH$. For this, we must therefore use the representations constructed previously to express these constraints as a function of $\bfH$.
We introduce the block decomposition of $\underline\bfx\in\RR^{n_x}$ and $[\psi]\in \MM_{n_x,\nu}$ such that
\begin{equation}
     \underline\bfx = \left [ \begin{array}{c}
                               \underline\bfx_q \\
                               \underline\bfx_w \\
                              \end{array} \,
                       \right ]
                                                \quad , \quad
             [\psi] = \left [
                               \begin{array}{c}
                               \lbrack \psi_q \rbrack \\
                               \lbrack \psi_w \rbrack \\
                               \end{array}\,
                       \right ]                                                                            \, , \label{eq7.1}
\end{equation}
in which $\underline\bfx_q \in \RR^{n_q}$, $\underline\bfx_w \in \RR^{n_w}$, $[\psi_q] \in \MM_{n_q,\nu}$,
and $[\psi_w] \in \MM_{n_w,\nu}$. Therefore, $\bfX^{(\nu)} = (\bfQ^{(\nu)} , \bfW^{(\nu)})$ defined by
Eq.~\eqref{eq5.4} can be rewritten as
\begin{equation}
\bfQ^{(\nu)} =  \underline\bfx_q +[\psi_q]\,[\xi]^{1/2}\, \bfH \quad , \quad
\bfW^{(\nu)} =  \underline\bfx_w +[\psi_w]\,[\xi]^{1/2}\, \bfH \, .                                           \label{eq7.2}
\end{equation}
Equation~\eqref{eq3.2} is $\bfY^{(n_q)}(t_n) = \underline\bfy(t_n)  + [V(t_n)]\, \bfQ^{(\nu)}$. Substituting in it the representation of $\bfQ^{(\nu)}$ given by Eq.~\eqref{eq7.2} yields
\begin{equation}
\bfY^{(n_q)}(t_n) = \underline\bfy_H(t_n)  + [V_H(t_n)]\, \bfH\, ,                                           \label{eq7.4}
\end{equation}
in which $\underline\bfy_H(t_n) = \underline\bfy(t_n)  + [V(t_n)]\, \underline\bfx_q$
and $[V_H(t_n)] = [V(t_n)]\, [\psi_q]\,[\xi]^{1/2}$.\\

\noindent\textit{(i) Constraint associated with the normalized random residual of the stochastic equation}.
Equation~\eqref{eq4.2} is written as $\bfcurR(t_n) = \bfcurN_t(\bfY^{(n_q)}(t_n),t_n,\bfW^{(\nu)})$. Substituting in its right-hand side the representations $\bfY^{(n_q)}(t_n)$ given by Eq.~\eqref{eq7.4} and $\bfW^{(\nu)}$ given by Eq.~\eqref{eq7.2} show that $\bfcurR(t_n)$ depends only on $t_n$ and $\bfH$. Consequently, using Eqs.~\eqref{eq4.8} with \eqref{eq4.6}, $\rho^2$  is a nonlinear mapping $\bfeta\mapsto h^c_\rho(\bfeta)$ of $\bfH$ such that $\rho^2 = h^c_\rho(\bfH)$. The constraint consists of imposing a reference value to $\Vert\,\rho\,\Vert_{L^2}^2 = E\big\{\rho^2\big\}$ that is chosen as
$b^c_\rho = E\big\{\rho_{\pref}^2\big\}$ with $\rho_{\pref}= \widehat\rho_\pref / \underline{\widehat\rho}_\pref$,  for which the mathematical expectation  is computed using $N_d = N_{d,\pref}$ realizations of the  reduced-order representation $\bfY^{(n_q)}$  and where $N_{d,\pref}$ is the given reference value of $N_d$ introduced in Section~\ref{Section4.1}-(v). The constraint on $\rho$ is then written as
\begin{equation}
E\big\{h^c_\rho(\bfH)\big\}  = b^c_\rho\, .                                                                                  \label{eq7.5}
\end{equation}
Note that the nonlinear application $h^c$ cannot be constructed explicitly. In addition we will need to calculate the gradient of $h^c(\bfeta)$ with respect to $\bfeta$. This important difficulty will be resolved by statistical conditioning (see Section~\ref{Section9}).\\

\noindent\textit{(ii) Constraints associated with the prior probability model of} $\bfW^{(\nu)}$.
In order to avoid a significant modification of the prior probability distribution of $\bfW^{(\nu)}$,  we impose the mean value and the second-order moment  of each component $W_j^{(\nu)}$ of $\bfW^{(\nu)}$.  For all $j\in\big\{1,\ldots,n_w\big\}$, we thus write $E\big\{W^{(\nu)}_j\big\} = b^c_{w,j}$ and $E\big\{(W^{(\nu)}_j)^2\big\} = b^c_{w,j+n_w}$ in which $b^c_{w,j}$ and $b^c_{w,j+n_w}$ are the mean value and the second-order moment of $W_j$ (and not of $W^{(\nu)}_j$), which are computed with the training dataset for $N_d=N_{d,\pref}$.
The use of the representation $\bfW^{(\nu)} =  \underline\bfx_w +[\psi_w]\,[\xi]^{1/2}\, \bfH$  allows for defining a nonlinear mapping $\bfeta\mapsto \bfh^c_w(\bfeta)$ from $\RR^\nu$ into $\RR^{2n_w}$ such that
$E\big\{h^c_{w,j}(\bfH) \big\} = b^c_{w,j}$ and $E\big\{h^c_{w,j+n_w}(\bfH) \big\} = b^c_{w,j+n_w}$. Introducing the vector
$\bfb^c_w = (b^c_{w,1},\ldots ,b^c_{w,n_w},b^c_{w,1+n_w},$ $\ldots ,b^c_{w,2 n_w}) \in \RR^{2 n_w}$, the  constraint on $\bfW^{(\nu)}$ is written as
\begin{equation}
E\big\{\bfh^c_w(\bfH)\big\}  = \bfb^c_w\, .                                                                                  \label{eq7.6}
\end{equation}

\noindent\textit{(iii) Global constraint imposed to random vector} $\bfH$.
Defining the vector $\bfb^c = (b^c_\rho, \bfb^c_w) \in \RR^{1+2n_w}$ and the mapping $\bfh^c = (h^c_\rho, \bfh^c_w)$ with values in $\RR^{1+2n_w}$,
Eqs.~\eqref{eq7.5} and \eqref{eq7.6} can be rewritten as
\begin{equation}
E\big\{\bfh^c(\bfH)\big\}  = \bfb^c\, .                                                                                  \label{eq7.7}
\end{equation}

\section{Probabilistic learning on manifolds under constraints for stochastic process $\bfY^c$}
\label{Section8}
For the PLoM under constraints, stochastic process $\bfY^{(n_q)}$ is denoted by $\bfY^c$.

\subsection{Methodology}
\label{Section8.1}
For the different random variables of the problem, when they will be relative to the PLoM under constraints, we will use an exponent $c$. Thus $\bfH$ relates to the training dataset and to the PLoM without constraint, while $\bfH$ will be noted $\bfH^c$ for the PLoM under constraints.
In these conditions, the random variable $\bfX^{(\nu)} = (\bfQ^{(\nu)} , \bfW^{(\nu)})$ defined by Eq.~\eqref{eq7.2} is denoted by
$\bfX^c = (\bfQ^c , \bfW^c)$  when the constraints are applied.
We have to find among all the probability density functions  that satisfy the constraint $E\big\{\bfh^c(\bfH^c)\big\}  = \bfb^c$, the pdf $p_{\bfH^c}$ that is closest to the pdf $p_\bfH$. The use of the  Kullback-Leibler minimum cross-entropy principle \cite{Bhattacharyya1943,Kullback1951,Kapur1992} yields
\begin{equation}
p_{\bfH^c} = \arg \min_{p \, \in \, \curC_{\ad,\bfH^c}} \int_{\RR^\nu} p(\bfeta)\,
                                       \log\frac{p(\bfeta)}{p_\bfH(\bfeta)}\, d\bfeta\, ,                                        \label{eq8.1}
\end{equation}
in which the admissible set is defined by
$\curC_{\ad,\bfH^c} = \big\{p: \RR^\nu\!\rightarrow \RR^+ \! ,
    \int_{\RR^\nu} p(\bfeta)\,d\bfeta = 1  ,  \int_{\RR^\nu}\! \bfh^c(\bfeta)\, p(\bfeta)\,d\bfeta = \bfb^c \big\}$.
\subsection{Expressing $p_{\bfH^c}$ as the solution of an optimization problem}
\label{Section8.2}
The optimization problem defined by Eq.~\eqref{eq8.1} can only be solved by using an adapted numerical method.
A major difficulty is due to the normalization constant of $p_{\bfH^c}$ that induces considerable difficulties
when $\nu$ is large, which is bypassed by using the following method.
Let $\lambda_0 > 0$ be the real Lagrange multiplier that allows for imposing the normalization constraint
$\int_{\RR^\nu} p(\bfeta) \, d\bfeta =1$. Let $\bflambda \in \curC_{\ad,\bflambda}\subset \RR^{1+2n_w}$ be the one
introduced for imposing the constraint $\int_{\RR^\nu}\! \bfh^c(\bfeta)\, p(\bfeta)\,d\bfeta = \bfb^c$ with
$\curC_{\ad,\bflambda} = \big\{\bflambda\in\RR^{1+2n_w} \, , \,  \int_{\RR^\nu} e^{-\phi(\bfeta)\, -
                                      <\bflambda\, ,\bfh^c(\bfeta)>}\, d\bfeta $ $< +\infty \big\}$,
in which $\phi$ is defined by Eq.~\eqref{eq5.9}.
For $\bflambda$ given in $\curC_{\ad,\bflambda}$, let $\bfH_\bflambda$ be the $\RR^\nu$-valued random variable whose pdf with respect to $d\bfeta$ on $\RR^\nu$ is written as
\begin{equation}
p_{\bfH_\bflambda}(\bfeta) = c_0(\bflambda) \, \exp\big\{-\phi(\bfeta) \, - <\! \bflambda\, ,\bfh^c(\bfeta)\!>\big\} \, ,                 \label{eq8.3}
\end{equation}
in which $c_0(\bflambda)$ is the normalization constant that depends on $\bflambda$.
Let $\bflambda\mapsto \Gamma(\bflambda)$ be the real function on $\curC_{\ad,\bflambda}$ such that
$\Gamma(\bflambda) = <\! \bflambda \, ,\bfb^c \!> - \log c_0(\bflambda)$ (see \cite{Agmon1979}).
The gradient of $\Gamma$ at point $\bflambda$ is
\begin{equation}
\bfnabla \Gamma(\bflambda) = \bfb^c - E\big\{\bfh^c(\bfH_\bflambda)\big\} \, ,                                                   \label{eq8.4}
\end{equation}
while its Hessian matrix $[\Gamma''(\bflambda)] = [\cov \big\{\bfh^c(\bfH_\bflambda) \big\} ]$
is the covariance matrix of random vector $\bfh^c(\bfH_\bflambda)$.
It will be assume that $[\Gamma''(\bflambda)]$ belongs to $\MM^+_{1+2n_w}$ for all $\bflambda\in\curC_{\ad,\bflambda}$.
This hypothesis is related to the algebraic independence of the constraints and will be checked in the algorithm for each application.
Consequently, $\bflambda\mapsto \Gamma(\bflambda)$ is a strictly convex function in $\curC_{\ad,\bflambda}$. Nevertheless, it cannot
be proven that admissible set $\curC_{\ad,\bflambda}$ is convex for the general case because the mapping $h^c_\rho$ depends on the application that is performed.
So, we will assume that the following optimization problem
\begin{equation}
\bflambda^\sol = \arg\, \min_{\bflambda\in\curC_{\ad,\bflambda}}  \, \Gamma(\bflambda)\, ,                           \label{eq8.6}
\end{equation}
has a unique solution $\bflambda^\sol$ that is such that $\bfnabla \Gamma(\bflambda^\sol) =\bfzero_{1+2n_w}$ and that
\begin{equation}
p_{\bfH^c}(\bfeta) =\big\{ p_{\bfH_\bflambda}(\bfeta)\big\}_{\bflambda = \bflambda^\psol} \quad , \quad
 p_{\bfH}(\bfeta) =\big\{ p_{\bfH_\bflambda}(\bfeta)\big\}_{\bflambda = \ppbfzero_{1+2n_w}}         \, .                      \label{eq8.7}
\end{equation}
It should be noted that $\Gamma(\bflambda)$ cannot be evaluated for any given $\bflambda$ because function $\Gamma$ depends on normalization constant $c_0(\bflambda)$ that involves an integral on $\RR^\nu$, which cannot be calculated when $\nu$ is large.
\subsection{Iterative algorithm for computing the Lagrange multipliers $\bflambda^\sol$}
\label{Section8.3}
Taking into account Eqs.~\eqref{eq7.7} and \eqref{eq8.4}, the solution $\bflambda^\sol$ of the optimization problem defined by Eq.~\eqref{eq8.6} is computed by solving the equation $\bfnabla \Gamma(\bflambda) =\bfzero_{1+2n_w}$ in $\bflambda$ using a Newton iterative method, which allows for constructing a solution $\bfH_{\lambda^\psol}$ that satisfies the constraint defined by Eq.~\eqref{eq7.7}.
The Lagrange multiplier $\bflambda$ is rewritten as $\bflambda = (\lambda_\rho, \bflambda_w)$ in which the real-valued Lagrange multiplier $\lambda_\rho$ is associated with the constraint $E\big\{h^c_\rho(\bfH)\big\}  = b^c_\rho$ (see Eq.~\eqref{eq7.5}) and where the vector-valued multiplier $\bflambda_w$ is associated with the constraint $E\big\{\bfh^c_w(\bfH)\big\}  = \bfb^c_w$  (see Eq.~\eqref{eq7.6}).
On this basis, three iteration algorithms, referred as $\algo~1$ to $\algo~3$,  have been thoroughly tested through the applications presented in Sections~\ref{SectionDO}, \ref{SectionNS}, and \ref{SectionSD}.

- The first algorithm, $\algo~1$, does not take into account the constraints associated with the prior probability distribution of $\bfW^{(\nu)}$. As we have explained in Section~\ref{Section7}, this algorithm can be not suitable.

- The two other algorithms, $\algo~2$ and $\algo~3$, take into account not only the constraint on $\rho$ but also the constraints on $\bfW^{(\nu)}$. Algorithm $\algo~2$ is the classical approach for which the constraints vector is globally taken. We have been able to verify that this $\algo~2$ is not always efficient, involving convergence difficulties. This phenomenon is linked to the fact that the two constraints (one on $\rho$ and the other on $\bfW^{(\nu)}$) play in opposite directions (which does not prevent obtaining the minimum, but with potential numerical difficulties). It is for this reason that we have introduced algorithm $\algo~3$ for which the constraint on $\rho$ and the constraints on $\bfW^{(\nu)}$ are decoupled for the computation of the Lagrange multipliers. The numerous numerical tests, which have been carried out, have shown that $\algo~3$ is often most efficient algorithm.\\

\noindent\textit{(i) Algorithm $\algo~1$}. The only constraint is on the normalized random residual $\rho$. The iteration algorithm is written, for $i \geq 1$,  as
$\lambda_\rho^{i+1} =  \lambda_\rho^{i}  - \Gamma_\rho''(\lambda_\rho^i)^{-1} \, \Gamma'_\rho(\lambda_\rho^i)$ with $\lambda_\rho^{1} =0$, where $\Gamma'_\rho(\lambda_\rho^i) = b^c_\rho - E\big\{ h^c_\rho( \bfH_{\lambda_\rho^{i}} )\big\}$ and where
$\Gamma_\rho''(\lambda_\rho^i) = \var(h^c_\rho( \bfH_{\lambda_\rho^{i}} ))$  is the variance.
At each iteration $i$, the convergence of the iteration algorithm is controlled  by calculating the error $\err_R(i)$ for $R= \rho^2$ and the corresponding error $\err_W(i)$ for $\bfW^{(\nu)}$, such that
\begin{equation}
\err_R(i) = \vert \, b^c_\rho - E\big\{ h^c_\rho( \bfH_{\lambda_\rho^{i}} )\big\}\,  \vert\,  /  \, b^c_\rho \quad , \quad
\err_W(i) = \Vert \, \bfb^c_w - E\big\{\bfh^c_w(\bfH_{\lambda_\rho^{i}})\big\}\, \Vert \, /  \,\Vert \,\bfb^c_w \,\Vert \, .          \label{eq8.9}
\end{equation}

%
\noindent\textit{(ii) Algorithm $\algo~2$}. The constraints are on the normalized random residual $\rho$ and on the random control parameter $\bfW^{(\nu)}$.
The iteration algorithm is written as
\begin{equation}
\bflambda^{i+1} =  \bflambda^{i}  - [\Gamma''(\bflambda^i)]^{-1} \,
                                            \bfGamma'(\bflambda^i) \quad , \quad i \geq 1 \, ,                                      \label{eq8.12}
\end{equation}
with $\bflambda^{1} =\bfzero_{1+2 n_w}$, where $\bfGamma'(\bflambda^i) = \bfb^c - E\big\{\bfh^c(\bfH_{\bflambda^{i}})\big\}$,
and where $[\Gamma''(\bflambda^i)] = [\cov\big\{\bfh^c(\bfH_{\bflambda^{i}})\big\}]$ is the covariance matrix.
 At each iteration $i$, by using $\bflambda^i = (\lambda_\rho^i , \bflambda_w^i)$, the convergence of the iteration algorithm is controlled  by calculating the error
\begin{equation}
\err_{RW}(i) = \sqrt{ \err_R(i)^2 + \err_W(i)^2}\, ,                                                                 \label{eq8.13}
\end{equation}
in which $\err_R(i)$ and $\err_W(i)$ are  defined by
\begin{equation}
\err_R(i)  = \vert \,b^c_\rho - E\big\{ h^c_\rho( \bfH_{\bflambda^{i}} )\big\} \,\vert\,  /  \, b^c_\rho    \quad , \quad
\err_W(i)  = \Vert \, \bfb^c_w - E\big\{\bfh^c_w(\bfH_{\bflambda^{i}})\big\} \, \Vert \, /  \,\Vert \,\bfb^c_w \,\Vert \, .                  \label{eq8.11}
\end{equation}
%

\noindent\textit{(iii) Algorithm $\algo~3$}. It is algorithm $\algo~2$ for which $\lambda_\rho$ and $\bflambda_w$ are decoupled in Eq.~\eqref{eq8.12}.
The iteration algorithm is then written, for $i \geq 1$, as
\begin{equation}
(\lambda^{i}_\rho, \bflambda^{i}_w) = \bflambda^{i} \quad , \quad
\lambda_\rho^{i+1} =  \lambda_\rho^{i}  - \Gamma_\rho''(\bflambda^i)^{-1} \,\Gamma_\rho'(\bflambda^i)  \quad , \quad
\bflambda_w^{i+1} =  \bflambda_w^{i}  - [\Gamma_w''(\bflambda^i)]^{-1} \,\bfGamma_w'(\bflambda^i)  \quad , \quad
\bflambda^{i+1} = (\lambda^{i+1}_\rho, \bflambda^{i+1}_w)  \, ,                                                          \label{eq8.16}
\end{equation}
with $\bflambda^{1} =\bfzero_{1+2 n_w}$, where $(\Gamma_\rho'(\bflambda^i),\bfGamma_w'(\bflambda^i)) = \bfGamma'(\bflambda^i)$,
and where
$\Gamma_\rho''(\bflambda^i) = [\Gamma''(\bflambda^i)]_{11}$ and
$[\Gamma_w''(\bflambda^i)]_{jj'} = [\Gamma''(\bflambda^i)]_{1+j,1+j'}$ for $j$ and $j'$ in $\big\{1,\ldots ,n_w\big\}$.
At each iteration $i$, the convergence of the iteration algorithm is controlled  by calculating the errors
$\err_R(i)$ and $\err_W(i)$ by Eq.~\eqref{eq8.11}, and $\err_{RW}(i)$ by Eq.~\eqref{eq8.13}.
\subsection{Additional realizations of $\bfH_{\bflambda^i}$ and learned dataset for $(\bfY^c,\bfW^c)$}
\label{Section8.4}
%

\noindent\textit{(i) Learned dataset} $\curY^{c,\ar}_{n_\ppMC}$ \textit{for} $(\bfY^c,\bfW^c)$.
For each $\bflambda^i$ of the iteration algorithm, $n_\pMC \gg N_d$ addition realizations of $\bfH_{\bflambda^i}$ are generated for estimating
$E\big\{\bfh^c(\bfH_{\bflambda^i})\big\}$ and $[\cov\big\{\bfh^c(\bfH_{\bflambda^i})\big\}]$. At convergence for $i = i_\opt$, we have
$\bflambda^\sol \simeq \bflambda^{i_\ppopt}$. Taking into account Eq.~\eqref{eq8.7}, we have
$p_{\bfH^c} \simeq  p_{\bfH_{\bflambda}}$ for $\bflambda= \bflambda^{i_\popt}$. Consequently, at the last iteration, the generator allows for computing $n_\pMC$ additional realizations
$\big\{\bfeta^1_{c,\ar},\ldots , \bfeta^\pMC_{c,\ar}\big\}$ of $\bfH^c$ for which the generator is described in Paragraphs (iv) to (vi) below.
The learned dataset $\curY^{c,\ar}_{n_\ppMC}$ under constraints is defined by
$\curY^{c,\ar}_{n_\ppMC}  = \big \{ (\bfy^1_{c,\ar},\bfw^1_{c,\ar}),\ldots , (\bfy^{n_\ppMC}_{c,\ar}, \bfw^{n_\ppMC}_{c,\ar}) \big\}$,
in which $\bfw^\ell_{c,\ar}$ and $\bfy^\ell_{c,\ar}= \big\{ \bfy^\ell_{c,\ar}(t_n)\in\RR^N\, , \, n = 1,\ldots, n_\ptime \big\}$ are given, using Eqs.~\eqref{eq7.2} and \eqref{eq7.4}, by $\bfw^\ell_{c,\ar}  =  \underline\bfx_w +[\psi_w]\,[\xi]^{1/2}\, \bfeta^\ell_{c,\ar}$  and for $n \in \big\{1,\ldots , n_\ptime\big\}$, by $\bfy^\ell_{c,\ar}(t_n) = \underline\bfy_H(t_n)  + [V_H(t_n)]\, \bfeta^\ell_{c,\ar}$. \\

%
\noindent \textit{(ii) Computing the random residual of the stochastic equation for the learned realizations under constraints}.
The $n_\pMC$ realizations $\big\{\widehat\rho_{c,\ar}^{\,\ell}\big\}_\ell$ of random residual $\widehat\rho$, corresponding
to the learned dataset $\curY^{c,\ar}_{n_\ppMC}$ under constraints, are computed (see Eq.~\eqref{eq4.11}), for $\ell\in\big\{1,\ldots,n_\pMC\big\}$, by
$(\widehat\rho_{c,\ar}^{\,\ell})^2 =  \big ( 1/ (N\, n_\psp)\big ) \sum_{n_s=1}^{n_\psp} \Vert\, \bfr_{c,\ar}^\ell(\tau_{n_s})\, \Vert^2$,
in which $\bfr_{c,\ar}^\ell(\tau_{n_s})\in\RR^N$ is computed (see Eq.~\eqref{eq4.3}), for $t=\tau_{n_s}$, by
$\bfr_{c,\ar}^\ell(\tau_{n_s}) = \bfcurN_t\big (\bfy_{c,\ar}^\ell(\tau_{n_s}),\tau_{n_s},\bfw_{c,\ar}^\ell\big )$.\\


\noindent\textit{(iii) Construction of the diffusion-maps basis}.
The PLoM \cite{Soize2016,Soize2020c} uses the diffusion-maps basis \cite{Coifman2005} for projecting the nonlinear ISDE constructed by the PLoM method. This projection allows for preserving the concentration of the probability measure of $\bfX^c = (\bfQ^c , \bfW^c)$ on the manifold (see the definition of $\bfX^c$ at the beginning of Section~\ref{Section8.1}). The diffusion-maps basis depends only on $\big\{\bfeta_d^1, \ldots, \bfeta_d^{N_d} \big\}$ defined by Eq.~\eqref{eq5.6} (initial dataset)  and is independent of the iteration algorithm. This basis  is represented by the matrix
$[g] = [\bfg^1 \ldots \bfg^{m}]\in\MM_{N_d,m}$ with  $1 < m \leq N_d $.
Let $[\KK]$ be the symmetric $(N\times N)$ real matrix such that
$[\KK]_{\ell\ell'} = \exp\big\{- (4\,\varepsilon_\pdiff)^{-1} \Vert\,\bfeta_d^\ell -\bfeta_d^{\ell'}\,\Vert^2\big\}$,
which depends on a real smoothing parameter $\varepsilon_\diff > 0$. Let $[\PP]$ be the transition matrix in $\MM_{N_d}$ of a Markov chain such that $[\PP] = [\bb]^{-1}\, [\KK]$ in which $[\bb]$ is the positive-definite diagonal real matrix such that $[\bb]_{i\ell} = \delta_{i\ell}\,\sum_{\ell'=1}^{N_d} [\KK]_{\ell\ell'}$.
For $m$ fixed in $\big\{1,\ldots , N_d\big\}$, let  $\bfg^1,\ldots ,\bfg^m$ be the right eigenvectors in $\RR^{N_d}$ of matrix $[\PP]$ such that $[\PP]\, \bfg^\alpha = \kappa_\alpha\, \bfg^\alpha$ in which the eigenvalues are sorted such that $1=\kappa_1  > \kappa_2 > \ldots > \kappa_m$. It can easily be proven that the eigenvalues are positive and the largest is $\kappa_1 = 1$ for which all the components of the corresponding eigenvector $\bfg^1$ are equal. Considering the normalization  $[g]^T\, [\bb]\, [g] = [I_{m}]$, the right-eigenvalue problem of the nonsymmetric matrix $[\PP]$ can then be done solving the eigenvalue problem $ [\bb]^{-1/2}\, [\KK]\, [\bb]^{-1/2}\, \bfgamma^\alpha = \kappa_\alpha\, \bfgamma^\alpha$ related to a positive-definite symmetric real matrix with the orthonormalization $<\! \bfgamma^\alpha \!, \bfgamma^\beta\!> = \delta_{\alpha\beta}$. Therefore, $\bfg^\alpha$ can be deduced from $\bfgamma^\alpha$ by $\bfg^\alpha =[\bb]^{-1/2}\, \bfgamma^\alpha$.
This construction introduces two hyperparameters: the dimension $m \leq N_d$ and the smoothing parameter $\varepsilon_\diff > 0$. We refer the reader to
\cite{Soize2020c,Soize2019b} for estimating their values.
We introduce the matrix $[a]$  defined by
\begin{equation}
[a] = [g] \, \big ([g]^T\, [g]\big  )^{-1} \in \MM_{N_d,m} \, .                                                                                         \label{eq8.19}
\end{equation}
%

\noindent\textit{(iv) Generator of random vector} $\bfH_{\bflambda^i}$ \textit{by using PLoM  algorithm under constraints}.
Now and throughout the rest of Section~\ref{Section8.4},  we remove the exponent $i$ on $\bflambda$.
Let $\big \{ \big ( [\bfcurZ_{\bflambda}(r)], [\bfcurX_{\bflambda}(r)]\big  ) , r \geq 0\big\}$ be the stochastic process with values in $\MM_{\nu,m}\times \MM_{\nu,m}$, which is the solution of the following ISDE,
\begin{align}
d[\bfcurZ_{\bflambda}(r)]  = & [\bfcurX_{\bflambda}(r)] \, dr  \, ,    &r > 0\, ,                                                                \label{eq8.20}\\
d[\bfcurX_{\bflambda}(r)]  = & [\curL_{\bflambda}([\bfcurZ_{\bflambda}(r)])]\, dr
       -\frac{1}{2} \,f_0 [\bfcurX_{\bflambda}(r)]\, dr
                              + \sqrt{f_0} \, d[\bfW^\wien(r)]\, [a]  \, , & r > 0\, ,                                                             \label{eq8.21}\\
[\bfcurZ_{\bflambda}(0)]  = & [\eta_d]\, [a] \quad , \quad [\bfcurX_{\bflambda}(0)]  = [v_0]\, [a] \quad  a.s. \, , & r =0 \, ,                    \label{eq8.22}
\end{align}
in which:
\begin{enumerate}
\item[(1)] $[a]$ is the matrix defined by Eq.~\eqref{eq8.19}, $[\eta_d] = [\bfeta_d^1 \ldots \bfeta_d^{N_d}] \in \MM_{\nu,N_d}$, and
          $[v_0] \in \MM_{\nu,N_d}$ is such that $\big\{[v_0]_{k\ell}\big\}_{k\ell}$  are $\nu\times N_d$ independent normalized Gaussian real-valued random variables
          that are fixed and independent of iteration $i$.
\item[(2)] $f_0 > 0$ is a free parameter allowing the dissipation to be controlled in the stochastic dynamical system and therefore to quickly reach the stationary solution corresponding to the invariant measure. This parameter is chosen such that $f_0 \leq  4/\widehat s _\nu$. The value $4$ is generally a good choice.
\item[(3)] $\big\{[\bfW^\wien(r)] , r\in\RR^+\big\}$ is the  stochastic process, defined on $(\Theta,\curT,\curP)$, indexed by $\RR^+$, with values in $\MM_{\nu,N_d}$, for which the columns of $[\bfW^\wien]$ are $N_d$ independent copies of the $\RR^{\nu}$-valued normalized Wiener process whose matrix-valued autocorrelation function is $\min (r,r')\, [I_{\nu}]$. This stochastic process is independent of iteration $i$ (this is the same for all the iterations) and is statistically independent of the random control parameter $\bfW$.
\item[(4)] the random matrix $[\curL_{\bflambda}([\bfcurZ_{\bflambda}(r)])]$ with values in $\MM_{\nu,m}$  is written as
$[\curL_{\bflambda}([\bfcurZ_{\bflambda}(r)])] =  [L_{\bflambda}([\bfcurZ_{\bflambda}(r)]\, [g]^T)] \, [a]$,
in which, for $k=1,\ldots ,\nu$, for $\ell=1,\ldots ,N_d$,  and for $[u] = [\bfu^1 \ldots \bfu^{N_d}]\in \MM_{\nu,N_d}$
with $\bfu^\ell = (u^\ell_1,\ldots , u^\ell_\nu)$, we have
\begin{equation}
[L_\bflambda([u])]_{k\ell} =  \frac{1}{\zeta(\bfu^\ell)} \frac{\partial\zeta(\bfu^\ell)}{\partial u^\ell_k}
                       -  <\! \bflambda \, , \frac{\partial\bfh^c(\bfu^\ell)}{\partial u^\ell_k} \! > \, ,                               \label{eq8.24}
\end{equation}
in which $\zeta$ is defined by Eq.~\eqref{eq5.8} and where the gradient of $\zeta$ is such that
\begin{equation}
\bfnabla_{\bfu^\ell}\zeta(\bfu^\ell) = \frac{1}{\widehat s_\nu^2} \frac{1}{N_d}
\sum_{j=1}^{N_d} \big (\frac{\widehat s_\nu}{s_\nu} \bfeta_d^j -\bfu^\ell \big )\,
\exp\big \{ -\frac{1}{2 \widehat s_\nu^2}\Vert \,\frac{\widehat s_\nu}{s_\nu} \bfeta_d^j -\bfu^\ell\,\Vert^2\big \}\, .                 \label{eq8.25}
\end{equation}
\end{enumerate}
For $f_0$ fixed, let $r_0$ be a sufficiently large positive real number in order that the stochastic process
$\big\{ ( [\bfcurZ_{\bflambda}(r)], [\bfcurX_{\bflambda}(r)]) ,$ $r \ge r_0\big\}$
is stationary. For given $r_\st$ such that  $r_\st \geq r_0$,  the probability distribution of random variable
$\big ( [\bfcurZ_{\bflambda}(r_\st)], [\bfcurX_{\bflambda}(r_\st)] \big ) $ is the invariant measure of
Eqs.~\eqref{eq8.20} and \eqref{eq8.21}. As proven in \cite{Soize2016,Soize2020c}, the $\RR^\nu$-valued random variable $\bfH_{\bflambda}$ is a column (see Paragraph (v) below) of random matrix $[\bfH_\lambda]$ with values in $\MM_{\nu,N_d}$  such that
\begin{equation}
[\bfH_\bflambda] = [\bfcurZ_{\bflambda}(r_\st)]\, [g]^T  \, .                                                                          \label{eq8.26}
\end{equation}
%

\noindent \textit{(v) Generation of} $n_\pMC$ \textit{realizations of} $\bfH_\bflambda$.
The St\"{o}rmer-Verlet algorithm (see Appendix~A) is used for solving Eqs.~\eqref{eq8.20} to \eqref{eq8.22} as proposed in \cite{Soize2015,Soize2016}, which allows for generating $n_\pMC$ realizations $\big\{ [z_\bflambda^1], \ldots , [z_\bflambda^{n_\ppMC}]\big\}$ in $\MM_{\nu,m}$ of random matrix $[\bfcurZ_\bflambda]$. It can then be deduced $n_\pMC$ additional realizations
$\big\{ \bfeta_\bflambda^1, \ldots , \bfeta_\bflambda^{n_\ppMC}\big\}$ in $\RR^\nu$ of random vector
$\bfH_\bflambda$ such that, for $k\in\big\{1,\ldots ,\nu\big\}$ and for $\ell\in\big\{1,\ldots ,n_\pMC\big\}$, we have
$\big\{\bfeta_\bflambda^\ell\big\}_k = [\eta^\ell_{\bflambda}]_{k\,j_0^{\,\ell}}$ in which
$[\eta^\ell_{\bflambda}] = [z^\ell_{\bflambda}] \, [g]^T  \in \MM_{\nu,N}$ and where $j_0^{\,\ell}$ is defined in Section~\ref{Section6}-(i).
\section{Computational statistics method for estimating the derivative of $\bfh^c$}
\label{Section9}
Equation~\eqref{eq8.24} shows that we have to compute, for any given $\bfu^\ell = (u^\ell_1,\ldots , u^\ell_\nu)$ in $\RR^\nu$, the partial derivative
$\partial\bfh^c(\bfu^\ell) /\partial u^\ell_k$ for $k=1,\ldots, \nu$ in which $\bfh^c(\bfu^\ell) = \big (h^c_\rho(\bfu^\ell), \bfh^c_w(\bfu^\ell)\big )
\in \RR^{1+2n_w}$ (see Section~\ref{Section7}-(iii)). The first component $h^c_\rho$ of $\bfh^c$ involves the nonlinear stochastic equation and since  $\nu$ can be large, the computation of the partial derivatives is a very difficult problem. The construction of a surrogate model of function $\bfu^\ell\mapsto h^c_\rho(\bfu^\ell)$ by using a deterministic approach, such as the meshless methods \cite{Nayroles1992,Belytschko1996,Duarte1996,Breitkopf2000,Rassineux2000,Zhang2000}, is not adapted to this  difficult and numerically expensive problem taking into account the dimension $\nu$ of $\bfu^\ell$ and the possible complexity of function $h_\rho^c$. We thus propose to develop a computational statistics method based on statistical conditioning for the first component $h^c_\rho$ of $\bfh^c$. For the other components, that is to say for function $\bfh^c_w$, there is no difficulty.
We then have to construct, for all $\bfeta$ in $\RR^\nu$, an estimation of the gradient
$\bfnabla h^c_\rho(\bfeta)$ of $h^c_\rho(\bfeta)$ and, for all $i$ in $\big\{1,\ldots, 2n_w\big\}$, the gradient
$\bfnabla h^c_{w,i}(\bfeta)$ of $h^c_{w,i}(\bfeta)$.
\subsection{Estimation of the gradient of $h_\rho^c(\bfeta)$}
\label{Section9.1}
The mapping $h^c_\rho$ is approximated by the equation
\begin{equation}
h^c_\rho(\bfeta) = E\big\{\,\rho^2 \, \vert \, \bfH =\bfeta\big\} \, ,                                                                                 \label{eq9.1}
\end{equation}
in which the conditional mathematical expectation of $\rho^2$ given $\bfH =\bfeta$ is estimated by using the $n_\pMC$ realizations
$\big\{\rho^1_\ar , \ldots , \rho^{n_\ppMC}_\ar \big\}$ of  $\rho$ (see Eq.~\eqref{eq6.7bis}) and the $n_\pMC$ realizations
$\big\{ \bfeta^1_\ar ,\ldots , \bfeta^\pMC_\ar\big\}$ of random vector $\bfH$ (see Section~\ref{Section6}-(i)), computed by the PLoM without constraint. These realizations have been chosen because the PLoM method without constraint already gives a learned dataset of dimension $n_\pMC \gg N_d$, which already verifies well the stochastic equation.\\

\noindent\textit{(i) Normalization and gradient of} $h^c_\rho$.
The scaling of $\bfH$ and of its realizations $\big\{\bfeta^\ell_\ar\big\}_\ell$ is necessary.
Let $\underline\bfeta_\ar =(\underline\eta_{\ar ,1},\ldots , \underline\eta_{\ar ,\nu})$ and $\bfsigma_\ar = (\sigma_{\ar,1},\ldots ,\sigma_{\ar,\nu})$ be the empirical estimates of the mean value and the standard deviation of $\bfH$, computed with $\big\{ \bfeta^1_\ar ,\ldots , \bfeta^\pMC_\ar\big\}$.
Let $\widehat\bfH = (\widehat H_1,\ldots ,\widehat H_\nu)$ and $\widehat\bfeta = (\widehat\eta_1,\ldots ,\widehat\eta_\nu)$ such that, for all $\bfeta =(\eta_1,\ldots ,\eta_\nu)$ in $\RR^\nu$ and for all $k\in\big\{1,\ldots ,\nu\big\}$,
$\widehat H_k =   (H_k - \underline\eta_{\,\ar , k}) / \sigma_{\ar,k}$ and
$\widehat\eta_k = (\eta_k - \underline\eta_{\,\ar ,k}) / \sigma_{\ar,k}$.
Consequently, Eq.~\eqref{eq9.1} can be rewritten as
\begin{align}
h^c_\rho(\bfeta)& = \widehat h^c_\rho(\widehat\bfeta)
                     \quad , \quad \eta_k = \underline\eta_{\ar ,k} + \sigma_{\ar,k} \widehat\eta_k  \, ,                \label{eq9.3}\\
\widehat h^c_\rho(\widehat\bfeta)& = E\big\{\, \rho^2 \, \vert \, \widehat\bfH =\widehat\bfeta\big\}\, ,                                                   \label{eq9.4}
\end{align}
in which $\widehat\bfeta = \widehat\bfeta(\bfeta)$ has to be seen as a function of $\bfeta$.
The realizations of $\widehat\bfH$ are $\big\{ \widehat\bfeta^1_\ar ,\ldots , \widehat\bfeta^{\,\pMC}_\ar\big\}$ such that, for $\ell=1,\ldots ,n_\pMC$
and $k=1,\ldots , \nu$, we have $\widehat\eta^{\,\ell}_{\ar,k} = (\eta^\ell_{\ar,k} - \underline\eta_{\,\ar , k}) / \sigma_{\ar,k}$.
From Eqs.~\eqref{eq9.3}, it can be deduced that, for all $k\in \big\{ 1,\ldots , \nu \big\}$, we have
$ {\partial h^c_\rho(\bfeta)} / {\partial\eta_k} = \sigma_{\ar,k}^{-1} \,  {\partial \widehat h^c_\rho(\widehat\bfeta)} / {\partial\widehat\eta_k} $.\\

%
\noindent\textit{(ii) Estimation of} $\widehat h^c_\rho(\widehat\bfeta)$. Let $(r,\widehat\bfeta)\mapsto p_{R,\widehat\bfH}(r,\widehat\bfeta)$ be the joint pdf on $\RR\times\RR^\nu$ of the positive-valued random variable $R = \rho^2$ and $\widehat\bfH$ with respect to the Lebesgue measure on $\RR\times\RR^\nu$. We then have
\begin{equation}
p_{\widehat\bfH}(\widehat\bfeta) = \int_\RR p_{R,\widehat\bfH}(r,\widehat\bfeta) \, dr \, .                                           \label{eq9.7}
\end{equation}
From Eq.~\eqref{eq9.4}, it can be deduced that
\begin{equation}
\widehat h^c_\rho(\widehat\bfeta) = \frac{1}{p_{\widehat\bfH}(\widehat\bfeta)} \int_\RR r\, p_{R,\widehat\bfH}(r,\widehat\bfeta) \, dr \, .   \label{eq9.8}
\end{equation}
Let $\big\{r_\ar^1, \ldots r_\ar^{\,n_{\ppMC}}\big\}$ be the realizations of $R$ such that $r_\ar^\ell = (\rho^\ell_\ar)^2$, and let $\sigma_R$ be its standard deviation estimated using $\big\{r_\ar^1, \ldots r_\ar^{\,n_\ppMC}\big\}$. For $n_\pMC$ sufficiently large, the Gaussian kernel-density estimation \cite{Bowman1997} of $p_{R,\widehat\bfH}(r,\widehat\bfeta)$ can be written as
\begin{equation}
p_{R,\widehat\bfH}(r,\widehat\bfeta) \simeq \frac{1}{n_\pMC} \sum_{\ell=1}^{n_\ppMC} \frac{1}{\sigma_R(\sqrt{2\pi} s_\pSB)^{\nu+1}}
       \exp \big \{ -\frac{1}{2 s_\pSB^2} \big (  (\frac{r^\ell_\ar \! - \!r}{\sigma_R} )^2 \!
                                          + \Vert \,\widehat\bfeta^{\,\ell}_\ar \! -\! \widehat\bfeta\,\Vert^2 \big ) \big \} \, ,                               \label{eq9.9}
\end{equation}
in which $s_\pSB = \big\{ 4 /\big(n_\pMC(2+\nu + 1)\big) \big\}^{1/(4 +\nu+1)}$ is the usual multidimensional optimal Silverman bandwidth.  Using Eqs.~\eqref{eq9.7} and \eqref{eq9.9}  yields
\begin{equation}
p_{\widehat\bfH}(\widehat\bfeta) \simeq \frac{1}{n_\pMC} \sum_{\ell=1}^{n_\ppMC} \frac{1}{(\sqrt{2\pi} s_\pSB)^{\nu}}
       \exp \big \{ -\frac{1}{2 s_\pSB^2} \Vert \,\widehat\bfeta^{\,\ell}_\ar - \widehat\bfeta\,\Vert^2  \big\} \, .                                           \label{eq9.10}
\end{equation}
From Eqs.~\eqref{eq9.8} to \eqref{eq9.10}, it can be deduced that
\begin{equation}
  \widehat h^c_\rho(\widehat\bfeta)  = \frac{\curA(\widehat\bfeta)}{\curB(\widehat\bfeta)} \, ,                                                           \label{eq9.11}
\end{equation}
in which $\curA(\widehat\bfeta)$ and $\curB(\widehat\bfeta)$ are defined by
\begin{equation}
  \curA(\widehat\bfeta) = \sum_{\ell=1}^{n_\ppMC} r^\ell_\ar \exp \big\{ \!-\frac{1}{2 s_\pSB^2} \Vert \,\widehat\bfeta^{\,\ell}_\ar - \widehat\bfeta\,\Vert^2  \big\}
  \quad , \quad
  \curB(\widehat\bfeta)  = \sum_{\ell=1}^{n_\ppMC}  \exp \big \{ \!-\frac{1}{2 s_\pSB^2} \Vert\, \widehat\bfeta^{\,\ell}_\ar - \widehat\bfeta\,\Vert^2  \big \} \, .  \label{eq9.12}
\end{equation}

%
\noindent\textit{(iii) Estimation of} $\bfnabla\widehat h^c_\rho(\widehat\bfeta)$. From Eqs.~\eqref{eq9.11} and \eqref{eq9.12}, it can be seen that
\begin{equation}
 \bfnabla_{\!\!\widehat\bfeta} \, \widehat h^c_\rho(\widehat\bfeta)  = \frac{1}{\curB(\widehat\bfeta)} \big (\bfnabla_{\!\!\widehat\bfeta}\,
      \curA(\widehat\bfeta) - \widehat h^c_\rho(\widehat\bfeta) \,\bfnabla_{\!\!\widehat\bfeta}\,\curB(\widehat\bfeta) \big ) \, ,                \label{eq9.14}
\end{equation}
in which $\bfnabla_{\!\!\widehat\bfeta} \,\curA(\widehat\bfeta)$ and $\bfnabla_{\!\!\widehat\bfeta} \,\curB(\widehat\bfeta)$ are given by
\begin{equation}
  \bfnabla_{\!\!\widehat\bfeta} \,\curA(\widehat\bfeta) = \frac{1}{s_\pSB^2}\sum_{\ell=1}^{n_\ppMC} r^\ell_\ar
    (\widehat\bfeta^\ell_\ar \! -\widehat\bfeta )\exp \big\{ \! -\frac{1}{2 s_\pSB^2} \Vert \,\widehat\bfeta^\ell_\ar \! -  \widehat\bfeta\,\Vert^2  \big\} \quad ,\quad
  \bfnabla_{\!\!\widehat\bfeta} \,\curB(\widehat\bfeta)  = \frac{1}{s_\pSB^2}\sum_{\ell=1}^{n_\ppMC} (\widehat\bfeta^\ell_\ar \! -\widehat\bfeta )
                                                    \exp \big\{\! -\frac{1}{2 s_\pSB^2} \Vert\, \widehat\bfeta^\ell_\ar \! - \widehat\bfeta\,\Vert^2  \big\} \, . \label{eq9.15}
\end{equation}

%
\noindent\textit{(iv) Numerical conditioning}. Equations~\eqref{eq9.11} to \eqref{eq9.15} show the presence of exponentials of $-o^\ell(\widehat\bfeta)$ with
$o^\ell(\widehat\bfeta) = \frac{1}{2 s_\pSB^2} \Vert\, \widehat\bfeta^\ell_\ar - \widehat\bfeta\,\Vert^2$. If, for a given $\widehat\bfeta$, $o^\ell(\widehat\bfeta)$ is large enough for all $\ell=1,\ldots ,n_\pMC$, then we can get a value close to $0/0$ or $c/0$ for the computation of $\bfnabla_{\!\!\widehat\bfeta} \, \widehat h^c_\rho(\widehat\bfeta)$, which is "almost" indeterminate or "almost" in overflow, since $\exp\big\{-o^\ell(\widehat\bfeta)\big\}$ can be very small for all $\ell$. It can be seen that if $\exp\big\{-o^\ell(\widehat\bfeta)\big\}$ is replaced by $\exp\big\{-(o^\ell(\widehat\bfeta)- \underline e(\widehat\bfeta)\big\}$ in Eqs.~\eqref{eq9.11} to \eqref{eq9.15} in which $\underline e(\widehat\bfeta)$ is independent of $\ell$, then $\widehat h^c_\rho(\widehat\bfeta)$ defined by Eq.~\eqref{eq9.11} and  $\bfnabla_{\!\!\widehat\bfeta} \, \widehat h^c_\rho(\widehat\bfeta)$ defined by Eq.~\eqref{eq9.14} are not modified. Therefore, for improving the numerical conditioning,
in Eqs.~\eqref{eq9.11} to \eqref{eq9.15}, $\exp\big\{-\frac{1}{2 s_\pSB^2} \Vert\, \widehat\bfeta^{\,\ell}_\ar - \widehat\bfeta\,\Vert^2\big\}$ is replaced by
$\exp\big\{- \big ( \frac{1}{2 s_\pSB^2} \Vert\, \widehat\bfeta^{\,\ell}_\ar - \widehat\bfeta\,\Vert^2 - \underline e(\widehat\bfeta) \big )\big\}$ in which
$\underline e(\widehat\bfeta) = (1/n_\pMC) \sum_{\ell=1}^{n_\ppMC}  \Vert\, \widehat\bfeta^{\,\ell}_\ar - \widehat\bfeta\,\Vert^2/(2 s_\pSB^2)$.
\subsection{Estimation of the gradient of $h_{w,i}^c(\bfeta)$ for $i=1,\ldots,2 n_w$}
\label{Section9.2}
From Section~\ref{Section7}-(ii) and for all $j\in\big\{1,\ldots , n_w\big\}$, we have
$h^c_{w,j}(\bfH) = W_j^{(\nu)}$ and $h^c_{w,j+n_w}(\bfH) = (W_j^{(\nu)})^2$.
Using Eq.~\eqref{eq7.2} yields, for all $\bfeta =(\eta_1,\ldots ,\eta_\nu)\in\RR^\nu$,
\begin{equation}
h^c_{w,j}(\bfeta) = \!\big\{ \underline\bfx_w \!+ \![\psi_w]\,[\xi]^{1/2}\, \bfeta\big\}_j \quad , \quad
h^c_{w,j+n_w}(\bfeta) = \!\big (\big\{ \underline\bfx_w\! + \![\psi_w]\,[\xi]^{1/2}\, \bfeta\big\}_j\big )^2 \, .                                         \label{eq9.17}
\end{equation}
It can easily be deduced that, for all $j\in \big\{1,\ldots , n_w\big\}$ and for all $k\in\big\{ 1,\ldots , \nu\big\}$,
\begin{equation}
\frac{\partial h^c_{w,j}(\bfeta)}{\partial \eta_k}  = \big \{[\psi_w]\,[\xi]^{1/2} \big \}_{jk} \quad , \quad
\frac{\partial h^c_{w,j+n_w}(\bfeta)} {\partial \eta_k}  =
       2 \big (\big\{ \underline\bfx_w \!+ \![\psi_w]\,[\xi]^{1/2}\, \bfeta\big\}_j\big  )  \big \{[\psi_w]\,[\xi]^{1/2} \big \}_{jk}\,  .                     \label{eq9.18}
\end{equation}
%
%
\section{Application 1: nonlinear stochastic dynamical system}
\label{SectionDO}
%
This application deals with the nonstationary stochastic response of a nonlinear dynamic system with one degree of freedom (Duffing oscillator) excited by a nonstationary stochastic process and with a random control parameter. This example was deliberately chosen to be very simple in order to quickly show to the reader the role played by the constraints on the stochastic equation during the learning carried out by the PLoM. For this application, we will not give the algebraic complements necessary for its implementation and we will not do the detailed analysis. These aspects will be presented  for Applications~2 and 3.
\subsection{Nonlinear stochastic equation}
\label{SectionDO1}

\noindent \textit{(i) Nonlinear stochastic dynamics equation with initial conditions}. We consider the Duffing oscillator with $N=1$,
\begin{align}
\ddot Y(t) & + 2\,\chi_d \, g_1(W_1)\, \dot Y(t) +  g_1(W_1)^2 \big ( 1  +  k_b\, Y(t)^2\big )Y(t) = \gamma(t,W_2) \quad , \quad t\in ]0,T] \, , \nonumber \\
Y(0) &= \dot Y(0) = 0 \quad  a.s. \, ,                                                                                                               \nonumber
\end{align}
in which $\bfW =(W_1,W_2)$ is the $\RR^{2}$-valued random control parameter ($n_w=2$) defined on the probability space $(\Theta,\curT,\curP)$ such that
$W_1$ and $W_2$ are independent normalized Gaussian random variables, where $\chi_d = 0.05$, $k_b = 5\times 10^8$, $T= 0.7325\, s$, and where $Y = \big\{Y(t),t\in [0,T]\big\}$ is the unknown real-valued stochastic process  defined on  $(\Theta,\curT,\curP)$.\\

\noindent \textit{(ii) Nonstationary stochastic excitation and random control parameters}. The stochastic excitation is defined  by
\begin{equation}
\gamma(t,W_2) = \big( \frac{t}{T} \, g_2(W_2)\big )^2\, \big( 1 + 0.05\, \sin(\omega_b \,t)\big)\, \exp\bigg \{ - \frac{1}{1-(\frac{2t}{T} -1)^4}\bigg \} \quad , \quad \forall \, t \geq 0 \, ,                                                                             \nonumber
\end{equation}
in which $\omega_b = 2\pi \!\times\! 100\, rad/s$. For $j=1,2$, we have
$g_j(W_j) = \underline{g}_j\,\big(1+\sqrt{3}\, \delta_j (2\, \curU_j-1)\big )$ in which $\curU_j  = \big (1+\hbox{erf}(W_j/\sqrt{2})\big )/2$,
with $\delta_j=0.2$ and where $\underline{g}_j = E\big\{g_j(W_j)\big\}$ with $\underline{g}_1  = \omega_b$ and $\underline{g}_2 =6$.\\

\noindent \textit{(iii) Values of the main parameters}. The time sampling is done with $n_\ptime =2\, 931$ sampling times for which the time step is $\Delta t = 2.5\!\times\! 10^{-4}\, s$. There is no time subsampling, that is to say, $n_\psp = n_\ptime$. The number of points in the training dataset is $N_d= 80$, the KL-expansion of stochastic process $Y$ has been performed with $\varepsilon_\pKL = 10^{-6}$ yielding $n_q =25$. The PCA of random vector $\bfX$ of dimension $n_x=n_q+n_w=27$ has been performed with $\varepsilon_\PCA = 10^{-6}$ yielding $\nu =27$. For $\varepsilon_\pdiff = 63$, the dimension of the diffusion-maps basis is $m=28$. In the ISDE, $f_0=4$ and the parameters for the St\"{o}rmer-Verlet algorithm (Appendix~A) are $\Delta r = 0.19914$, $l_0 = 100$, and $M_0 = 20$.
\subsection{Reference stochastic solution}
\label{SectionDO2}
\begin{figure}[h!]
  \centering
  \includegraphics[width=5.5cm]{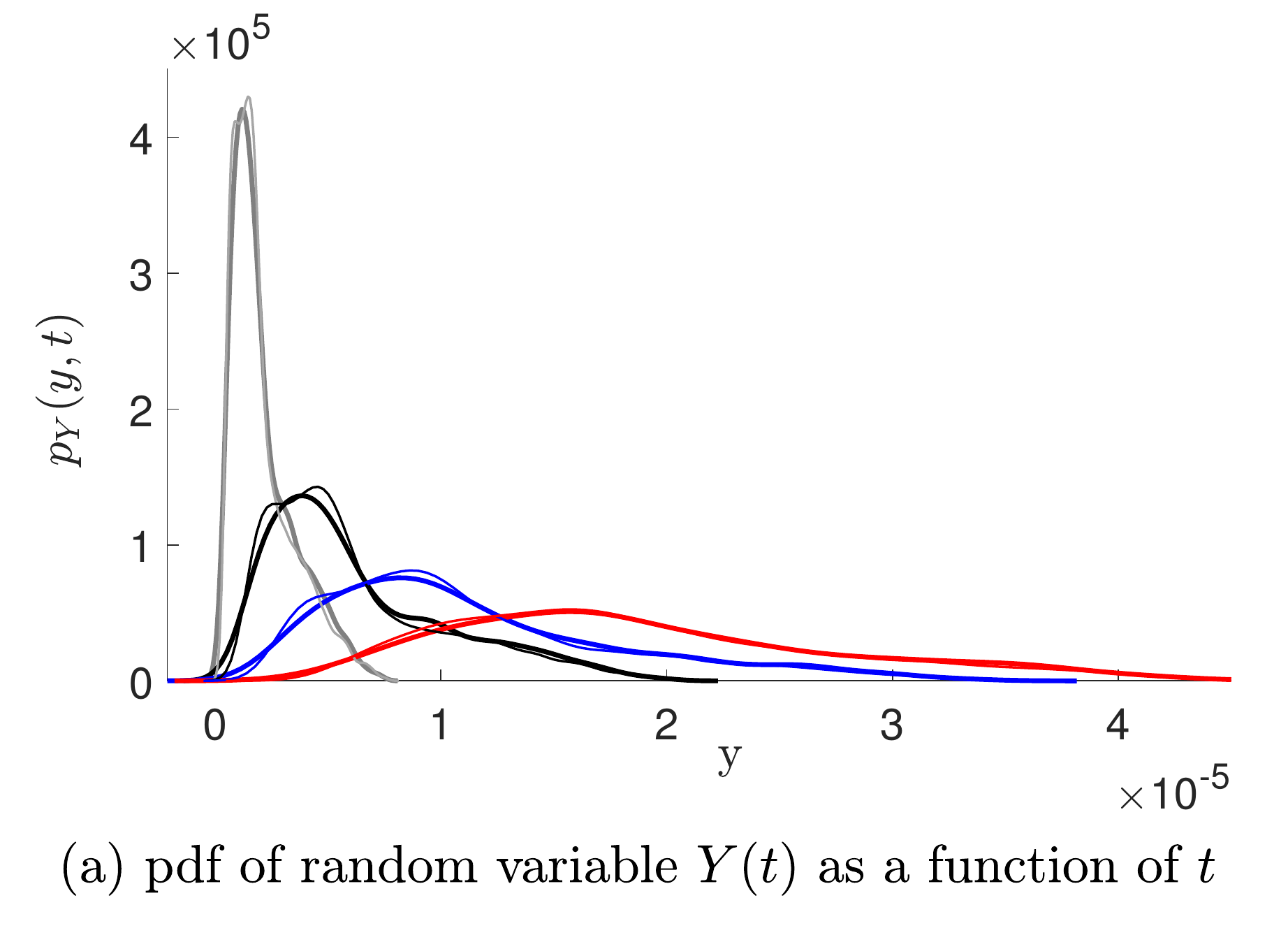}  \includegraphics[width=5.5cm]{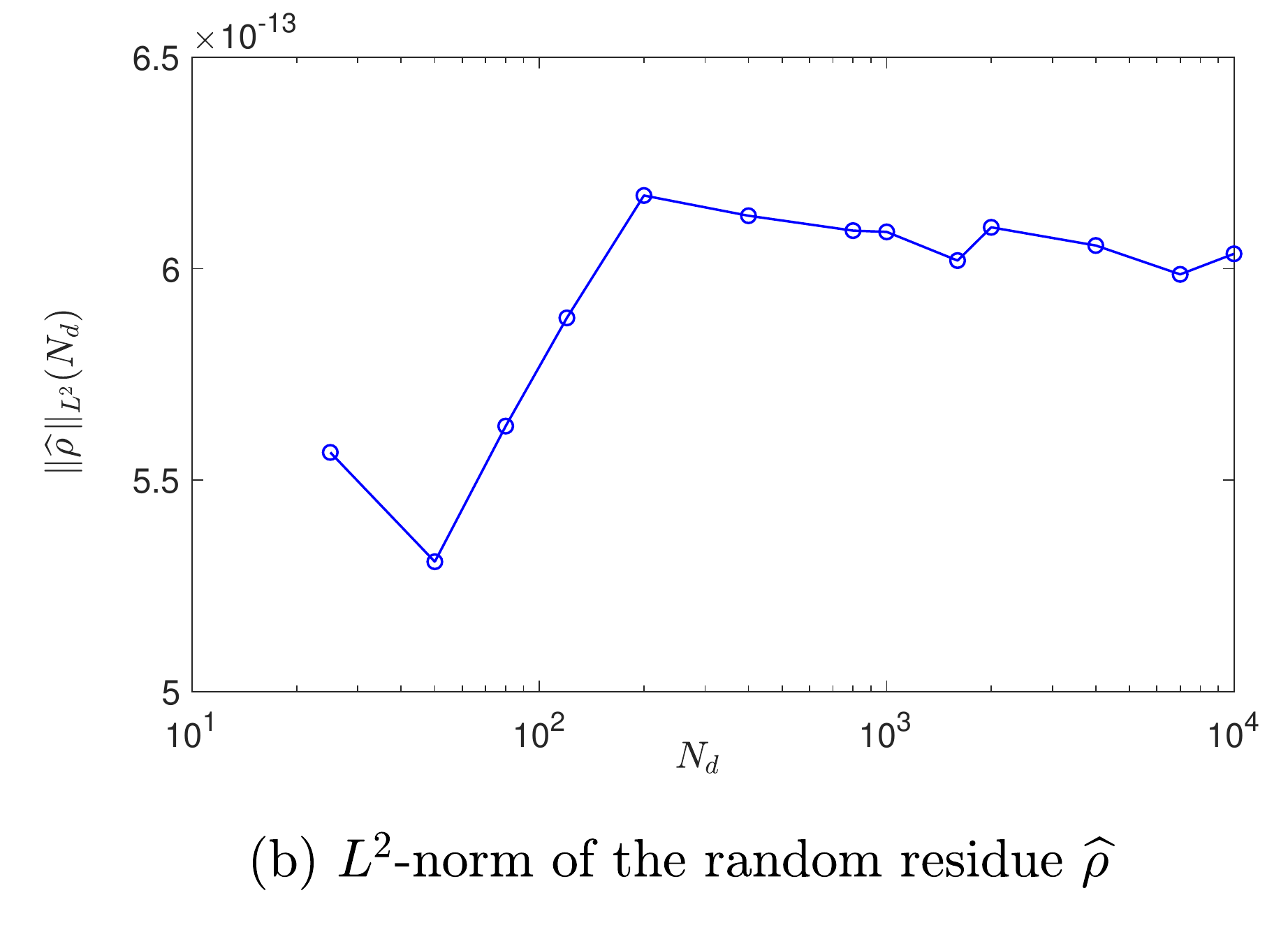}
\caption{(a) time evolution of the pdf $y\mapsto p_Y(y,t)$ for $t = 0.1785\, s$ (green), $t = 0.2977\, s$ (black),  $t = 0.3960\, s$ (blue), and $t = 0.5535\, s$ (red) estimated with $N_d =1\,000$ points (thick line) and with $N_d =10\,000$ points (thin line) in the training dataset; (b) graph of $N_d \mapsto \Vert\,\widehat\rho\, \Vert_{L^2}(N_d)$.}
\label{figureDO1}
\end{figure}

Figure~\ref{figureDO1} is related to the reference solution that is constructed using a training dataset with $N_d=1\,000$ points. Figure~\ref{figureDO1}-(a) displays the time evolution of the pdf $y\mapsto p_Y(y,t)$ of random variable $Y(t)$ for $t = 0.1785$, $0.2977$, $0.3960$, and $0.5535\, s$, which are estimated with $N_d =1\,000$ and $10\,000$ points in the training dataset. It can be seen that convergence with respect to $N_d$ is obtained for $N_d=1\,000$ (the curves for $10\,000$ are very close to those for $1\,000$). The convergence of the reference solution constructed with  $N_d=1\,000$ can also be viewed in Fig.~\ref{figureDO1}-(b) that shows the graph of function $N_d \mapsto \Vert\,\widehat\rho\, \Vert_{L^2}(N_d)$ of the random residual $\widehat\rho$. The values of $\Vert\,\widehat\rho\, \Vert_{L^2}(N_d)$ are close for $N_d=1\,000$ and $N_d=10\, 000$.
\subsection{Analysis of the role played by theconstraints on the probability distribution of $\bfW$}
\label{SectionDO3}
\begin{figure}[h!]
  \centering
  \includegraphics[width=5.5cm]{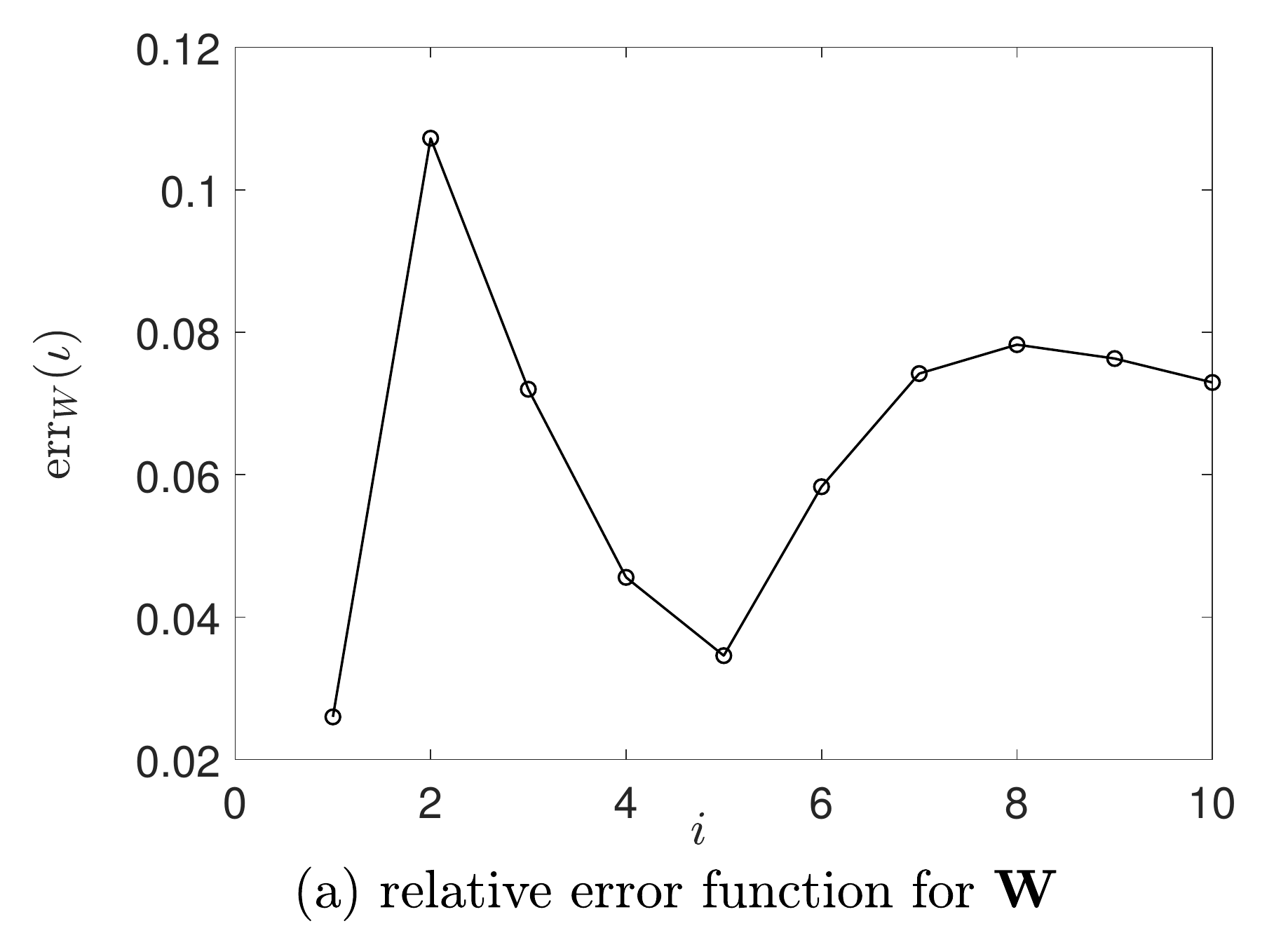}  \includegraphics[width=5.5cm]{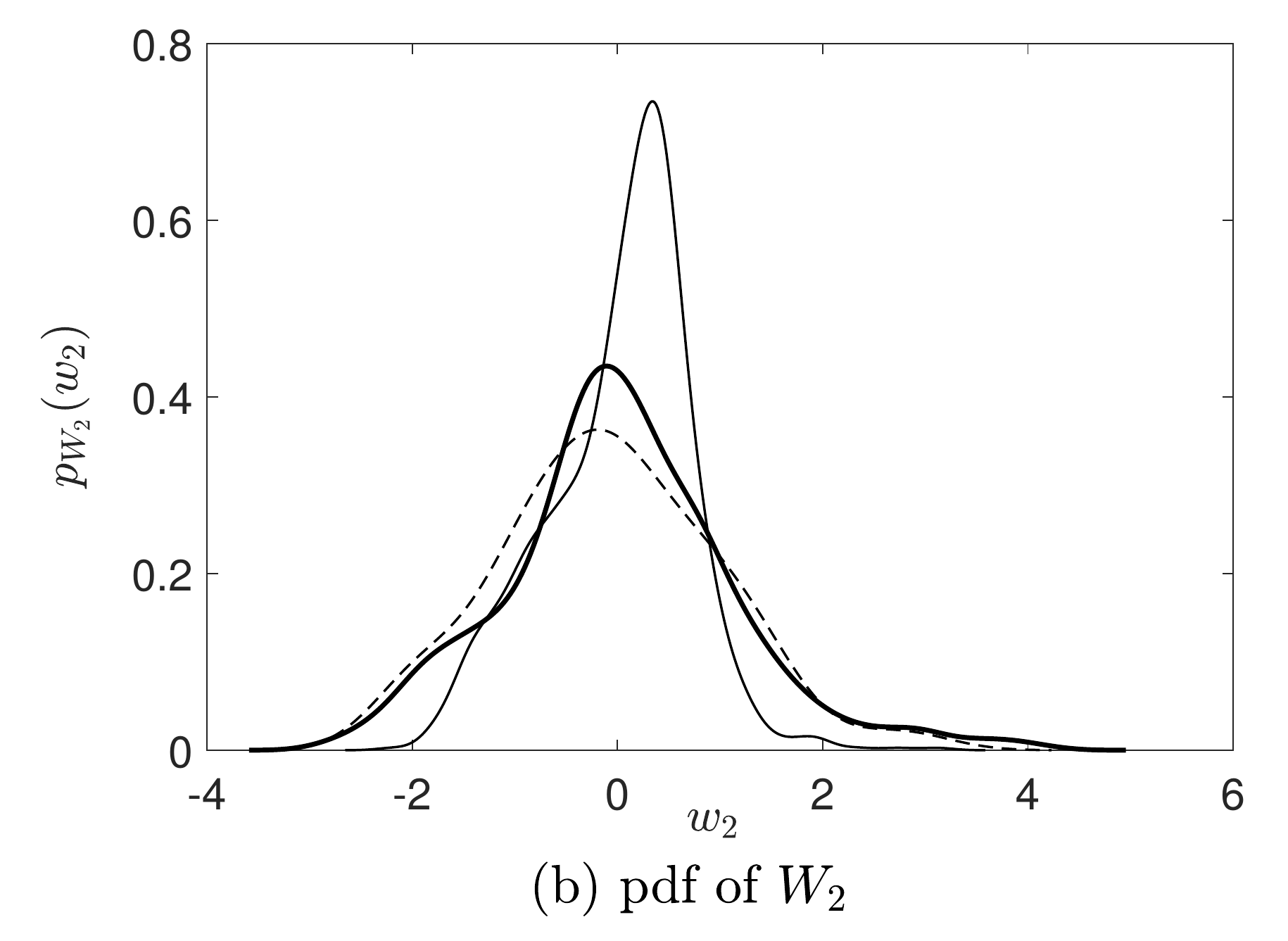}
\caption{For $N_d = 80$ and $n_\pMC = 1\,000$: (a) relative error function $i\mapsto {\rm{err}}_W(i)$ computed with  $\algo~1$ of the PLoM under the constraint on $\rho$ but without the constraints on $\bfW$;  (b) pdf $w_2\mapsto p_{W_2}(w_2)$ of $W_2$ estimated with the training dataset (dashed line), with  $\algo~1$ of  the PLoM under  the constraint on $\rho$ but without the constraints on $\bfW$ (thin solid line), and with $\algo~3$ of the PLoM under the constraints on $\rho$ and  $\bfW$ (thick solid line).}
\label{figureDO2}
\end{figure}

For simplifying the writing, $\bfW^{(\nu)}$ is simply noted $\bfW$.
Figure~\ref{figureDO2} allows for analyzing the role played by the constraints on the probability distribution of $\bfW$ during the minimization of the normalized random residual. The computation is performed with $N_d=80$ points in the training dataset and $n_\pMC=1\,000$ additional realizations for the PLoM under constraints. Figure~\ref{figureDO2}-(a) displays the graph of error function $i\mapsto {\rm{err}}_W(i)$ (see Eq.~\eqref{eq8.9}) computed with  $\algo~1$ of the PLoM under the constraint on $\rho$ but without the constraints on $\bfW$. At iteration $i_\opt=10$, the relative error ${\rm{err}}_R(i_\opt)$
(see Eq.~\eqref{eq8.9}) is minimum while $i\mapsto {\rm{err}}_W(i)$ (see Eq.~\eqref{eq8.9}) is increasing from $i =5$ and does not reach a minimum for $i_\opt=10$. This phenomenon can also be seen of Figure~\ref{figureDO2}-(b) that displays the pdf $w_2\mapsto p_{W_2}(w_2)$ of $W_2$ estimated with the training dataset, with  $\algo~1$ of  the PLoM under the constraint on $\rho$ but without the constraints on $\bfW$, and with $\algo~3$ of the PLoM under the constraints on $\rho$ and $\bfW$. It can be seen that $\algo~3$ allows for obtaining a pdf of $W_2$, which is close to the pdf that corresponds to the training dataset.
\subsection{Results of the PLoM without and under constraints for a given dimension of the training dataset}
\label{SectionDO4}
\begin{figure}[h!]
  \centering
  \includegraphics[width=5.0cm]{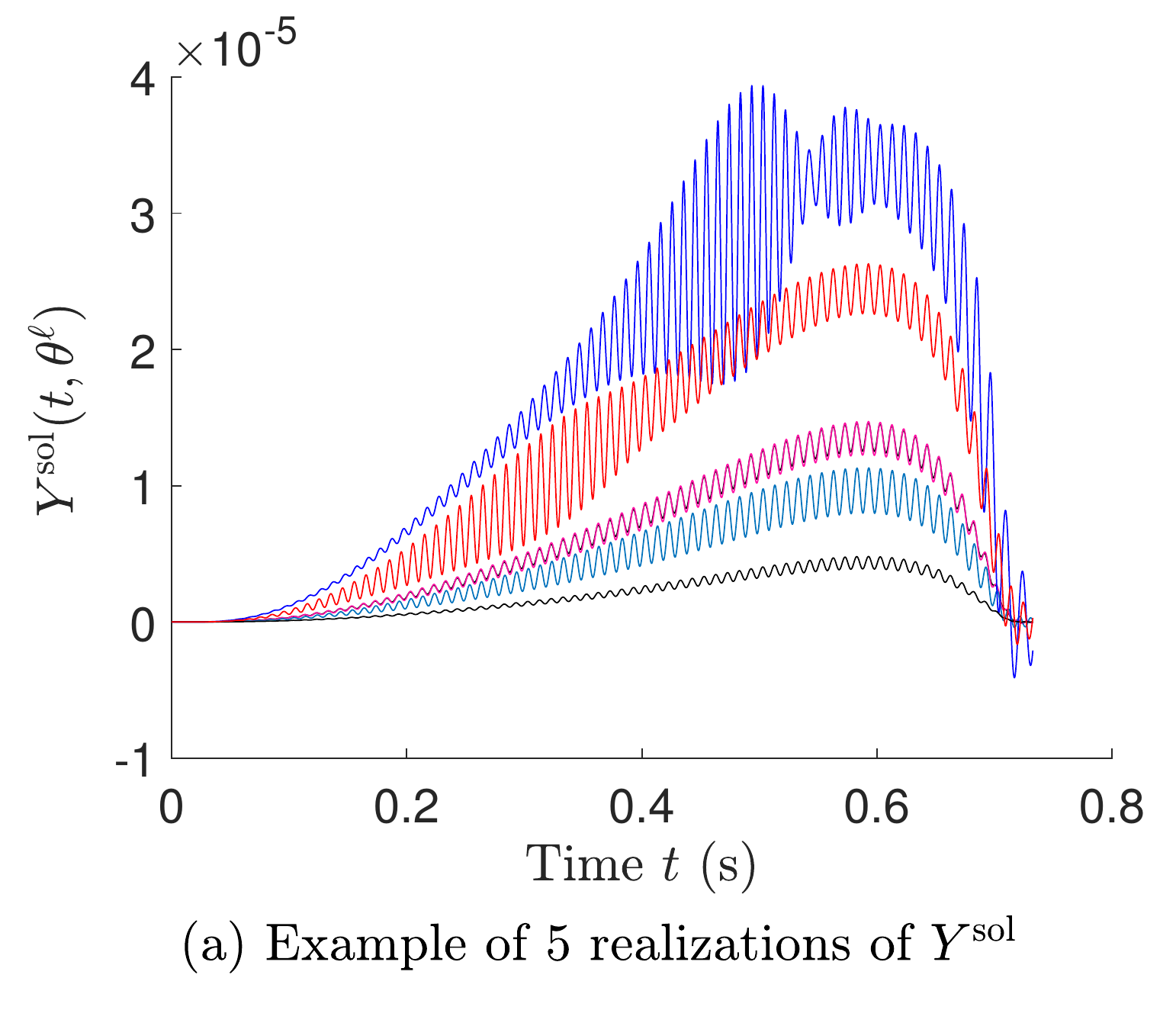} \hfil \includegraphics[width=5.5cm]{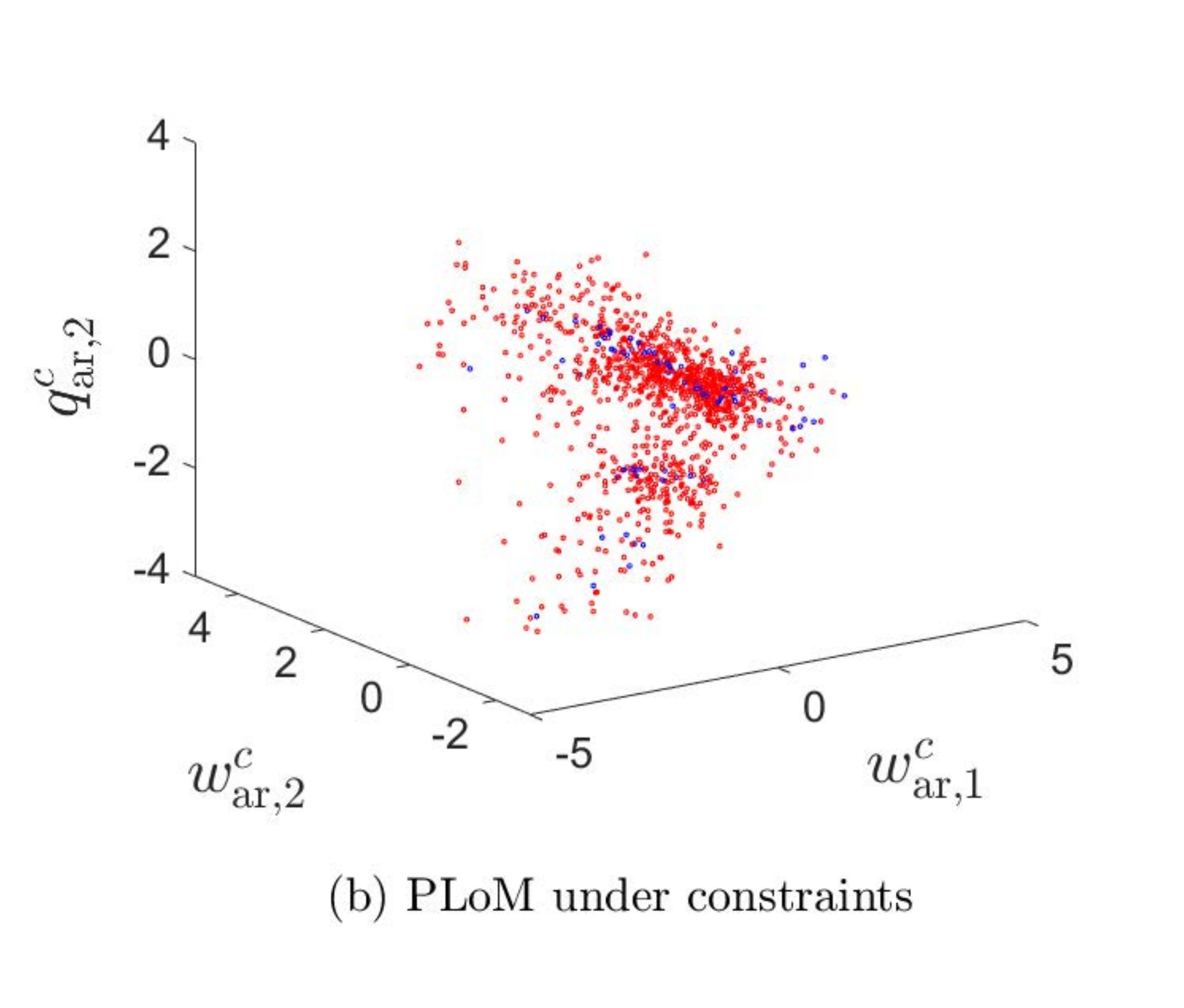}
\caption{(a) Example of $5$ realizations $\bfy^\ell$ of $Y^\sol$. (b) 3D-plot of realizations $(w^c_{\arp,1},w^c_{\arp,2},q^c_{\arp,2})$ of $(W^c_1,W^c_2,Q^c_2)$  computed with the PLoM under the constraints: $n_\pMC = 1\,000$ additional realizations (red symbols) and $N_d = 80$ points from the training dataset (blue symbols).}
\label{figureDO3}
\end{figure}

In this section, the computation is performed with $N_d=80$ points in the training dataset and $n_\pMC=1\,000$ additional realizations for the PLoM under constraints.
In order to visualize the type of sample path of stochastic process $Y^\sol$, Fig.~\ref{figureDO3}-(a) displays the graph of $5$ realizations $\bfy^\ell = Y^\sol(\cdot,\theta^\ell)$ of stochastic process $Y^\sol$.
Fig.~\ref{figureDO3}-(b) displays $n_\pMC = 1\,000$ additional realizations of components $(W^c_1,W^c_2,Q^c_2)$ of random vector $\bfX^c = (\bfQ^c,\bfW^c)$. A concentration of these realizations can be observed.
\begin{figure}[h!]
  \centering
  \includegraphics[width=4.0cm]{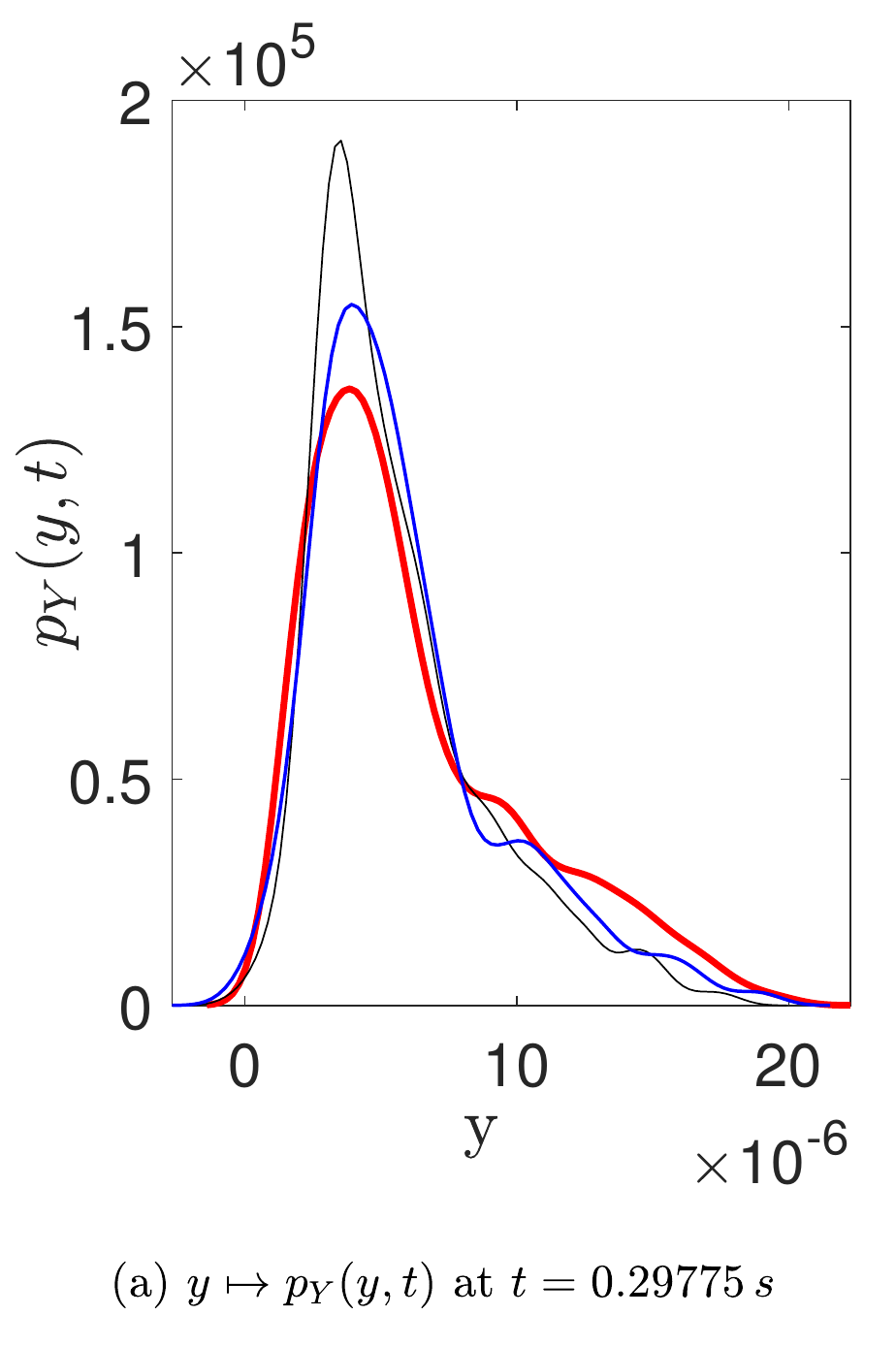}  \includegraphics[width=4.0cm]{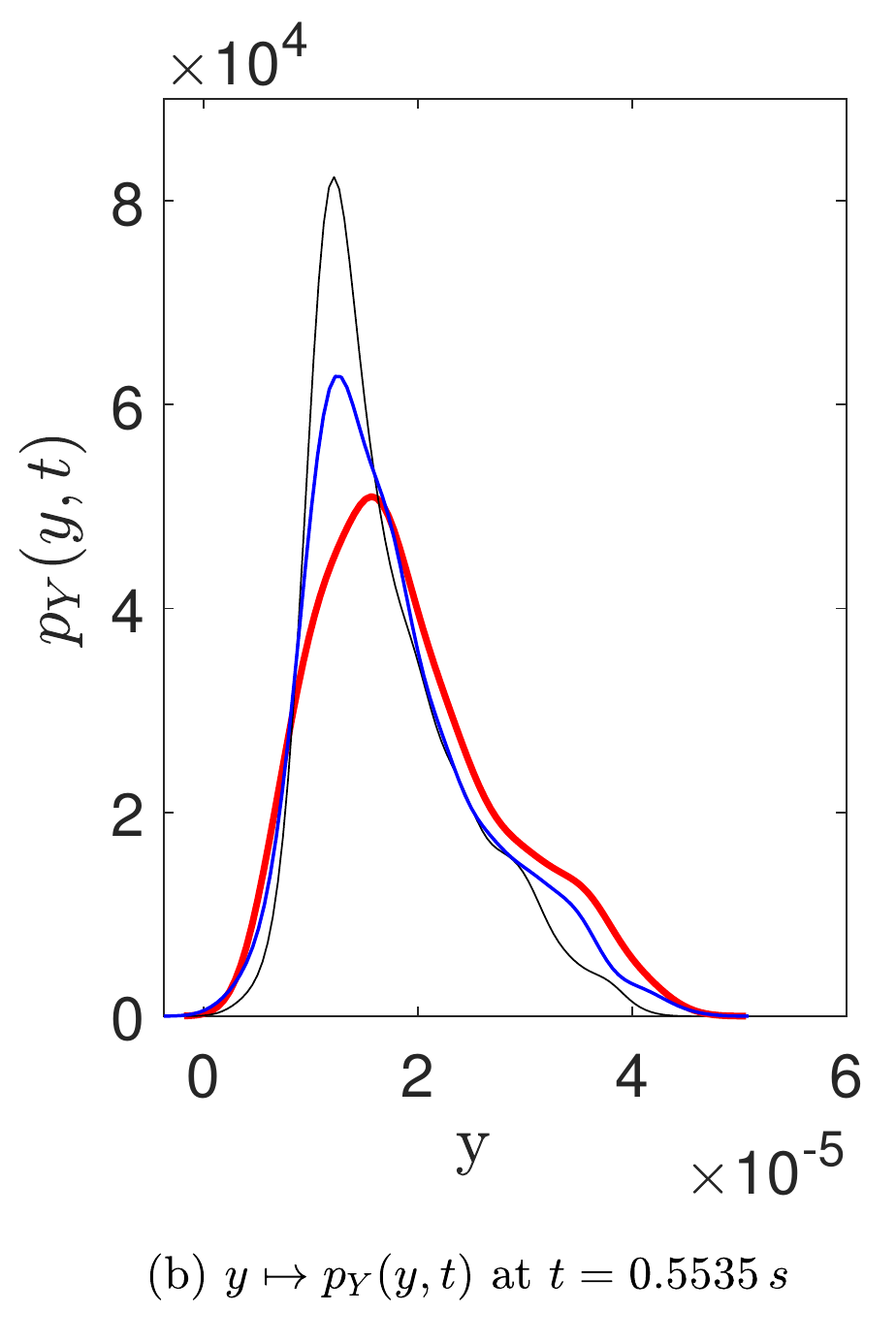}
  \includegraphics[width=4.0cm]{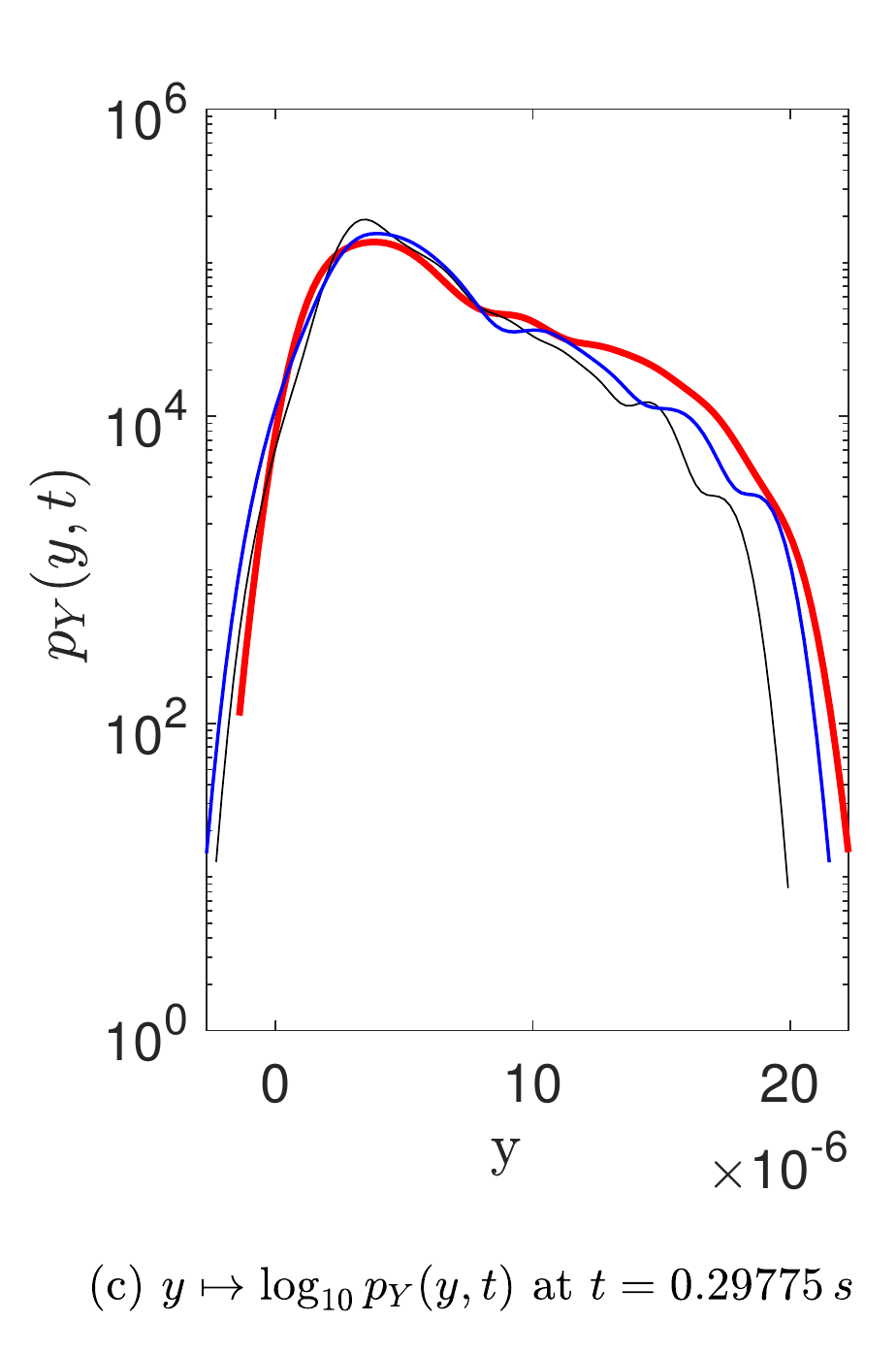}  \includegraphics[width=3.9cm]{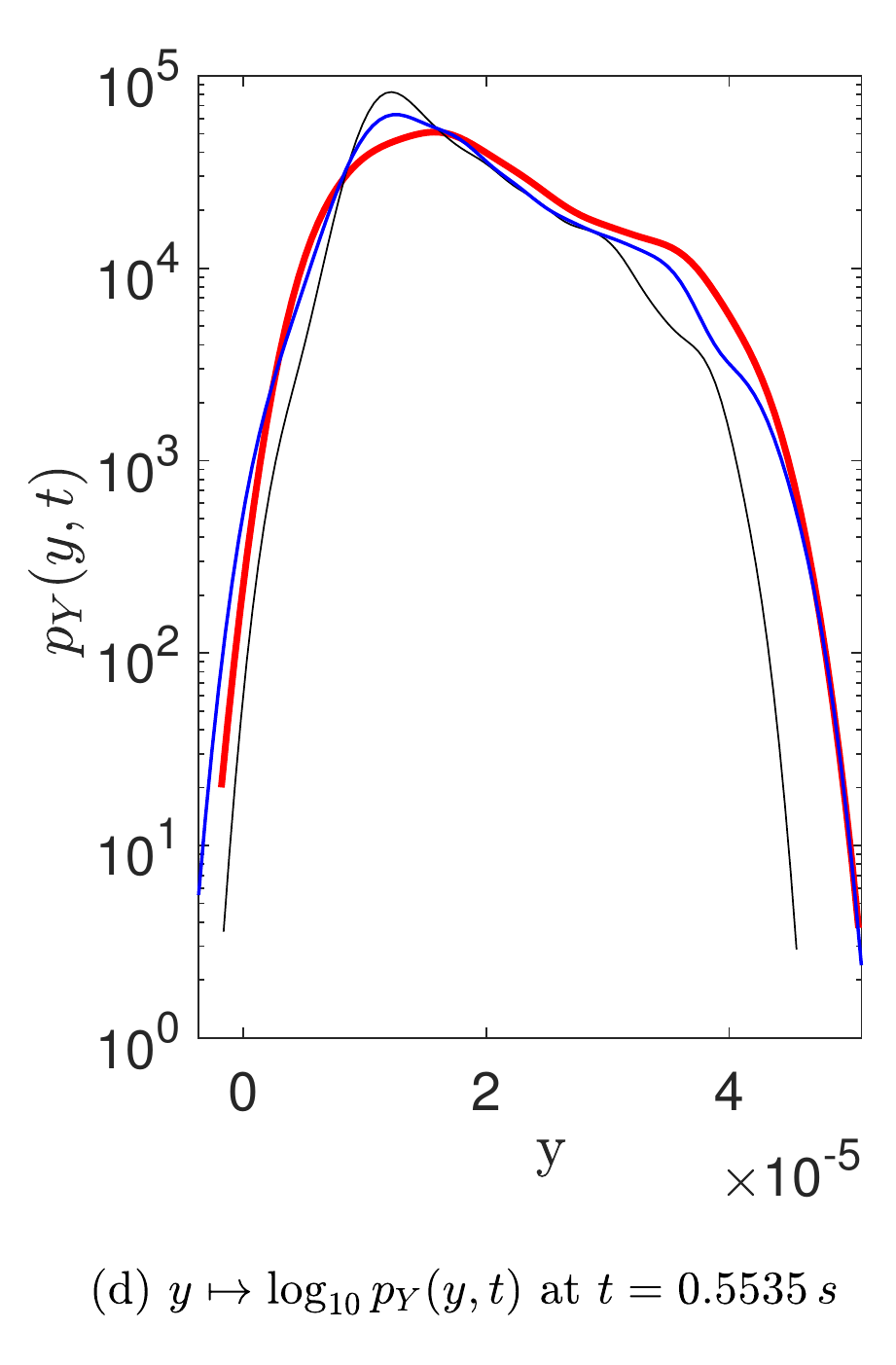}
\caption{For $N_d=80$ and $n_\pMC=1\,000$:  pdf $y\mapsto p_Y(y,t)$ (a)-(b) and $y\mapsto \log_{10} p_Y(y,t)$ (c)-(d)   for $t = 0.2977\, s$ (a)-(c) and for $t = 0.5535\, s$ (b)-(d), corresponding to the reference (red thick lines), to the PLoM without constraint (thin black lines), and to the PLoM under constraints (blue med line).}
\label{figureDO4}
\end{figure}
Concerning the pdf $y\mapsto p_Y(y,t)$ of random variable $Y(t)$, Fig.~\ref{figureDO4} shows, for the two instants $t=0.2977\, s$ and $t=0.5535\, s$, the reference constructed with $N_d=1\, 000$  and, for $N_d=80$ points and $n_\pMC=1\,000$ additional realizations, the estimation constructed with the PLoM without constraint and the one with the PLoM under constraints using $\algo~3$ for which the minimum is obtained for
$i_\opt=8$. This figure already shows that the PLoM without constraint gives good results but that the PLoM under constraints
significantly improves the result. In particular, it can be seen in figures (c) and (d), which are in a logarithmic scale for the ordinates, that the PLoM under constraints makes it possible to correctly predict the tails of the probability density function of the reference.
%
\section{Application 2: unsteady 2D Navier-Stokes equation for incompressible fluid}
\label{SectionNS}
%
\subsection{Boundary value problem}
\label{SectionNS1}
We consider a dimensionless 2D formulation in  a Cartesian coordinate system $(O, x_1, x_2)$, whose generic  point is $\bfx = (x_1,x_2)$ . The  2D open domain containing the fluid is $\Omega = ]0,\ell_{x_1}[\times ]0,\ell_{x_2}[$ with $\ell_{x_1}=0.9$ and $\ell_{x_2} = 1$. The boundary is written as
$\partial\Omega = \Gamma_0 \cup \Gamma_u$ in which $\Gamma_u = \big\{ x_1\in [0,\ell_{x_1}],x_2=\ell_{x_2} \big\}$ and where $\Gamma_0 = \partial\Omega \backslash \Gamma_u$.
Let $\bfu(\bfx,t) = (u_1(\bfx,t),u_2(\bfx,t))$ be the dimensionless velocity field and $p(\bfx,t)$ be the dimensionless pressure field, in which $t$ is the dimensionless time. These  quantities are related to the non dimensionless quantities (denoted with a tilde) as follows:
$\bfx = \widetilde\bfx / \widetilde\ell_{\pref}$, $t = \widetilde t \, \widetilde V_{\pref} / \widetilde\ell_{\pref}$ in which $\widetilde\ell_{\pref}$ is a given reference length and where $\widetilde V_{\pref}$ is the reference velocity in $m/s$ (independent of $\bfx$ and $t$), $\bfu = \widetilde\bfu / \widetilde V_{\pref}$, and $p = \widetilde p / (\widetilde \rho_{\pref} \widetilde V_{\pref}^2)$ with $\widetilde \rho_{\pref}$ the mass density in $kg/m^3$ (independent of $\bfx$ and $t$). Let $\curR e = \widetilde V_{\pref}\widetilde\ell_{\pref}/\nu_\pCFD$ be the Reynolds number in which $\nu_\pCFD$ is the kinematic viscosity such that $\nu_\pCFD = \widetilde\eta_{\pref}/\widetilde\rho_{\pref}$ where
$\widetilde \eta_{\pref}$ is the dynamic viscosity in $Pa\times s$. It is assumed that $\nu_\pCFD$ is spatially uniform (independent of $\bfx$ and $t$) and consequently, the energy equation is not required. The boundary conditions on the non dimensionless boundary $\partial\widetilde\Omega = \widetilde\Gamma_0 \cup\widetilde\Gamma_u$ are written, for all $t$ in $[t_0,T]$, as
$\widetilde\bfu(\cdot , t) = (0,0)$ on $\widetilde\Gamma_0$ and $\widetilde\bfu(\cdot,t) = (\widetilde V_\pCFD\, , 0 )$ on $\widetilde\Gamma_u$ where
$\widetilde V_\pCFD$ in $m/s$ is written as $\widetilde V_\pCFD = V_\pCFD\times\widetilde V_{\pref}$ in which $V_\pCFD$ is the given dimensionless velocity  that is assumed independent of $\bfx$ and $t$.
The dimensionless boundary value problem for the unsteady incompressible flow in $\Omega$ is defined as follows. For all $t\in ]t_0,T]$ and for all $\bfx \in \Omega$,  we have the following dimensionless 2D Navier-Stokes equations for an incompressible flow,
\begin{align}
& \frac{\partial\bfu}{\partial t}  +\bfu\cdot\bfnabla\bfu +\bfnabla p - \frac{1}{\curR e} \bfnabla^2\bfu =\bfzero \, , \label{eqNS1.1}\\
& \bfnabla\cdot \bfu  = 0 \, ,                                                                                         \label{eqNS1.2}
\end{align}
with the boundary conditions that are written for all $t\in [t_0,T]$,
\begin{equation}
\bfu(\bfx , t) = (0,0) \, ,\, \bfx\in\Gamma_0 \quad , \quad \bfu(\bfx,t) = (V_\pCFD, 0 )\, , \, \bfx\in \Gamma_u \, , \label{eqNS1.2b}
\end{equation}
and  with the following initial condition at time $t_0$, $\bfu(\bfx , t_0) = (0,0)$ for all $\bfx\in\Omega$. We choose
$\widetilde\ell_{\pref} = 0.9~m$ and $\widetilde V_\pref = \widetilde v\, V_\pCFD$ with $\widetilde v = 1~m/s$. Consequently, the Reynolds can be rewritten as $\curR e = 0.9 V_\pCFD / \nu_\pCFD$.
\subsection{Prior probability model of uncertainties}
\label{SectionNS2}
Parametric probabilistic uncertainties and their quantification in computational fluid  dynamics have extensively been studied (see for instance \cite{LeMaitre2010,Bijl2013}).
The present application is developed to validate the PLoM under constraints. We must therefore introduce uncertainties and we therefore choose a parametric probabilistic approach. The uncertain control parameters are $V_\pCFD$ and $\nu_\pCFD$. They are modeled by two independent positive-valued random variables, defined on the probability space $(\Theta,\curT,\curP)$, with values in $[ V_\pmin , V_\pmax]$ and $[\nu_\pmin , \nu_\pmax]$, written as
$V_\pCFD = V_\pmin(1+\varepsilon_V\,\curU_1)$ and $\nu_\pCFD = \nu_\pmin(1+\varepsilon_\nu\,\curU_2)$,
in which $\varepsilon_V = V_\pmax /V_\pmin -1$ and $\varepsilon_\nu = \nu_\pmax / \nu_\pmin -1$, and where $\curU_1$ and $\curU_2$
are independent uniform random variables on $[0,1]$. The considered numerical values are $V_\pmin =0.1\, m/s$, $V_\pmax =0.3\, m/s$,
$\nu_\pmin = 10^{-5}$, and $\nu_\pmax = 3\times 10^{-5}$.

Consequently, the support of the probability distribution of the random Reynolds number $\curR e$
is the interval $[3\, 000 , 27\, 000]$. The uniform random variables are classically written as
$\curU_j  = \big (1+\hbox{erf}(W_j/\sqrt{2})\big )/2$, in which $W_1$ and $W_2$ are independent normalized Gaussian random variables.
Let $\bfW = (W_1,W_2)$ be the $\RR^{n_w}$- valued random variable with $n_w=2$. For $j=1,2$, let $\big\{ \bfw_{d,j}^\ell ,\ell=1,\ldots,N_d \big\}$ be $N_d$ realizations
generated with the prior probability model for constituting the training dataset.
The random Reynolds number $\curR e$ is defined by a given real-valued mapping $f_{\curR e}$  on $\RR^2$ such that
\begin{equation}
 \curR e =  f_{\curR e}(\bfW) \, .                                                                                        \nonumber 
\end{equation}
\subsection{Computational model, KL expansion, random residual, and PLoM parameters}
\label{SectionNS3}
%

\noindent \textit{(i) Time discretization scheme}. We use the time discretization scheme proposed in \cite{Griebel1998}.
We choose $t_0 = 0$. Let $\Delta t$ be the  time step, $\big\{ t_n = n\, \Delta t ,\, n= 1,\ldots n_\ptime \big\}$ be the  time sampling, and $T =n_\ptime \, \Delta t$ be the  final time.
Let us assumed that the velocity field $\bfx\mapsto \bfu^{(n)}(\bfx) = \bfu(\bfx,t_n)$ and the pressure field $\bfx\mapsto p^{(n)}(\bfx) = p(\bfx,t_n)$ are known at the $n$-th sampling time  and satisfy Eqs.~\eqref{eqNS1.1} and \eqref{eqNS1.2}. The following scheme allows for calculating the approximation of the solution at time $t_{n+1}$ (below, the $\bfx$ dependence is removed). In Eq.~\eqref{eqNS1.1}, the partial time derivative of $\bfu$ at time $t_{n+1}$, which is discretized as $(\bfu^{(n+1)} - \bfu^{(n)})/\Delta t$, is written as the sum of three terms: the nonlinear force that is explicitly treated, the viscosity force that is implicitly treated, and the pressure correction term that is implicitly treated for enforcing incompressibility defined by to Eq.~\eqref{eqNS1.2},
\begin{equation}
 \frac{\bfu^\pNL \! - \bfu^{(n)} }{\Delta t }  =  - (\bfu\cdot \bfnabla \bfu)^{(n)}  \quad , \quad
 \frac{\bfu^{\rm v} \! - \bfu^\pNL } {\Delta t }  =  \frac{1}{\curR e} (-\bfnabla^2 \bfu^{\rm v})  \quad , \quad
 \frac{ \bfu^{(n+1)} \! - \bfu^{\rm v} } {\Delta t }  =  -\bfnabla p^{(n+1)}  \quad , \quad
 \bfnabla\cdot \bfu^{(n+1)}   =  0\, .                                                                                                     \label{eqNS3}
\end{equation}
The implicit equation for the calculation of $p^{(n+1)}$ is  obtained by substituting $\bfu^{(n+1)}$ given by the third Eq.~\eqref{eqNS3} into the fourth and gives the elliptic equation,
\begin{equation}
  -\bfnabla^2(\Delta t \, p^{(n+1)}) = -\bfnabla\cdot \bfu^{\rm v} \, .                                                                      \label{eqNS3.5}
\end{equation}

\noindent \textit{(ii) Spatial discretization}. The spatial discretization, the discretization of the  spatial derivatives, and the implementation of the boundary conditions, are performed as explained in \cite{Griebel1998}. A  grid is defined using $n_{x_1}$ grid points in the $x_1$-axis with a spatial step
$h_{x_1} = \ell_{x_1} /n_{x_1}$  and $n_{x_2}$ grid points in the $x_2$-axis with $h_{x_2} = \ell_{x_2} /n_{x_2}$. There are $n_{x_1} \times n_{x_2}$ cells.
The spatial discretization is performed on a staggered grid for which $u_1$ is placed on the $x_2$ (vertical) cell interfaces, $u_2$ is placed on the $x_1$ (horizontal) cell interfaces, and $p$ is placed in the cell midpoints. We then have,
\begin{itemize}

\item[$\triangleright$] for the interior points,  $N_{u_1} = (n_{x_1}-1) \times n_{x_2}$ points for velocity component $u_1$,
$N_{u_2} = n_{x_1} \times (n_{x_2}-1)$ points for velocity component $u_2$,
and $N_p = n_{x_1} \times n_{x_2}$ points for pressure $p$.
\item[$\triangleright$] for the total number of points allowing the boundary conditions to be implemented,
$(n_{x_1}+1) \times (n_{x_2}+2)$ points for velocity component $u_1$,
$(n_{x_1} +2)\times (n_{x_2}+1)$ points for velocity component $u_2$,
and $(n_{x_1}+2) \times (n_{x_2}+2)$ points for pressure $p$.
\end{itemize}
This staggered grid allows the boundary conditions to be easily implemented. Let be $N_u= N_{u_1} + N_{u_2}$ and $N=N_u+N_p$. At time $t$, the spatial discretization of random fields $\bfu(\cdot,t) = \big ( u_1(\cdot,t),u_2(\cdot,t) \big )$ and $p(\cdot,t)$ are the $\RR^{N_u}$- and $\RR^{N_{p}}$-valued random variables $\bfU(t)$ and $\bfP(t)$, and we define the $\RR^N$-valued random variable $\bfY(t)$ by
\begin{equation}
\bfY(t) = \big ( \bfU(t),\bfP(t) \big ) \, .                                                                                                                       \label{eqNS3.6}
\end{equation}
%

\noindent \textit{(iii) Computational model}. The time discretization scheme and the spatial discretization  allow for computing the time-spatial-discre\-ti\-za\-tion $\bfY(t_n) = \big(\bfU(t_n)$, $\bfP(t_n)\big )$, and the time derivative $\dot\bfU(t_n)$, of the solution of the boundary value problem, at time $t_n$, for all $n$ in $\big\{1,\ldots , n_\ptime\big\}$. In that condition,  Eq.~\eqref{eq2.1} is rewritten, at sampling time $t_n$,  as
\begin{align}
 & \bfcurN^{\,\pNS}\big (\dot\bfU(t_n),\bfU(t_n),\bfP(t_n),\bfW \big) = \bfzero_{N_u} \quad a.s. \, ,                                               \label{eqNS3.7}\\
 & \bfcurN^{\,\pDIV}\big (\bfU(t_n)\big ) = \bfzero_{N_p} \quad a.s.\, ,                                                                             \label{eqNS3.8}
\end{align}
in which  $\bfcurN^{\,\pNS}$ is a nonlinear operator  and where $\bfcurN^{\,\pDIV}$ is a linear operator.\\

\noindent \textit{(iv) Values of the parameters for the Navier-Stokes solver}. The staggered grid is such that
$n_{x_1}= 100$ and $n_{x_2}=111$. We then have $N_{u_1} = 10\ 989$, $N_{u_2} = 11\,000$, $N_u = 21\,989$, $N_p = 11\,100$, and $N   = 33\,089$.
The time step has been chosen to $\Delta t = 0.03$ and leads to a convergent solution up to the steady (stationary) stochastic flow with a $L^2$-norm of the residual (level of approximation) that is acceptable (see Section~\ref{SectionNS4}). Clearly, this approximation could be improved in decreasing the value of $\Delta t$ but there is no point in unnecessarily increasing the computational costs, because that does not change anything on the validation of the PLoM under constraints.
(i) For computation of the steady (time-stationary) stochastic flow (see Section~\ref{SectionNS4}-(i)), $T = 300$ and $n_\ptime = 10\, 000$.
For computing the random residual, $n_\psp = 500$.
(ii) For computation of the reference unsteady (time-nonstationary) stochastic flow (see Section~\ref{SectionNS4}-(ii)),
$t =30$ is chosen for analyzing the performance of the proposed PLoM method under constraints, which is an instant corresponding to the unsteady flow regime. We then have $T=30$ and $n_\ptime = 1\,000$. \\

\begin{figure}[h!]
  \centering
  \includegraphics[width=4.5cm]{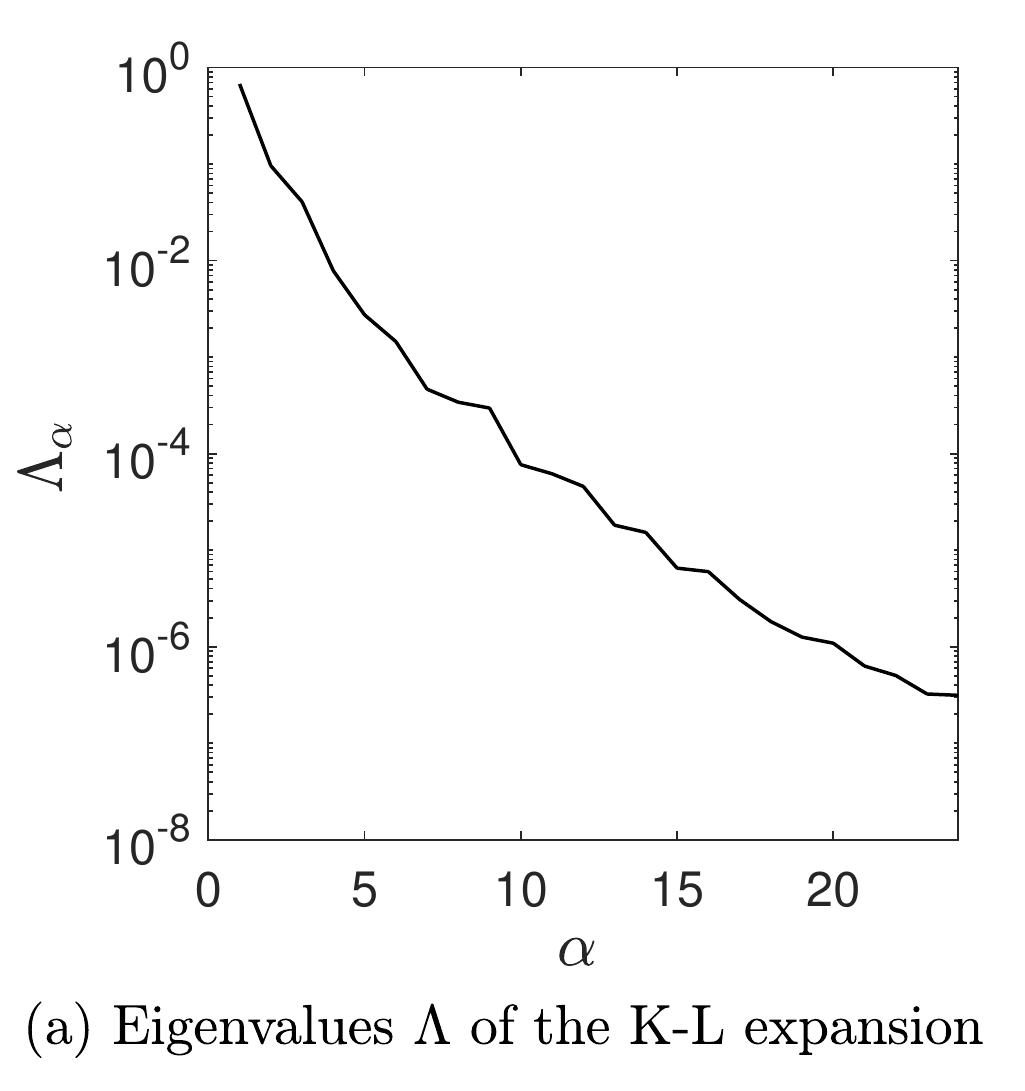}  \includegraphics[width=4.5cm]{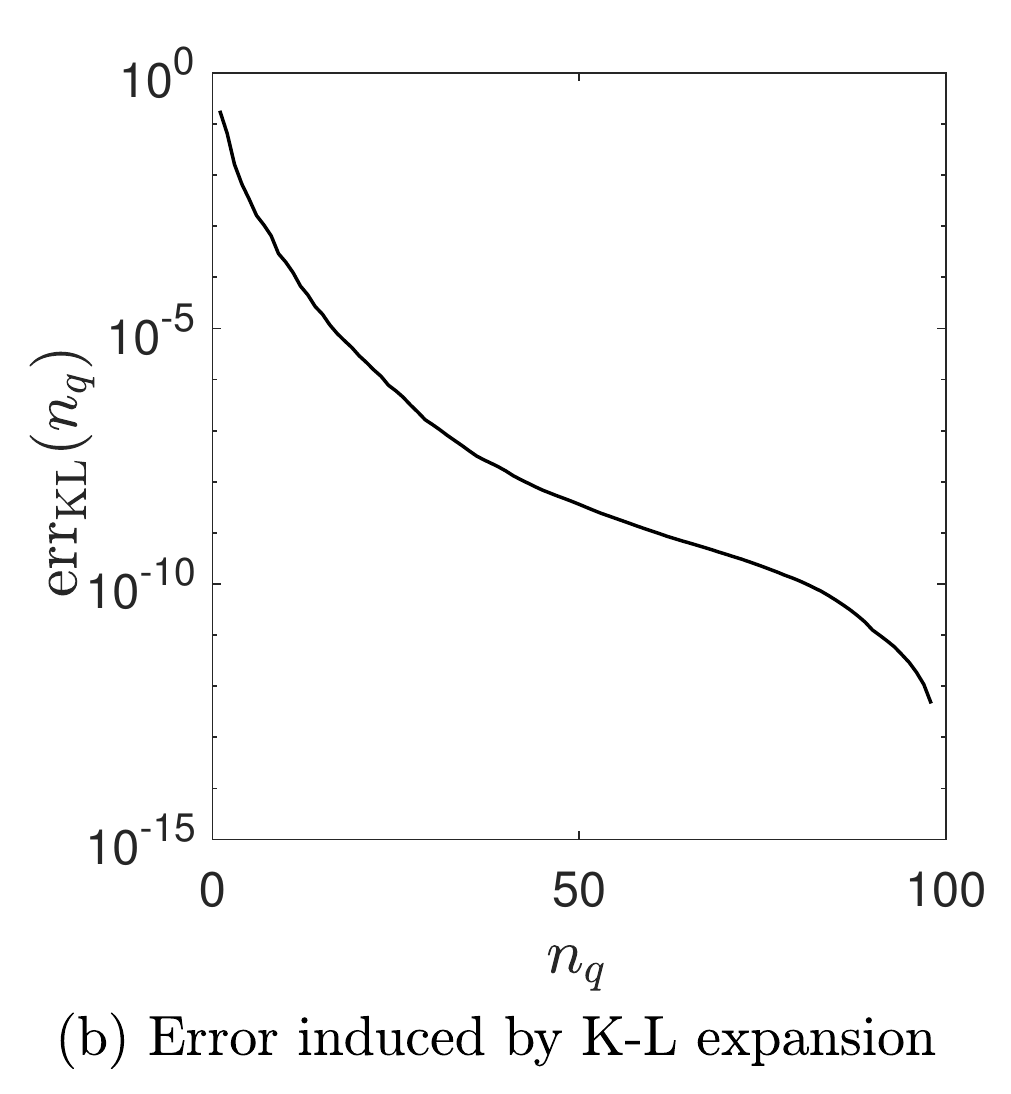} \includegraphics[width=4.5cm]{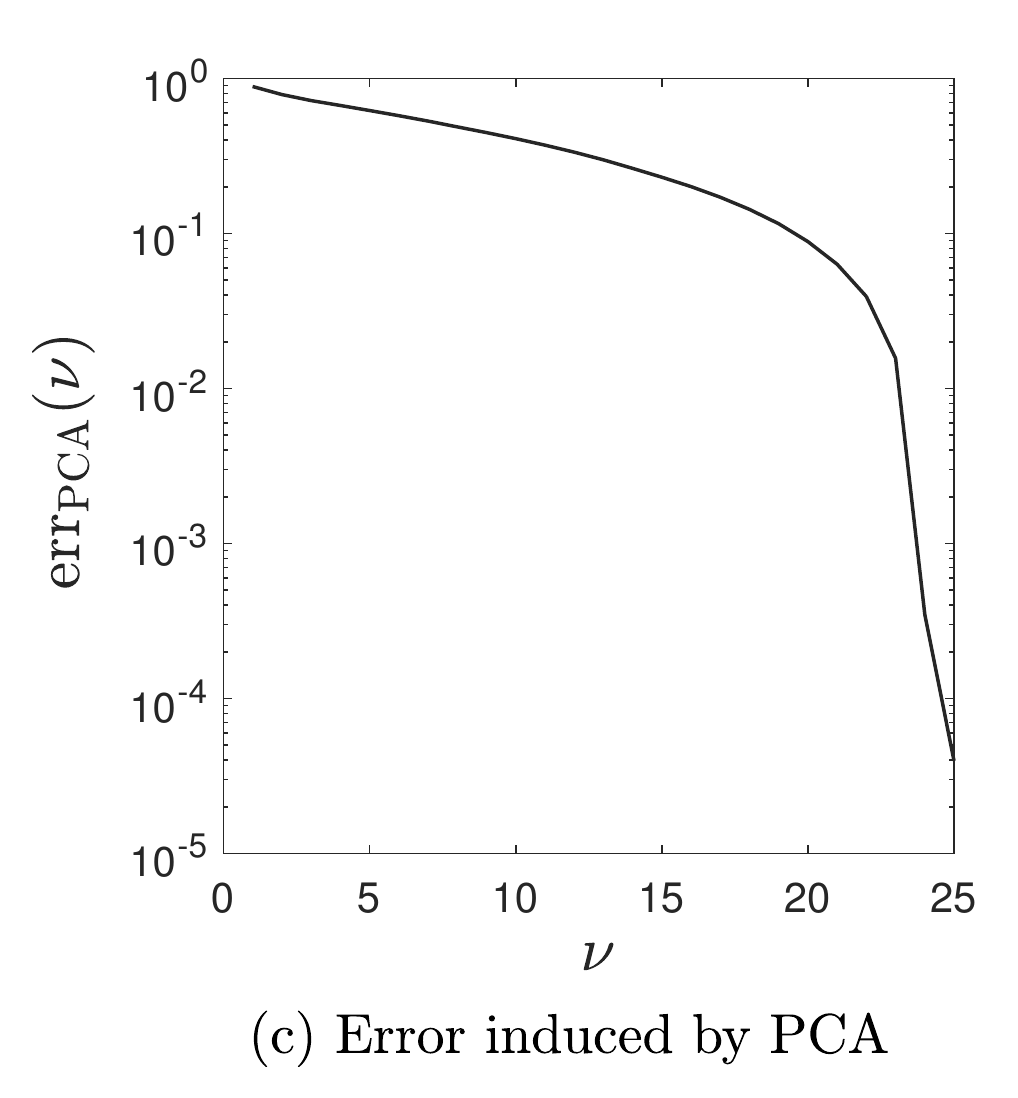}
\caption{For $N_d = 100$, $T= 30$, and $n_\ptime=1\,000$, KL expansion of nonstationary stochastic process $\bfY$:  (a) distribution of the eigenvalues $\alpha\mapsto \Lambda_\alpha$ and (b) error function $n_q\mapsto \errp_\KL(n_q)$. PCA of random vector $\bfX$: (c) error function $\nu\mapsto \errp_\PCA(\nu)$. }
\label{figureNS0}
\end{figure}

\noindent \textit{ (v)  Reduced-order representations (KL expansions) of the nonstationary stochastic processes} $\bfY=(\bfU,\bfP)$ \textit{and} $\dot\bfU$.
The KL expansion of the nonstationary $\RR^N$-valued stochastic process
$\big\{\bfY(t), t\in[t_0, T]\big\}$ is performed at sampling times $t_1,\ldots, t_{n_\pptime}$, yielding the representation $\bfY(t_n) = \underline\bfy(t_n) + [V(t_n)] \,\bfQ$. Since $\bfY=(\bfU,\bfP)$, we have
\begin{align}
& \bfU(t_n) = \underline\bfu(t_n) + [V_u(t_n)] \,\bfQ \, ,                                                         \label{eqNS3.9}\\
& \bfP(t_n) = \underline\bfp(t_n) + [V_p(t_n)] \,\bfQ \,  ,                                                       \label{eqNS3.10}
\end{align}
in which $\underline\bfu(t_n)\in \RR^{N_u}$, $[V_u(t_n)]\in\MM_{N_u,n_q}$, $\underline\bfp(t_n)\in \RR^{N_p}$, and $[V_p(t_n)]\in\MM_{N_p,n_q}$.
The time derivative of $\bfU(t)$ at $t_n$ is calculated by $\dot\bfU(t_n) = (\bfU(t_n) - \bfU(t_{n-1}))/\Delta t$, which allows for deducing $\underline{\dot\bfu}(t_n)$ and $[\dot V_u(t_n)]$. The KL expansion of $\dot\bfU$ can then be written as,
\begin{equation}
\dot\bfU(t_n) = \underline{\dot\bfu}(t_n) + [\dot V_u(t_n)] \,\bfQ \, .                                                             \label{eqNS3.11}\\
\end{equation}
For $N_d = 100$ and $T= 30$, Figs.~\ref{figureNS0}-(a) and -(b) show the distribution of the eigenvalues $\alpha\mapsto \Lambda_\alpha$ of the KL-expansion of stochastic process $\big\{\bfY(t)=\big (\bfU(t),\bfP(t)\big), t\in [t_0,T]\big\}$ and the error function
$n_q\mapsto \err_\KL(n_q)$ defined by Eq.~\eqref{eq3.7}.  Fixing the tolerance $\varepsilon_\KL$ to the value $10^{-6}$ yields the optimal value $n_q=24$.\\

\noindent \textit{(vi) Calculation of the random residual} $\widehat\rho$ \textit{using the subsampling times}.
The subsampling times $\big\{\tau_{n_s}, n_s = 1,\ldots , n_\psp\big\}$ (see Section~\ref{Section4.2}) is built  with a constant step $\Delta\tau = (T-t_0)/n_\psp$ and
$n_\psp = 50$. The family $\big \{ \dot\bfU(\tau_{n_s}),$ $\bfU(\tau_{n_s}),$ $\bfP(\tau_{n_s}),$  $n_s=1,\ldots ,n_\psp \big\}$ is extracted from the family $\big \{\dot\bfU(t_n),$ $\bfU(t_n),$ $\bfP(t_n),$  $n=1,\ldots ,n_\ptime \big\}$ computed with Eqs.~\eqref{eqNS3.9} to \eqref{eqNS3.11}.
Taking into account Eqs.~\eqref{eqNS3.7} and \eqref{eqNS3.8} at subsampling time $\tau_{n_s}$, the $\RR^N$-valued random residual $\bfcurR(\tau_{n_s})$,
defined by Eq.~\eqref{eq4.2}, is rewritten as $\bfcurR(\tau_{n_s}) = \big (\bfcurR^{\,\pNS}(\tau_{n_s}),\bfcurR^{\,\pDIV}(\tau_{n_s})\big)$ in which
the $\RR^{N_u}$-valued random variable $\bfcurR^{\,\pNS}(\tau_{n_s})$  and the $\RR^{N_p}$-valued random variable $\bfcurR^{\,\pDIV}(\tau_{n_s})$ are written, for all
$n_s\in\big\{1,\ldots , n_\psp\big\}$, as
\begin{align}
\bfcurR^{\,\pNS}(\tau_{n_s})  = & \bfcurN^{\,\pNS}(\dot\bfU(\tau_{n_s}),\bfU(\tau_{n_s}),\bfP(\tau_{n_s}),\bfW) \quad a.s. \, ,                    \label{eqNS3.12}\\
\bfcurR^{\,\pDIV}(\tau_{n_s}) = & \bfcurN^{\,\pDIV}(\bfU(\tau_{n_s})) \quad a.s.\, .                                                                \label{eqNS3.13}
\end{align}
Let $\bfcurR^{\,\pNS}_{u_1}(\tau_{n_s})$ and $\bfcurR^{\,\pNS}_{u_2}(\tau_{n_s})$ with values in $\RR^{N_{u_1}}$ and $\RR^{N_{u_2}}$ be the block decomposition of
$\bfcurR^{\,\pNS}(\tau_{n_s})$ associated with the components relative to the $x_1$- and $x_2$-axes.
We then have  $\bfcurR^{\,\pNS}(\tau_{n_s}) = \big (\bfcurR^{\,\pNS}_{u_1}(\tau_{n_s}), \bfcurR^{\,\pNS}_{u_2}(\tau_{n_s})\big)$ with values in $\RR^{N_u} = \RR^{N_{u_1}} \times \RR^{N_{u_2}}$.
The positive-valued random variable $\curR_\pnorm(\tau_{n_s})$ is defined by Eq.~\eqref{eq4.4}, but  the Euclidean norm is adapted to the structure of random vector $\bfcurR(\tau_{n_s})$ as
\begin{equation}
\curR_\pnorm(\tau_{n_s}) = \big \{ \,\Vert \,\bfcurR^{\,\pNS}_{u_1}(\tau_{n_s})\, \Vert^2 \, / \, N_{u_1}^2
                                                  + \Vert\, \bfcurR^{\,\pNS}_{u_2}(\tau_{n_s})\, \Vert^2 \, / \,N_{u_2}^2
                                                + \Vert\, \bfcurR^{\,\pDIV}(\tau_{n_s})\,\Vert^2\, /\, N_p^2 \,\big\}^{1/2} \, .                \label{eqNS3.14}
\end{equation}
The random residual $\widehat\rho$ is then calculated by using  Eq.~\eqref{eq4.10}.  The mean value $\underline{\widehat\rho}_\pref$ of the reference random residual $\widehat\rho_\pref$ is calculated for the fixed value $N_{d,\pref}= 100 $ of the number $N_d$ of training points in the training dataset.\\

\noindent \textit{(vii) Values of the parameters for the PLoM without and under constraints}. We consider the case $N_d = 100$ and $T= 30$ for which the KL expansions of stochastic processes $\dot\bfU$, $\bfU$, and $\bfP$  have been performed. Random vector $\bfX = (\bfQ,\bfW)$ has dimension $n_x = n_q + n_w = 26$.
Concerning the  PCA of $\bfX$, Figure~\ref{figureNS0}-(c) shows the error function $\nu\mapsto \err_\PCA(\nu)$ defined by Eq.~\eqref{eq5.3}. Fixing the tolerance $\varepsilon_\PCA$ to the value $10^{-6}$ yields the optimal value $\nu=26$.
For the PLoM without constraint or under constraints, the diffusion-maps basis is computed with $\varepsilon_\pdiff = 95$, which yields the optimal dimension $m = 27$.
Concerning the St\"{o}rmer-Verlet algorithm for solving the nonlinear ISDE: $f_0 =4$ and $\Delta r = 0.1974$.
The convergence analysis with respect to $n_\pMC$ is performed by considering the following values $100$, $400$, 800$, 1\,200$, $1\,600$, $2\,000$, and $4\,000$ of $n_\pMC$.
\subsection{Steady and unsteady stochastic flow}
\label{SectionNS4}
\noindent {\textit{(i)- Steady (time-stationary) stochastic flow}. For all the possible values of random control parameters $\nu_\pCFD\in [1\, , 3]\times 10^{-5} $ and $V_\pref\in [0.1\, , 0.3]\, m/s$ of the prior probability model, the steady flow (time-stationary regime of the stochastic flow) has been  obtained at $t = 300$ (almost-sure stochastic solution of the Navier-Stokes equations). For $t\ge 300$,  the stochastic solution is stationary in time. The mean value $m_P(\bfx,t)$ and the standard deviation $\sigma_P(\bfx,t)$ of the pressure random field $\big\{ P(\bfx,t),\bfx\in\Omega,t\ge 300 \big\}$  are independent of $t$.
For $N_d = 50$ realizations of the training dataset, Fig.~\ref{figureNS1} shows the estimates of the mean value and standard deviation  of the pressure random field at $t= 300$.\\
\begin{figure}[h!]
  \centering
  \includegraphics[width=4.5cm]{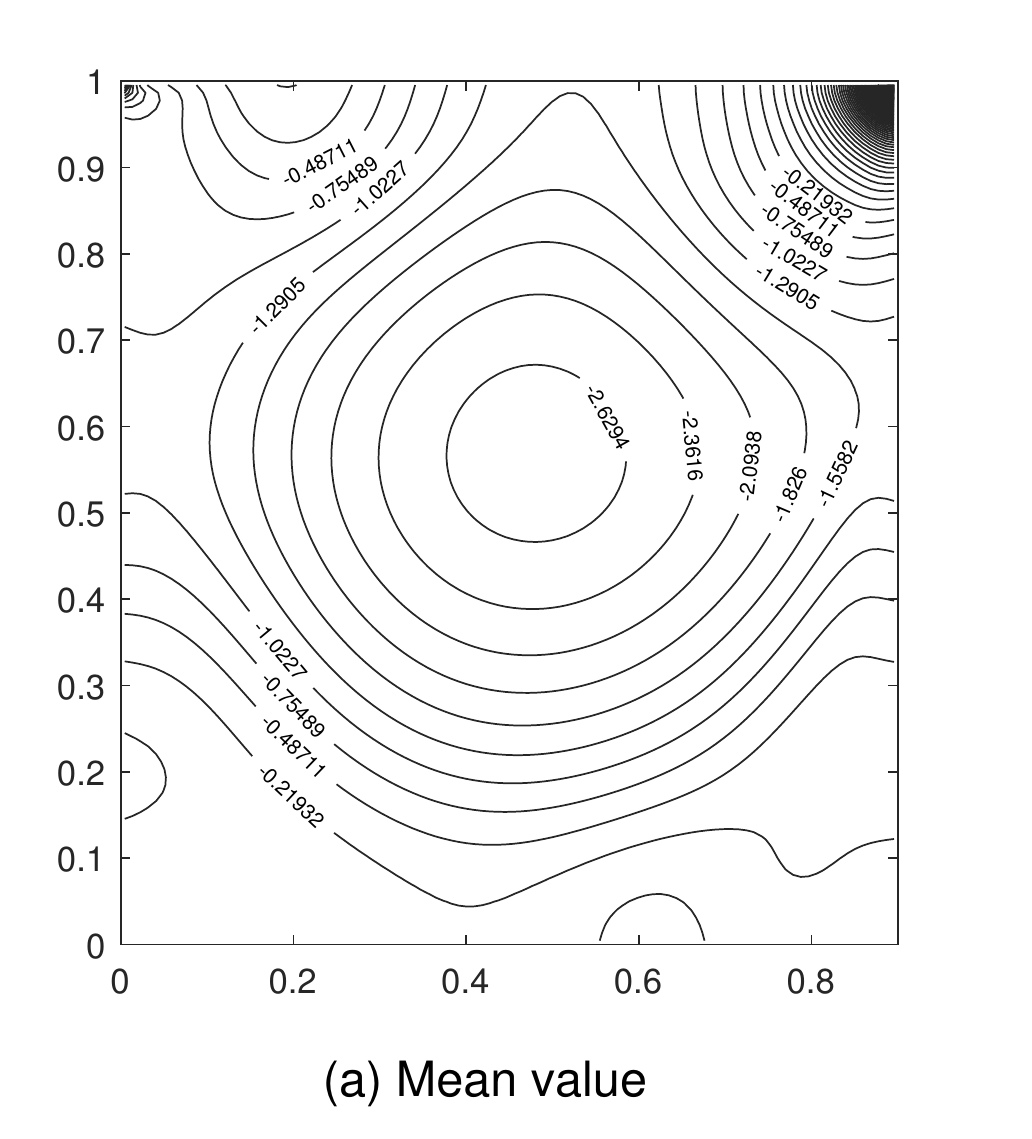}  \includegraphics[width=4.5cm]{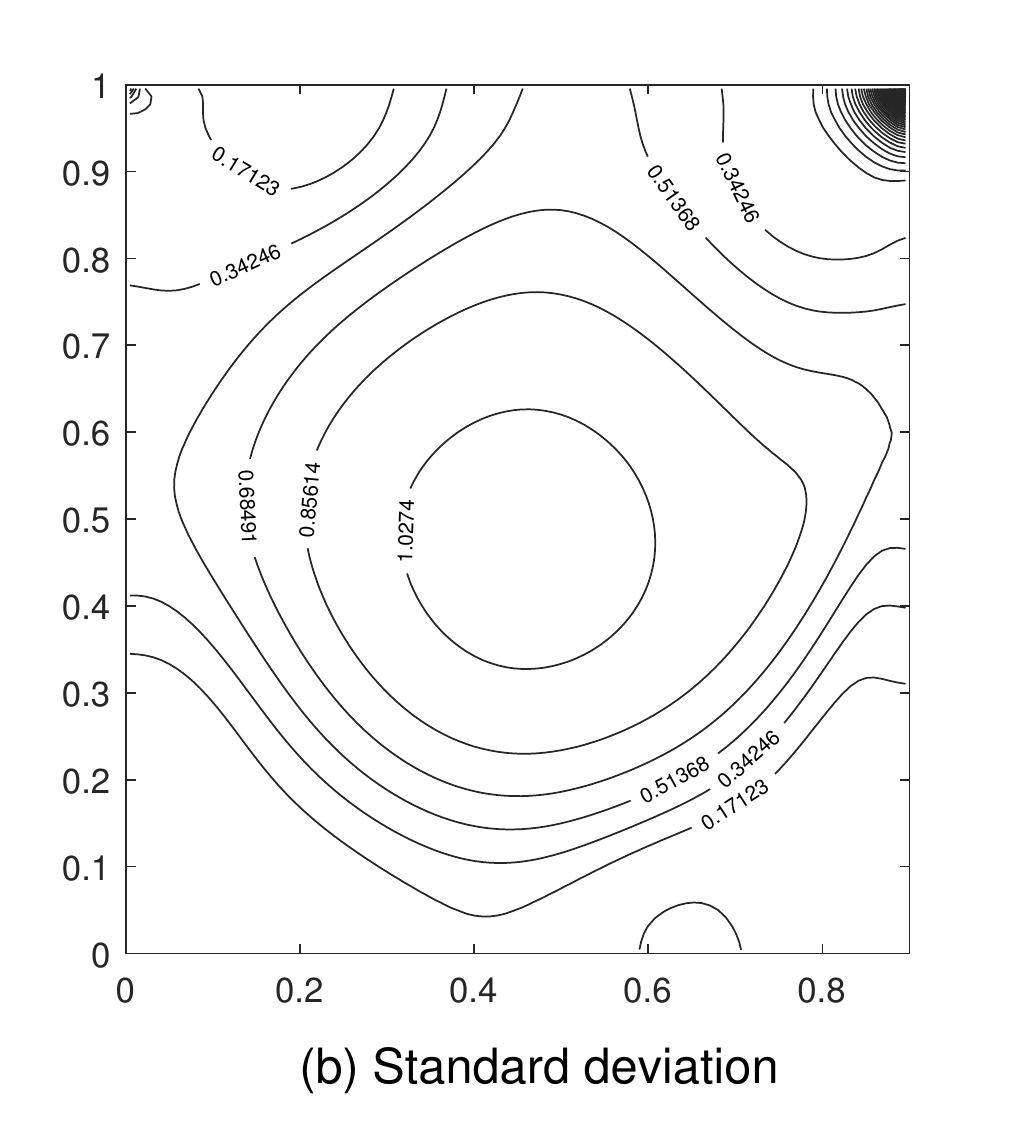}
\caption{Mean value $\bfx\mapsto 1\, 000 \times  m_P(\bfx,t)$ (a) and standard deviation $\bfx\mapsto 1\, 000 \times \sigma_P(\bfx,t)$ (b) estimated with $N_d = 50$ realizations of the training dataset, at $t=300$ of the time-stationary pressure random field
$\{ P(\bfx,t),\bfx\in\Omega,t\ge 300 \}$.}
\label{figureNS1}
\end{figure}

\noindent {\textit{(ii)- Reference unsteady (time-nonstationary) stochastic flow}. Instant $t =30$ is selected for analyzing the performance of the proposed method  in the unsteady flow regime. A reference unsteady solution has been  computed with $N_d=4\,000$  realizations from the training dataset requiring a significant CPU effort. This reference solution is used for analyzing all the convergence aspects of the PLoM without  and under constraints. Figures~\ref{figureNS2}-(a) and (b) show the estimates of the mean value  and standard deviation  of the pressure random field at $t= 30$.
The estimate of  $E\big\{\widehat\rho\big\}$  is $1.53 \times 10^{-3}$ while the estimate of
$\Vert \, \widehat\rho\,\Vert_{L^2}$ is $1.64 \times 10^{-3}$.
\begin{figure}[h!]
  \centering
  \includegraphics[width=3.95cm]{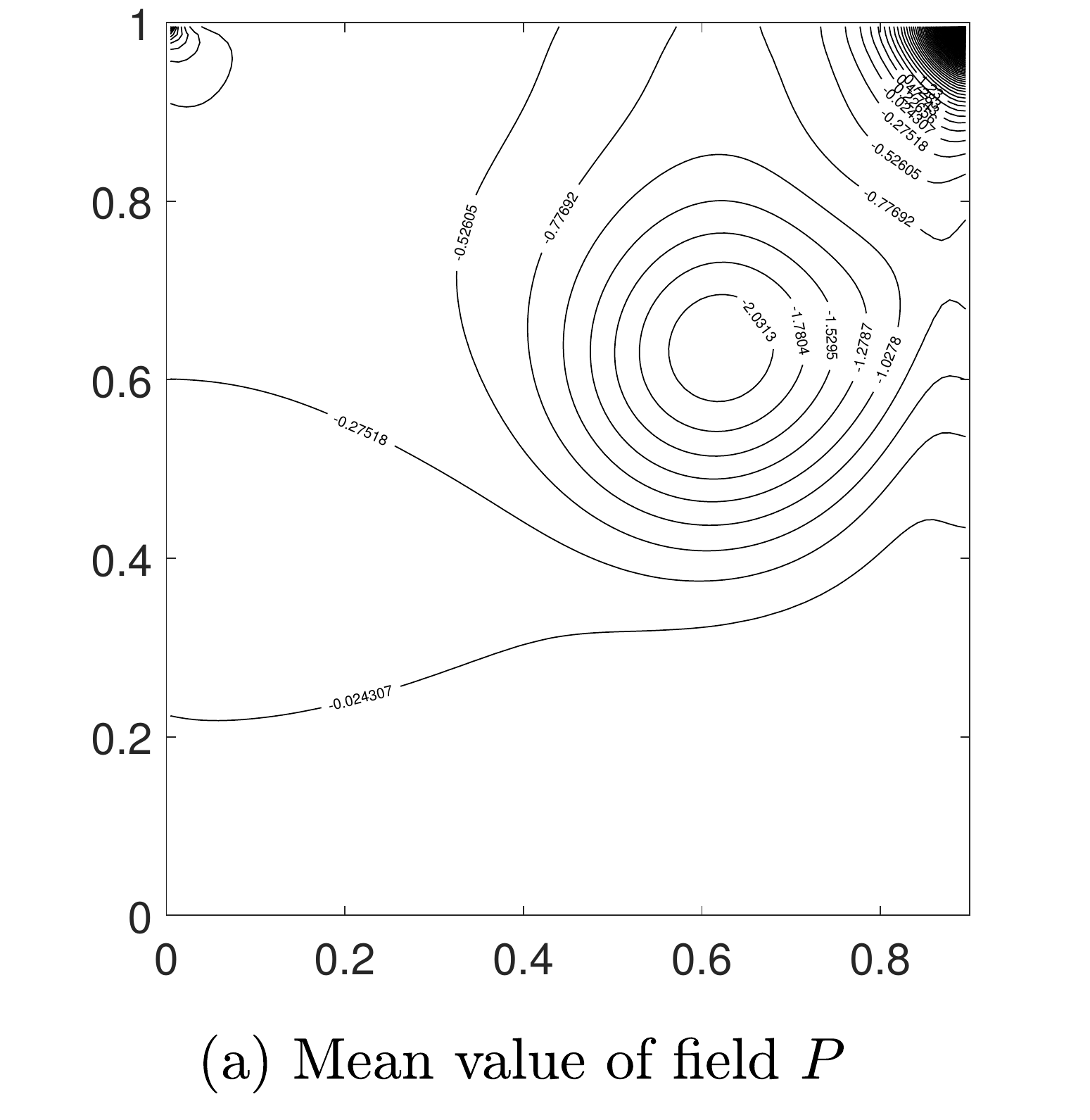}  \includegraphics[width=3.95cm]{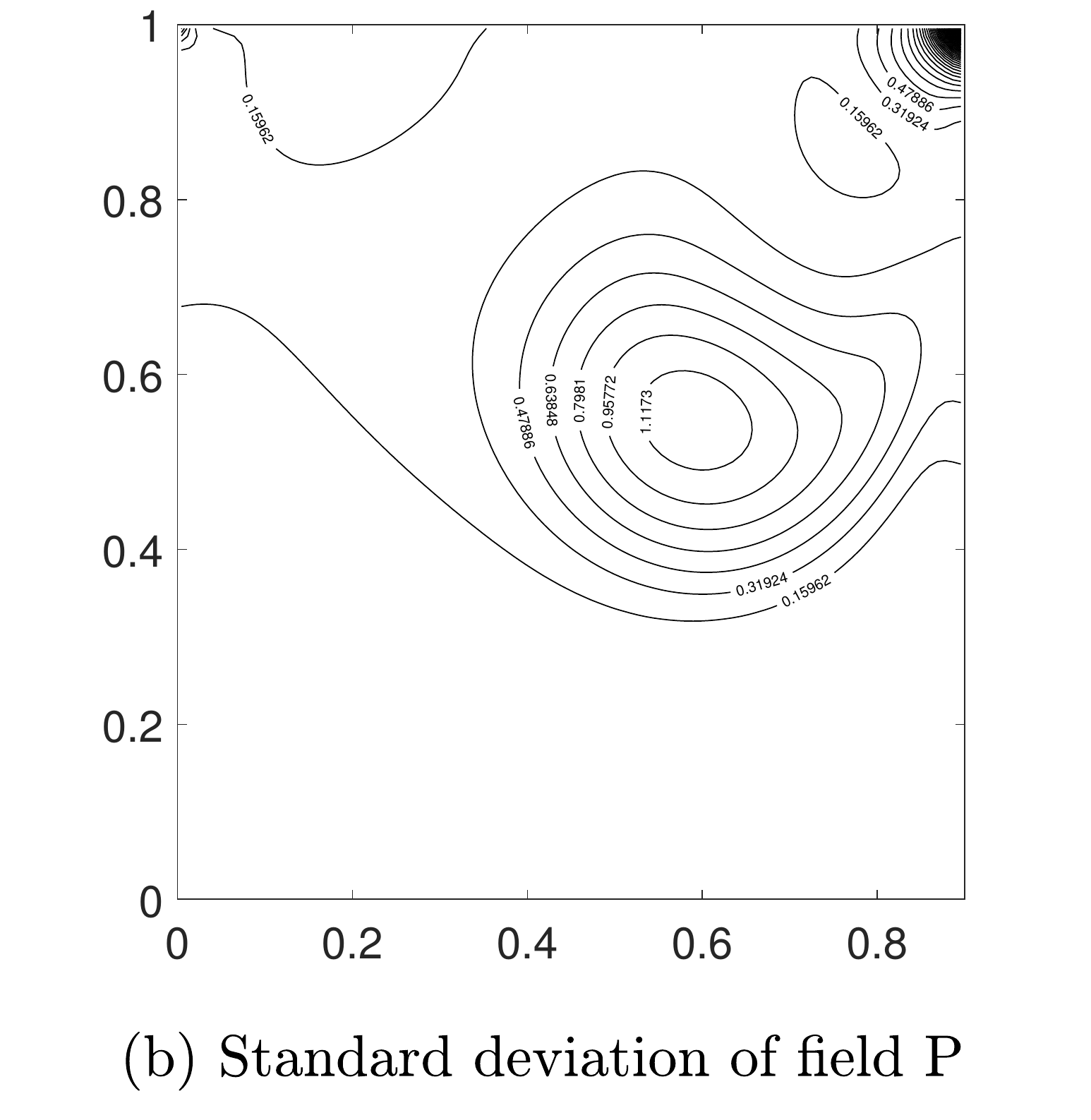}
  \includegraphics[width=4.1cm]{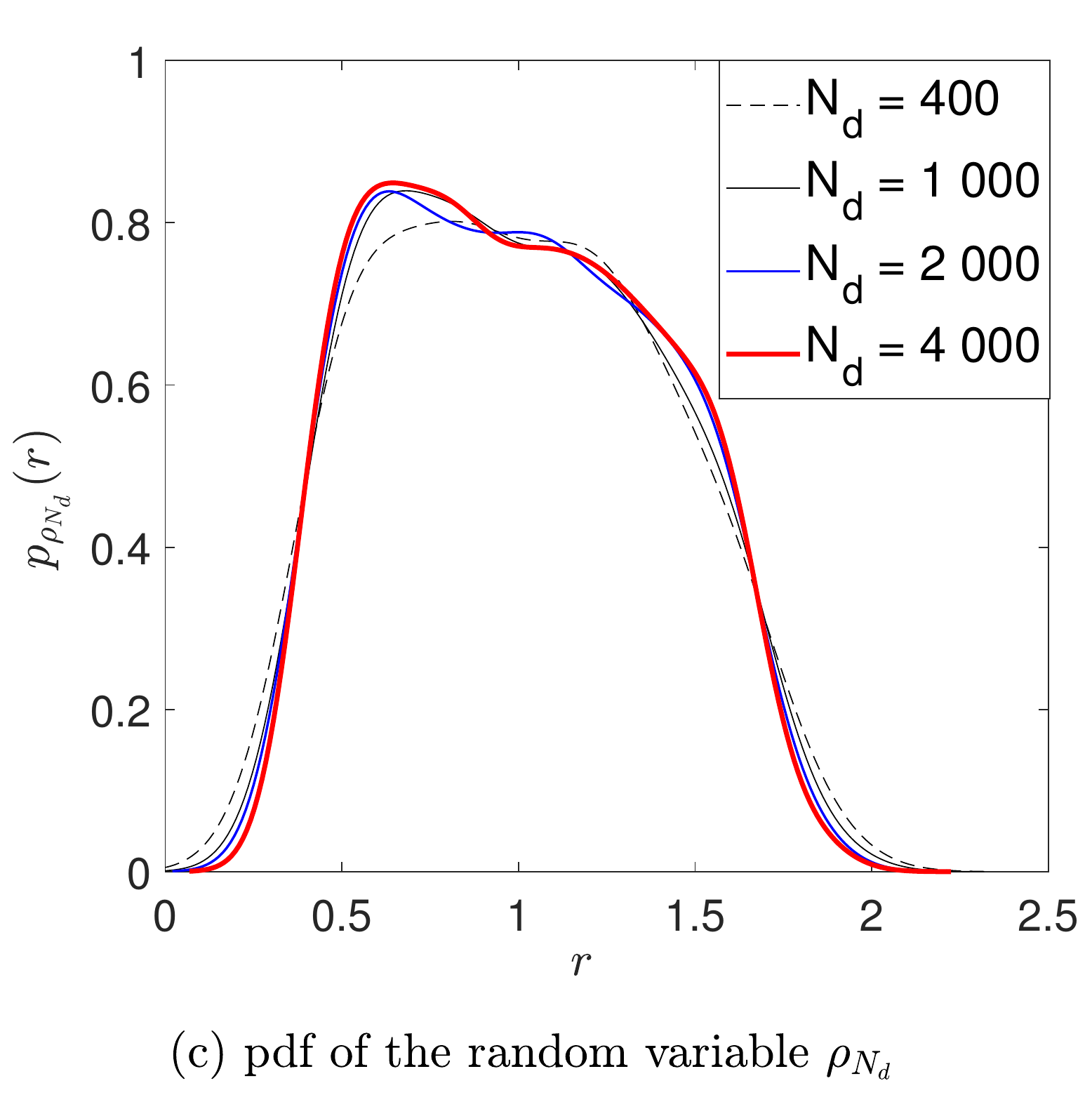}  \includegraphics[width=4.1cm]{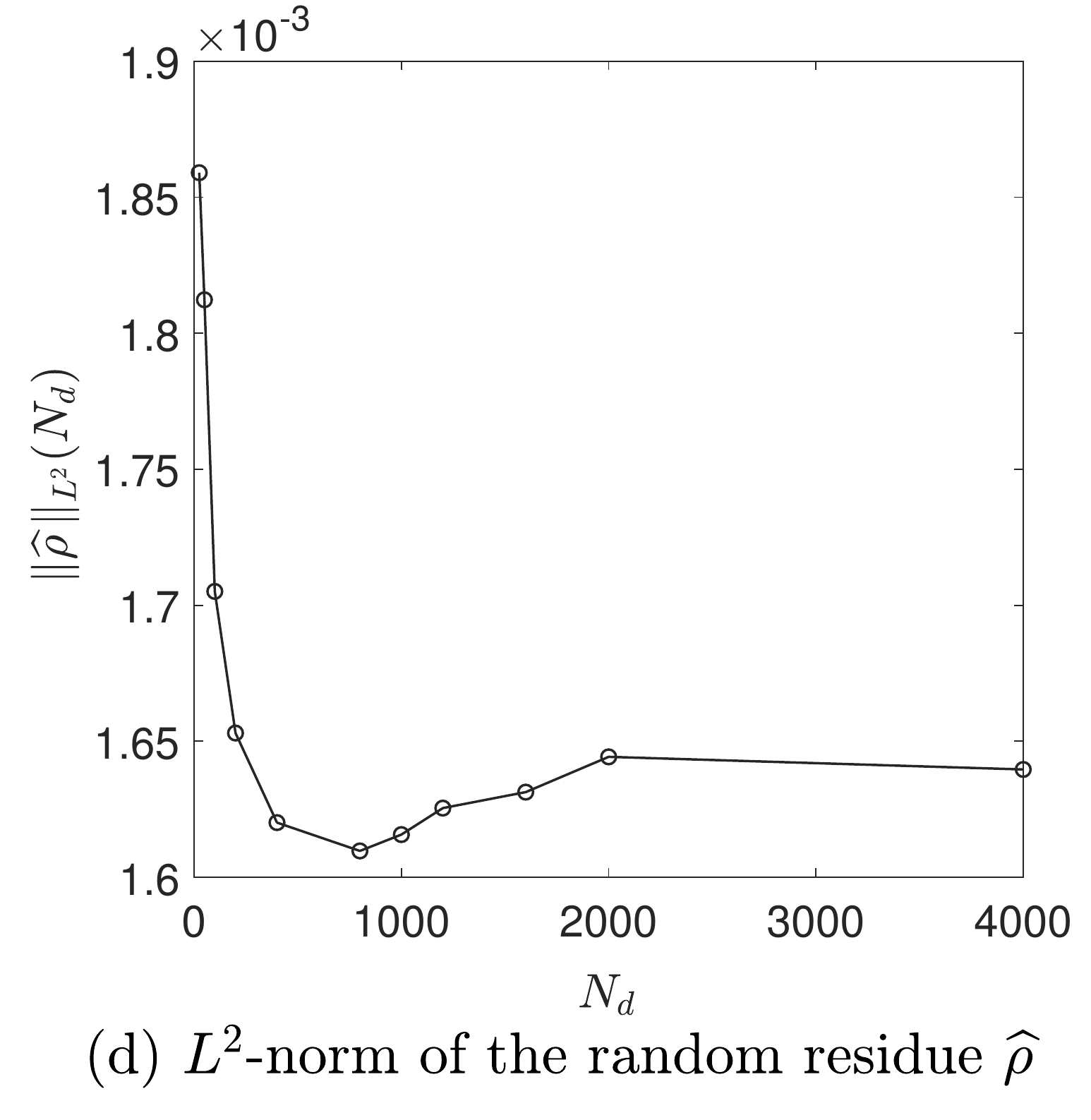}
\caption{Reference unsteady stochastic flow at $t=30$: (a) mean value $\bfx\mapsto 1\,000 \times  m_P(\bfx,t)$; (b) standard deviation $\bfx\mapsto 1\,000 \times  \sigma_P(\bfx,t)$; (c) pdf $r\mapsto p_{\rho_{N_d}}(r)$ of the random variable $\rho_{N_d}$; (d) $L^2$-norm $N_d\mapsto \Vert \,\widehat\rho\,\Vert_{L^2}(N_d)$ of the random residual $\widehat\rho$.}
\label{figureNS2}
\end{figure}
Figure~\ref{figureNS2}-(c) displays the graphs of the pdf $r\mapsto p_{\rho_{N_d}}(r)$ of the random variable
$\rho_{N_d} = \widehat\rho / {\underline{\widehat\rho}}_{N_d}$ in which ${\underline{\widehat\rho}}_{N_d} = E\big\{\widehat\rho\big\}$ is estimated using $N_d$ realizations. Therefore, $E\big\{\rho_{N_d}\big\} = 1$. For $N_d=400$, $1\,000$, $2\,000$, and $4\,000$, the standard deviation of $\rho_{N_d}$ is equal to $0.396$, $0.383$, $0.379$, and $0.377$, respectively. Figure~\ref{figureNS2}-(d) displays the graph $N_d\mapsto\Vert \,\widehat\rho\, \Vert_{L^2}(N_d)$.
These results show that the convergence with respect to $N_d$ is reasonably obtained for $N_d=4\,000$.
\subsection{Analysis of the role played by the constraints on the probability distribution of $\bfW$}
\label{SectionNS5}
As in Application~1, for simplifying the writing, $\bfW^{(\nu)}$ is simply noted $\bfW$.
We consider the training dataset of dimension $N_d = 100$ and  $n_\pMC = 400$ additional realizations (learning) are computed by using the PLoM under constraints. Figures~\ref{figureNS5}-(a) and (b)  correspond to $\algo~1$ of the PLoM under the constraint on $\rho$ but without the constraints on $\bfW$. The mean value ${\underline{\widehat\rho}}_{\,\pref} = E\big\{\widehat\rho_\pref\big\}$ is estimated using the $N_{d,\pref}=100$ realizations.
For $N_d = 100$, $n_\pMC = 400$, $t=30$, and using $\algo~1$, Figs.~\ref{figureNS5}-(a) and (b) show the error functions $i\mapsto \err_R(i)$ and  $i\mapsto \err_W(i)$ defined by Eq.~\eqref{eq8.9}.
As expected, Fig.~\ref{figureNS5}-(a) shows the decrease in the error function of $\rho$, which shows that the optimization process decreases the norm of $\rho$. On the other hand, since there is no constraint on $\bfW$, the optimization process to decrease the random residual of the stochastic PDE, modifies the prior probability distribution of $\bfW$, which is seen by the growth of function $i\mapsto\err_W(i)$ that is plotted in Fig.~\ref{figureNS5}-(b).
\begin{figure}[h!]
  \centering
  \includegraphics[width=3.8cm]{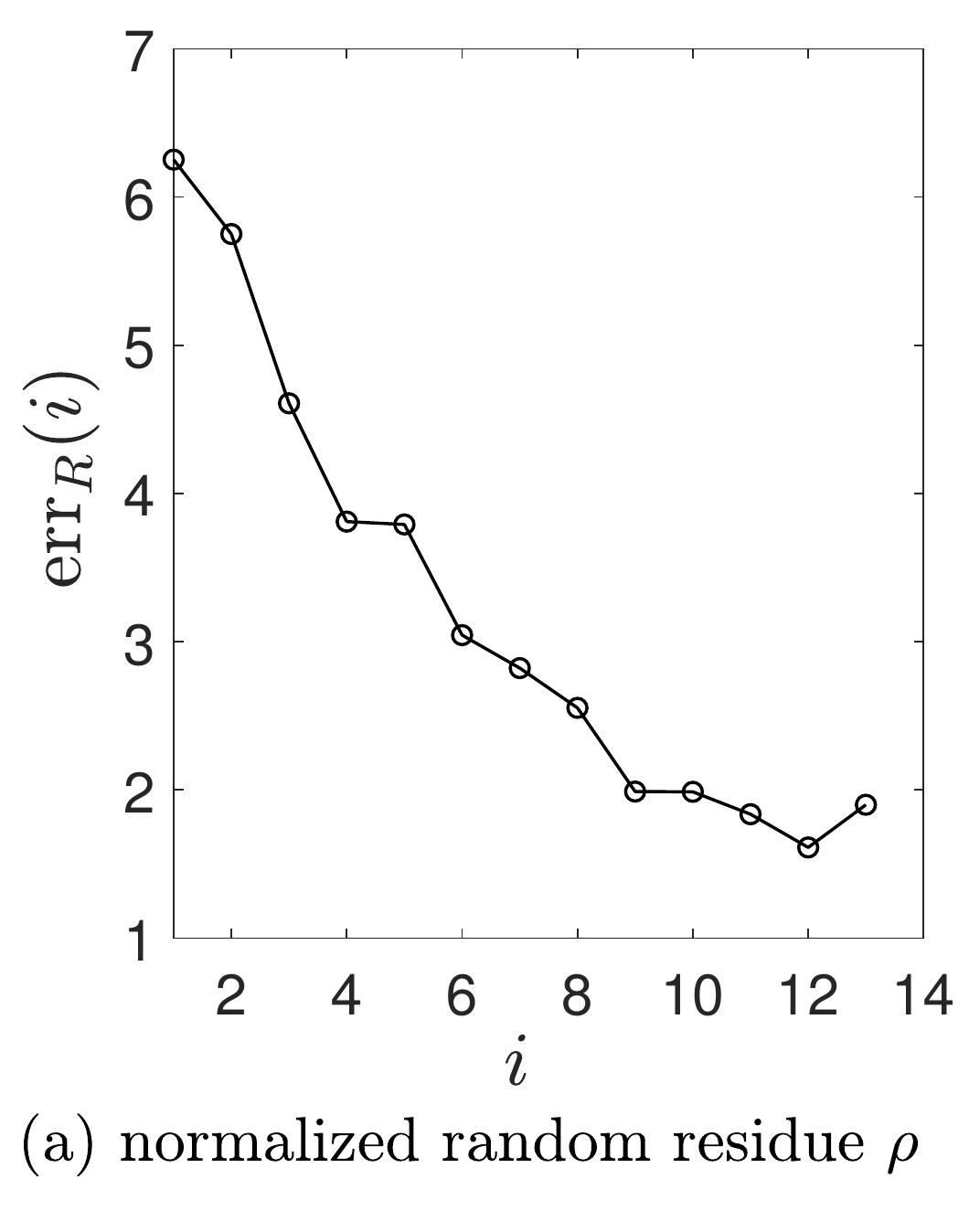}  \includegraphics[width=3.8cm]{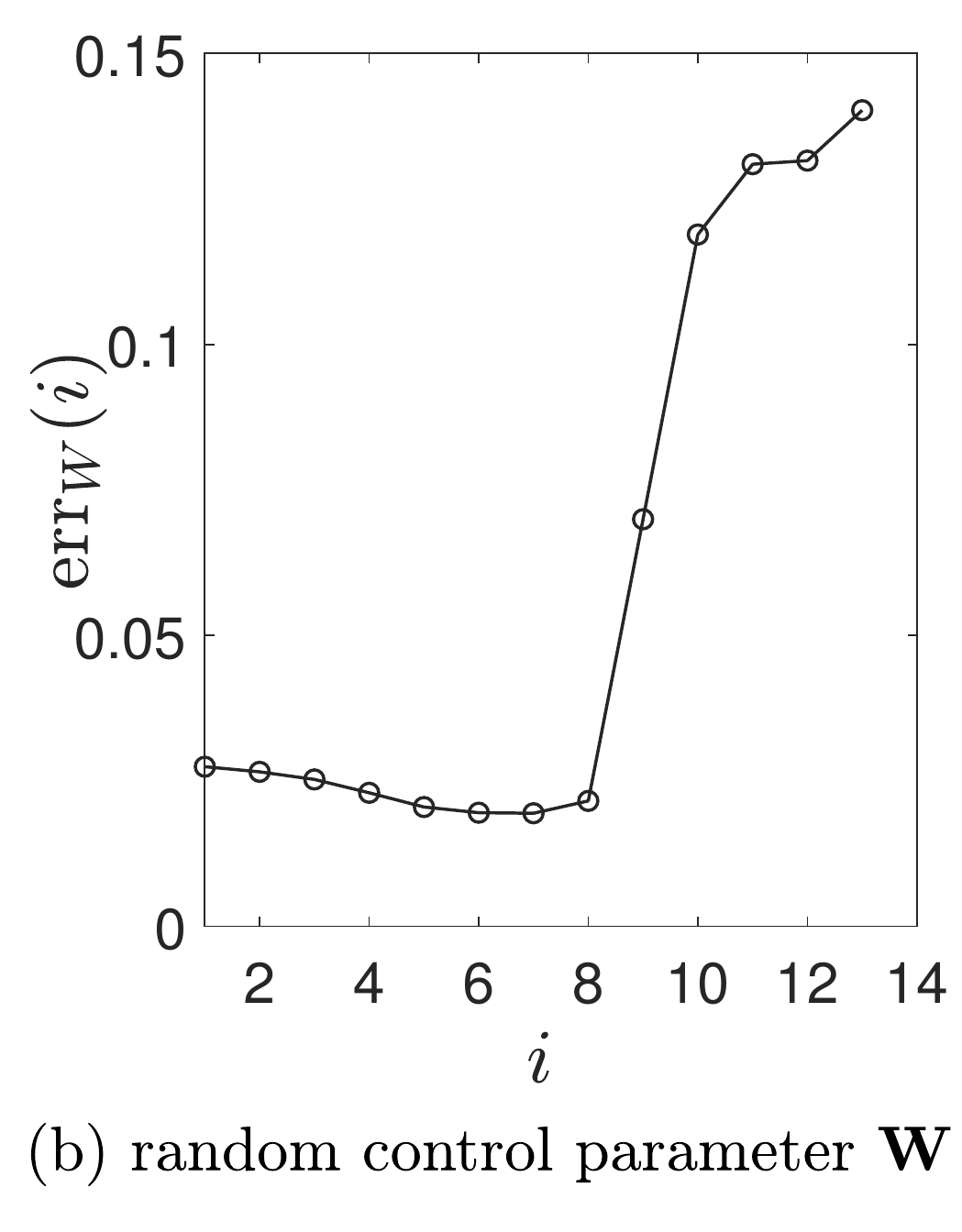}
   \includegraphics[width=4.2cm]{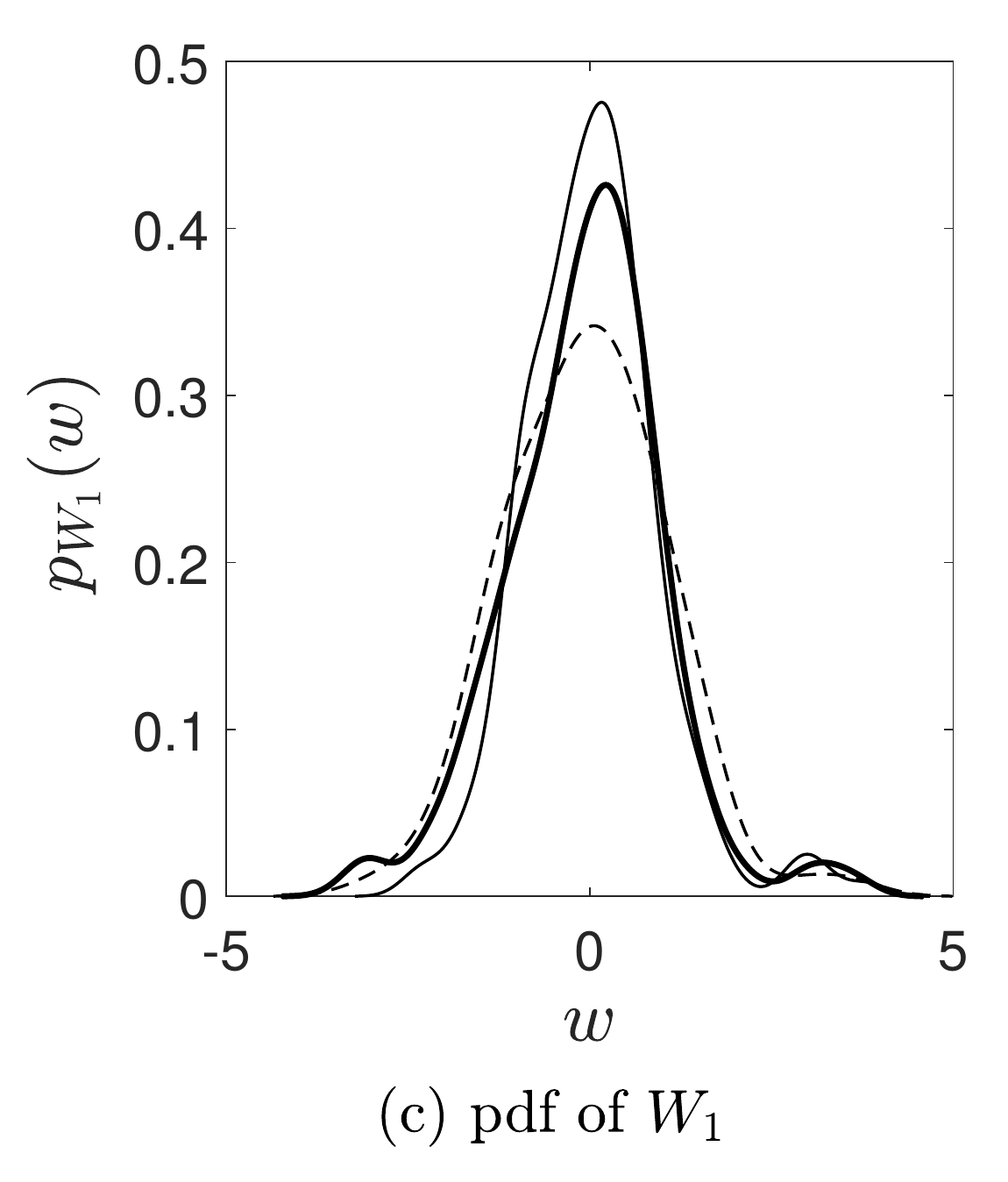}  \includegraphics[width=4.2cm]{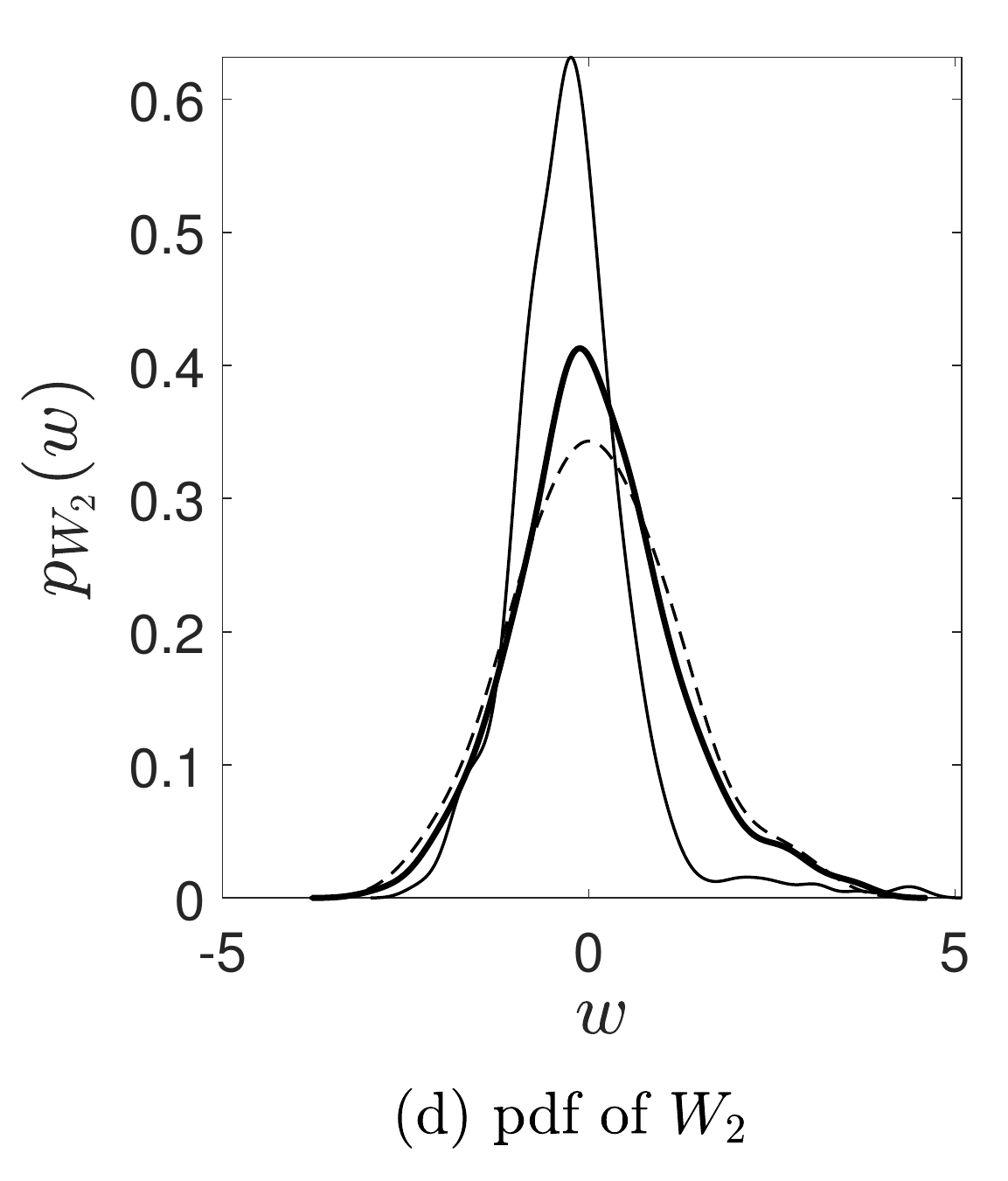}
\caption{For $N_d = 100$, $n_\pMC = 400$,  unsteady stochastic flow at $t=30$, and for the PLoM computation under the constraint on $\rho$ but without the constraints on $\textbf{W}$: (a) function $i\mapsto \errp_R(i)$; (b) function $i\mapsto \errp_W(i)$;
(c) pdf $w\mapsto p_{W_1}(w)$ and (d) pdf $w\mapsto p_{W_2}(w)$ estimated with $N_d = 100$ realizations (prior model) (dashed line), with $n_\pMC = 400$ additional realizations (learning) calculated using $\algo~1$ of the PLoM under the constraint on $\rho$ but without the constraints on $\bfW$ (thin solid line), and with $n_\pMC = 400$ additional realizations (learning) calculated using $\algo~3$ of the PLoM under the  constraints on $\rho$ and  $\bfW$ (thick solid line).}
\label{figureNS5}
\end{figure}
Figures~\ref{figureNS5}-(c) and (d) quantify this modification of the prior probability density function of each component of random vector $\bfW = (W_1,W_2)$. Figure~\ref{figureNS5}-(c) shows the pdf $w\mapsto p_{W_1}(w)$ of $W_1$ while Fig.~\ref{figureNS5}-(d) the pdf $w\mapsto p_{W_2}(w)$ of $W_2$. On each of these two figures, it is plotted the estimated pdf, on the one hand with the $N_d = 100$ realizations (prior pdf estimated with the training), on the other hand with the $n_\pMC = 400$ additional realizations (learning) calculated using $\algo~1$ (converged for $i_\opt=12)$  and using $\algo~3$ (converged for $i_\opt=5)$ of the PLoM under the constraints on $\rho$ and $\bfW$. It can be seen notable improvement of the probability density function of $\bfW = (W_1,W_2)$ when the constraints on $\bfW$ are taken into account.
\subsection{Analysis of the behavior of the iterative algorithms as a function of $n_\pMC$ and convergence criteria}
\label{SectionNS6}
The three algorithms, $\algo~1$ to $\algo~3$  have deeply been tested.
For the reasons given in Section~\ref{Section8.3}, we have used  and we present the results obtained with $\algo~3$ for the iterative algorithm.
\begin{figure}[h!]
  \centering
  \includegraphics[width=4.5cm]{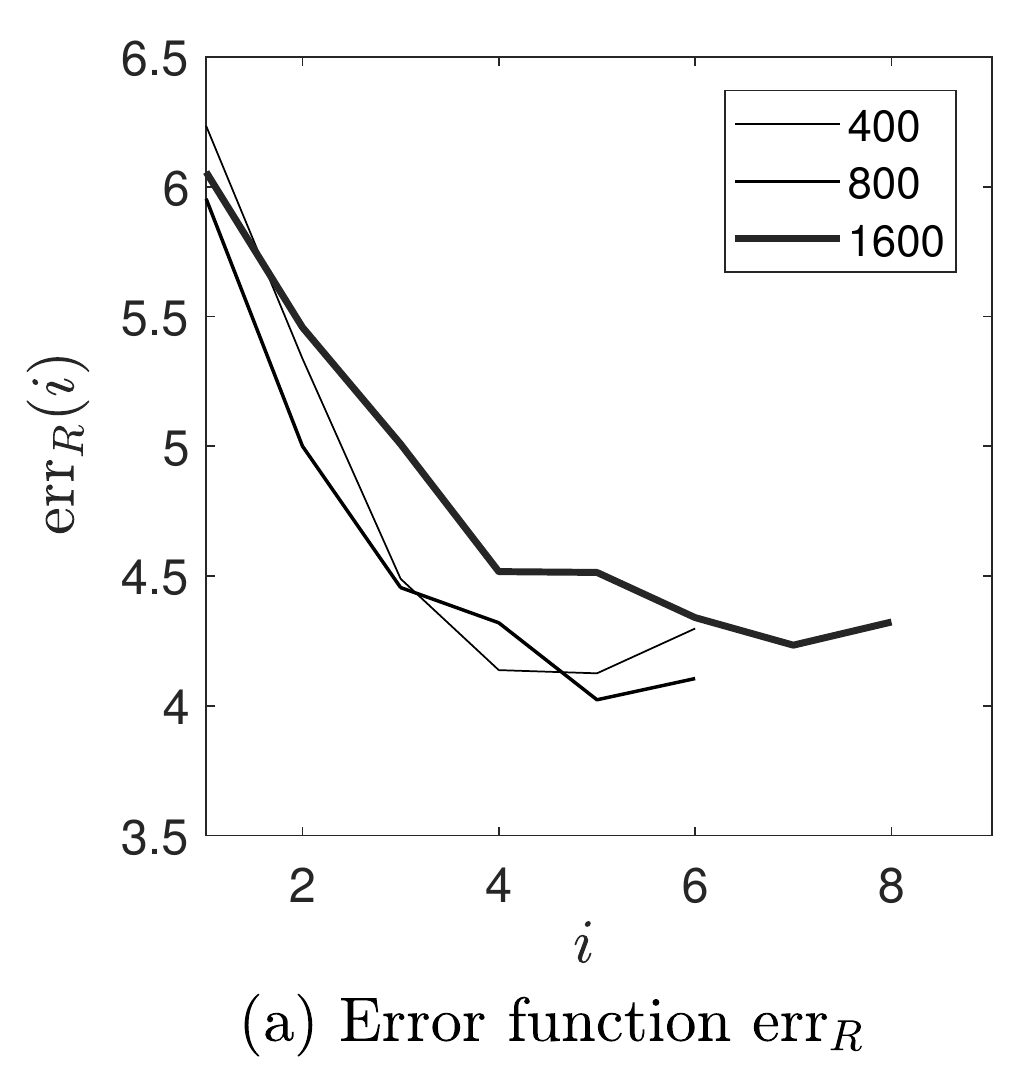}  \includegraphics[width=4.5cm]{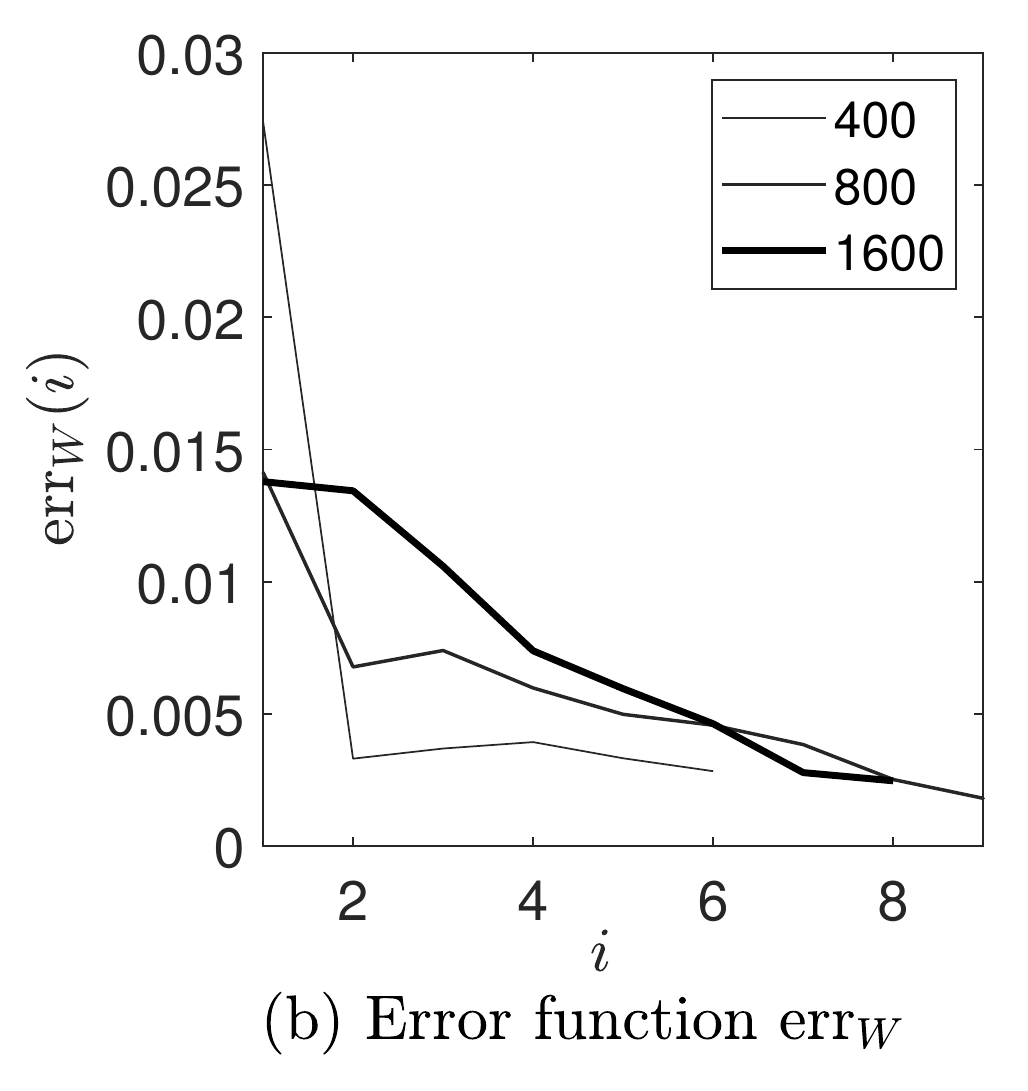} \includegraphics[width=4.5cm]{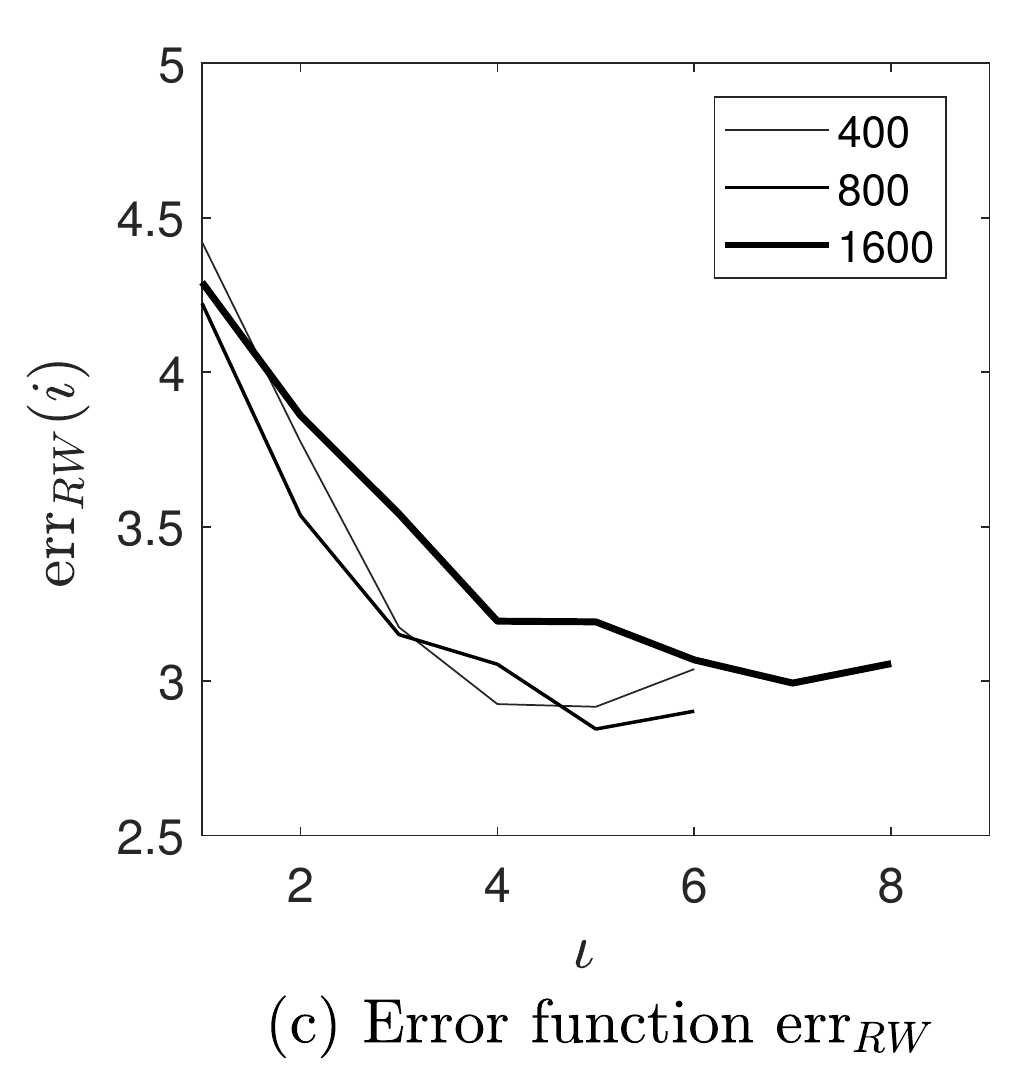}
\caption{For unsteady stochastic flow at $t=30$, for $N_d = 100$, and using $\algo~3$ for the iteration algorithm, evolution of the error functions as a function of $n_\pMC$: $i\mapsto\errp_R (i)$ (a), $i\mapsto\errp_W (i)$ (b), and $i\mapsto\errp_{RW} (i)$ (c).}
\label{figureNS7}
\end{figure}
For unsteady stochastic flow at $t=30$ and for $N_d = 100$, Fig.~\ref{figureNS7} shows the evolution of the error functions $i\mapsto\err_R (i)$ , $i\mapsto\err_W (i)$, and $i\mapsto\err_{RW} (i)$ defined in $\algo~3$, for $n_\pMC = 400$, $800$, and $1\,600$.
For $n_\pMC = 400$, a minimum of $\err_{RW}$ is reached for $i_\opt=5$ with $\err_{RW}(i_\opt)= 2.92$, while, for $n_\pMC = 1\,600$, a minimum of $\err_{RW}$
is reached for $i_\opt=7$ with $\err_{RW}(i_\opt)= 2.99$. Although the two values are close, one might think that the solution obtained for $n_\pMC = 400$ is better than that obtained for $n_\pMC = 1\,600$. In fact it is not the case  because the larger $n_\pMC $ the better the convergence of the statistical estimators of the solution. The effects of the small difference between $2.92$ and $2.99$ are negligible.
\subsection{Results of the PLoM without and under constraints for a given dimension of the training dataset}
\label{SectionNS7}
We consider the training dataset with $N_d = 100$. A large number $n_\pMC\gg N_d$ of additional realizations is generated with the PLoM under the constraints on $\rho$ and $\bfW$, by using algorithm $\algo~3$. The unsteady stochastic flow is observed at $t=30$. The challenge is therefore, for this small  value $N_d$ of points in the training dataset, to show that, for a  large value of $n_\pMC$ (with convergence with respect to $n_\pMC$), a good approximation of the reference is obtained whereas the use of the PLoM without the constraints gives an acceptable approximation, which can therefore be improved.  For this application, the PLoM method under constraints of the stochastic PDE allows an improvement of the PLoM without constraint \cite{Soize2016}.\\

(i) Figure~\ref{figureNS8} shows the probability density function $r\mapsto p_{\rho}(r)$ of $\rho$ as a function of the number $n_\pMC$ of additional realizations generated with the PLoM without constraint and under constraints. Figure~\ref{figureNS8}-(c) shows, for $n_\pMC=4\,000$, the comparison of this pdf estimated with the PLoM without constraint and under constraints.
The analysis of these figures shows a good convergence of the pdf of $\rho$ with respect to $n_\pMC$.\\
\begin{figure}[h!]
  \centering
  \includegraphics[width=4.5cm]{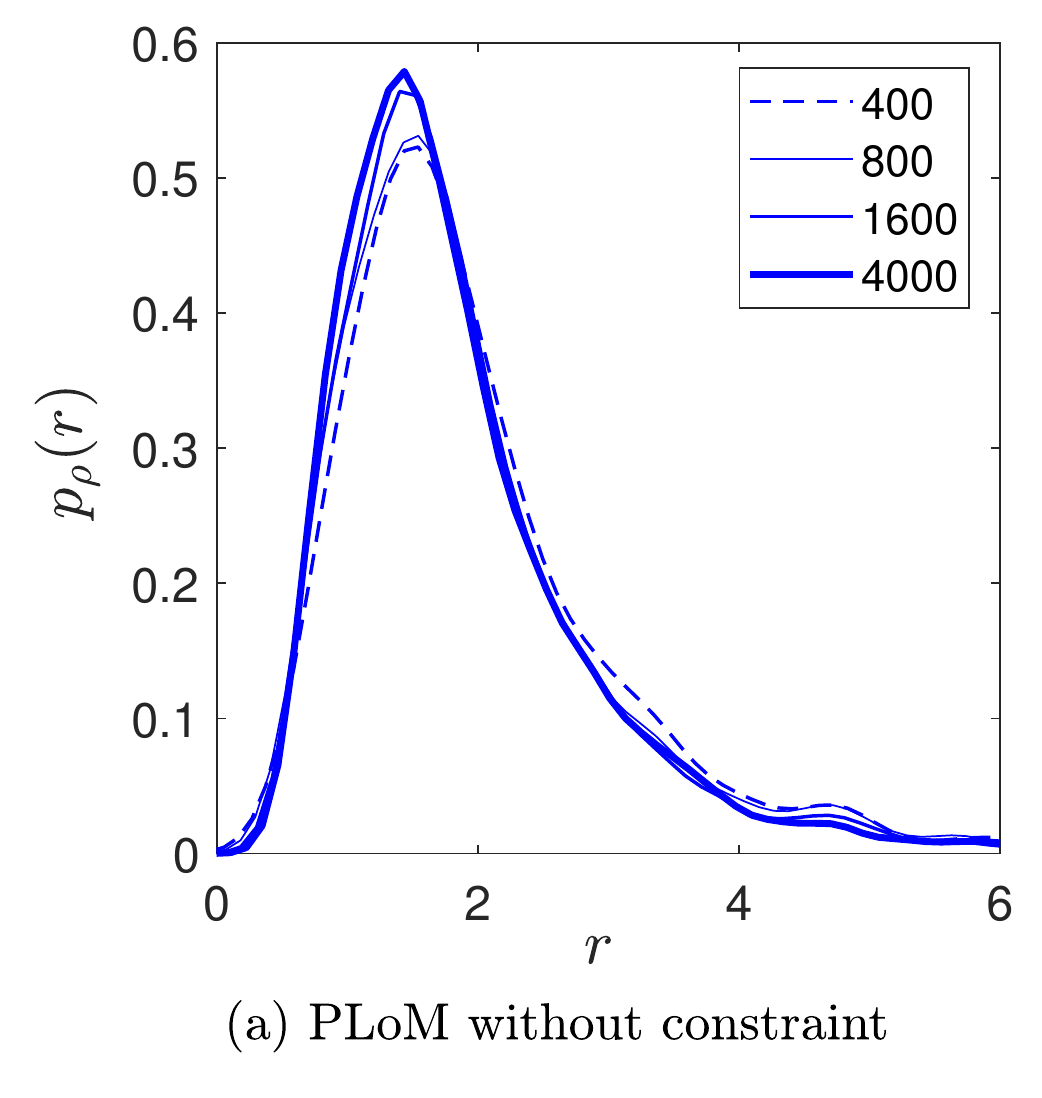}  \includegraphics[width=4.5cm]{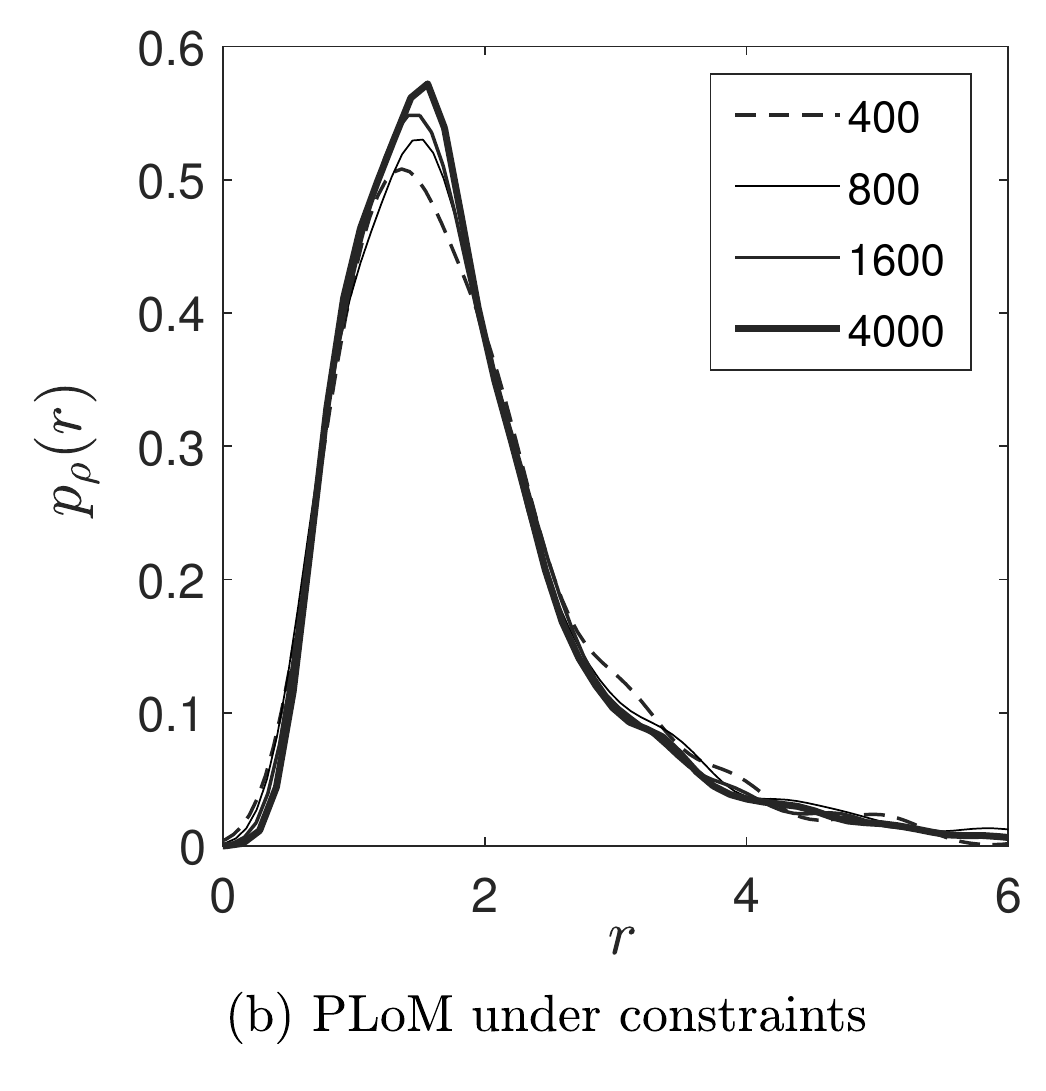} \includegraphics[width=4.5cm]{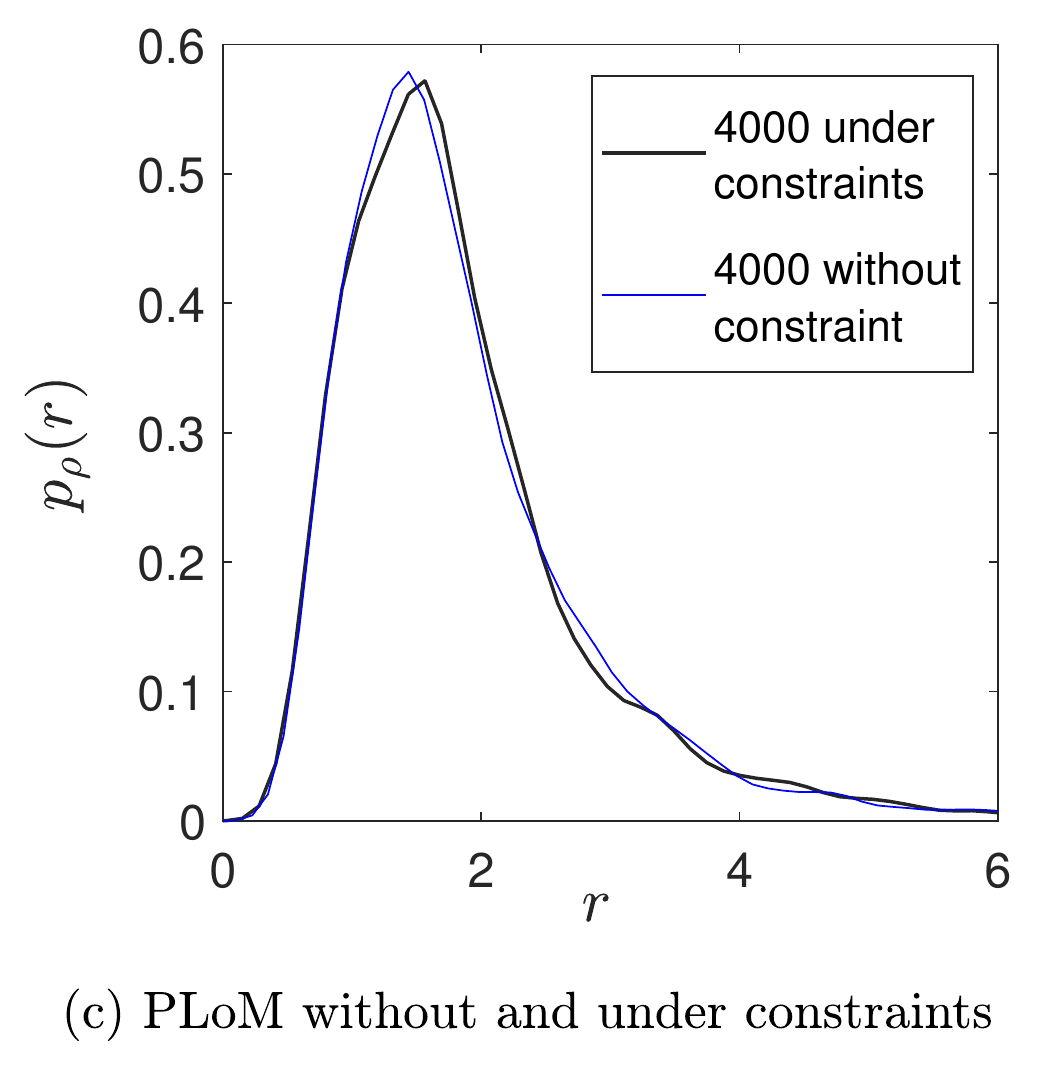}
\caption{For unsteady stochastic flow at $t=30$ and for $N_d = 100$, graph of pdf $r\mapsto p_{\rho}(r)$ of $\rho$ as a function of $n_\pMC$, estimated (a) with the PLoM without constraint, (b) with the PLoM under constraints with $\algo~3$, and (c) for $n_\pMC=4\,000$, the PLoM without and under constraints.}
\label{figureNS8}
\end{figure}

(ii) For the unsteady stochastic flow at $t=30$, for $N_d = 100$ and $n_\pMC=4\,000$, Fig.~\ref{figureNS9} shows the evolution of the error functions $i\mapsto\err_R (i)$, $i\mapsto\err_W (i)$, and $i\mapsto\err_{RW} (i)$ computed with $\algo~3$.
The iterative algorithm is stopped at iteration $14$ for which there is a local minimum that is $\err_{RW}(14)= 3.65$ and for which $\err_R(14)= 5.16$ and $\err_W(14)= 0.0015$.
\begin{figure}[h!]
  \centering
  \includegraphics[width=4.5cm]{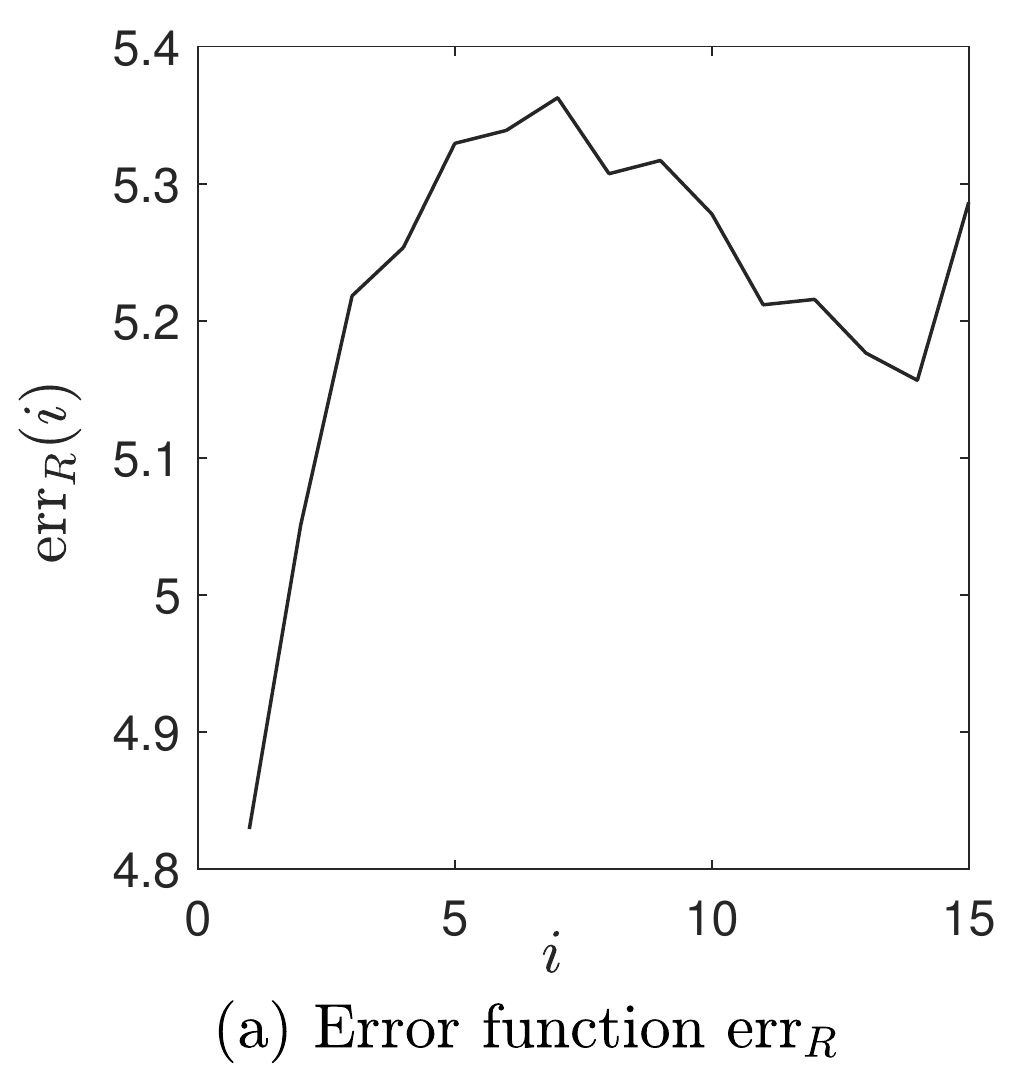}  \includegraphics[width=4.5cm]{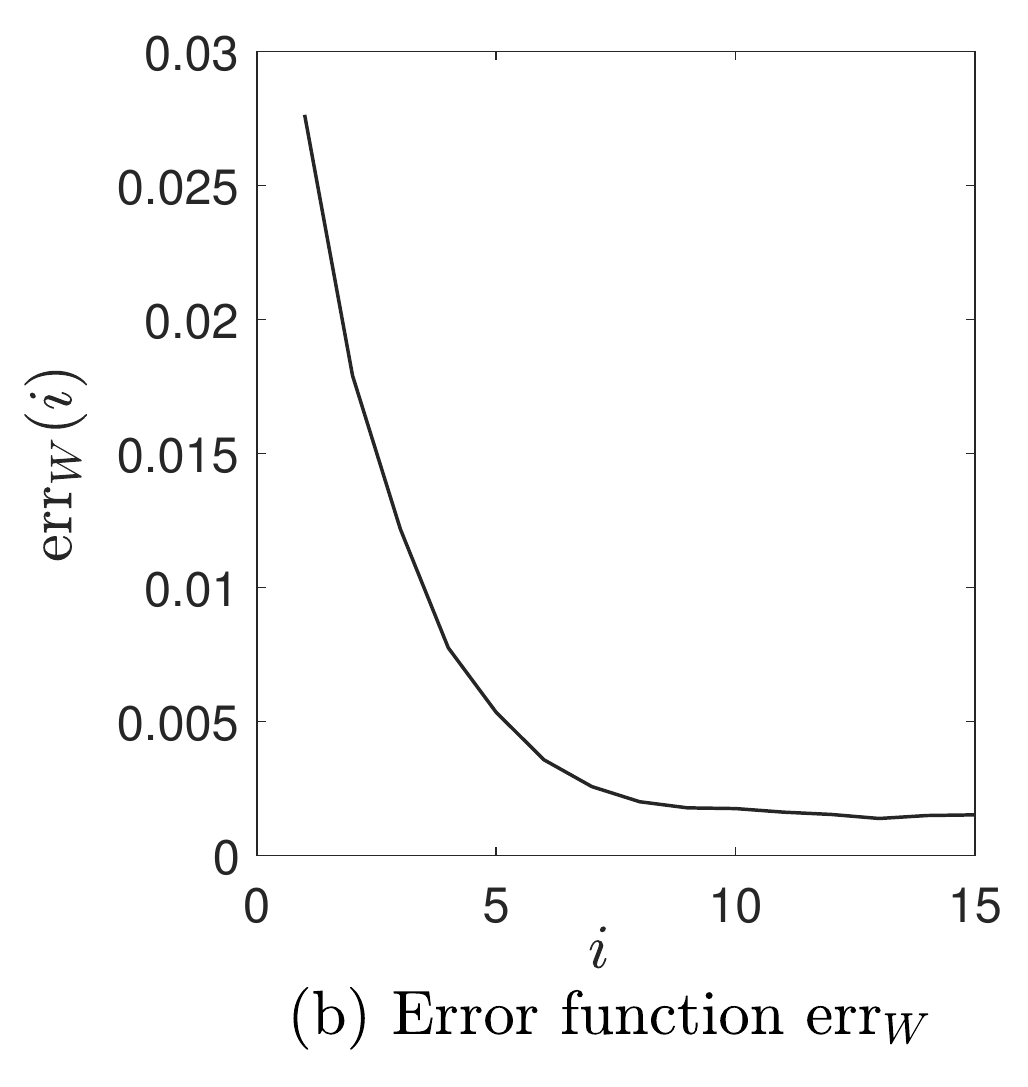} \includegraphics[width=4.5cm]{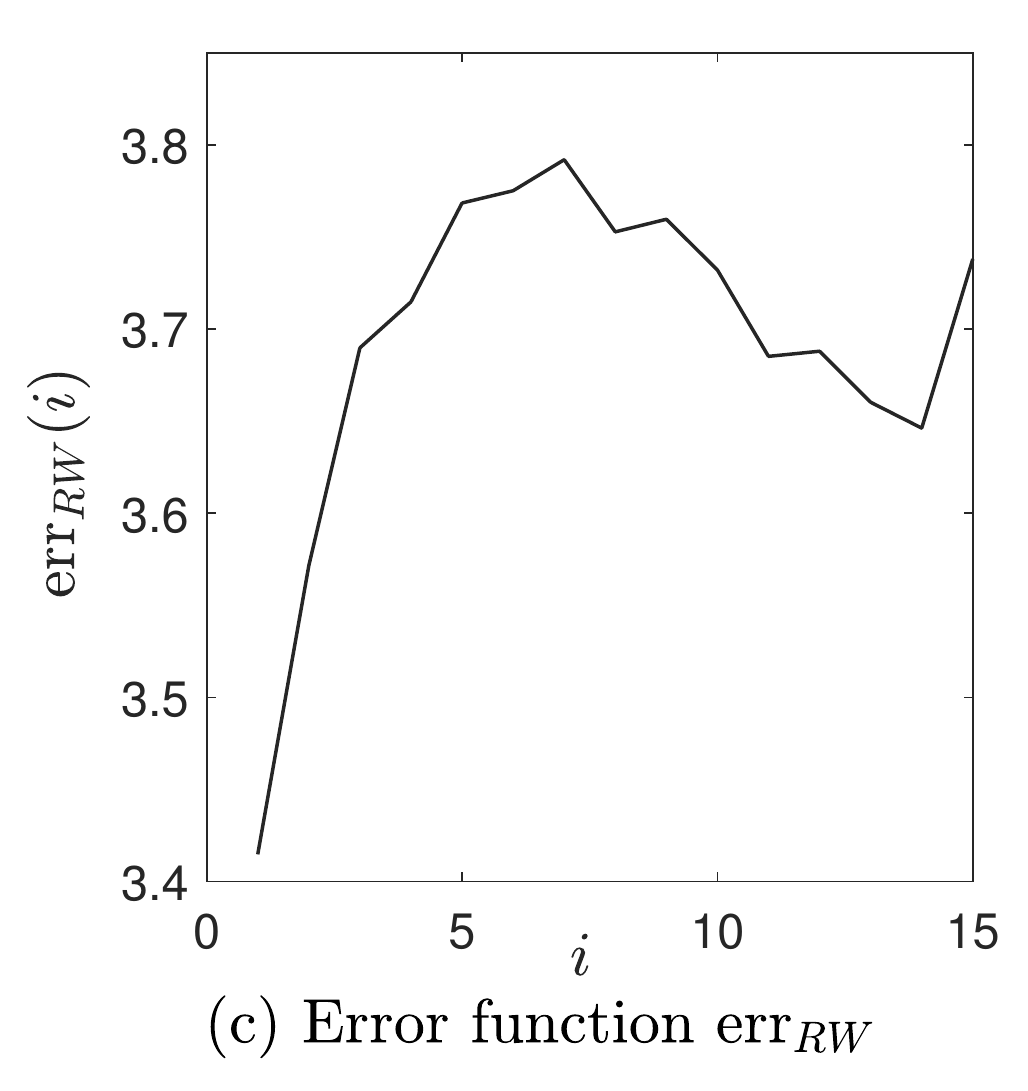}
\caption{For unsteady stochastic flow at $t=30$, $N_d = 100$, $n_\pMC = 4\,000$, and using $\algo~3$, evolution of the error functions $i\mapsto\errp_R (i)$ (a), $i\mapsto\errp_W (i)$ (b), and $i\mapsto\errp_{RW} (i)$ (c).}
\label{figureNS9}
\end{figure}
\begin{figure}[h!]
  \centering
  \includegraphics[width=5.0cm]{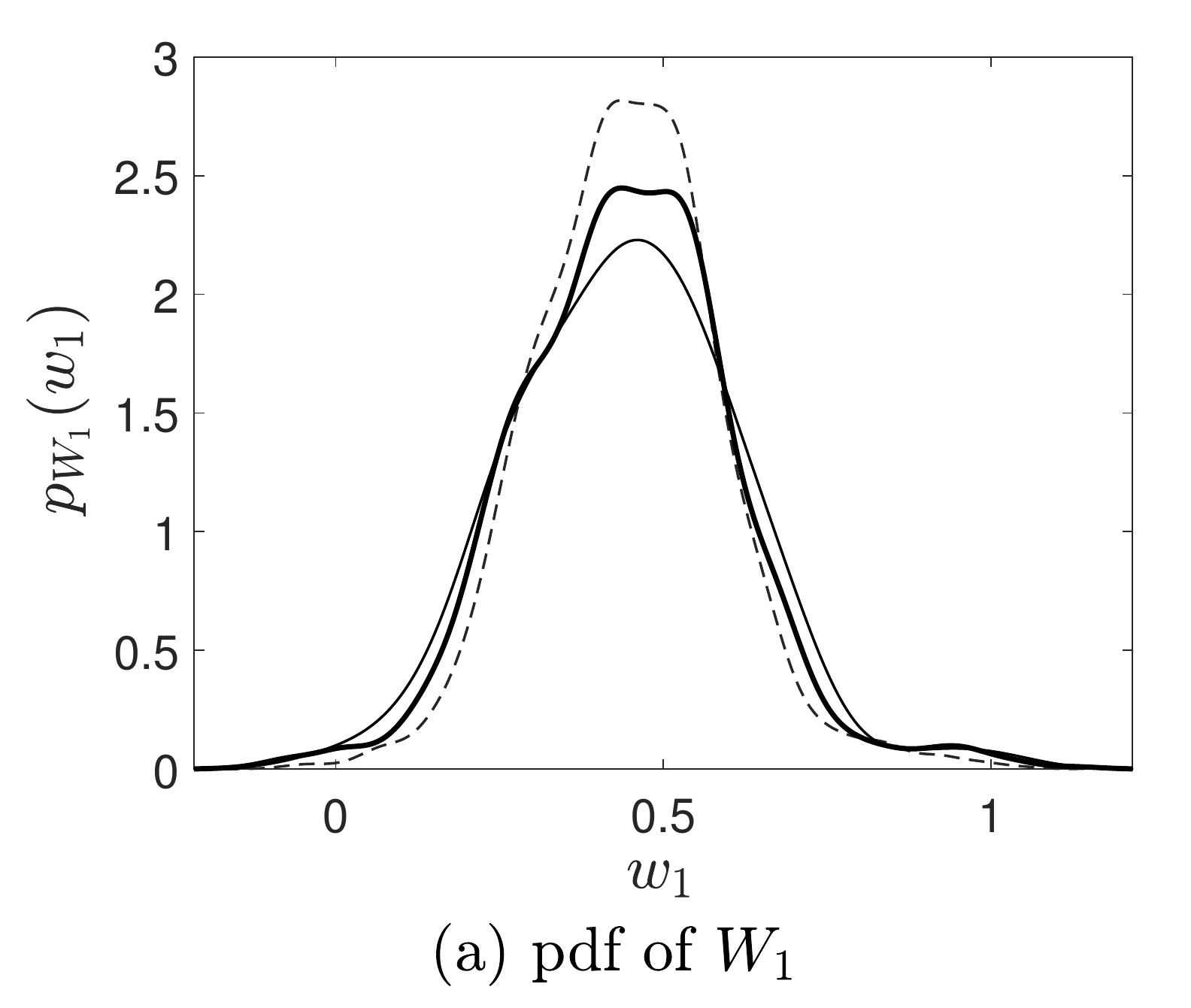}  \includegraphics[width=5.0cm]{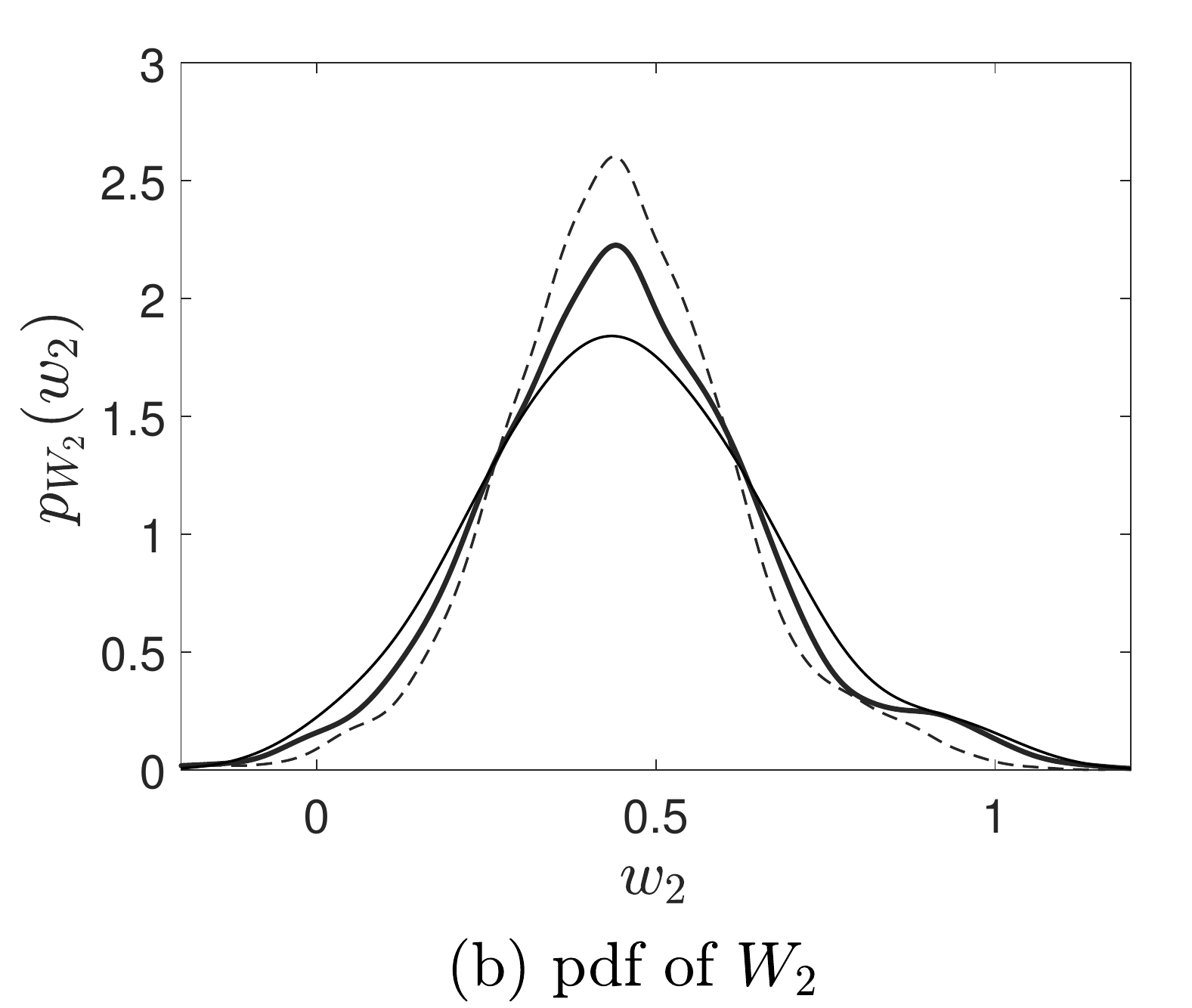}
\caption{For $N_d = 100$ and unsteady stochastic flow at $t=30$, pdf $w\mapsto p_{W_1}(w)$ (a) and pdf $w\mapsto p_{W_2}(w)$ (b), estimated with the $N_d = 100$ realizations (prior model) from the training dataset (solid line), with $n_\pMC = 4\, 000$ additional realizations (learning) estimated with the PLoM without constraint (dashed line) and under the constraints (thick solid line).}
\label{figureNS10}
\end{figure}
Figure~\ref{figureNS10} shows the prior probability density function $w\mapsto p_{W_1}(w)$ and $w\mapsto p_{W_2}(w)$ of each component $W_1$ and $W_2$ of random vector $\bfW$. In each figure is plotted the estimated pdf, on the one hand with the $N_d = 100$ realizations from the training dataset (prior pdf estimated with the training), on the other hand with the $n_\pMC = 4\,000$ additional realizations (learning) estimated with the PLoM without constraint and with the PLoM under the constraints on $\rho$ and $\bfW$.
The mean value $E\big\{\widehat\rho\big\}$ of  $\widehat\rho$ is $3.23 \times 10^{-3}$, its standard deviation is $2.74 \times 10^{-3}$, and $\Vert\, \widehat\rho\,\Vert_{L^2}$ is $4.23 \times 10^{-3}$. These values must be compared with those of the reference given in Section~\ref{SectionNS4}-(ii).

To illustrate the distribution of the realizations generated by the PLoM under constraints, Fig.~\ref{figureNS11} displays the $4\,000$ additional realizations  of components of $\bfX^c = (\bfQ^c,\bfW^c)$. A concentration of these realizations can be observed.
\begin{figure}[h!]
\centering
\includegraphics[width=6.2cm]{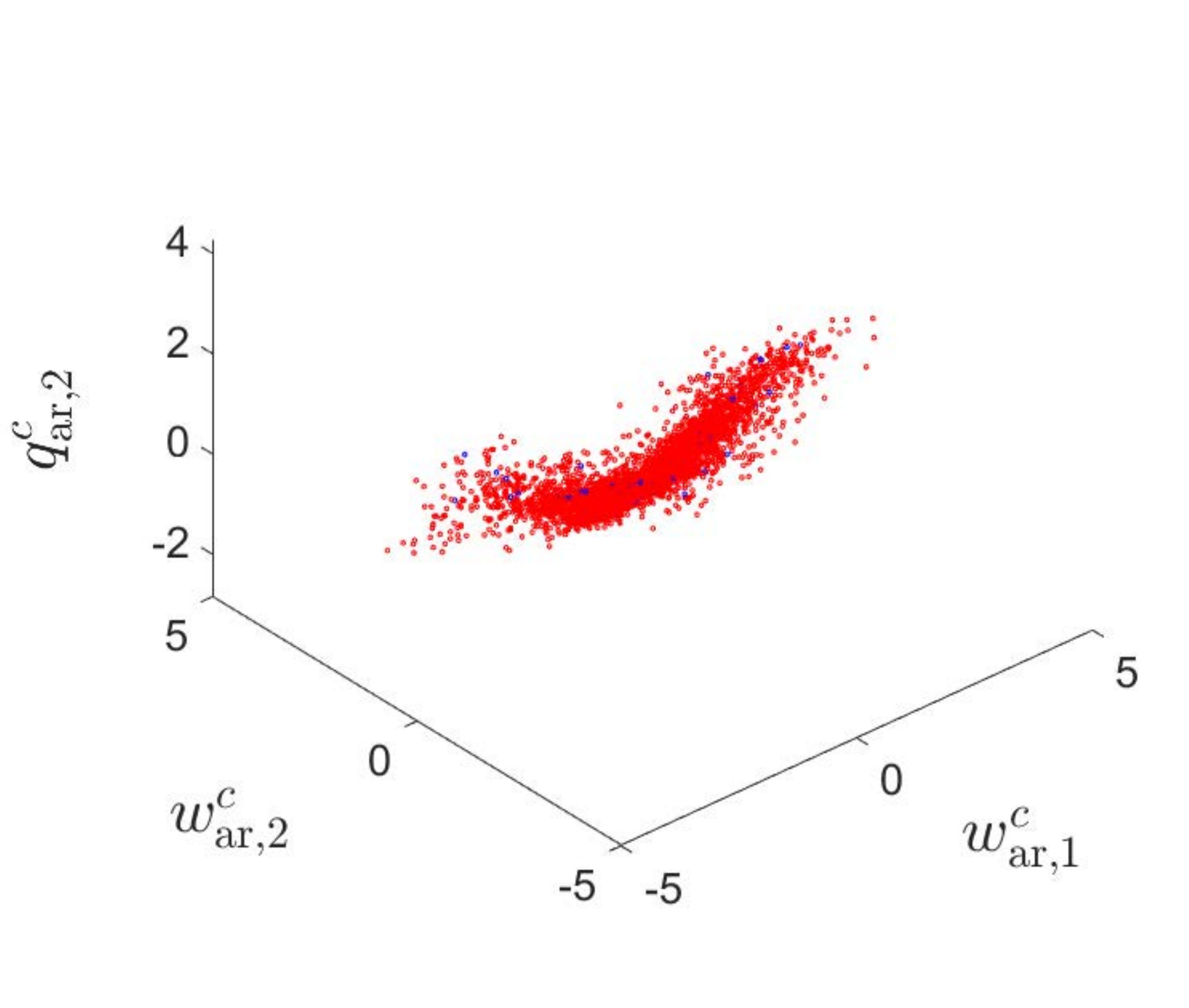}  \includegraphics[width=6.2cm]{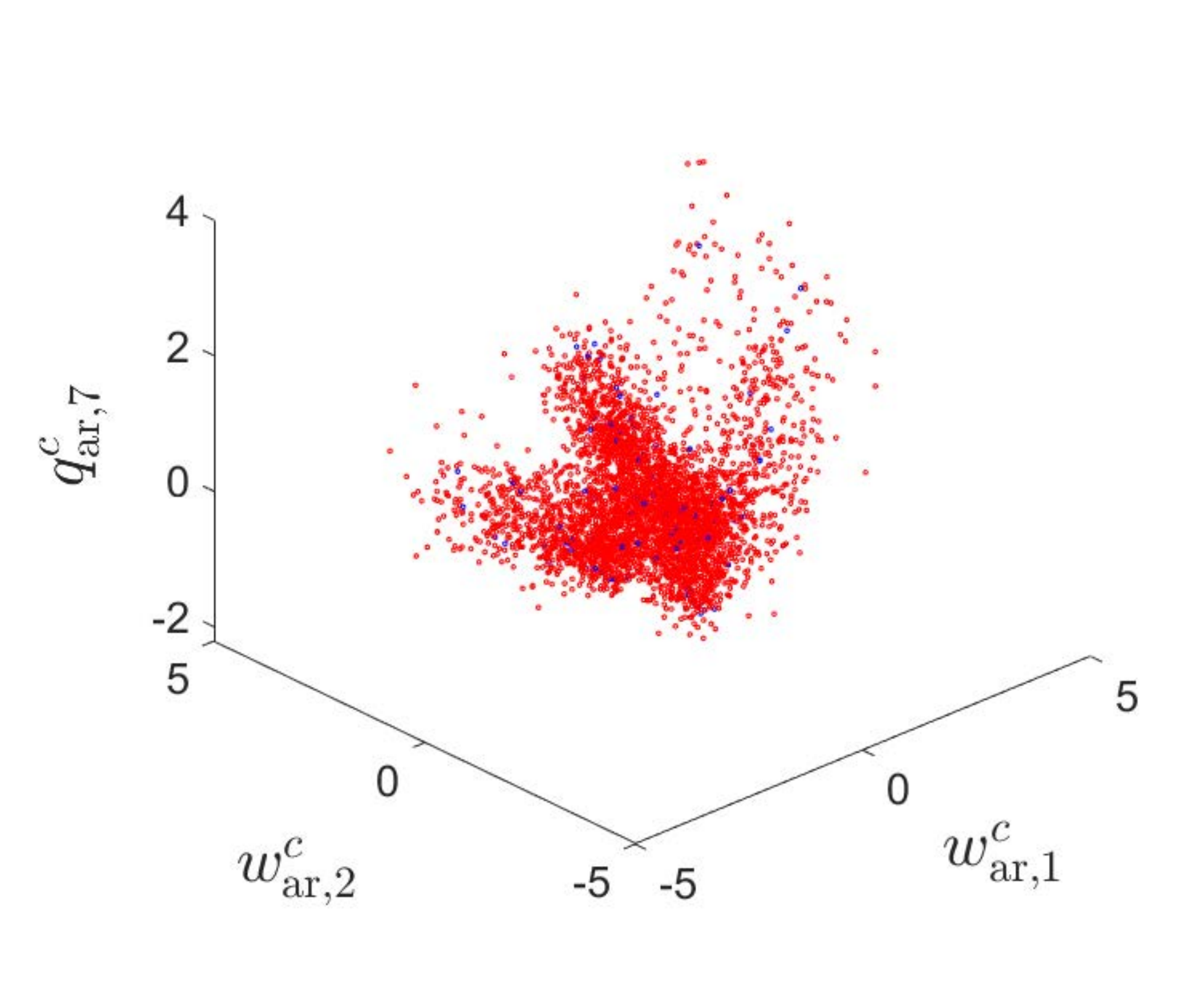}
\caption{3D-plot of realizations $(w^c_{\arp,1},w^c_{\arp,2},q^c_{\arp,2})$ of $(W^c_1,W^c_2,Q^c_2)$ (left figure) and $(w^c_{\arp,1},w^c_{\arp,2},q^c_{\arp,7})$ of $(W^c_1,W^c_2,Q^c_7)$ (right figure), computed with the PLoM under constraints: $n_\pMC = 4\,000$ additional realizations (red symbols) and $N_d = 100$ points from the training dataset (blue symbols).}
\label{figureNS11}
\end{figure}
\begin{figure}[h!]
\centering
\includegraphics[width=4.5cm]{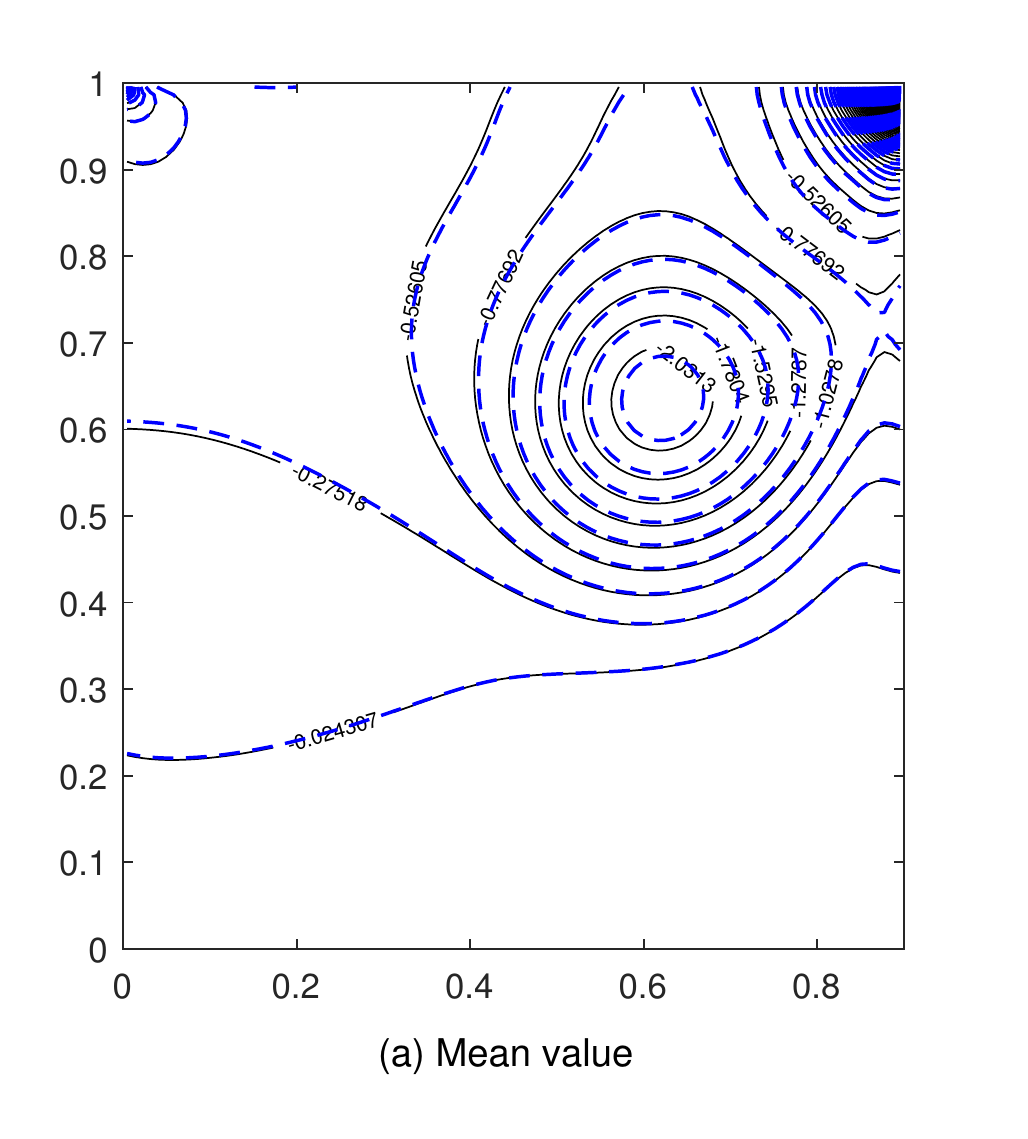}  \includegraphics[width=4.5cm]{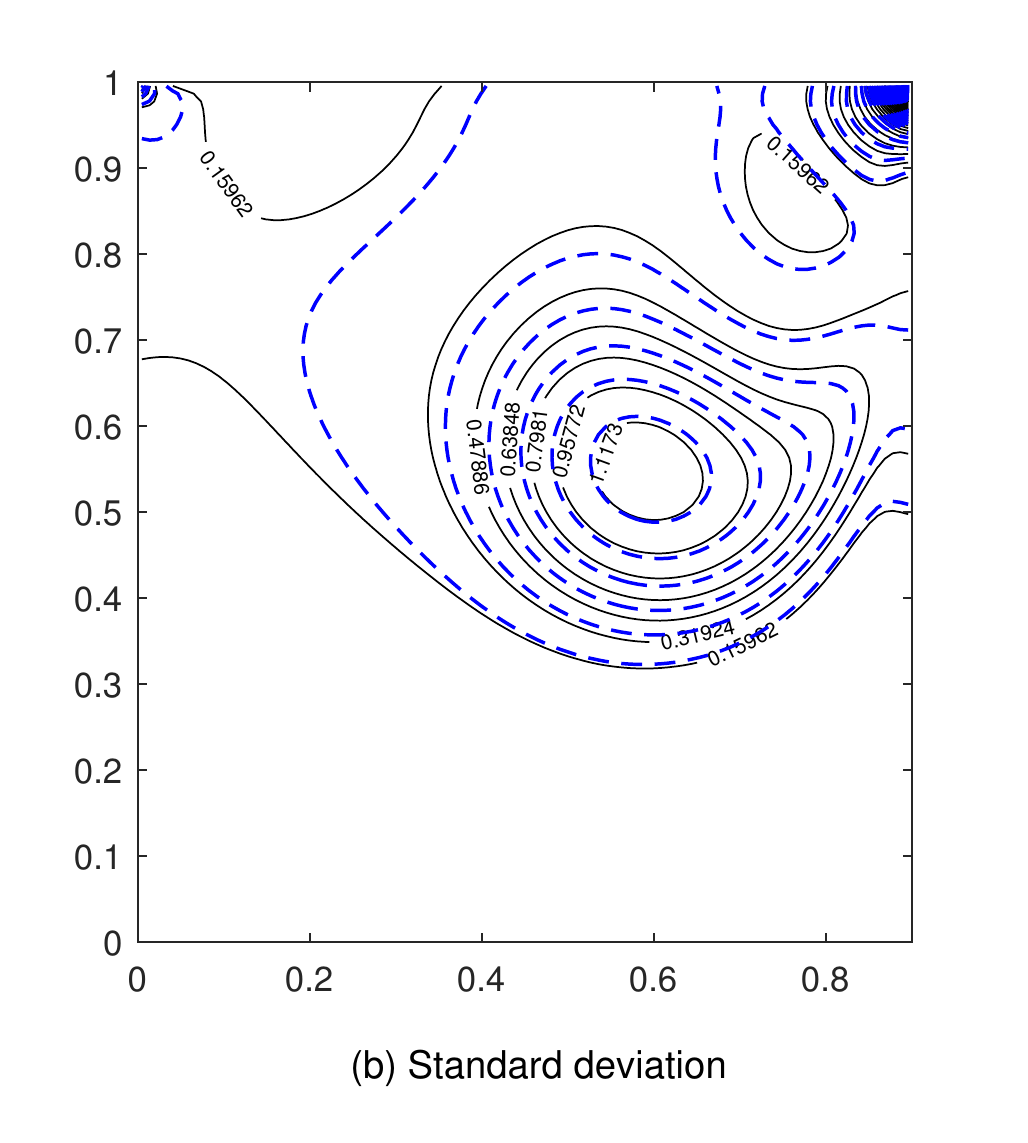}
\caption{For unsteady stochastic flow at $t=30$ and for $N_d = 100$, 2D-plot of mean value $\bfx\mapsto 1\,000 \times  m_P(\bfx,t)$ (a) and of standard deviation $\bfx\mapsto 1\,000 \times  \sigma_P(\bfx,t)$ (b) of the pressure random field corresponding to the reference (black thin solid lines) and estimated with $n_\pMC=4\,000$ additional realizations computed with the PLoM without constraint (blue thick dashed lines).}
\label{figureNS12}
\end{figure}
\begin{figure}[h!]
\centering
\includegraphics[width=3.9cm]{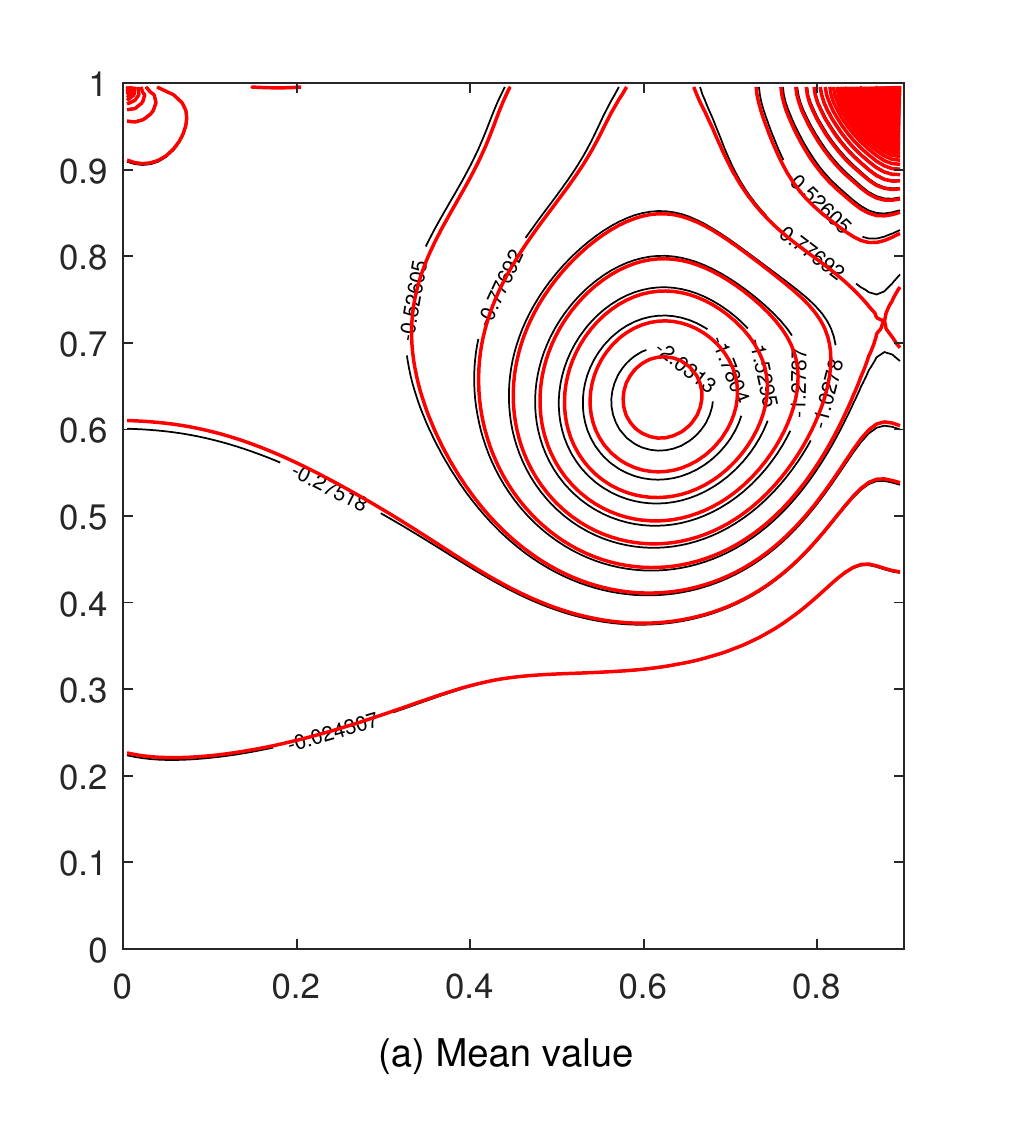}  \includegraphics[width=3.9cm]{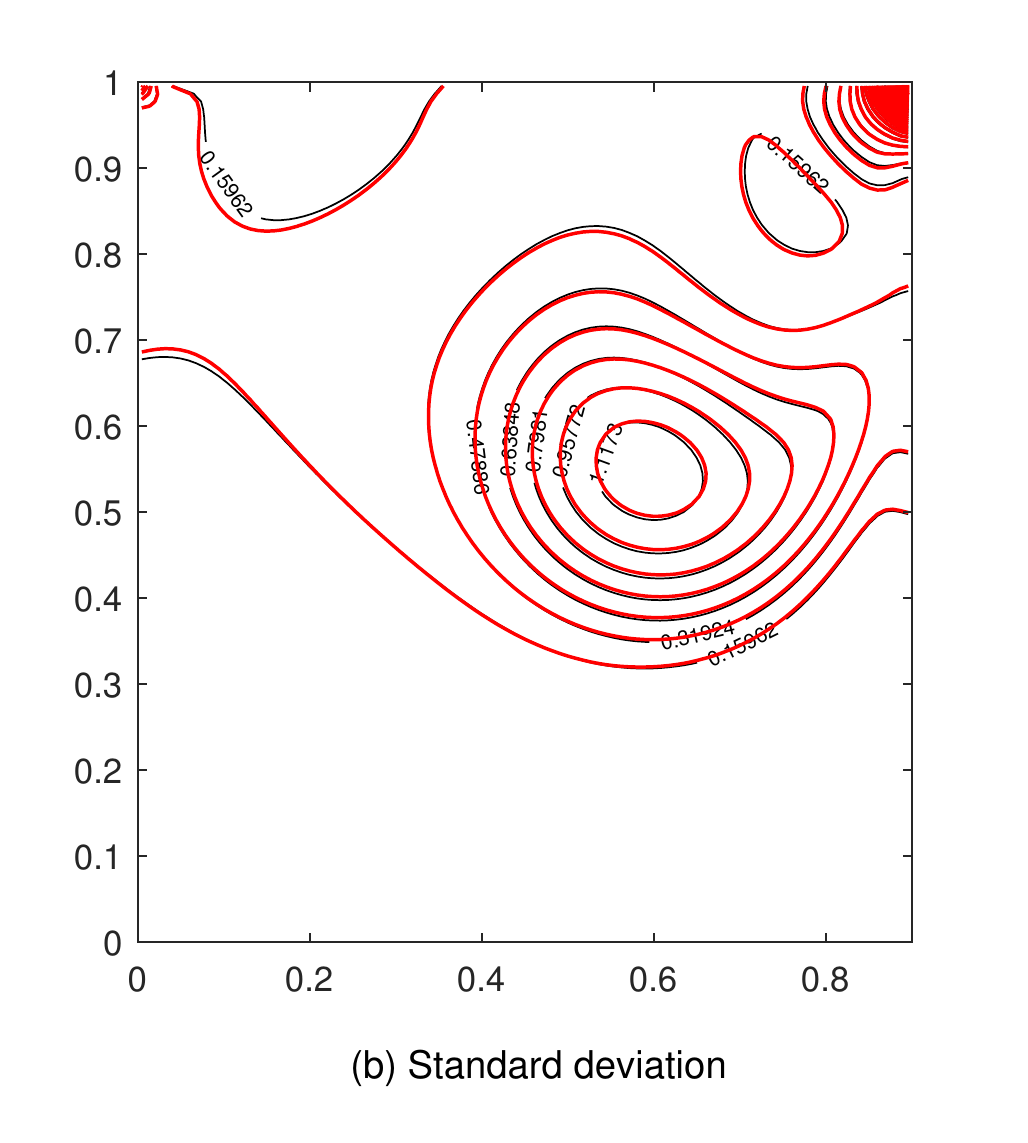}
\includegraphics[width=4.1cm]{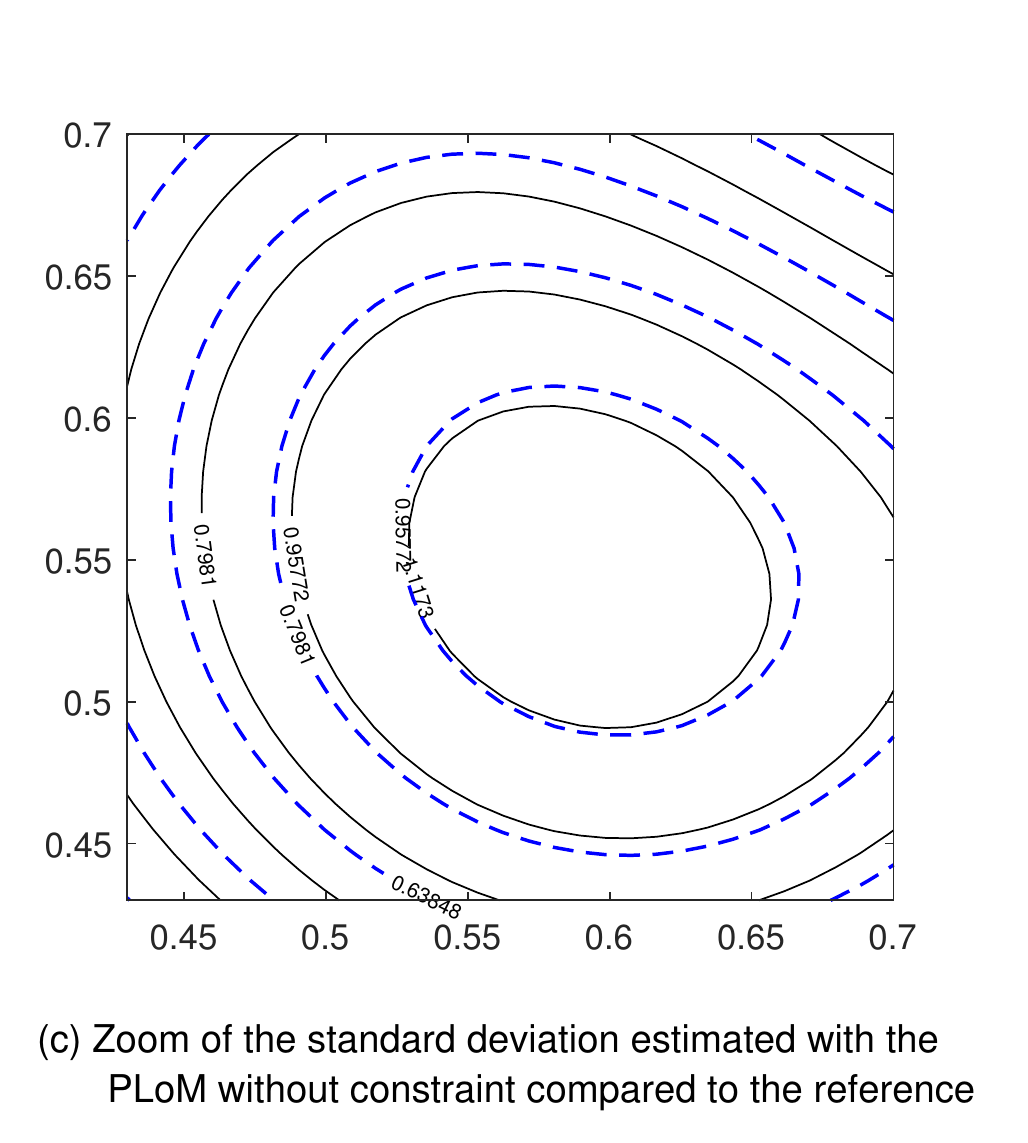}  \includegraphics[width=4.1cm]{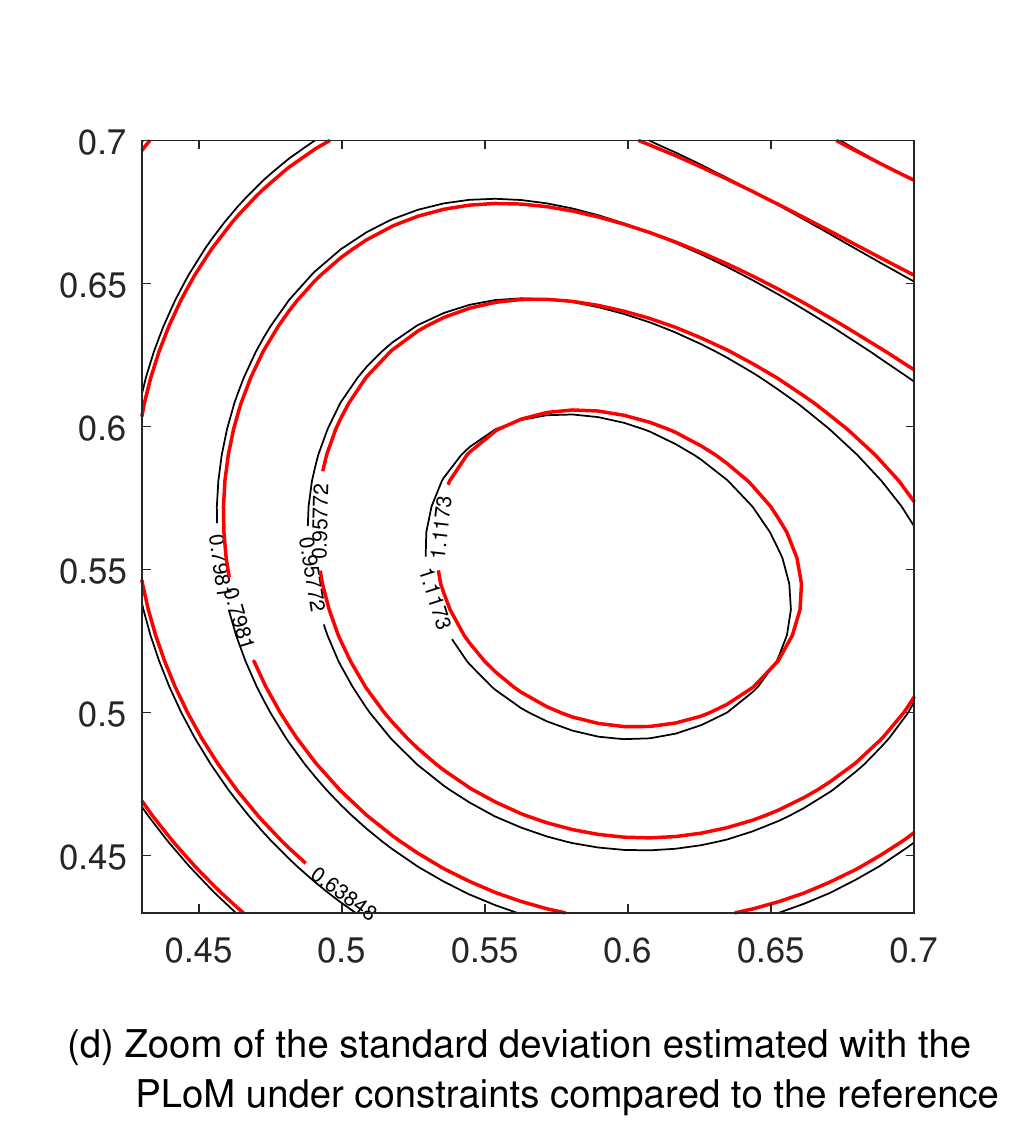}
\caption{For unsteady stochastic flow at $t=30$ and for $N_d = 100$, 2D-plot of mean value $\bfx\mapsto 1\,000 \times  m_P(\bfx,t)$ (a) and of standard deviation $\bfx\mapsto 1\,000 \times  \sigma_P(\bfx,t)$ (b) of the  pressure random field corresponding to the reference (black thin solid lines) and estimated with $n_\pMC=4\,000$ additional realizations computed with the PLoM under constraints (red thick solid lines).
Fig.~(c) is a zoom of Fig.~\ref{figureNS12}-(b). Fig.~(d) is a zoom of Fig.~(b).}
\label{figureNS13}
\end{figure}

Figure~\ref{figureNS12} displays the 2D-plot of  $\bfx\mapsto 1\,000 \times  m_P(\bfx,t)$  and $\bfx\mapsto 1\,000 \times  \sigma_P(\bfx,t)$  estimated with $n_\pMC=4\,000$ additional realizations generated by the PLoM without constraint and are compared to the reference (see Fig.~\ref{figureNS2}). These results show that the standard PLoM method (that is to say without constraint) is already good. It allows for generating a large number $n_\pMC$ of additional realizations (learning) from a small number $N_d$ of points of the training dataset. These realizations allow for estimating the second-order quantities of the random field,  which are good for the mean and less good for the standard deviation (see the zoom and the associated comment of Fig.~\ref{figureNS13}-(c)).

Similarly, Fig.~\ref{figureNS13} displays the 2D-plot of $\bfx\mapsto 1\,000 \times  m_P(\bfx,t)$  and  $\bfx\mapsto 1\,000 \times  \sigma_P(\bfx,t)$  estimated with $n_\pMC=4\,000$ additional realizations computed with the PLoM under constraints and shows the comparison with  the reference (see Fig.~\ref{figureNS2}). These results show that the PLoM under constraints give an excellent prediction of the second-order quantities, for both the mean and the standard deviation of the random fields.
Figure~\ref{figureNS13}-(c) corresponds to a zoom of Fig.~\ref{figureNS12}-(b).  At the center of the vortex, it can be seen that the value of the innermost level line (in the center of the vortex) is 0.95772 for the reference solution (solid line) while that which corresponds for the PLoM without constraint, is not the proximal dashed line, but the second deviating from the center. The value of the proximal dashed line is $1.1173$ instead of  $0.95772$. In contrast, Fig.~\ref{figureNS13}-(d) that is a zoom of Fig.~\ref{figureNS13}-(b) shows an excellent coincidence of the level lines for the same region of the fluid domain.
\subsection{Discussion}
\label{SectionNS8}
The results obtained for this application devoted to unsteady nonlinear stochastic computational fluid dynamics show that the PLoM without constraint is already giving  good results (as it was expected, taking into account the previous works). However the novel proposed extension, consisting in constraining the PLoM method by the stochastic PDE allows to improve the results, which are then quite excellent (see zoom in Fig.~\ref{figureNS13}-(d)).

\section{Application 3: nonlinear 3D solid dynamics}
\label{SectionSD}
%
%
\subsection{Description of the nonlinear 3D dynamical system}
\label{SectionSD1}
\noindent \textit{(i) System definition and geometry}. It is a 3D MEMS accelerometer whose scheme is shown in Fig.~1-(A) of Ref.~\cite{Soize2019a}. The mobile part is constituted of  a deformable rectangular frame in silicon with one  vertical deformable central beam attached to it. The suspended part is made up of a deformable parallelepipedic solid with two attached vertical deformable beams that are around the central beam, and is attached to the mobile part by a 3D suspension constituted of $20$ linear springs. The MEMS is represented by an 3D-domain $\Omega \subset \RR^3$ and is described in a Cartesian coordinate system $Ox_1x_2x_3$ that is attached to its mobile part and adopted as the reference coordinate system. The origin $O$ is located at the bottom left corner of the device. The axis $Ox_1$ is horizontal and oriented positively from left to right. The axis $Ox_2$ is vertical and oriented positively from bottom to top. The axis $Ox_3$ is perpendicular to the plane $Ox_1x_2$. The overall dimensions in microns ( $\times 10^{-6}\, m$ ) of the rectangular frame are $30\times 31\times 4$. \\

\noindent \textit{(ii) Materials}. The suspended and mobile parts are made of a homogeneous, orthotropic, linear elastic material, which is a standard $(100)$ silicon wafer~\cite{Hopcroft2010},
whose Young moduli are $\underline{E}_{11} = \underline{E}_{22} =169\times 10^9~N/m^2$, $\underline{E}_{33} =130 \times 10^9~N/m^2$, with Poisson ratios
$\underline{\nu}_{23} =0.36$, $\underline{\nu}_{31} =0.28$, $\underline{\nu}_{12} =0.064$, and with shear moduli $\underline{G}_{23} = \underline{G}_{31} = 79.6\times 10^9~N/m^2$ and $\underline{G}_{12} = 50.9\times 10^9~N/m^2$. Let $[\underline{C}]\in\MM_6^+$ be the Kelvin representation of the fourth-order elasticity tensor of this medium, which belongs to the class of orthotropic symmetry. The mass density is $2\,330~Kg/m^3$. There are three types of spring. Each spring acts in one direction.
In $x_1$-, $x_2$-, and $x_3$-directions, the stiffnesses are $\underline{k}^s_{1} = 4~N/m$, $\underline{k}^s_{2} = 6~N/m$, and $\underline{k}^s_{3} = 1.5~N/m$. A nonlinear elastic material is inserted between the two vertical beams of the suspended part and the vertical central beam of the mobile part. Its constitutive equation corresponds to a cubic, elastic, restoring force whose elastic constant is $k_b = 2\times 10^{12}~N/m$.\\

\noindent \textit{(iii) Time-dependent excitation}. The base $\Gamma_0$ of the mobile part is submitted to a given acceleration in the $x_1$-direction. The time-dependent modulation is
$\gamma(t) = \gamma_0 \, \big\{ \sin\big (t(\omega_c +\Delta\omega_c/2 )\big) -  \sin\big (t(\omega_c -\Delta\omega_c/2 )\big )\big\} / ( {\pi\, t})$,
in which $\gamma_0 = 120~m/s^2$, $\omega_c = 2\, \pi \times 13 \times 10^6~rad/s$ is the central angular frequency and where $\Delta\omega_c = 2\, \pi \times  10 \times 10^6~rad/s$ is the bandwidth. In the positive-frequency domain,  the energy of signal $\gamma(t)$ is mainly concentrated in the frequency band $[\omega_\pmin, \omega_e]$, with $\omega_\pmin = 2\pi\times  8\times 10^6~rad/s$ and $\omega_e = 2\pi\times  18\times 10^6~rad/s$.\\

\noindent \textit{(iv) Time interval of analysis, frequency band of analysis, and observations}.
This nonlinear dynamical system is analyzed for $t\in [t_0, T]$ with $t_0 = -1.3889$ $\times 10^{-5}~s$ and $T = 3.1014\times 10^{-5}~s$. At time $t_0$, the device is at rest.
The frequency band of analysis is $[0\, , \omega_\pmax ]$ with $\omega_\pmax  = 2\pi\times  72\times 10^6~rad/s$. The frequency band of observation is $\curB_o = [0, \omega_o]$, where $\omega_o=2\,\pi\times 70\times 10^6 ~rad/s$. If the dynamical system was linear, the energy of the response signal would be concentrated in the same frequency band $[\omega_\pmin, \omega_e]$ as that of the energy of the excitation signal. Due to the nonlinearity however, part of the energy of the excitation signal is transferred outside its frequency band and consequently, the frequency band of the response is not $[\omega_\pmin, \omega_e]$ but $[0, \omega_\pmax]$, where $\omega_\pmax  > \omega_e$.
The results will be presented for two selected observations $\pObs_1$ and $\pObs_6$.  The first one is the $x_1$-dof, denoted as $\dof_1$, of the node of coordinates $(14.5,8.0,0.0)\times 10^{-6}\, m$ located on the vertical beam of the mobile part. The second one is the $x_1$-dof, denoted as $\dof_6$,  of the node of coordinates $(23.0,17.0,0.0)\times 10^{-6}\, m$ located at the down right corner of the suspended part. The results will be presented in the frequency domain (in $Hz$) for each one of the two observations and will be the random function $\nu_\pfreq\mapsto \pdB_j(\nu_\pfreq) = \log_{10}(\vert\, \widehat Y_{\pdof_j}(2\pi\nu_\pfreq )\,\vert)$ for $j=1$ and $j=6$, in which $\omega \mapsto \widehat Y_{\pdof_j}(\omega)$ is the Fourier transform of the displacement $t \mapsto Y_{\pdof_j}(t)$.
\subsection{Prior probability model of uncertainties}
\label{SectionSD2}
%
\noindent \textit{(i) Prior probability model of the springs stiffnesses}.
Let $\bfK^s = (K_1^s,K_2^s,K_3^s)$ be the $\RR^3$-valued random variable defined on a probability space $(\Theta,\curT,\curP)$.
For $j\in\{1,2,3\}$, $K_j^s$ is the random stiffness of the spring of type $j$, with
mean value $\underline{k}^s_j$ and with a uniform probability distribution on $[\underline{k}_j^s(1-\sqrt{3}\, \delta_s) \, , \underline{k}_j^s(1+\sqrt{3}\, \delta_s)]$, in which $\delta_s =0.2$ for all $j$. Random variable $K^s_j$ can then be written as
$ K^s_j =  \underline{k}_j^s \big ( 1 + \sqrt{3}\, \delta_s(2\,\curU^s_j -1) \big )$,  in which $\curU^s_j$ is a uniform random variable on $[0,1]$ such that
$\curU^s_j  = \big (1+\hbox{erf}(W^s_j/\sqrt{2})\big )/2$, where $W^s_j$ is a normalized Gaussian random variable. The random variables $W^s_1$, $W^s_2$, $W^s_3$ are assumed to be independent and we introduce the $\RR^3$-valued random variable $\bfW^s = (W^s_1, W^s_2, W^s_3)$.\\

\noindent \textit{(ii) Prior probability model of the elasticity tensor of the silicon}.
The material in $\Omega$ is a random homogeneous medium. The Kelvin matrix representation $[\bfC]$ of the random fourth-order  elasticity tensor is defined on $(\Theta,\curT,\curP)$, independent of $\bfW^s$, with values in $\MM_6^+$. Its  mean value is matrix $[\underline{C}]\in \MM_6^+$ whose Cholesky factorization is $[\underline{C}] = [\underline{L}_C]^T\, [\underline{L}_C]$.  The statistical fluctuations of $[\bfC]$ around $[\underline{C}]$ can be constructed either in the class of orthotropic symmetry \cite{Guilleminot2013} or both in the orthotropic symmetry class and the anisotropic symmetry class \cite{Guilleminot2013a,Soize2017b}. Here, we consider the statistical fluctuations in the anisotropy symmetry class \cite{Soize2006,Soize2017b} and consequently, $[\bfC] = [C_\epsilon]+ [\underline{L}_\epsilon]^T\, [\bfG]\, [\underline{L}_\epsilon]$
in which $[C_\epsilon] = \epsilon (1+\epsilon)^{-1}[\underline{C}]$ and $[\underline{L}_\epsilon] = (1+\epsilon)^{-1/2}[\underline{L}_C]$
with $\epsilon = 10^{-6}$.
The $\MM_6^+$-valued random matrix $[\bfG]$ is written  as $[\bfG] =  [\bfL_G]^T \, [\bfL_G]$ in which $[\bfL_G]$ is a random $(6\times 6)$ upper triangular matrix defined as follows \cite{Soize2017b}.
Let $\{\curW_{ij}\}_{1\leq i \leq j\leq 6}$ be $21$ independent normalized Gaussian real-valued random variables defined on $(\Theta,\curT,\curP)$, independent of $\bfW^s$. The upper part of matrix $[\bfL_G]$ is such that
$[\bfL_G]_{ij} = (\delta_C /\sqrt{7})\,\curW_{ij}$ for $1 \leq i < j \leq 6$ and $[\bfL_G]_{ii} = (\delta_C /\sqrt{7}) \big ( 2\beta(\,\curW_{ii},\alpha_i)\big )^{1/2}$ for $i \in \big\{ 1,\ldots , 6 \big\}$ in which
$\alpha_i =7/(2\delta_C^2) + (1-i)/2$ and where $w \mapsto \beta (w,\alpha_i)$ is such that $\Gamma_{\alpha_i} =\beta(\,\curW_{ii},\alpha_i)$
is the Gamma random variable with parameter $\alpha_i$. We have  $E\big\{[\bfG]\big\} = [I_6]$.
The statistical fluctuations are controlled by the hyperparameter
$\delta_C = \big\{ (1/6) \, E \{ \Vert \,[\bfG] - [I_6]\,\Vert^2_F\} \big\}^{1/2}$ set here to the value $0.1$ in which $\Vert\cdot\Vert_F$ is the Frobenius norm.
Finally, we introduce the random vector $\bfW^C =(W^C_1,\ldots,W^C_{21})$ such that
$W_i^C = \curW_{ii}$ for $i \in \big\{ 1,\ldots , 6 \big\}$ and
$W_{6+k}^C = \curW_{ij}$ for $1\leq i < j\leq 6$ with $k =(i,j)\in \big\{ 1,\ldots , 15\big\}$.
\\

\noindent \textit{(iii) Random vector} $\bfW$.
Let $\bfW = (W_1,\ldots, W_{n_w})$ be the $\RR^{n_w}$-valued random variable defined by $\bfW = (\bfW^s,\bfW^C)$ with $n_w = 3 + 21 = 24$.
For constituting the training dataset, the $N_d$ realizations $\big\{ \bfw_{d}^\ell ,\ell=1,\ldots,N_d \big\}$ of $\bfW$ are generated with its prior probability model.
The random stiffnesses $\bfK^s$ and the random elasticity matrix $[\bfC]$  are defined by a given $\RR^{n_w}$-valued mapping $f_{\pparam}$  on $\RR^{n_w}$ such that
\begin{equation}
 \big\{\bfK^s,[\bfC]\big\} =  f_{\pparam}(\bfW) \, .                                                                                                  \label{eqSD2.1}
\end{equation}
\subsection{Computational model,  KL expansion, random residual, and PLoM parameters}
\label{SectionSD3}
\textit{(i) Discrete nonlinear dynamical equation}.  The semidiscretization of the boundary value problem of the nonlinear 3D dynamical system  is performed using the Finite Element (FE) method. The FE mesh is shown in  Fig.~1-(B) of Ref.~\cite{Soize2019a}. There are $7\,328$ eight-nodes solid elements, $10\,675$ nodes, and $N = 31\, 410$ free dofs.
In the reference coordinate system, the semidiscretized stochastic dynamical system is written as
\begin{align}
& [M]\,\ddot\bfY(t) + [D]\,\dot\bfY(t) + [K(\bfW)]\,\bfY(t) +\bff\pNL(\bfY(t)) = \gamma(t)\,\bff_M\,,
                                                                     \quad   t\in]t_0,T]\, ,                            \label{eqSD3.1}\\
& \bfY(t_0) = \bfzero_N \,\, , \,\, \dot\bfY(t_0) = \bfzero_N \, .                                                      \label{eqSD3.2}
\end{align}
In Eq.~\eqref{eqSD3.1}, $\bfY(t) = (Y_1(t),\ldots ,Y_N(t))$ is the vector of the $N$ free degrees of freedom of the displacements in the
reference coordinate system. The corresponding velocity and acceleration vectors are $\dot\bfY(t)$ and $\ddot\bfY(t)$. The matrices $[M]$, $[D]$, and $[K(\bfW)]$ are the FE mass, damping, and stiffness matrices with values in $\MM_N^+$.  The stiffness matrix depends on $\bfW$ via the use of Eq.~\eqref{eqSD2.1} and $\bff\pNL(\bfY(t))$ is the $\RR^N$-vector of the nonlinear internal forces at time $t$ induced by the nonlinear material. Finally, $\bff_M$ is the $\RR^N$ vector independent of time $t$, which represents the spatial part of the external forces in the reference coordinate system, due to the imposed acceleration of boundary $\Gamma_0$. Note that, for the dynamical system under consideration, $\bff_M$ is independent of the damping and stiffness matrices. The time-dependent load vector is such that $\Vert\,\bff_M\Vert = 9.64\times 10^{-12}$, $\max_t \vert\, \gamma(t)\,\vert$ $ = 2\times 10^7$, and consequently, $\max_t \Vert \,\gamma(t)\bff_M\Vert = 5.59\times 10^{-6}$.\\

\noindent \textit{(ii) Damping modeling}. Let $\bfvarphi^1, \ldots , \bfvarphi^{n_\pmode}$ be the first $n_\mode = 40$ elastic modes, normalized with respect to the mass matrix, associated with the first $n_\mode$ lowest eigenfrequencies  $0 < \omega_1 \leq \ldots \leq  \omega_{n_\pmode}$ such that
$[K(\underline \bfw)]\, \bfvarphi^{\,\beta} = \omega_\beta^2\, [M]\, \bfvarphi^{\,\beta}$
in which  $\underline \bfw = E\big\{\bfW\big\}$. We have  $\omega_1 = 2\pi\times  4.1341\times 10^5~rad/s$ and $\omega_\pmode = 2\pi\times  1.2400\times 10^8~rad/s$. There are $25$ elastic modes in the frequency band $\curB_o$. The computational model of matrix $[D]\in \MM_N^{+}$ is derived from the one  proposed Page 93 in \cite{Ohayon1998}, and is written as
$[D] = \sum_{\beta=1}^{n_\pmode} 2\,\chi_d\,(\omega_\beta - \omega_{n_\pmode})\, [M]\, \bfvarphi^{\,\beta}\, ([M]\, \bfvarphi^{\,\beta})^T
                     + 2\, \chi_d\, \omega_{n_\pmode} \, [M]$,
in which $\chi_d = 0.02$. This damping model yields a constant damping rate $\chi_d$ for the first $n_\mode$ elastic modes and yields the damping rates $\big\{ \chi_d \, \omega_{n_\pmode} / \omega_{n_\pmode+1},\ldots , $ $\chi_d\, \omega_{n_\pmode}/\omega_N \big\}$ for the $N-n_\mode$ higher elastic modes.\\

\noindent \textit{(iii) Time sampling and discretization scheme}.
Let $\Delta t = 6.9444\times 10^{-9} s$ be the time step. The sampling times are $t_1,\ldots, t_{n_\pptime}$ such that
$ t_n = t_0 + n \, \Delta t$ for $n= 1,\ldots n_\ptime$
in which $n_\ptime = 6\,466$ and $T - t_0 = n_\ptime \, \Delta t = 4.4903\times 10^{-5} s$.  Using Eqs.~\eqref{eqSD3.1} and
\eqref{eqSD3.2} yields, because $\bff_\pNL(\bfzero_N) = \bfzero_N$,
\begin{equation}
 \bfY(t_0) = \bfzero_N\,\, , \,\, \dot\bfY(t_0) = \bfzero_N \,\, , \,\, \ddot\bfY(t_0) = \ddot\bfy(t_0) \,\, , \,\,
                              \ddot\bfy(t_0) = \gamma(t_0)\, [M]^{-1} \,\bff_M   \, .                                                         \label{eqSD3.3}
 \end{equation}
The centered Newmark integration method (implicit scheme) is used. At each sampling time, the nonlinear algebraic equation is solved by using a fixed point method with a relative precision of $10^{-6}$. In order to guarantee the convergence of the fixed point method, a local adaptive time step is used.
Knowing $\bfY(t_{n-1})$, $\dot\bfY(t_{n-1})$, and $\ddot\bfY(t_{n-1})$ at sampling time $t_{n-1}$, and knowing $\bfY(t_{n})$, the following equations are used  for calculating $\ddot\bfY(t_{n})$ and $\dot\bfY(t_{n})$,
\begin{align}
& \ddot\bfY(t_{n}) = a_0 (\bfY(t_{n}) - \bfY(t_{n-1})) - a_1\dot\bfY(t_{n-1}) - a_2\ddot\bfY(t_{n-1}) \, ,                                          \label{eqSD3.4}\\
& \dot\bfY(t_{n})  = \dot\bfY(t_{n-1}) + a_3(\ddot\bfY(t_{n-1}) +  \ddot\bfY(t_{n})) \, ,                                                         \label{eqSD3.5}
\end{align}
in which $a_0 = 4/\Delta t^2$, $a_1 = 4/\Delta t$, $a_2=1$, and $a_3 = \Delta t/2$.\\

\noindent \textit{(iv) Computational model}.
The FE discretization and time discretization scheme previously defined allow for computing the $\RR^N$-valued time series
$\big\{ (\bfY(t_n), \dot\bfY(t_n), \ddot\bfY(t_n)) , n =1,\ldots , n_\ptime \big\}$.  In that condition,  Eq.~\eqref{eq2.1} is rewritten, at sampling time $t_n$,  as
\begin{equation}
 \bfcurN^{\,\pSD}(\ddot\bfY(t_n),\dot\bfY(t_n),\bfY(t_n),t_n,\bfW) = \bfzero_{N} \quad a.s. \, ,                                                  \label{eqSD3.6}
\end{equation}
in which  $\bfcurN^{\,\pSD}$ is a nonlinear operator with values in $\RR^{N}$.\\

\noindent \textit{ (v)  KL expansion of the nonstationary stochastic processes} $\bfY$.
The KL expansion of the nonstationary $\RR^N$-valued stochastic process
$\big\{\bfY(t), t\in[t_0, T]\big\}$ is performed at the sampling times, $t_1,\ldots, t_{n_\pptime}$, yielding the representation
$\bfY(t_n) = \underline{\bfy}(t_n) + [V(t_n)] \,\bfQ$.
Knowing $\underline{\bfy}(t_n)$ and $[V(t_n)]$ for $n\in\big\{1,\ldots ,n_\ptime\big\}$, their first and second time derivatives are computed by the following recurrences
based on the centered Newmark integration method used in Eqs.~\eqref{eqSD3.4} and \eqref{eqSD3.5}.

(a) Initialization of the recurrences.   $\bfY(t_0) = \dot\bfY(t_0) = \bfzero_N$  and $\ddot\bfY(t_0) =\ddot\bfy(t_0)$ in which
$\ddot\bfy(t_0)$ is defined in Eq.~\eqref{eqSD3.3}. Choosing $\underline{\bfy}(t_0) = \underline{\dot\bfy}(t_0) = \bfzero_N$  and $\underline{\ddot\bfy}(t_0) =\ddot\bfy(t_0)$, it can then be deduced that $[V(t_0)]= [\dot V(t_0)]=[\ddot V(t_0)]=[0_{N,n_q}]$.

(b) Recurrences.  For computing the mean function $t\mapsto \underline{\bfy}(t)$, the recurrence is written,  for $n\in\big\{1,\ldots ,n_\ptime\big\}$,
as $\underline{\ddot\bfy}(t_n) = a_0 \, (\underline{\bfy}(t_n) - \underline{\bfy}(t_{n-1}) )
                   - a_1\, \underline{\dot\bfy}(t_{n-1}) - a_2\,\underline{\ddot\bfy}(t_{n-1})$ with
$ \underline{\dot\bfy}(t_n)  = \underline{\dot\bfy}(t_{n-1}) + a_3\, (\underline{\ddot\bfy}(t_{n-1}) +  \underline{\ddot\bfy}(t_n) )$,
while for the function $t\mapsto [V(t)]$, the recurrence is
$[\ddot V(t_n)] = a_0 \, ([V(t_n)] - [V(t_{n-1})] )- a_1\, [\dot V(t_{n-1})] - a_2\,[\ddot V(t_{n-1})]$
with $[\dot V(t_n)]  = [\dot V(t_{n-1})] + a_3\, ([\ddot V(t_{n-1})] +  [\ddot V(t_n)] )$. Consequently, the KL expansions of stochastic processes
$\bfY$, $\dot\bfY$, and $\ddot\bfY$ at the sampling times $t_1,\ldots, t_{n_\pptime}$ are
\begin{equation}
 \bfY(t_n) = \underline{\bfy}(t_n) + [V(t_n)] \,\bfQ   \quad , \quad
 \dot\bfY(t_n) = \underline{\dot\bfy}(t_n) + [\dot V(t_n)] \,\bfQ   \quad , \quad
 \ddot\bfY(t_n) = \underline{\ddot\bfy}(t_n) + [\ddot V(t_n)] \,\bfQ   \, .                                                                  \label{eqSD3.7}
\end{equation}
For $N_d=120$, Figs.~\ref{figureSD1}-(a) and -(b) show the distribution of the eigenvalues $\alpha\mapsto \Lambda_\alpha$ of the KL-expansion of stochastic
process $\big\{\bfY(t), t\in [t_0,T]\big\}$ and the error function $n_q\mapsto \err_\KL(n_q)$ defined by Eq.~\eqref{eq3.7}. Fixing the tolerance $\varepsilon_\KL$ to the value $10^{-6}$ yields the optimal value $n_q=107$.\\
\begin{figure}[h!]
  \centering
  \includegraphics[width=4.5cm]{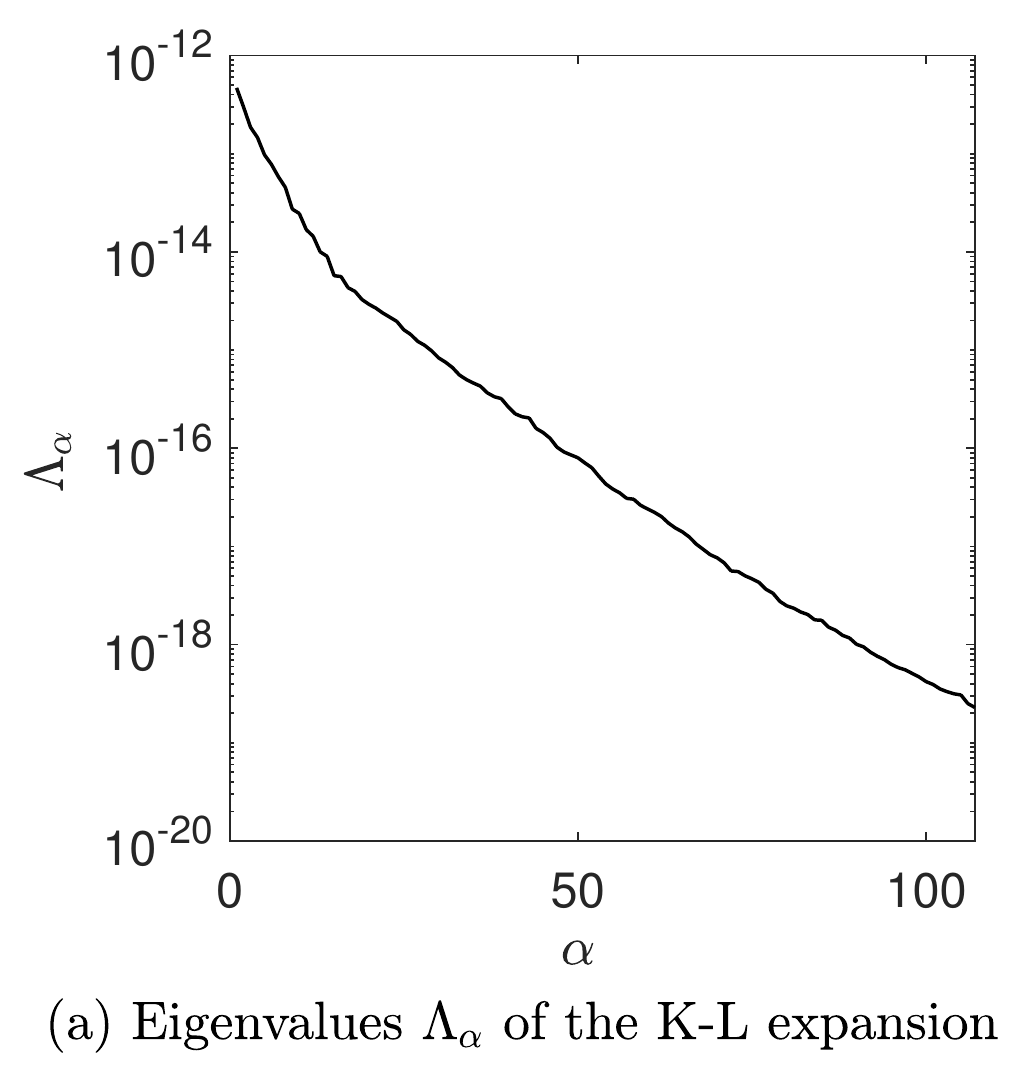}  \includegraphics[width=4.5cm]{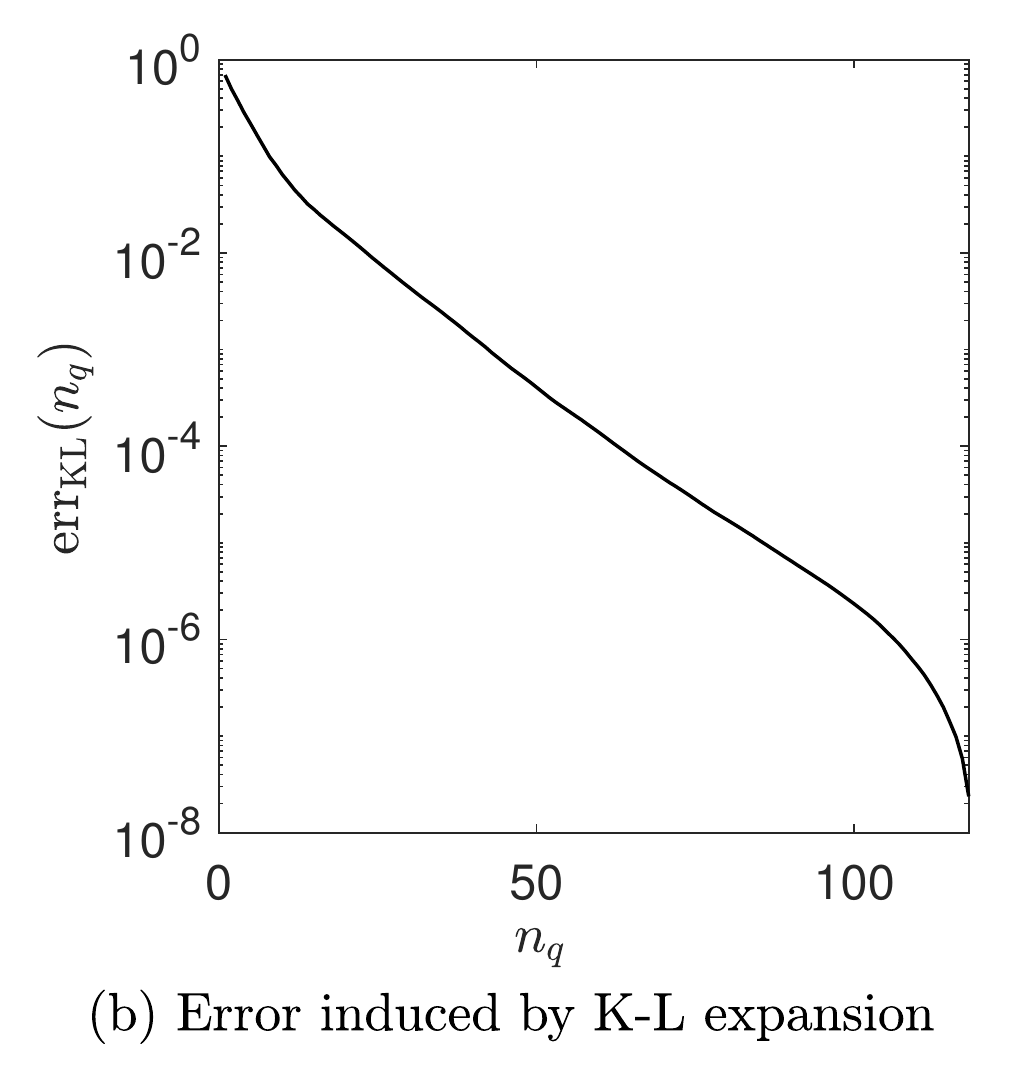} \includegraphics[width=4.5cm]{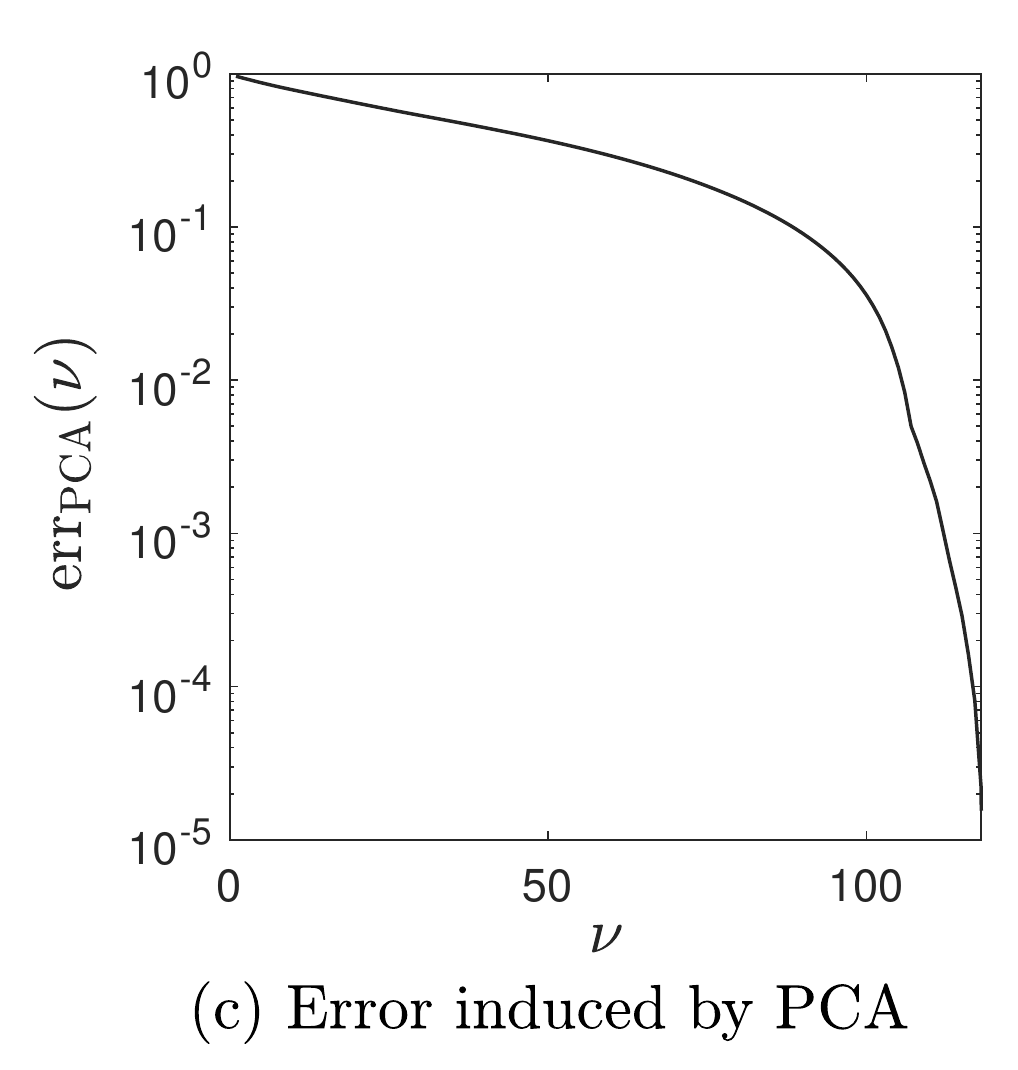}
\caption{For $N_d = 120$, for $T-t_0 =4.4903\times 10^{-5} s$, and for $n_\ptime=6\,466$,  KL expansion of nonstationary stochastic process $\bfY$:  (a) distribution of the eigenvalues $\alpha\mapsto \Lambda_\alpha$ and (b) error function $n_q\mapsto \errp_\KL(n_q)$. PCA of random vector $\bfX$: (c) error function $\nu\mapsto \errp_\PCA(\nu)$. }
\label{figureSD1}
\end{figure}
%

\noindent \textit{(vi) Calculation of random residual $\widehat\rho$ using the subsampling times} .
Let $b(t_n) = (1/N_d) \sum_{\ell=1}^{N_d} \Vert\,\bfy^\ell(t_n)\,\Vert$  be the estimate of the mean value of $\Vert\,\bfY(t_n)\,\Vert$ computed with the training dataset.  The subsampling times  $\big \{\tau_{n_s}, n_s = 1,\ldots , n_\psp\big \}\subset \big \{t_n, n = 1,\ldots , n_\ptime\big \}$ are chosen as the $n_\psp$ sampling times for which
$b(\tau_1), \ldots ,$ $b(\tau_{n_\psp})$ correspond to the $n_\psp$ largest values of the set $\big \{ b(t_1), \ldots , b(t_n)\big \}$.
For calculating the random residual with the KL expansions, the family $\big \{ \bfY(\tau_{n_s}),$ $\dot\bfY(\tau_{n_s}),$ $\ddot\bfY(\tau_{n_s}),$ $n_s=1,\ldots ,n_\psp\big \}$ is extracted from the family $\{ \bfY(t_n),$ $\dot\bfY(t_n),$ $\ddot\bfY(t_n),$ $n=1,\ldots ,n_\ptime\}$ computed with Eq.~\eqref{eqSD3.7}.
Taking into account Eq.~\eqref{eqSD3.6} at subsampling time $\tau_{n_s}$, the $\RR^N$-valued random residual $\bfcurR(\tau_{n_s})$,
 defined by Eq.~\eqref{eq4.2}, is rewritten as
\begin{equation}
\bfcurR(\tau_{n_s})  =  \bfcurN^{\,\pSD}(\ddot\bfY(\tau_{n_s}),\dot\bfY(\tau_{n_s}),\bfY(\tau_{n_s}),\tau_{n_s},\bfW) \quad a.s.                \label{eqSD3.10}
\end{equation}
The positive-valued random variable $\curR_\pnorm(\tau_{n_s})$ is defined by Eq.~\eqref{eq4.4} and writes
$\curR_\pnorm(\tau_{n_s}) = N^{-1/2}\, \Vert \,\bfcurR(\tau_{n_s})\,\Vert$.
The random residual $\widehat\rho$ is then calculated using  Eq.~\eqref{eq4.10} with $n_\psp =60$.
The mean value $\underline{\widehat\rho}_{\pref}$  is calculated for the fixed value $N_{d,\pref}= 120 $.
For $N_d=120$ and using the KL expansion, the mean value of  $\widehat\rho$ is $5.48 \times 10^{-7}$ and $\Vert \,\widehat\rho\,\Vert_{L^2} = 6.93\times 10^{-7}$.\\

\noindent \textit{(vii) Values of the parameters for the PLoM without and under constraints}. We consider always the case $N_d = 120$ for which the KL expansion of stochastic process $\bfY$ has been performed. The dimension of $\bfX = (\bfQ,\bfW)$ is $n_x = n_q + n_w = 107 + 24 = 131$.
Concerning the PCA of $\bfX$, Fig.~\ref{figureSD1}-(c) shows the error function $\nu\mapsto \err_\PCA(\nu)$ defined by Eq.~\eqref{eq5.3}.
Fixing the tolerance $\varepsilon_\PCA$ to the value $10^{-3}$ yields the optimal value $\nu=112$.
For the PLoM without constraint or under constraints, the diffusion-maps basis is computed with $\varepsilon_\pdiff = 240$, which yields  the optimal dimension $m=113$.
Concerning the St\"{o}rmer-Verlet algorithm for solving the nonlinear ISDE: $f_0 =4$ and $\Delta r = 0.2147$.
\subsection{Stochastic solution of reference and predictions by the PLoM without constraint}
\label{SectionSD4}
%

\noindent\textit{(i) Solution of reference}.
A reference solution has been computed using $N_d=1\,000$  realizations from the training dataset requiring a significant CPU effort.  Figure~\ref{figureSD2} shows, for each one of the two observations, two types of graphs. The first type corresponds to the deterministic function $\nu_\pfreq\mapsto \pdB_{\linear,j}(\nu_\pfreq)$ computed by using the deterministic linear computational model for which the control parameters are fixed to $\underline{\bfw}$ and  the nonlinear material between the beams are removed. The second type corresponds, for the probability level $p_c=0.98$, to the confidence region  of the random function $\nu_\pfreq\mapsto \pdB_j(\nu_\pfreq)$ computed using with the nonlinear stochastic computational model estimated using the $N_d=1\,000$ realizations. It can be observed that
the effect of the structural nonlinearity is very important outside the frequency band $[\omega_\pmin, \omega_e]$ of the excitation.
\begin{figure}[h!]
  \centering
  \includegraphics[width=6.0cm]{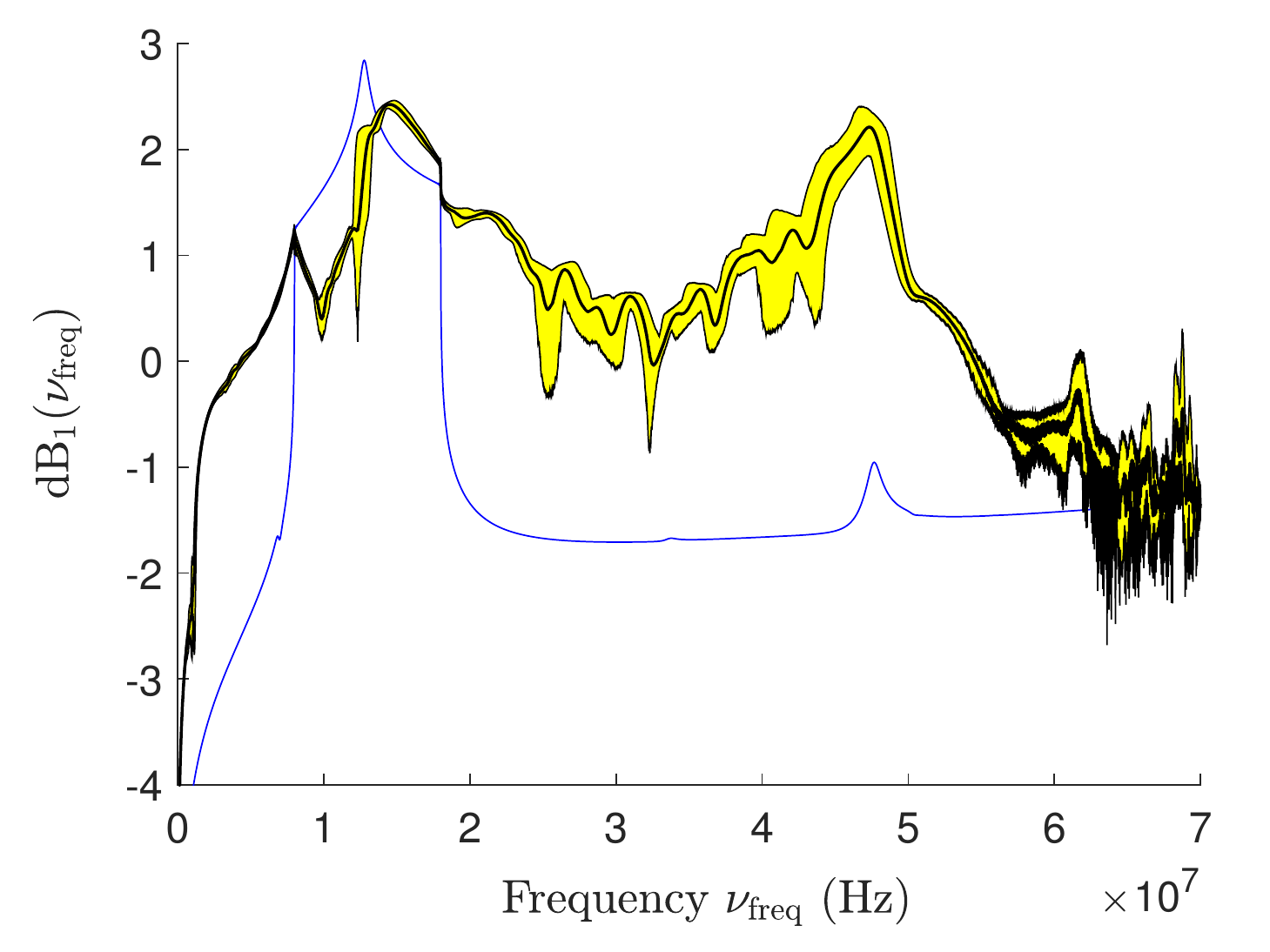} \hfil \includegraphics[width=6.0cm]{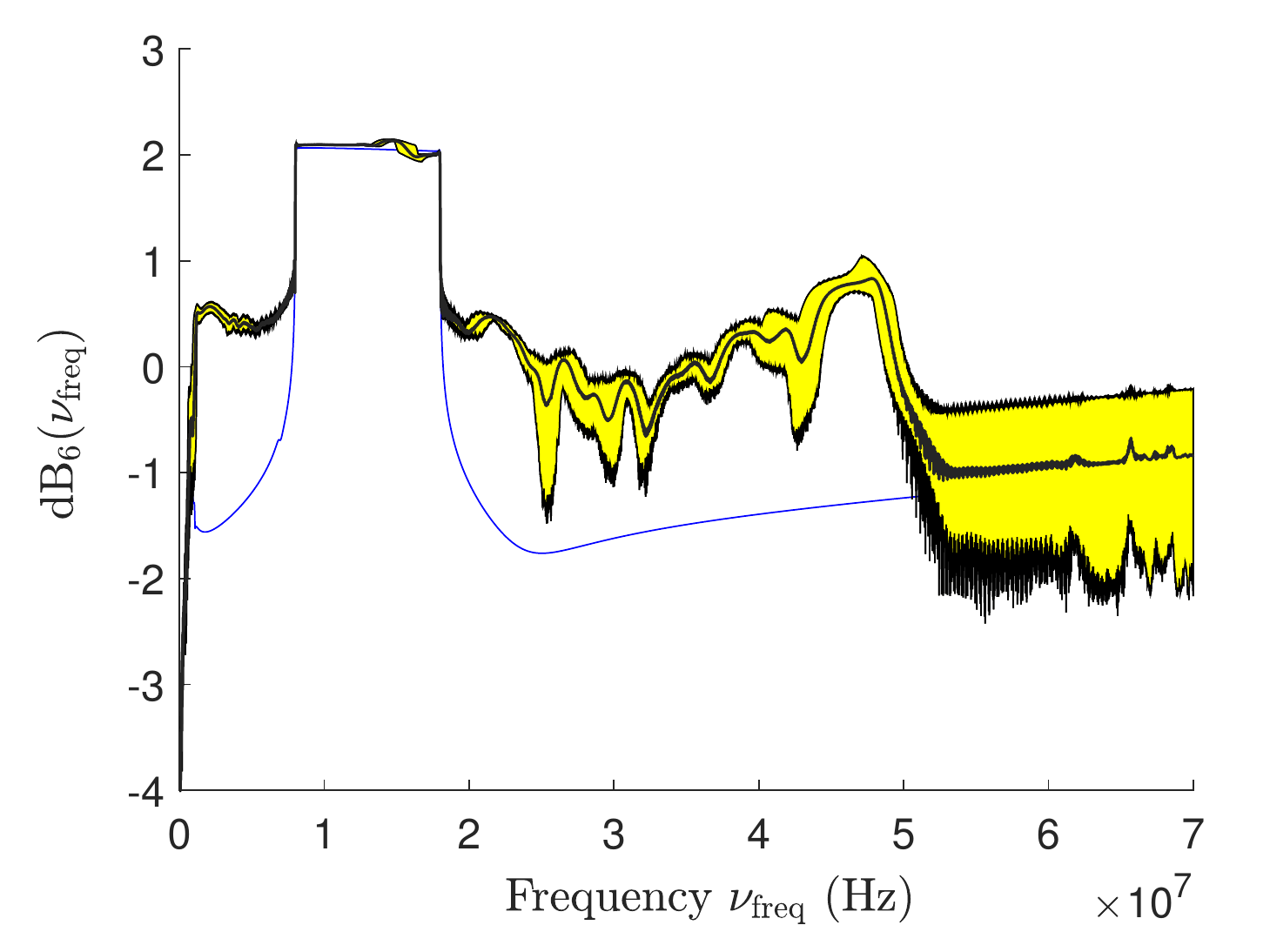}
\caption{Reference stochastic solution for $N_d=1\,000$: for $\ppObs_j$, $j=1$ (left figure) and $j=6$ (right figure), graph of the deterministic function $\nu_\pfreq\mapsto \pdB_{\linear,j}(\nu_\pfreq)$ (blue thin line),  and graph of the confidence region (yellow region with black thin lines for the upper and lower envelopes) of the random function $\nu_\pfreq\mapsto \pdB_j(\nu_\pfreq)$. The black thick line inside the yellow region is the estimate of mean value $\nu_\pfreq\mapsto E\{\pdB_j(\nu_\pfreq)\}$.}
\label{figureSD2}
\end{figure}
Figure~\ref{figureSD3}-(a) displays the graphs of the pdf $r\mapsto p_{\rho_{N_d}}(r)$ of $\rho_{N_d}= \widehat\rho / {\underline{\widehat\rho}}_{N_d}$ in which ${\underline{\widehat\rho}}_{N_d} = E\big\{\widehat\rho\big\}$ is estimated using $N_d$ realizations. We then have $E\big\{\rho_{N_d}\big\} = 1$ and the standard deviations are, for $N_d=30$, $60$, $120$, $440$, and $1\,000$, respectively, $0.080$, $0.094$, $0.091$, $0.087$, and $0.087$. Figure~\ref{figureSD3}-(b) shows the graph  $N_d\mapsto\Vert \,\widehat\rho\, \Vert_{L^2}(N_d)$.
These results show that the convergence with respect to $N_d$ is reasonably obtained for $N_d=1\,000$, that is the value selected for the reference.\\
\begin{figure}[h!]
  \centering
  \includegraphics[width=5.0cm]{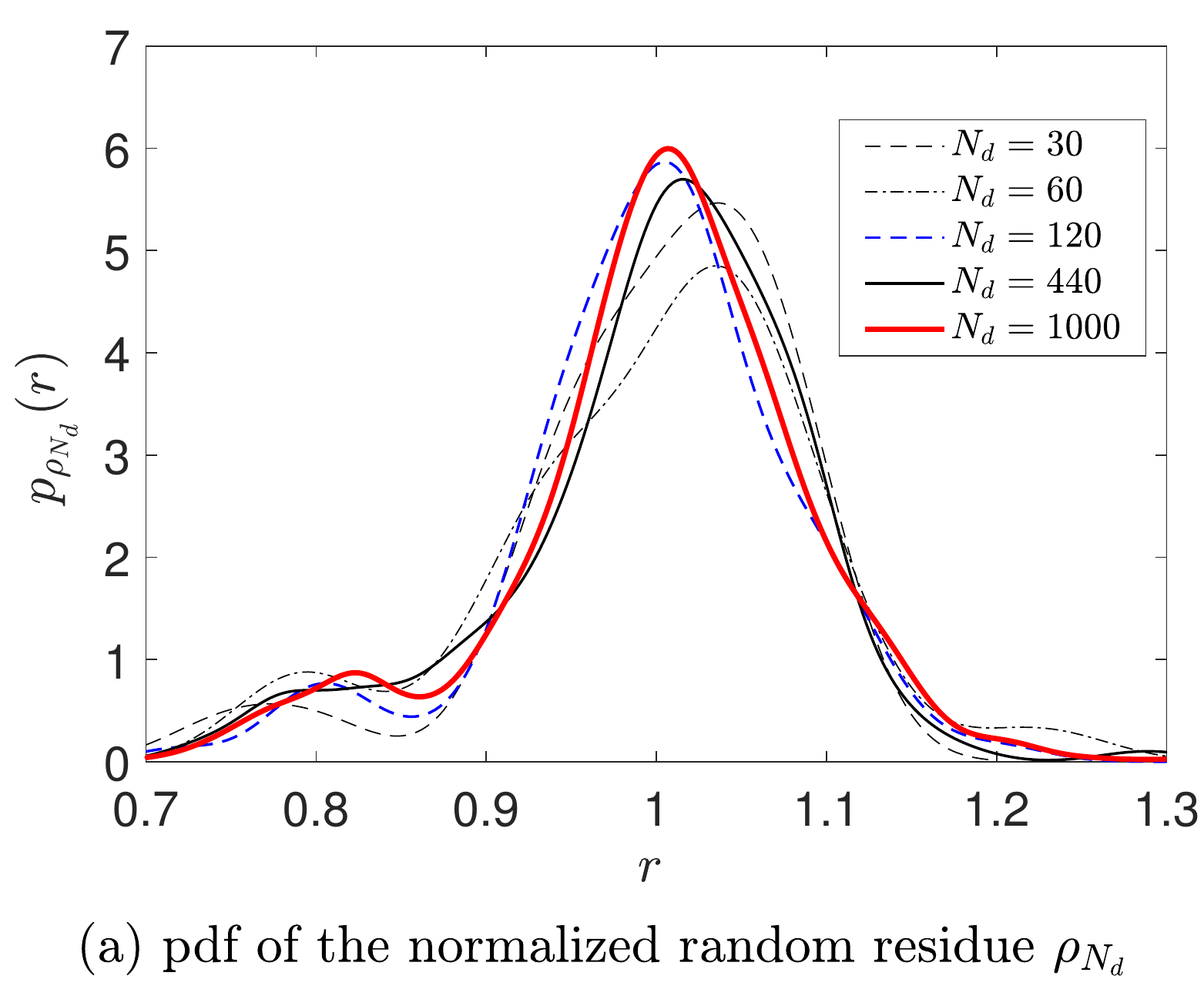}\hfil \includegraphics[width=5.0cm]{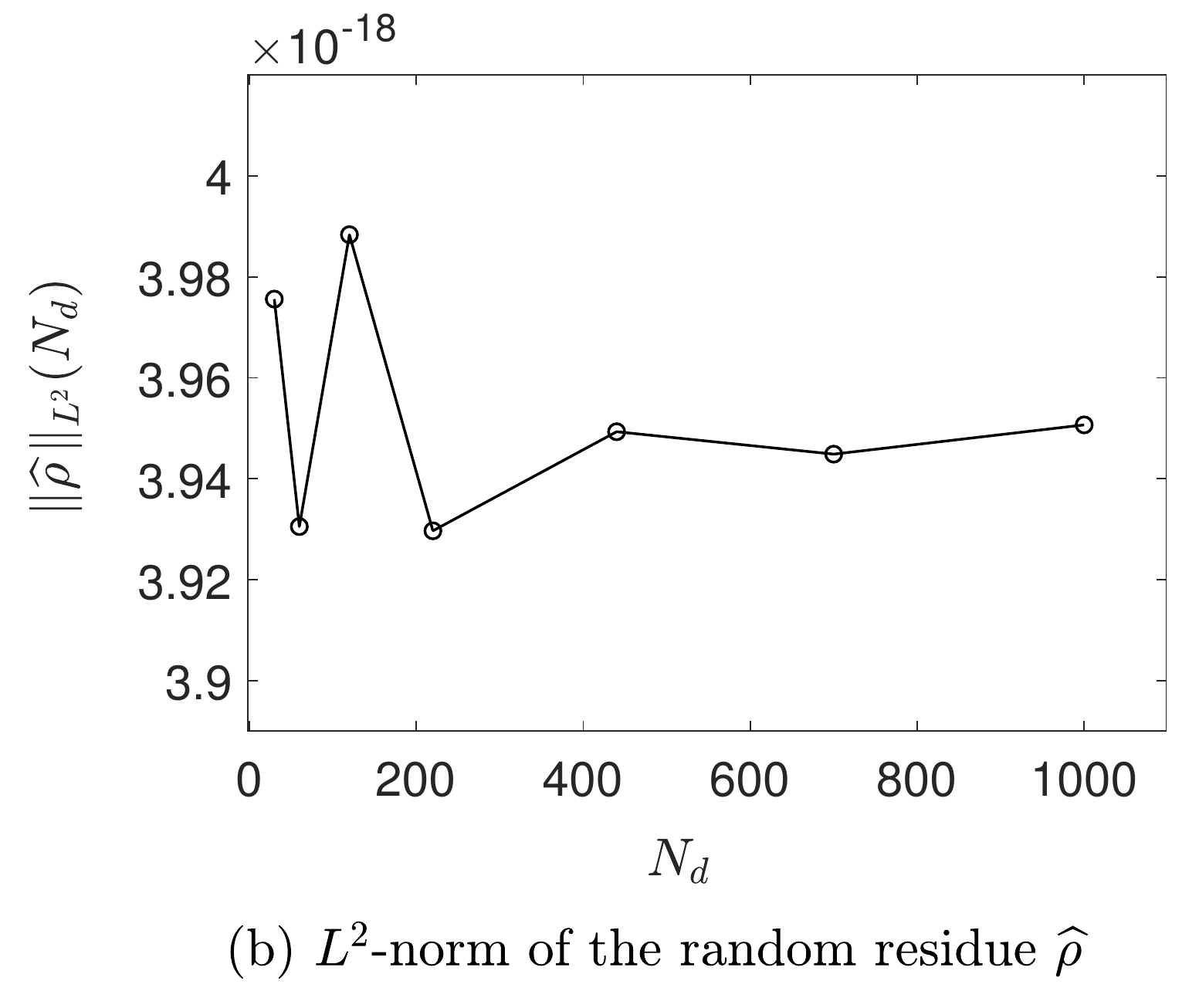}
\caption{Reference stochastic solution: (a) evolution of the pdf $r\mapsto p_{\rho_{N_d}}(r)$ of $\rho_{N_d}$ as a function of the number $N_d$ of realizations in the training dataset;  (b)  $N_d\mapsto \Vert \,\widehat\rho\,\Vert_{L^2}(N_d)$ of the random residual $\widehat\rho$.}
\label{figureSD3}
\end{figure}
%

\noindent\textit{(ii) Predictions using the PLoM without constraint}.
For $N_d=120$ and using the nonlinear stochastic computational model, for each one of the two observations and
for the probability level $p_c=0.98$, Fig.~\ref{figureSD4} shows the confidence region of the random function $\nu_\pfreq\mapsto \pdB_j(\nu_\pfreq)$ estimated using $n_\pMC=1\,000$ additional realizations generated with the PLoM without constraint. These confidence regions are almost identical to those of the reference stochastic solution shown in Fig.~\ref{figureSD2} (see also the discussion given in Section~\ref{SectionSD7}).
\begin{figure}[h!]
  \centering
  \includegraphics[width=6.0cm]{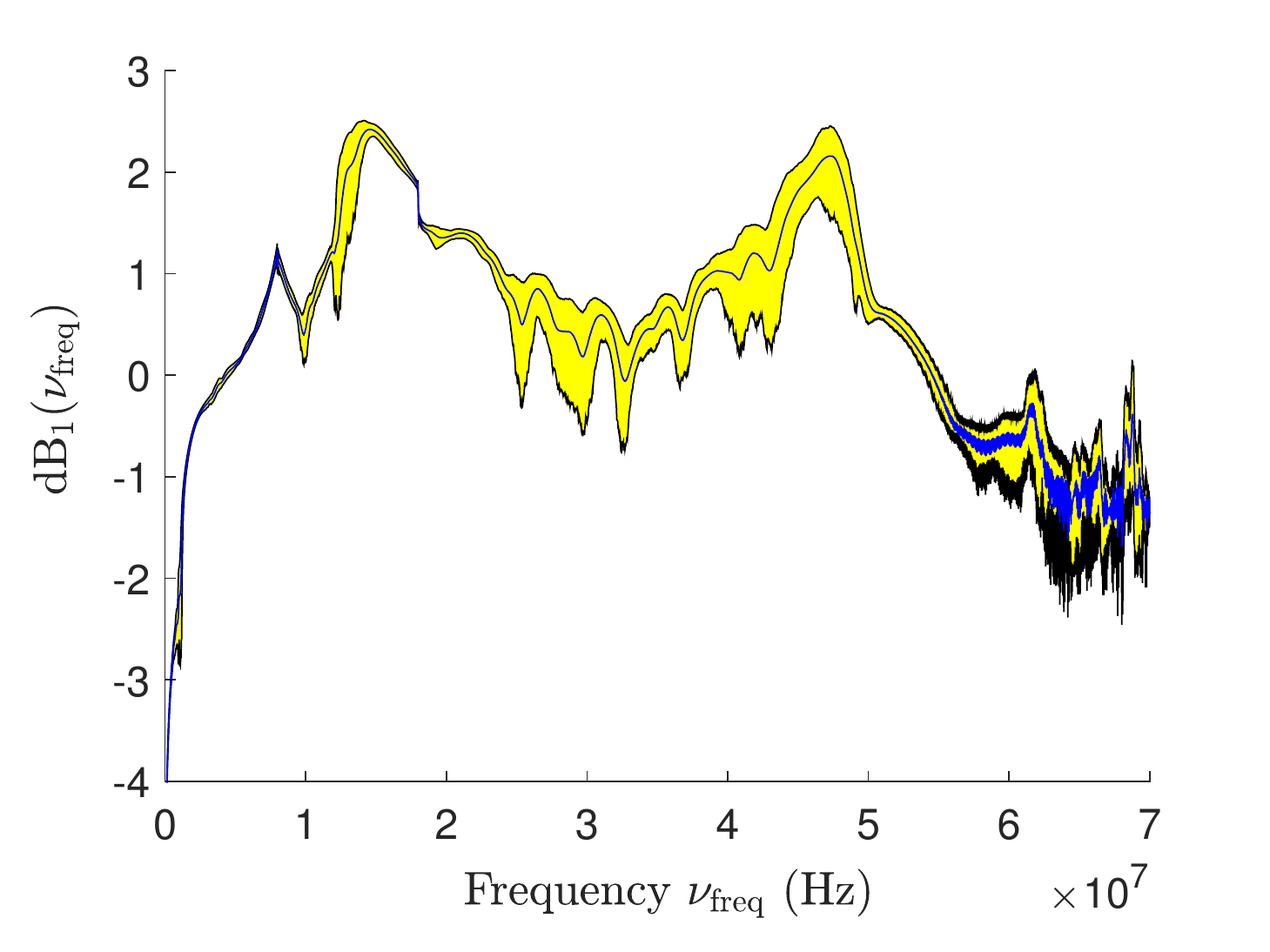} \hfil \includegraphics[width=6.0cm]{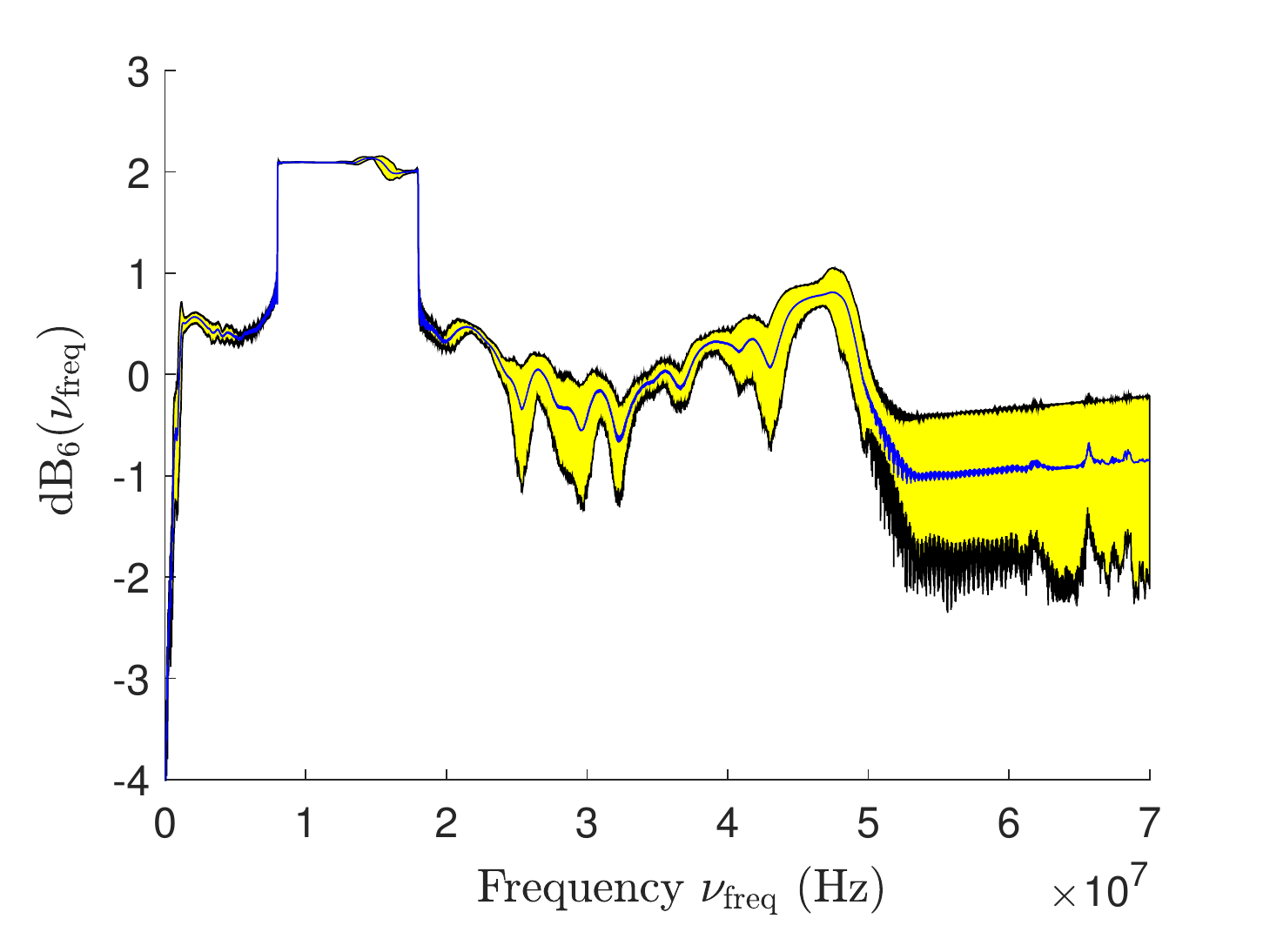}
\caption{For $N_d=120$, for $n_\pMC=1\,000$ additional realizations generated by the PLoM without constraints, for $\ppObs_j$, $j=1$ (left figure) and $j=6$ (right figure),  graph of the confidence region (yellow region with thin lines for the upper and lower envelopes) of the random function $\nu_\pfreq\mapsto \pdB_j(\nu_\pfreq)$. The thick line inside the yellow region is the estimate of mean value $\nu_\pfreq\mapsto E\{\pdB_j(\nu_\pfreq)\}$.}
\label{figureSD4}
\end{figure}

\subsection{Analysis of the role played by constraints on the probability distribution of random control parameter $\bfW$}
\label{SectionSD5}
As previously, for simplifying the writing, $\bfW^{(\nu)}$ is simply noted $\bfW$.
 For $N_d = 120$, $n_\pMC = 1\,000$, and using $\algo~1$, Fig.~\ref{figureSD5}-(a) shows the decrease in the error function $i\mapsto \err_R(i)$ while Fig.~\ref{figureSD5}-(b) shows the growth of function $i\mapsto\err_W(i)$ for $i > 24$, proving that the constraint on $\bfW$ must be taken into account.
\begin{figure}[h!]
  \centering
  \includegraphics[width=5.0cm]{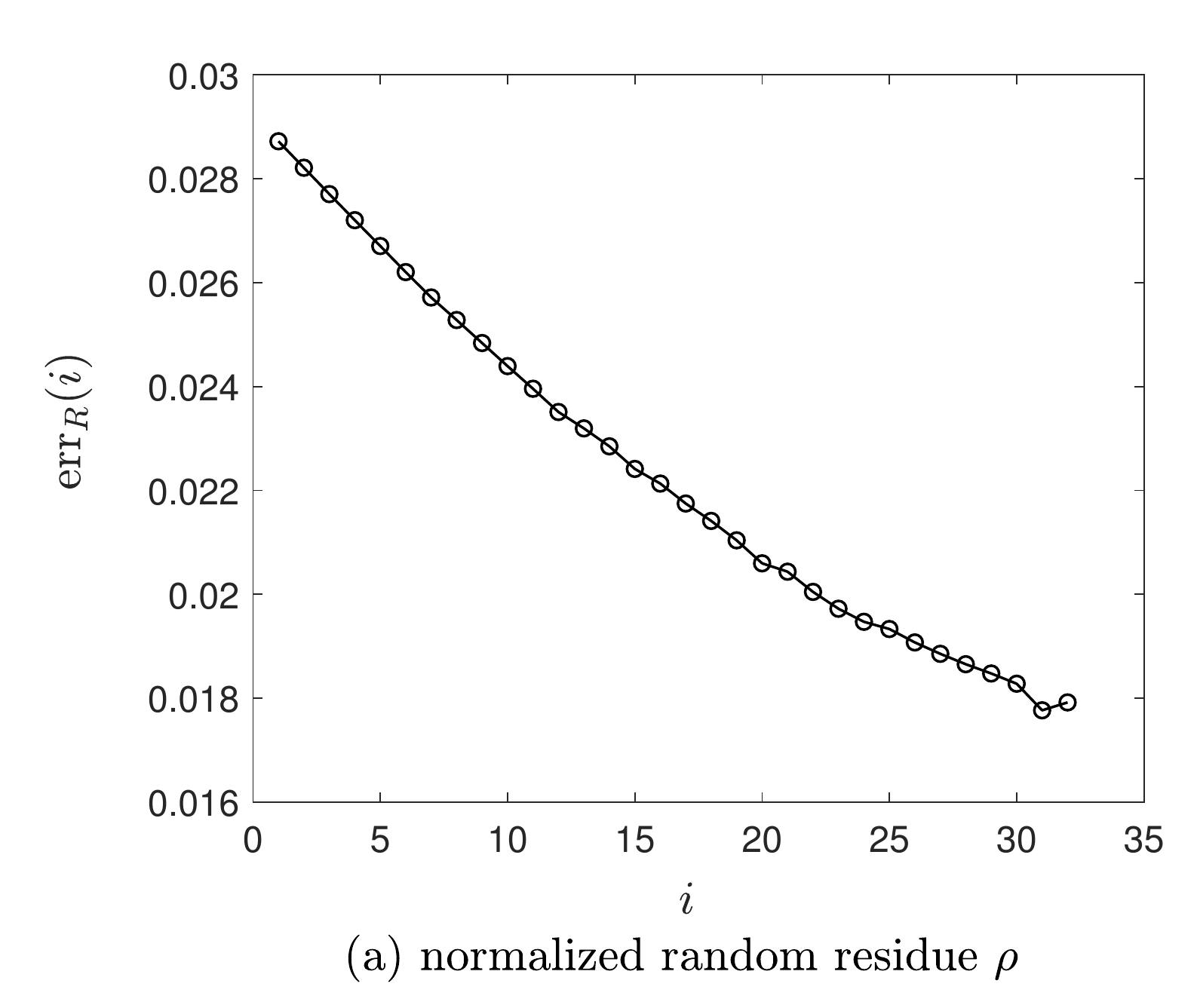} \hfil \includegraphics[width=5.0cm]{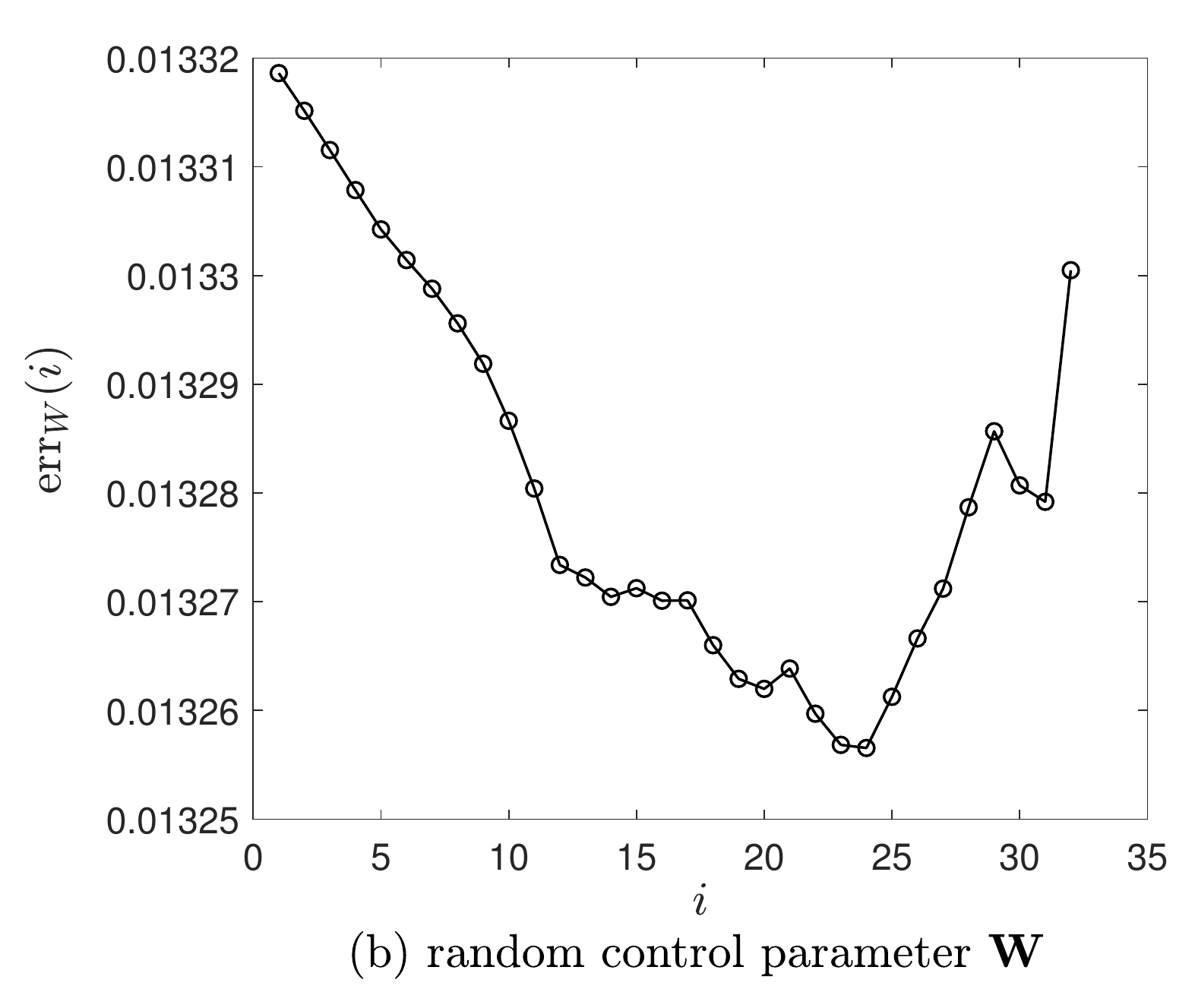}
\caption{For $N_d = 120$ and with $n_\pMC = 1\,000$ for the PLoM computation under the constraint on $\rho$ but without constraint on $\bfW$ ($\algo~1$): (a) function $i\mapsto \errp_R(i)$ and (b) function $i\mapsto \errp_W(i)$.}
\label{figureSD5}
\end{figure}
\subsection{Results of the PLoM without and under constraints for a given dimension of the training dataset}
\label{SectionSD6}
The training dataset has $N_d = 120$ realizations and  $n_\pMC\gg N_d$ additional realizations are generated with the PLoM under constraints using algorithm $\algo~3$.

(i) Figure~\ref{figureSD7} shows the pdf $r\mapsto p_{\rho}(r)$ of $\rho$ as a function of the number $n_\pMC$ of additional realizations generated with the PLoM without any constraint and under the constraints. The analysis of these figures shows a good convergence of the pdf of $\rho$ with respect to $n_\pMC$.
\begin{figure}[h!]
  \centering
  \includegraphics[width=4.5cm]{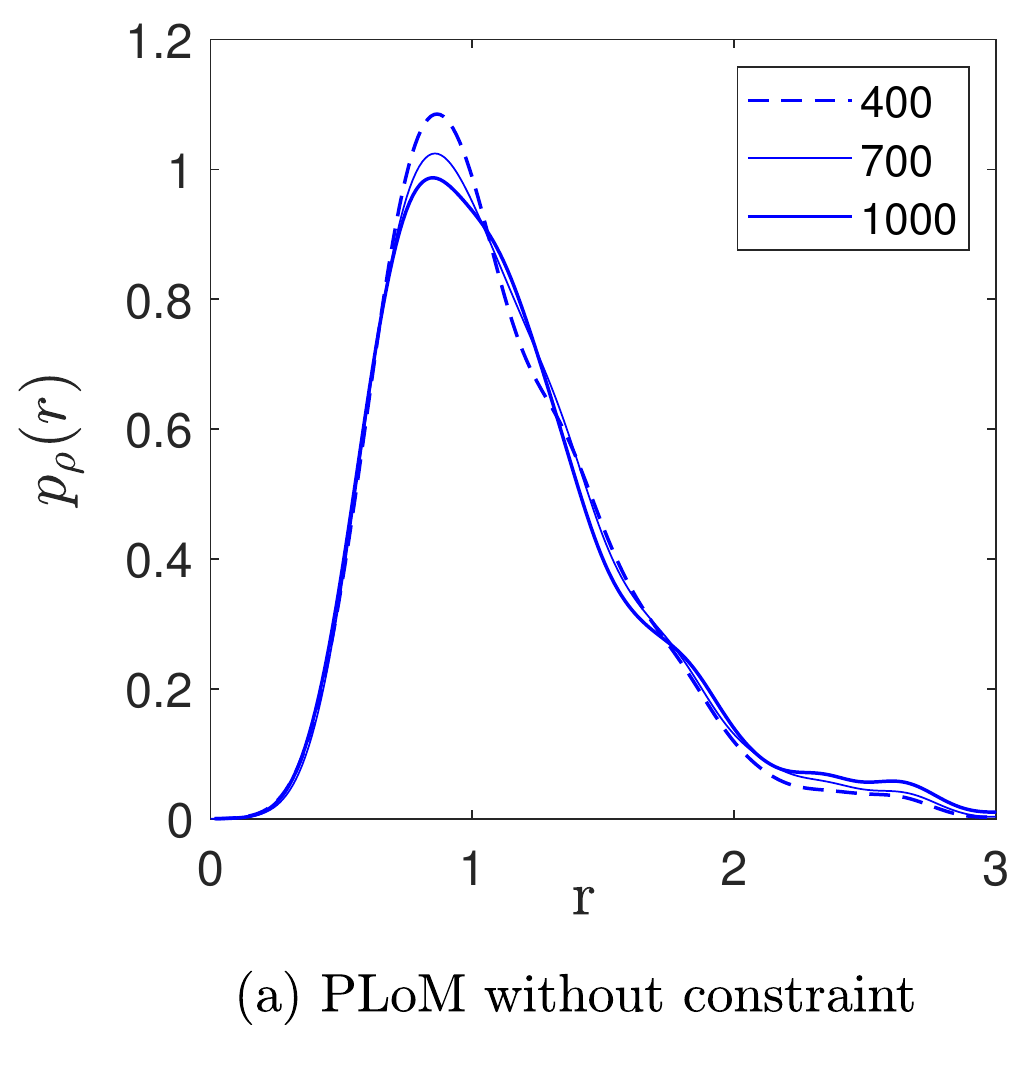}  \includegraphics[width=4.5cm]{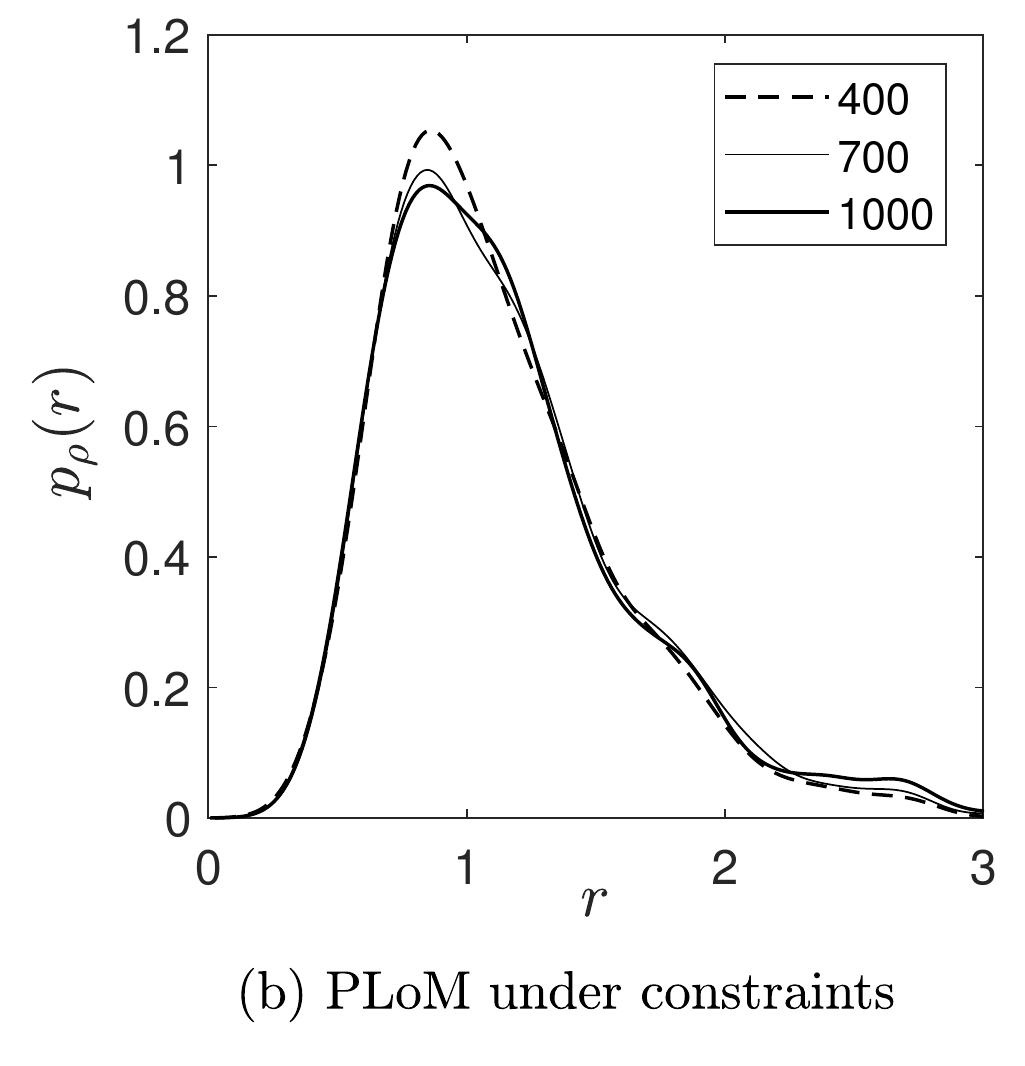} \includegraphics[width=4.5cm]{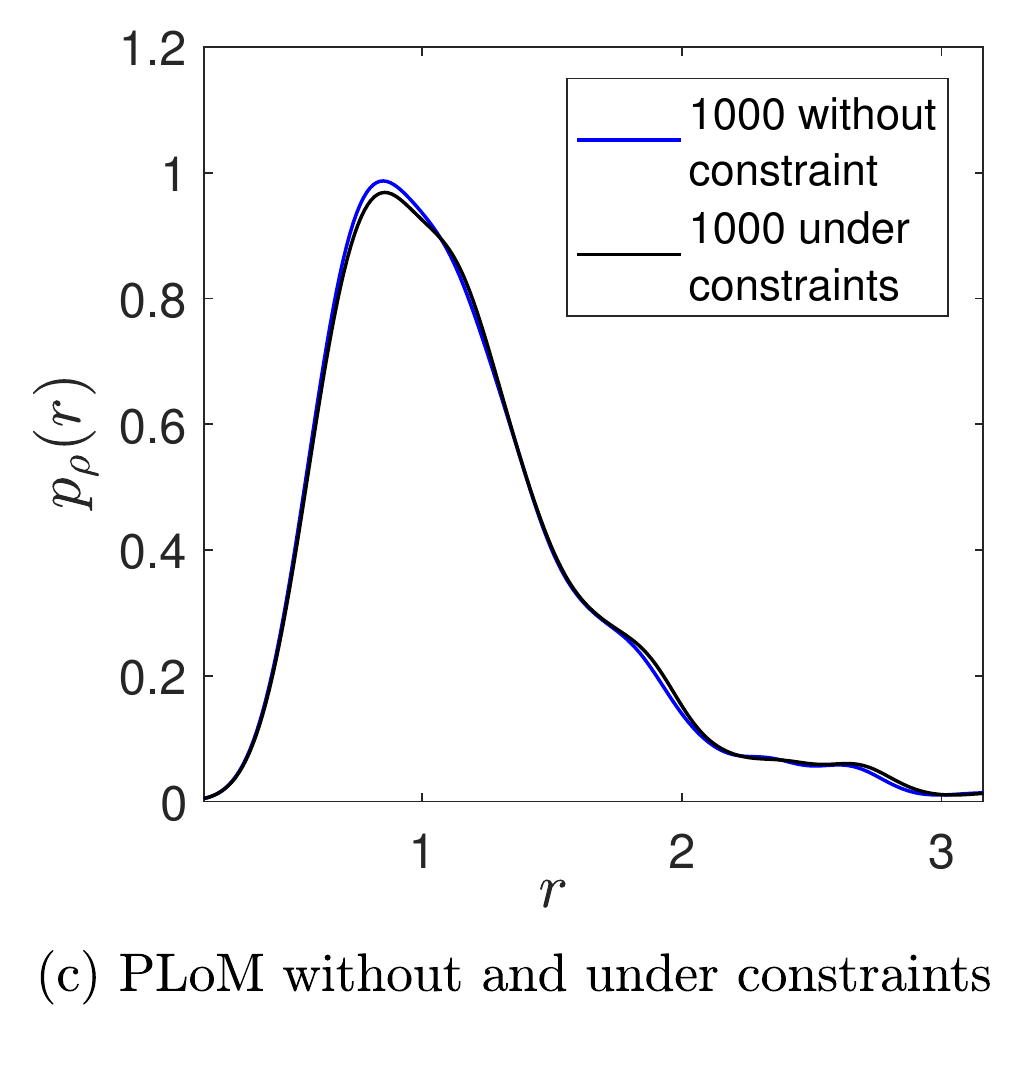}
\caption{For $N_d = 120$ and using $\algo~3$, graph of pdf $r\mapsto p_{\rho}(r)$ of $\rho$  as a function of $n_\pMC$, estimated with the PLoM without constraint (a), under constraints (b), and for $n_\pMC=1\,000$, comparison of the estimates obtained with the PLoM without and under constraints (c).}
\label{figureSD7}
\end{figure}

(ii) The case $N_d = 120$ is analyzed with $n_\pMC=1\,000$ additional realizations generated with the PLoM under constraints.  Figure~\ref{figureSD8} shows the evolution of the error functions $i\mapsto\err_R (i)$, $i\mapsto\err_W (i)$, and $i\mapsto\err_{RW} (i)$.
There is a local minimum for $i_\opt = 72$ such that $\err_{RW}(72)= 0.0032$,  for which $\err_R(72)= 8.9\times 10^{-6}$ and $\err_W(72)= 0.0023$.
\begin{figure}[h!]
  \centering
  \includegraphics[width=4.5cm]{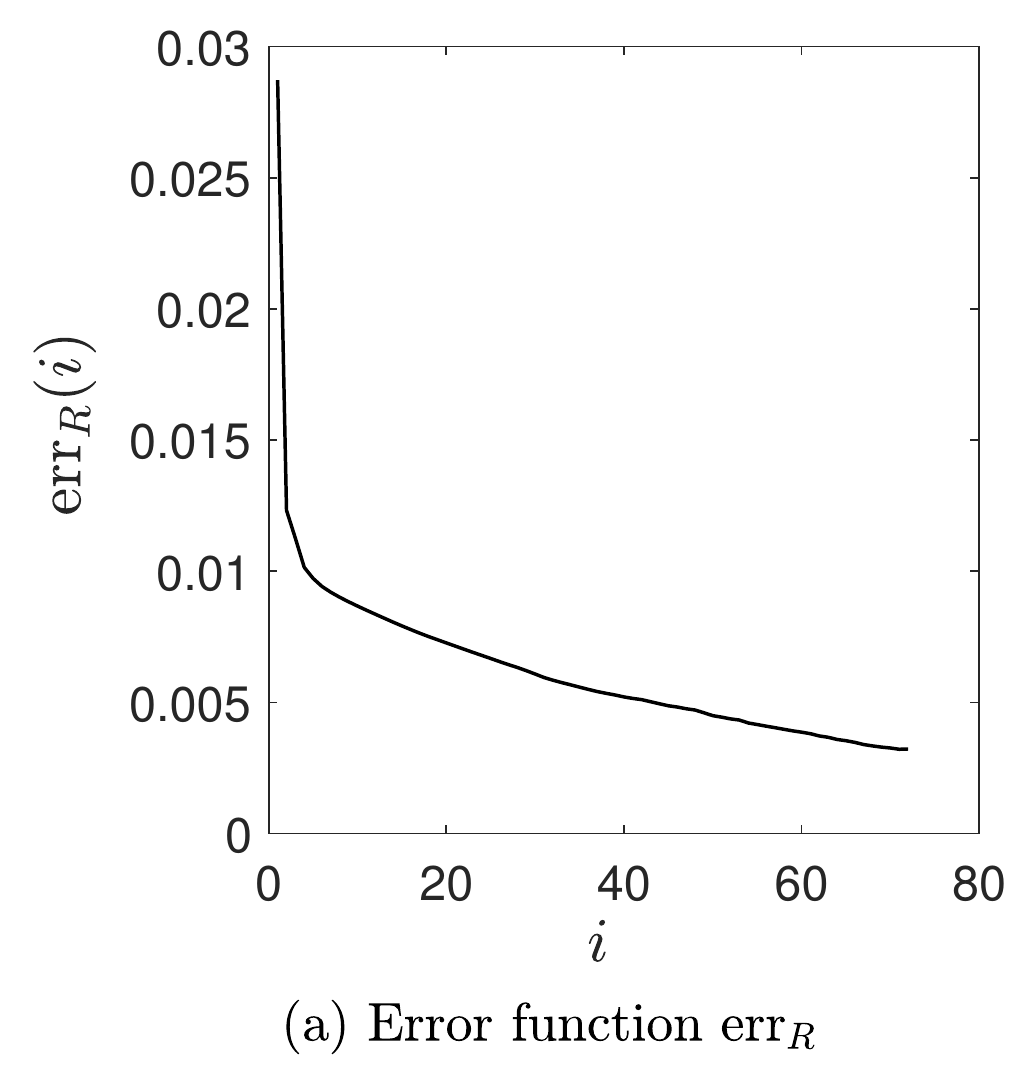}  \includegraphics[width=4.5cm]{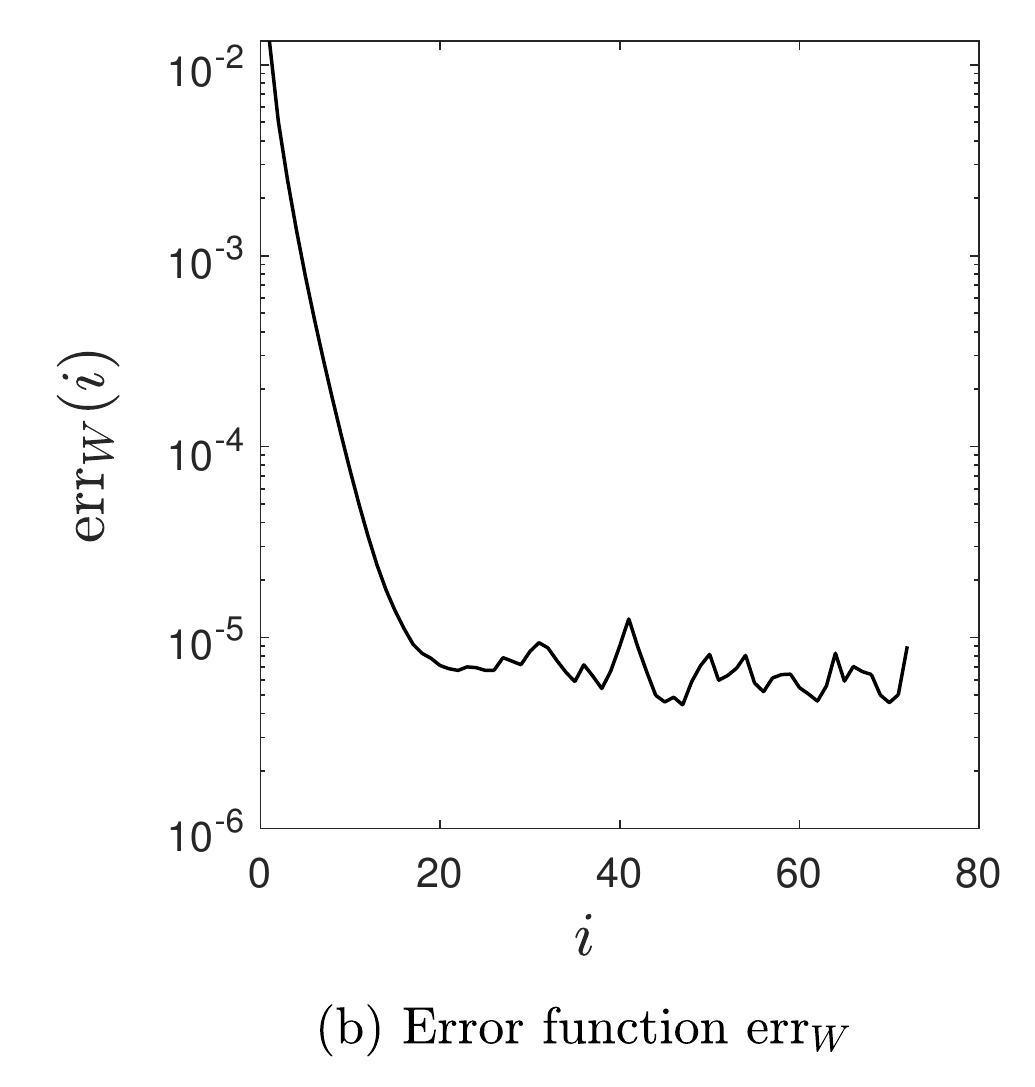} \includegraphics[width=4.5cm]{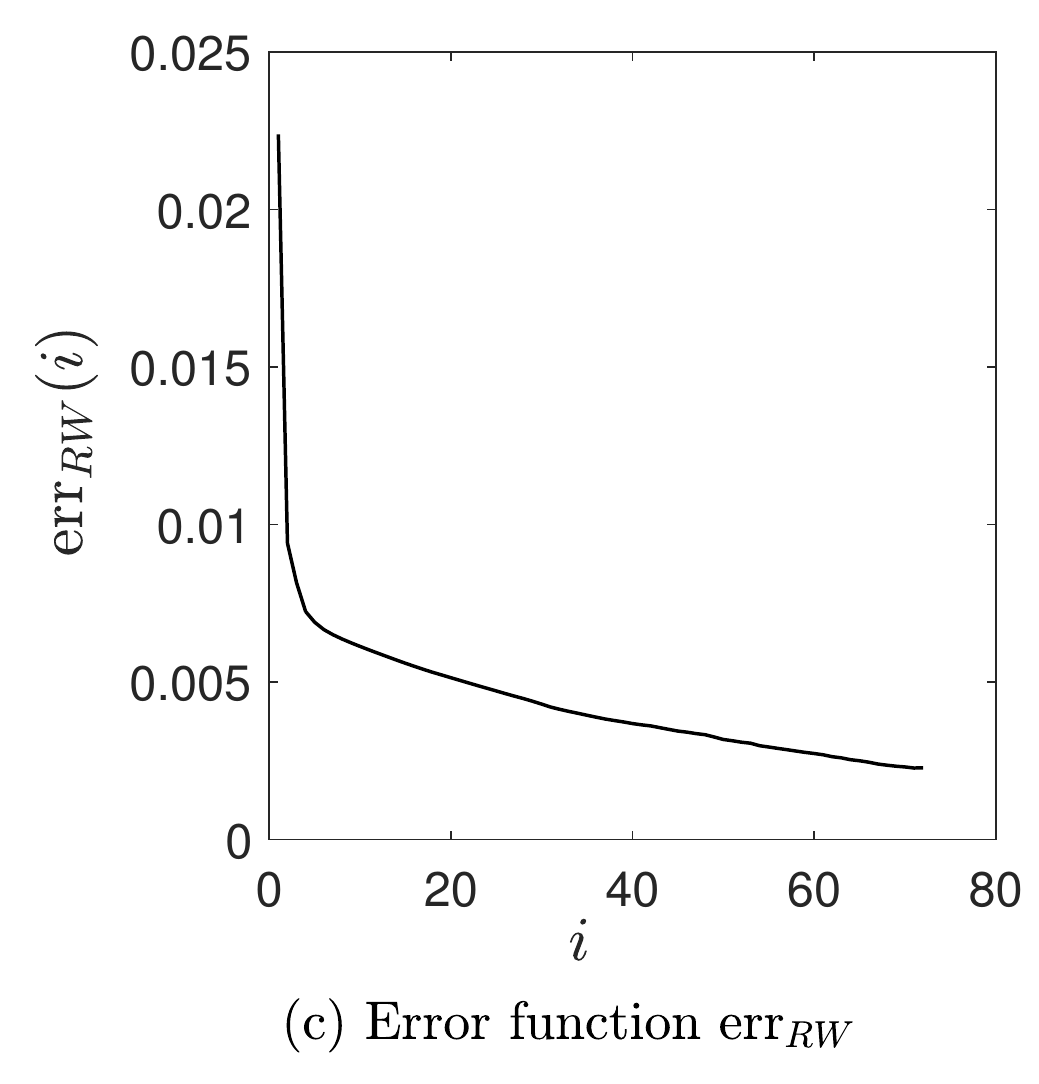}
\caption{For $N_d = 120$ with $n_\pMC = 1\,000$, and using $\algo~3$, evolution of the error functions $i\mapsto\errp_R (i)$ (a), $i\mapsto\errp_W (i)$ (b), and $i\mapsto\errp_{RW} (i)$ (c).}
\label{figureSD8}
\end{figure}
Figure~\ref{figureSD9} shows the pdf $w\mapsto p_{W_2}(w)$ and the pdf $w\mapsto p_{W_5}(w)$. In each figure is plotted the estimated pdf, on the one hand with the $N_d = 120$ realizations from the training dataset, on the other hand with the $n_\pMC = 1\,000$ additional realizations (learning) estimated with the PLoM without constraint and with the PLoM under the constraints on $\rho$ and $\bfW$ by using $\algo~3$.
The figures show that the estimation performed with the PLoM without constraint is already good. Indeed,
Fig.~\ref{figureSD8}-(b) shows that the error $\err_W(1) = 0.013$ at iteration $i = 1$, which corresponds to PLoM without constraint, is already small. So, for this case, the introduction of the constraint on $\bfW$ is not decisive, although it allows for improving the little that there is to improve.
The mean value  of $\widehat\rho$ is $6.35 \times 10^{-7}$, its standard deviation is $2.74 \times 10^{-7}$, and
$\Vert \,\widehat\rho\,\Vert_{L^2} = 6.95 \times 10^{-7}$.
In order to illustrate the distribution of the additional realizations generated by the PLoM under constraints, Fig.~\ref{figureSD10} displays the $n_\pMC = 1\,000$ additional realizations of components of random vector $\bfX^c = (\bfQ^c,\bfW^c)$. The probability distribution of  $(W^c_1,W^c_2,Q^c_1)$ is concentrated while the one for $(W^c_4,W^c_{10},Q^c_1)$ is not. For $N_d=120$ and using the nonlinear stochastic computational model,  for each one of the two observations, and for the probability level $p_c=0.98$, Fig.~\ref{figureSD11} shows the confidence region of the random function $\nu_\pfreq\mapsto \pdB_j(\nu_\pfreq)$ estimated using $n_\pMC=1\,000$ additional realizations generated with the PLoM under constraints. These confidence regions are almost identical to those of the reference stochastic solution shown in Fig.~\ref{figureSD2}.
\begin{figure}[h!]
  \centering
  \includegraphics[width=6.0cm]{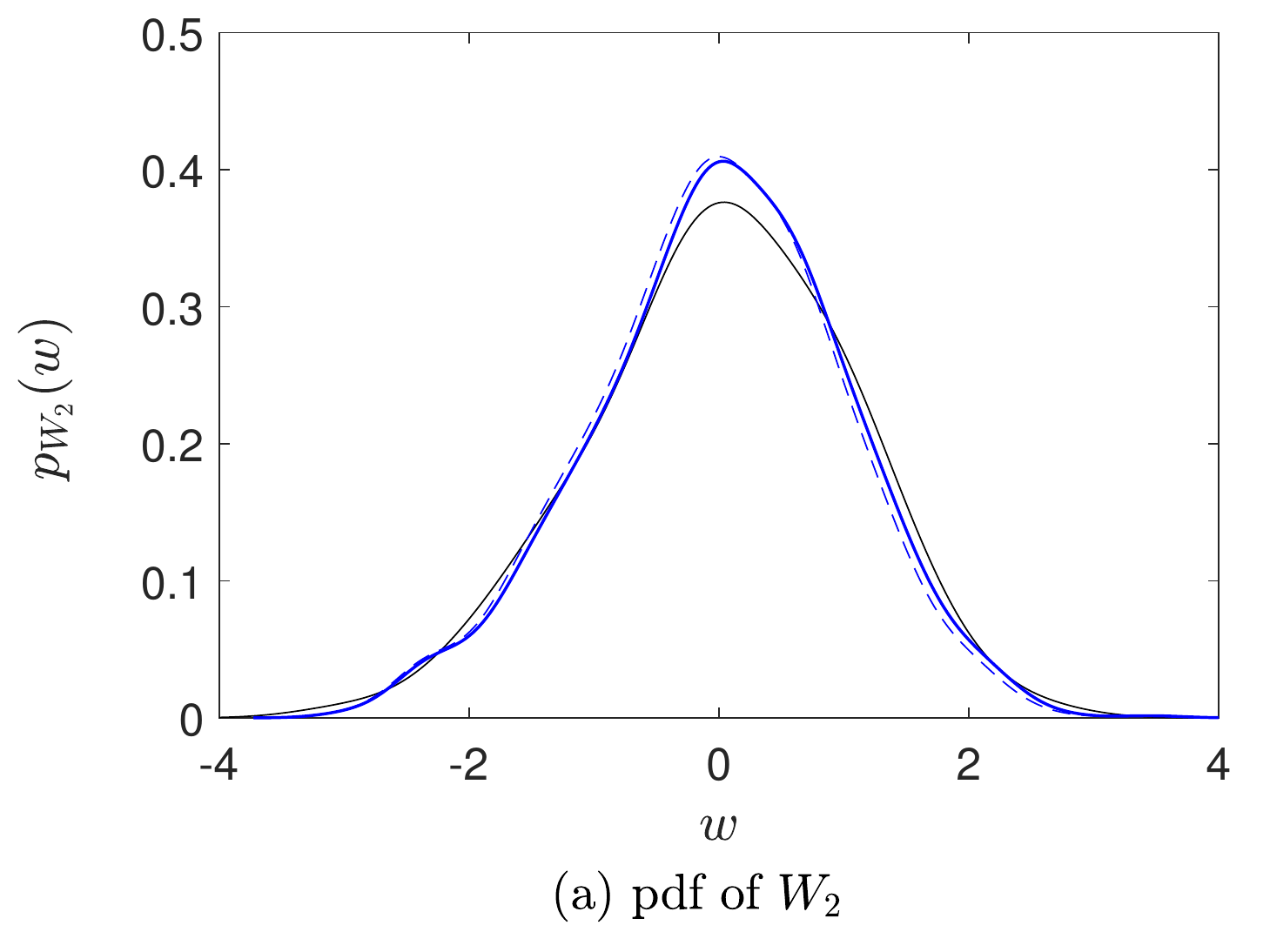} \hfil \includegraphics[width=6.0cm]{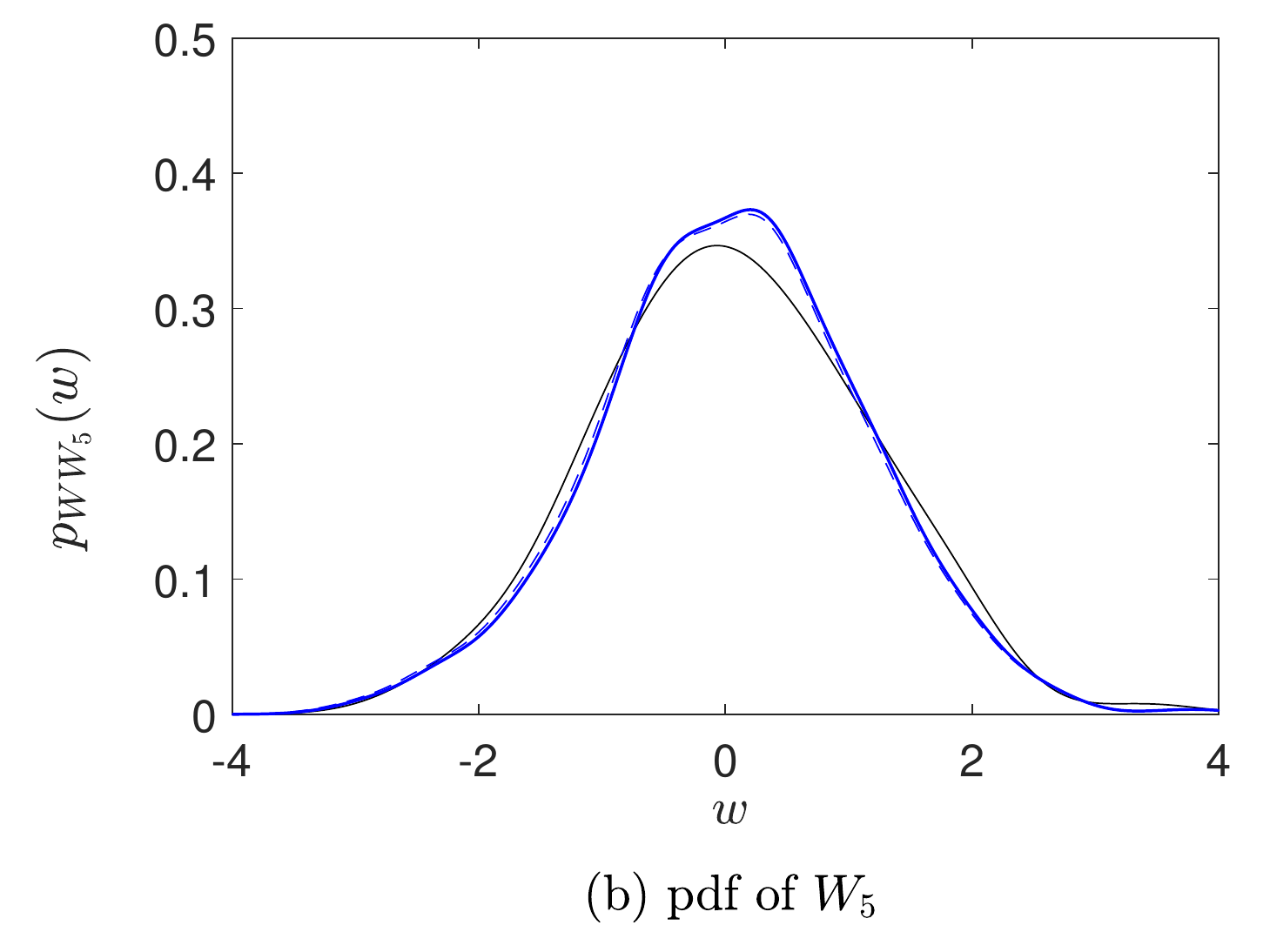}
\caption{For $N_d = 120$, pdf $w\mapsto p_{W_2}(w)$ (a) and pdf $w\mapsto p_{W_5}(w)$ (b), estimated with the $N_d = 120$ realizations (prior model) from the training dataset (black thin solid line), with $n_\pMC = 1\, 000$ additional realizations (learning) estimated with the PLoM without constraint (blue dashed line) and under the constraints by using $\algo~3$ (blue thick solid line).}
\label{figureSD9}
\end{figure}
\begin{figure}[h!]
\centering
\includegraphics[width=6.2cm]{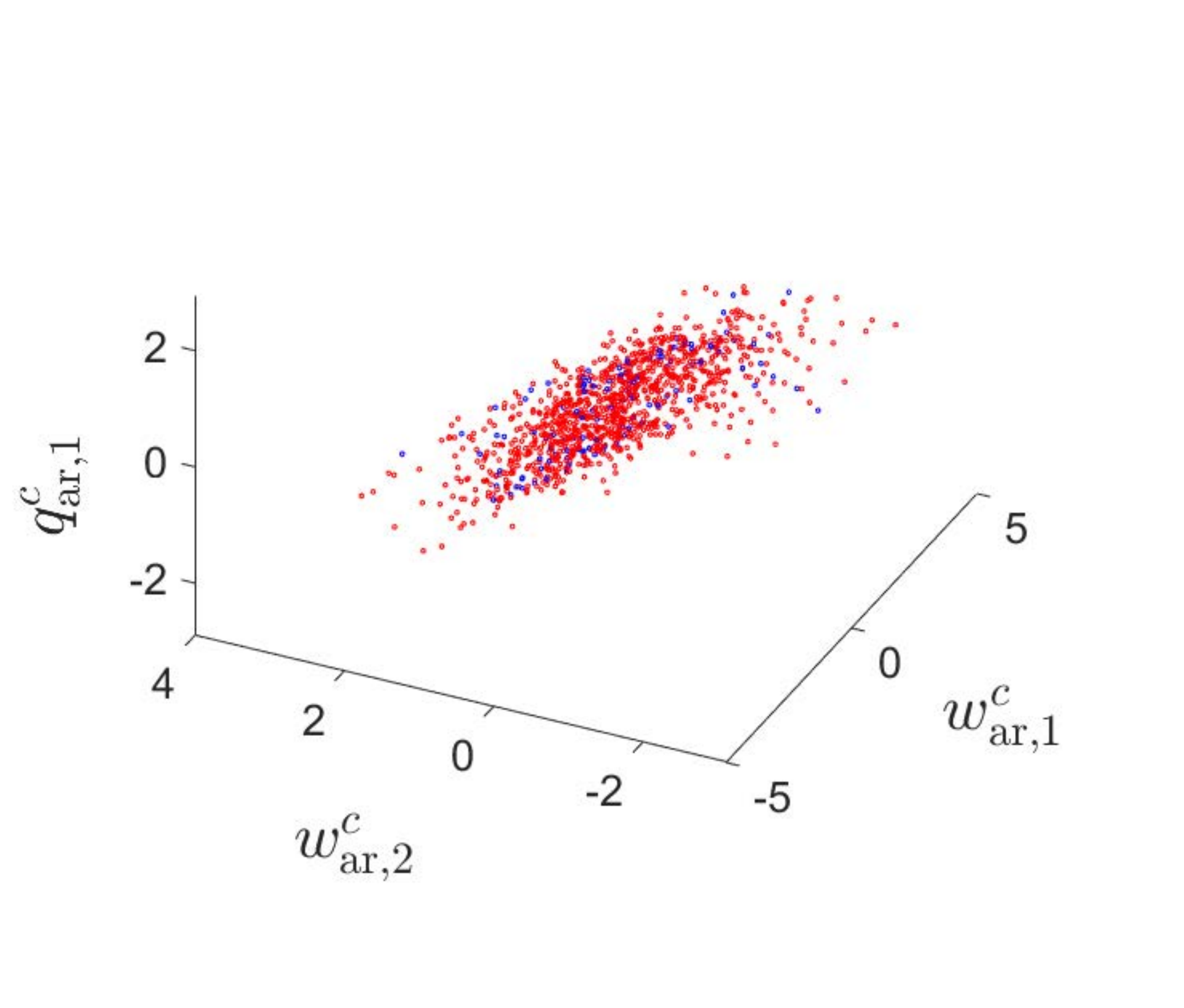} \hfil \includegraphics[width=6.2cm]{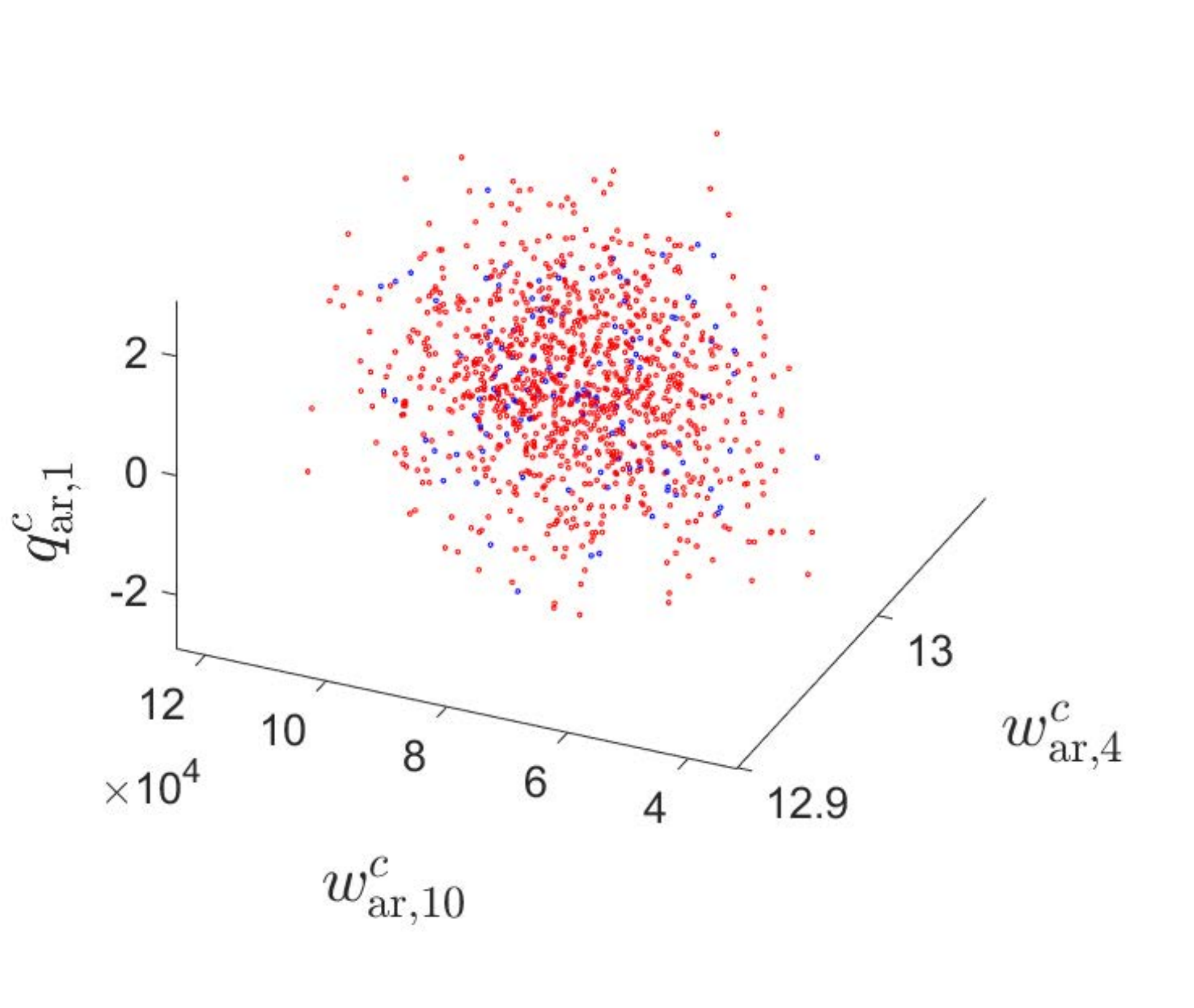}
\caption{3D-plot of realizations $(w^c_{\arp,1},w^c_{\arp,2},q^c_{\arp,1})$ of $(W^c_1,W^c_2,Q^c_1)$ (left figure) and $(w^c_{\arp,4},w^c_{\arp,{10}},q^c_{\arp,1})$ of $(W^c_4,W^c_{10},Q^c_1)$ (right figure), computed with the PLoM under constraints: $n_\pMC = 1\,000$ additional realizations (red symbols) and $N_d = 120$ points from the training dataset (blue symbols).}
\label{figureSD10}
\end{figure}
\begin{figure}[h!]
  \centering
  \includegraphics[width=6.0cm]{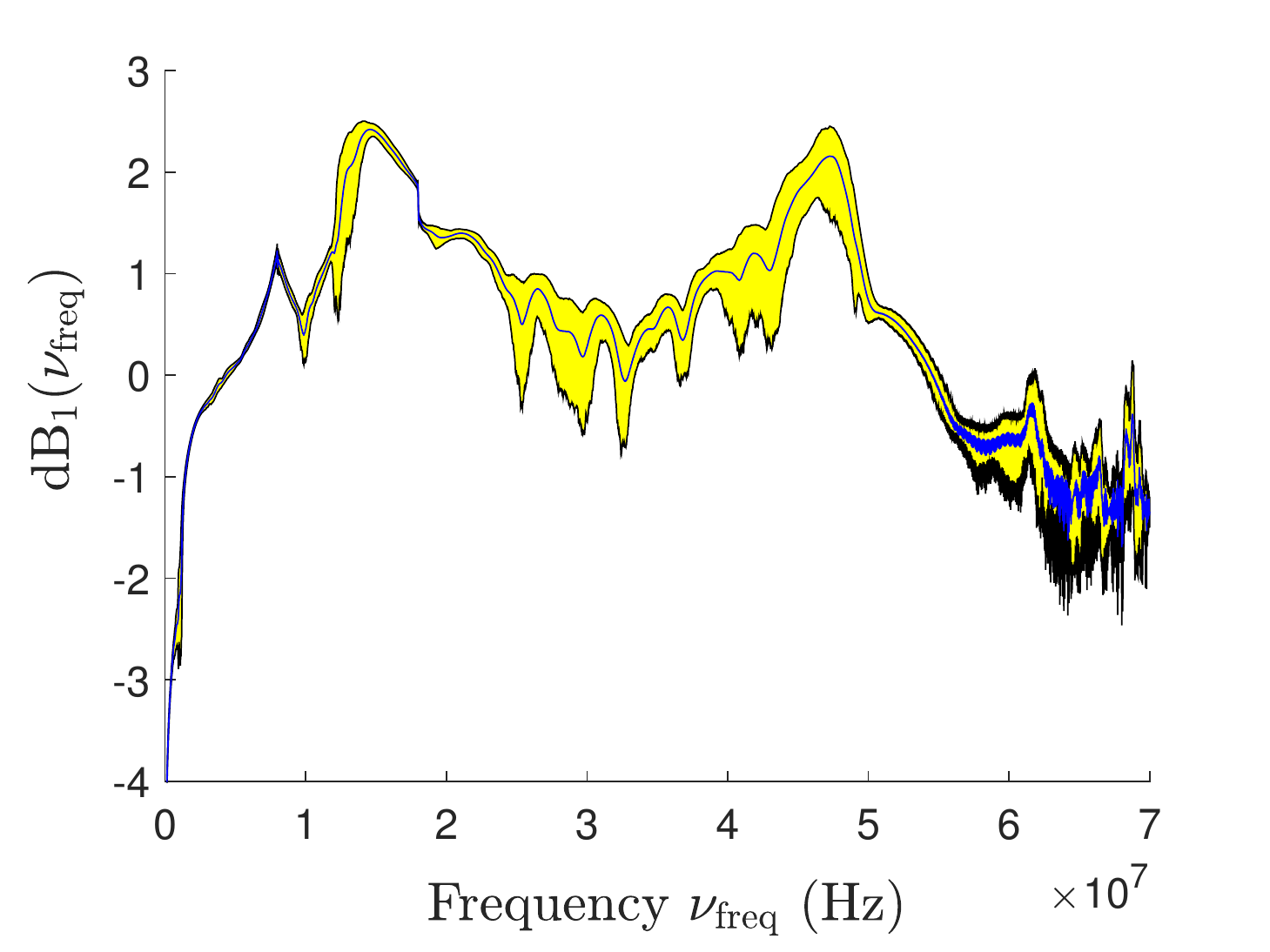} \hfil \includegraphics[width=6.0cm]{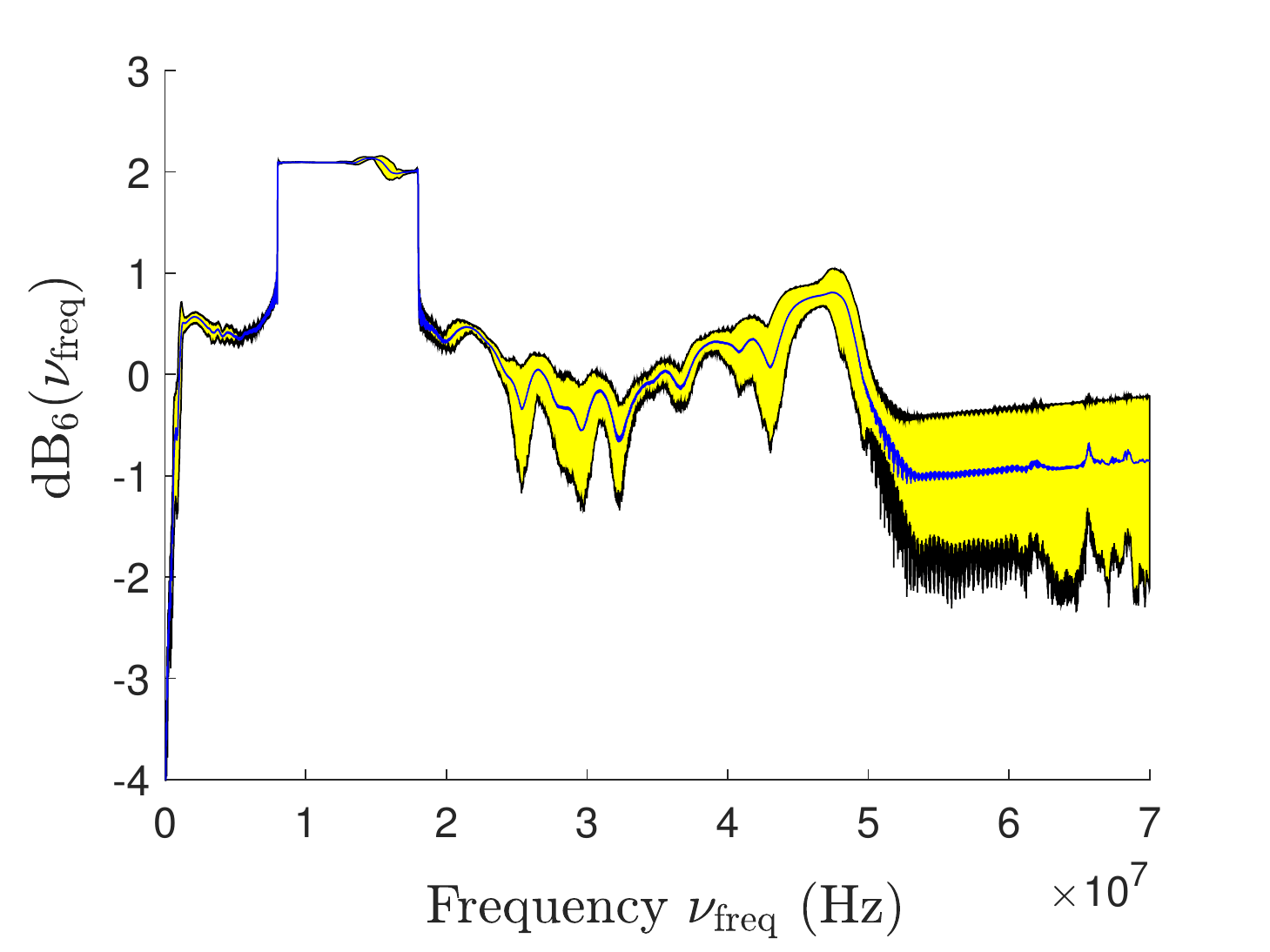}
\caption{For $N_d=120$, for $n_\pMC=1\,000$ additional realizations generated by the PLoM under constraints, for $\ppObs_j$, $j=1$ (left figure) and $j=6$ (right figure),  graph of the confidence region (yellow region with thin lines for the upper and lower envelopes) of the random function $\nu_\pfreq\mapsto \pdB_j(\nu_\pfreq)$. The thick line inside the yellow region is the estimate of mean value $\nu_\pfreq\mapsto E\{\pdB_j(\nu_\pfreq)\}$.}
\label{figureSD11}
\end{figure}
\subsection{Discussion}
\label{SectionSD7}
Similarly to the CFD application, the results obtained for this application devoted to the nonlinear computational stochastic dynamics of elastic solids show that the PLoM method without constraints gives good results, which was expected for but not at this level of quality. The use of the proposed extension, consisting in constraining the PLoM method by the stochastic PDE allows to verify the convergence and the robustness of the algorithms.
The examination of the confidence domains of the stochastic responses in the frequency domain does not show any significant differences (compare Fig.~\ref{figureSD4} with Fig.~\ref{figureSD11}). This application therefore also provides an additional validation of the PLoM method without constraint.
\section{Conclusion and discussion}
A novel extension of the PLoM method has been presented. It makes it possible to generate a large number of time-dependent additional realizations that verify the stochastic PDE  used to generate the training dataset. This is paramount for the case where only a small number of solutions can be secured from the numerical model due to computational cost. The method is general and can be applied to linear and nonlinear time-dependent stochastic boundary value problems.
Its implementation requires specialized algorithms related to the Karhunen-Lo\`eve expansions of high-dimension nonstationary stochastic processes, in order to circumvent potential difficulties associated with RAM limitations, thus providing an efficient alternative to I/O (Input-Output) on external memory.
Taking into account the scalar constraint related to the random residual of the stochastic PDE,  a novel statistical formulation and its  implementation have been developed for estimating the  gradient of the potential that appears in the drift vector of the reduced nonlinear It\^o stochastic differential equation, which is the MCMC generator.
The prior probability model of the vector-valued random control parameter of the stochastic boundary value problem is given and must not be modified by the PLoM under the constraint related to the random residual. In order to guarantee this property, we have had to introduce a vector constraint on the random control parameters.
The results obtained through the presented applications in the field of nonlinear computational fluid and solid stochastic dynamics, allow the following conclusions to be drawn.

- The PLoM constrained by PDE provides further validation of the PLoM method without constraint, proposed in \cite{Soize2016} for which a complete mathematical analysis has been proposed in \cite{Soize2020c}.  Although the obtained  results concerning the PLoM without constraint were expected, they are particularly good and interesting in terms of statistical methods of learning with  small data.

- The PLoM constrained by stochastic PDE allows a direct control of the random residual of the stochastic PDE and thus one is assured that the probability measure constructed by PLoM under constraint, checks better the stochastic boundary value problem that is considered.
This novel approach  makes it possible to improve learning from small data when the random residual of the stochastic PDE computed with the  PLoM without constraint is not sufficiently small. We  observe this for the CFD application for which the  PLoM under constraints gives  excellent results and improve the PLoM without constraint. It should also be noted that the robustness of the PLoM algorithms under constraints allows for maintaining the quality of the results given by the PLoM without constraint when these results are already good, as  can be seen in application~3 devoted to nonlinear solid dynamics.

\section*{Acknowledgments}
This material is based upon work supported by the Defense Advanced
Research Projects Agency (DARPA) under Agreement No. HR00111990032 and
by FASTMATH, a SciDAC Institute (Scientific Discovery through Advanced Computing) funded by the U.S. Department of Energy (DOE), Office of Science, Advanced Scientific Computing Research (ASCR).

\appendix
%
%
%
\section{St\"{o}rmer-Verlet algorithm for solving Eqs.~\eqref{eq8.20} to \eqref{eq8.22}}
\label{AppendixA}
The ISDE defined by Eqs.~\eqref{eq8.20} to \eqref{eq8.22} is solved for $r\in [0,r_\pmax]$ with $r_\pmax =(\ell_0 + n_\pMC\!\times\! M_0)\, \Delta r$ in which $\Delta r$ is the sampling step, where $\ell_0$ is chosen in order that the solution of Eqs.~\eqref{eq8.20} to \eqref{eq8.22} has reached the stationary regime, and where $M_0$ allows for controlling the number of sampling steps between two consecutive  realizations for obtaining a quasi-independence of the computed realizations.
Taking as the origin of $\ell$, the value $\ell_0$, for $\ell=0,1,\ldots, n_\pMC\!\times\! M_0$, we consider the sampling points $r_\ell = \ell\, \Delta r$ and the following notations:
$[\bfcurZ_\ell] = [\bfcurZ_{\bflambda}(r_\ell)]$, $[\bfcurX_\ell] = [\bfcurX_{\bflambda}(r_\ell)]$, and $[\bfW^\wien_\ell] = [\bfW^\wien(r_\ell)]$.
The St\"{o}rmer-Verlet scheme is written, for $\ell=0,1,\ldots, n_\pMC\!\times\! M_0$, as
\begin{equation}
[\bfcurZ_{\ell+\frac{1}{2}}]     =    [\bfcurZ_{\ell}] +\frac{\Delta r}{2} \, [\bfcurX_{\ell}] \, ,    \nonumber
\end{equation}
\begin{equation}
[\bfcurX_{\ell+1}]   =    \frac{1-\beta}{1+\beta}\, [\bfcurX_{\ell}] + \frac{\Delta r}{1+\beta}\, [\bfcurL_{\ell+\frac{1}{2}}] +
       \frac{\sqrt{f_0}}{1+\beta}\, [\Delta\bfW^\wien_{\ell+1}]\, [a]\, ,                                                          \nonumber
\end{equation}
\begin{equation}
[\bfcurZ_{\ell+ 1}]  =    [\bfcurZ_{\ell+\frac{1}{2}}] +\frac{\Delta r}{2} \, [\bfcurX_{\ell+1}]\, ,                                \nonumber
\end{equation}
with the initial condition defined by Eq.~\eqref{eq8.22}, where
$\beta=f_0\, \Delta r\, / 4$, and where $[\bfcurL_{\ell+\frac{1}{2}}]$ is the $\MM_{\nu,m}$-valued random variable such that
$[\bfcurL_{ \ell+\frac{1}{2} }]  = [\curL_{\bflambda}([\bfcurZ_{\ell+\frac{1}{2}}])]
                                 =  [L_{\bflambda}([\bfcurZ_{\ell+\frac{1}{2}}]\, [g]^T)] \, [a]$.
In the above equation, $[\Delta\bfW^\wien_{\ell+1}]$
is a random variable with values in $\MM_{\nu,N_d}$, in which the  increment
$[\Delta\bfW^\wien_{\ell+1}] = [\bfW^\wien (r_{\ell+1})] - [\bfW^\wien (r_{\ell})]$. The increments are statistically independent.
For all $\alpha=1,\ldots , \nu$ and for all $j=1,\ldots , N_d$, the real-valued random variables $\big\{[\Delta\bfW^\wien_{\ell+1}]_{\alpha j}\big\}_{\alpha j}$ are independent, Gaussian, second-order, and centered random variables such that
$E\big\{[\Delta\bfW^\wien_{\ell+1}]_{\alpha j}[\Delta\bfW^\wien_{\ell+1}]_{\alpha' j'}\big\}=\Delta r \,\delta_{\alpha\alpha'}\,\delta_{jj'}$.
%
%

\bibliographystyle{elsarticle-num}
\bibliography{references}

\end{document}